\documentclass[12pt, a4paper]{report}

\usepackage[utf8]{inputenc}
\usepackage[T1]{fontenc}
\usepackage[english]{babel} 
\usepackage[a4paper, margin=2.5cm]{geometry}
\usepackage{graphicx}
\usepackage[hidelinks]{hyperref}

\usepackage{palatino}
\usepackage[scaled=0.8]{beramono}

\usepackage{amsmath, amsfonts, amssymb, mathrsfs, bbold}
\usepackage{gensymb, cancel}

\usepackage{booktabs, array, multirow}
\usepackage{caption}
\usepackage{float}
\usepackage{enumitem}
\usepackage[usenames,dvipsnames]{xcolor} 

\usepackage{comment}
\usepackage{listings}
\usepackage[backend=bibtex]{biblatex}
\usepackage[title]{appendix}

\newenvironment{dedication}
{
   \cleardoublepage
   \thispagestyle{empty}
   \vspace*{\stretch{1}}
   \hfill\begin{minipage}[t]{0.6\textwidth}
   \raggedleft 
}
{
   \end{minipage}
   \vspace*{\stretch{3}}
   \clearpage
}

\newcommand{\gab}[1]{\textcolor{red}{\small[Gabriele: #1]}}

\begin{document}

\pagenumbering{roman} 

\begin{titlepage}
    \begin{center}
        {\large Università degli Studi di Trieste \par}
        {\large \bfseries Dipartimento di Matematica, Informatica e Geoscienze \par}
        
        \vspace{1cm}
        
        \includegraphics[width=0.25\textwidth]{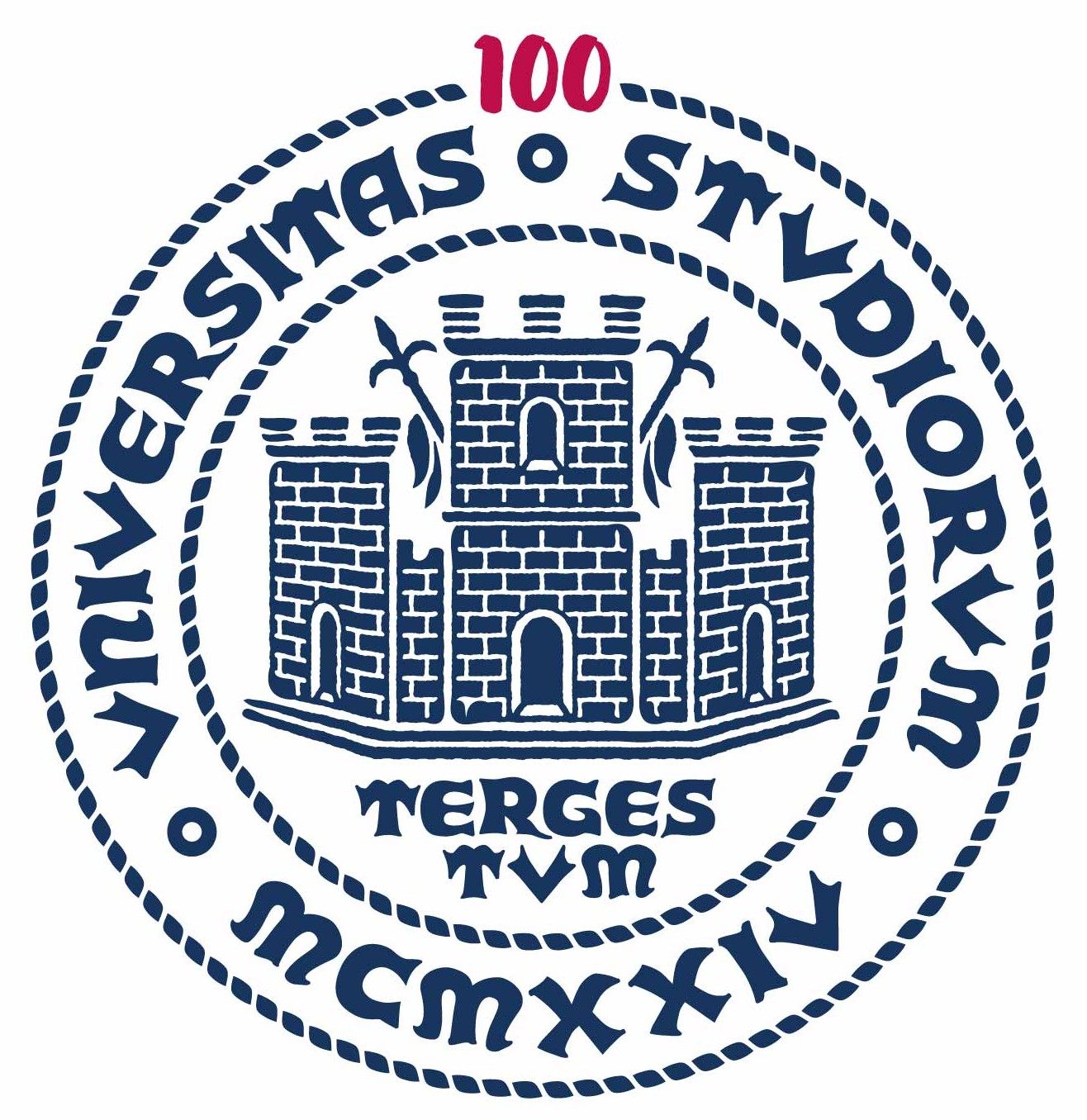}
        
        \vfill 
        
        {\Large \bfseries Corso di Laurea Magistrale in \\ Data Science and Scientific Computing \par}
        \vspace{0.5cm}
        {\large TESI DI LAUREA MAGISTRALE \par}
        
        \vspace{1.5cm}
        
        {\LARGE \bfseries A Dialectic Pipeline for Improving LLM Robustness \par}
        
        \vfill \vfill 
    \end{center}

    \noindent
    \begin{minipage}[t]{0.5\textwidth}
        {\large Relatore: \\ \textbf{Prof. Luca Bortolussi} \par \medskip}
        {\large Correlatore: \\ \textbf{Dott. Gabriele Sarti}}
    \end{minipage}
    \hfill
    \begin{minipage}[t]{0.4\textwidth}
        \raggedleft
        {\large Candidata: \\ \textbf{Sara Candussio}}
    \end{minipage}

    \vfill
    
    \hrulefill
    \begin{center}
        {\large \textbf{Anno Accademico 2023/2024}}
    \end{center}
\end{titlepage}

\begin{dedication}
    \textit{L'inferno dei viventi non è qualcosa che sarà; se ce n'è uno, è quello che è già qui, l'inferno che abitiamo tutti i giorni, che formiamo stando insieme.} 
    
    \textit{Due modi ci sono per non soffrirne.} 
    
    \textit{Il primo riesce facile a molti: accettare l'inferno e diventarne parte fino al punto di non vederlo più.} 
    
    \textit{Il secondo è rischioso ed esige attenzione e apprendimento continui:}
    
    \textit{cercare e saper riconoscere chi e cosa, in mezzo all'inferno, non è inferno, e farlo durare, e dargli spazio.}
    
    \vspace{2mm}
    Le Città Invisibili, Italo Calvino
\end{dedication}

\thispagestyle{plain}
\begin{center}
    \Large
    \textbf{Abstract}
    \vspace{0.5cm}
\end{center}
Assessing ways in which Language Models can reduce their hallucinations and improve the outputs' quality is crucial to ensure their large-scale use. \\

However, methods such as fine-tuning on domain-specific data or the training of a separate \textit{ad hoc} verifier require demanding computational resources (not feasible for many user applications) and constrain the models to specific fields of knowledge. \\

In this thesis, we propose a dialectic pipeline that preserves LLMs' generalization abilities while improving the quality of its answer via self-dialogue, enabling it to reflect upon and correct tentative wrong answers. \\

We experimented with different pipeline settings, testing our proposed method on different datasets and on different families of models. All the pipeline stages are enriched with the relevant context (in an oracle-RAG setting) and a study on the impact of its summarization or its filtering is conducted.\\

We find that our proposed dialectic pipeline is able to outperform by significative margins the standard model answers and that it consistently achieves higher performances than Chain-of-Thought only prompting.

\begin{center}
    \Large
    \vspace{0.5cm}
    \textbf{Sommario}
    \vspace{0.5cm}
\end{center}

Al fine di garantire l'uso su larga scala del Language Models, è fondamentale trovare delle strategie con cui è possibile ridurre le loro allucinazioni e quindi migliorare la qualità dei loro outputs. \\ 

Tuttavia, metodi come il fine-tuning su dati specifici o l'addestramento di verifiers \textit{ad hoc} richiedono risorse computazionali elevate (non accessibili per molte applicazioni utili all'utente) e limitano i modelli ad alcuni precisi ambiti di conoscenza. \\

In questa tesi, proponiamo una pipeline dialettica che preserva le capacità di generalizzazione dei LLMs migliorando al contempo la qualità delle loro risposte attraverso l'auto-dialogo, consentendo al modello di riflettere e correggere risposte inizialmente errate. \\

Abbiamo sperimentato diverse configurazioni di auto-dialogo, testando il nostro metodo su vari dataset e su diverse famiglie di Language Models. Tutte le fasi di questa pipeline dialettica sono state arricchite con il contesto rilevante per rispondere al prompt (in un setting oracle-RAG) ed è stato condotto uno studio sull'impatto del riassunto o del filtraggio di questo.\\ 

Abbiamo riscontrato che la pipeline dialettica proposta è in grado di superare con margini significativi le risposte standard del modello e che riesce ad ottenere in maniera consistente prestazioni superiori rispetto al solo prompting con Chain-of-Thought.

\newpage
\pagenumbering{arabic}
\setcounter{page}{1}

\chapter{Introduction}
In 2017, the field of Natural Language Processing had been revolutionized by Vaswani et al. \cite{vaswani2023}, leading to a mass-scale interest towards neural networks applied to text generation. The era of Large Language Models (LLMs) began with the intuition that a recurrent structure to is not mandatory to face the given task; what is necessary is just the \textit{attention mechanism} (and a supporting complex model structure, as we will show in section \ref{transformer-block}).\\

Many efforts were made by the researchers' community from that moment on in order to improve the quality of the produced text: these models were often object of the so-called \textit{hallucinations}, i.e. the generation of false or misleading information. This phenomenon is due to many reasons, often caused by data issues: incomplete, noisy, biased or not updated training data tend to drive the auto-regressive towards the wrong output. Additionally, an unusual prompting method or an imprecise instruction could confuse the model. Different strategies are used nowadays to reduce the \textit{hallucinations} frequency in LLMs: curated pre-training datasets and fine-tuning, RAG-powered applications (Section \ref{rag}), human-based alignment, Chain-of-Thought prompting (\ref{cot-method}) are all aimed at improving the models' prediction quality. \\

We propose a new solution to face this problem, consisting in prompting the model to reason multiple times on the answer to a question before definitively choosing the correct option. Our approach aims at improving the quality of the \textit{spontaneous} answer by incrementally checking whether and why it is correct. This approach constructs a \textit{dialectic pipeline} for generating the final output, where the candidate answer is checked and examined by two steps before being effectively chosen.
The term \textit{dialectic} is due to the fact that the model is dialoguing with itself in three different stages: the candidate answer production, the first comment on the correctness of it, and the final decision on which should be the final answer. We will refer to the first stage as the \textit{thesis}, to the second as \textit{antithesis} and to the latter as \textit{synthesis}. Differently from the Hegelian dialectic \cite{dialectic}, the \textit{antithesis} step is not forced to refute the \textit{thesis}' one, neither is the \textit{synthesis} to compromise between the other two. However, we find these names as representative for the role that they have in the process: the \textit{thesis} makes a first guess, the \textit{antithesis} has to check whether or not it is correct (and why), and the \textit{synthesis} further merges these opinions into a final, reasoned answer. \\

We are going to test this method on \textit{multi-hop} question-answering datasets, namely on tasks that require to properly merge multiple knowledge sources to answer the question. This kind of problems requires both content extraction and reasoning abilities, thus are more challenging than standard RAG-problems. We are going to consider different sub-tasks inside this broader class, consisting of multiple passages (\ref{wikihop-hops}) to exploit in slightly different ways, for example by comparing them or by constructing links between them (\ref{comparison-vs-bridge}).  In all the pipeline steps, we allow the model to access the relevant context in order to properly answer to the question while dealing with (implicit) challenging context pre-processing. \\

The research questions for which we search an answer through our experiments are oriented to assess whether or not this \textit{dialectic} pipeline can compete against well-established other methods in terms of accuracy and reliability (\ref{cot-results}). 
We want to test whether the pipeline is able to reach good accuracy values and whether this is maintained when we consider different models inside the pipeline. We test both the robustness across different families of models (\ref{gemma-2}, \ref{phi-mini}, \ref{llama}) and with respect to the same model but with different sizes (\ref{model-robust}). \\

We also try this pipeline on different datasets to assess the non-specificity of the proposed method: we will consider two \textit{multi-hop} datasets with various numbers of \textit{hops} and requiring different reasoning strategies to merge them (\ref{dataset-robust}). 
We are going to define some pipeline \textit{variations} that induce the models to behave differently when asked to solve a task (\ref{variations-overview}). \\

We check whether the \textit{synthesis} step is necessary to improve the models' performances and by which margin; this is due to the idea that the \textit{antithesis} step could be misleaded by the context or simply wrong. Consequently, we run experiments comparing the \textit{antithesis}' proposed answer and the \textit{synthesis}' ones, in order to assess whether the latter is in fact useful in prediction terms (\ref{ant-vs-syn}). \\

Finally, we test whether the \textit{way} in which the context is provided to the pipeline affects the answers' quality: works such as S2A (\ref{s2a}) and RE-RAG (\ref{re-rag}) spot a light on the beneficial effect of carefully selecting only the relevant parts of it and use only these to generate the output text for a given prompt. Consequently, we compare the answers obtained by passing the original, the summarized (\ref{summ}) and the filtered context to the pipeline. The filtering approach exploits a gradient-based attribution method (\ref{pecore}) that is used to select the passages' sentences that are found to be influential for the output generation (\ref{my-pecore}). \\

Our work is structured as follows. 
\begin{itemize}
\item The first section (\ref{literature}) is dedicated to a literature overview. It delves into the technical details of Transformer models and explains the differences among those we will consider in our experiments. Then it focuses on the existing approaches to overcome the problem of \textit{hallucinations}, such as SFT and RAG (\ref{rag}), training logical verifiers to check the correctness of the answer before outputting it (\ref{nli}) and prompting strategies aimed at showing the expected behaviour of the model before executing it. We consequently introduce possible alternatives to our approach, such as Chain-of-Thought prompting (\ref{cot-method}) and self-refinement methods (\ref{self-refinement}). Finally, we inspect ways in which the context can improve the answers' quality (\ref{context}, \ref{reasoning-on-context}).
    
\item We then present the data (\ref{data}) and highlight the pre-processing used to make these \textit{multi-hop} datasets also multiple choice: this is done to avoid approximate, LLM-based evaluations on the correctness of the answer.

\item The methods section (\ref{methods}) explains largely how we constructed the pipeline (\ref{dialectic-pipeline}), how practically we filtered (\ref{my-pecore}) and summarized (\ref{summ}) the context and how we assessed performances (\ref{guidance}, \ref{performances}). 

\item Finally, we run answer the research question described before in section (\ref{results}), decreeing whether or not the proposed method works and in which cases it performs better or worse. Final conclusions are drawn in section (\ref{conclusions}). 

\end{itemize} 

\newpage 

\chapter{Literature review} \label{literature}

In this section we go through a set of relevant works for our study. We begin by studying how the models that we use are structured, architecturally speaking; we then highlight the differences between the classes of LLMs exploited for our experiments. \\

Subsequently, we provide an overview of methods aimed at improving the output quality in terms of factuality, style and consistency with the given examples. \\

Finally, we highlight how these discoveries and proposed methods are related to our new approach, and how much it owes to other researchers' work.

\section{Transformer models} \label{transformers}

Language Models (LMs) are inherently sequential, processing input data in a step-by-step manner. Early LMs leveraged this sequential structure through Recurrent Neural Network (RNN) architectures, which were designed to capture dependencies in data sequences. However, RNNs faced limitations in scalability due to their reliance on sequential processing and their fixed-size context representations. To address these challenges, the \textit{attention mechanism} was introduced, providing a more flexible way to handle context by allowing models to focus on different parts of the input sequence as needed.

\subsection{The Self-Attention mechanism} \label{sa}

The \textit{attention mechanism}, originally proposed by Bahdanau, Cho and Bengio \cite{bahdanau2016} to improve machine traslation with RNNs, has assumed a central role when Vaswani et al. \cite{vaswani2023} published a new Language Models architecture based on it.

From that moment on, this approach has been exploited massively, displacing RNNs from the state-of-the-art techniques and becoming the new paradigm for text generation tasks.\\

The key point is to represent each token in the sequence with respect of the other tokens present in it, i.e. produce a meaningful \textit{contextual representation} of the words. By comparing an item of interest to a collection of other items, we can reveal their \textit{relevance} in the context in which they are placed. The results of these comparisons are then used to compute an output sequence for the current input sequence. \\

Diving into mathematical aspects, the comparison is performed by considering the dot product between two word vector representations and produces a score representing the relevance of one with respect to the other: 
$$
\text{score}(x_i, x_j) = x_i \cdot x_j
$$
and this quantity takes values in the $ \Bigl[ - |x_i| \cdot |x_j|,  |x_i| \cdot |x_j| \Bigr]$  interval\footnote{If $x_i$, $x_j \in \mathbb{R}^n$ (i.e. they are finite-dimensional), the bounds are finite. Otherwise, $ \text{score}(x_i, x_j) \in \left[ -\infty, +\infty \right]$.}, where the magnitude of the score is greater when the relevance of the tokens is greater. \\

Once these scores are computed, we are interested in translating them in relative relevance scores, i.e. finding a function that maps $\mathbb{R} \rightarrow [ 0, 1]$. One of the most popular choices in the machine learning field is the $ \text{softmax}(\cdot)$ function: 
$$
\alpha_{ij} = \text{softmax}(\text{score}(x_i, x_j)) = \frac{\text{exp}(\text{score}(x_i, x_j))}{\sum_j \text{exp}(\text{score}(x_i, x_j))}
$$

where the normalization is performed with respect to all the elements $x_j$ in the sequence. Thus each word $x_i$ will have an array of vector weights $\alpha^i = \left[ \alpha_{i1}, ..., \alpha_{iJ}\right]$ associated to it. \\

Now that we have the proportional relevance of each word in the sequence \textit{with respect to} $x_i$, we can compute a linear combination between each proportional relevance $\alpha_{ij}$ and each word $x_j$ in the sequence, resulting in the relevance of word $x_i$ in the context in which is immersed:
$$
a_i = \sum_j \alpha_{ij} \cdot x_j
$$
this is the so-called \textit{attention weights} associated to word $x_i$.\\

The process may appear non-trivial due to the shift of focus that happens in the last step. Firstly we focus on one word, consider the relevance of the other words with respect to it; then we use this information to compose a relevance score for the considered word. \\

The big implicit step in this procedure is the fact that if $x_j$ is a sequence of characters from a certain alphabet we need a further step that embeds each token into a numerical vector, such that it is possible to perform the previous computations. We will not dig deeper than this on the embedding details in this paragraph, but we will talk about this in section (\ref{embed}). \\

However, word embedding is not the only transformation imposed to tokens. The \textit{self-attention mechanism} represents differently the same word with respect to the \textit{role} that it plays in the algorithm. The three roles that each word may have are: 

\begin{itemize}
\item the \textbf{query}: the current focus of attention, i.e. the word compared to the others;
\item the \textbf{key}: the \textit{other} word compared to the current focus of attention to produce the \textit{proportional relevance weights};
\item the \textbf{value}: the \textit{other} word used to compute each \textit{attention weight} for the current word.
\end{itemize}

To produce these representations of the same word $x_j$, it is necessary to apply three different transformations that can be represented as matrices: 
$$
q_i = x_i \cdot W_Q \,; \quad k_j = x_j \cdot W_K \,; \quad v_j = x_j \cdot W_V
$$
being the query $W_Q \in \mathbb{R}^{d \times d_k}$, key $W_K \in \mathbb{R}^{d \times d_k}$ and value $W_V \in \mathbb{R}^{d \times d_v}$ matrices. \\

We need to transform $x_i$ and $x_j$ using the matrices $W^Q$, $W^K$ and $W^V$ since, if we do not account for these projections, the model would be forced to use the same hidden vector for these three different tasks. But the query, the key and the value roles require that the same token performs different tasks: consequently, the vectors have to be transformed in order to let the matrices $W^Q$, $W^K$ and $W^K$ absorb this role in their place. \\

With this in mind, we can rephrase the previous self-attention mechanism as: 
$$
\text{score}(x_i, x_j)  = q_i \cdot k_j \,; \quad a_i = \sum_j \alpha_{ij} \cdot v_j
$$

A final precaution has to be made: the dot product between two vectors can assume arbitrarily large values, leading to numerical issues and hindering the effective propagation of the gradients during the training. To avoid this scenario, a re-scaling of the dot product is performed:
$$
\text{score}(x_i, x_j)  = \frac{q_i \cdot k_j}{\sqrt{d_k}}
$$
where $d_k$ is the dimensionality of the query and the key vectors. \\

Transformers are able to take in input, as preceding context, a maximum number of tokens. This quantity is called \textit{context length} and currently takes values 4K or 8K, even though many modern models are more capacious. 

Previously, we considered one token in the sequence at a time; more precisely, $x_i$ is the embedding of a word in the sequence, and it has dimension $1 \times d$, where $d$ is the embedding dimensionality. We can now assume that the input of the self-attention operation is the matrix $X \in \mathbb{R}^{N \times d}$, containing the $d$-dimensionality embeddings of all the $N$ words in the context.  In the self-attention operation of the first model layer, this input has to be multiplied by the key, query and value matrices to produce the following matrices:
$$
Q = X W_Q \; \quad K = X W_K \; \quad V = X W_V
$$
where $Q \in \mathbb{R}^{N \times d_k}$, $K \in \mathbb{R}^{N \times d_k}$ and $V \in \mathbb{R}^{N \times d_v}$. The entire process described above can be summarized as follows
$$
X' = \text{SelfAttention}(Q, \, K, \, V) = \text{softmax} \Biggl( \frac{QK^T}{\sqrt{d_k}} \Biggr) V
$$
where $X'$ is the contextualized representation for the sequence produced in the first Self-Attention operation. $X'$ will undergo further refinement and contextualization in successive layers of the Transformer model.\\

The great advantage of the attention mechanism is each token's the attention weight can be computed independently of the others, given the weight matrices. This means that the self-attention computation can be performed in parallel, reducing by a large factor the training and the inference time with respect to the RNNs alternatives. In practice, this means that the loss for a full sequence can be computed in a single forward pass, rather than one pass per word in the sequence like in RNNs. 

\subsection{Multi-Head Self-Attention} \label{mhsa}

It may be difficult for a single triplet of weight matrices $(W_Q, \, W_K, \, W_V)$ "to capture all of the different kinds of parallel relations among its inputs"\cite{Jurafsky2009}.\\ 

The solution proposed to this limitation is to create a set of triplets $(W_Q^h \, W_K^h, \, W_V^h)$, $h \in \{1, ..., H \}$, each of them referred to as $head^h$.

The self-attention computation is performed independently in parallel for every attention \textit{head} (placed at the same depth in the model), and results are finally aggregated in a unified output for further processing.
The reason to have so many parameters is that each head will project the input in a different representation highlighting some characteristics allowing to perform a more varied analysis.\\

This choice implies that each head will output a $N \times d_v$ matrix\footnote{This is due to the fact that, given $Q^h \in \mathbb{R}^{N \times d_k}$, $K^h \in \mathbb{R}^{N \times d_k}$ and $V^ \in \mathbb{R}^{N \times d_v}$, we perform the following operations: 
$$
\text{softmax}\Biggl(\frac{Q^h {K^h}^T}{\sqrt{d_k}} \Biggr) \in \mathbb{R}^{d_k \times d_k} \, ; \quad \text{softmax} \Biggl(\frac{Q^l {K^l}^T}{\sqrt{d_k}} \Biggr) V \in \mathbb{R}^{d_k \times d_k} \times \mathbb{R}^{N \times d_v} \in \mathbb{R}^{N \times d_v}
$$
}, thus we will have $H$ matrices of that shape. In order to obtain an output of the same dimension of the input matrix $X \in \mathbb{R}^{N \times d_v}$, we have to project the outputs of all the heads into a single $\mathbb{R}^{N \times d_v}$ matrix. 

This is obtained by first concatenating the outputs of each head (in a $\mathbb{R}^{N \times H \cdot d_v}$ matrix) and then by projecting it using an output matrix $W_O \in \mathbb{R}^{H \cdot d_v \times d}$:
$$
( head_1 \oplus head_2 \oplus ... \oplus head_H) \in \mathbb{R}^{N \times H \cdot d_v} \, ; \quad ( head_1 \oplus head_2 \oplus ... \oplus head_H) W_O \in \mathbb{R}^{N \times d_v}
$$

\begin{figure}[H]
    \centering
    \includegraphics[width=0.7\linewidth]{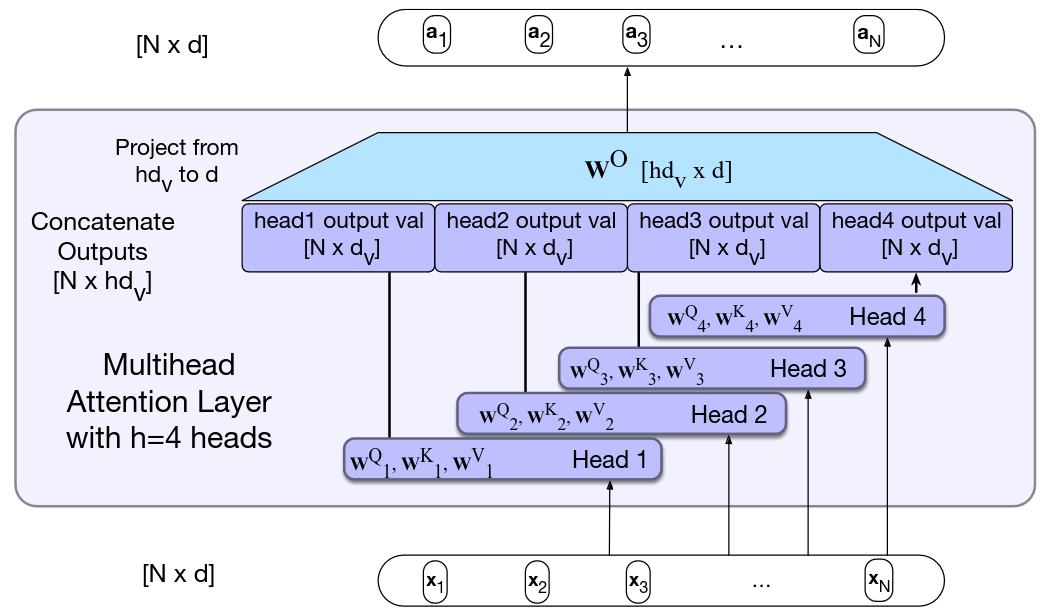}
    \caption{A schematic representation of Multi-Head Attention (MHA); each \textit{head} is provided with its own set of key, query and value matrices. The outputs of all the heads are first concatenated and consequently projected to $W_O$ (taken from \cite{Jurafsky2009}).}
\end{figure}

In matrix terms, the Multi-Head Self-Attention mechanism can be expressed as:
$$
Q^h = X W_Q^h \; \quad  K^h = X W_K^h \; \quad  V^h = X W_V^h 
$$
$$
head^h = \text{SelfAttention}(Q^h, K^h, V^h)
$$
$$
A = \text{MultiHeadAttention}(X) = ( head_1 \oplus head_2 \oplus ... \oplus head_H) W_O
$$

\subsection{The Transformer block}

In this section we are going to describe the structure of the Transformer unit, exploiting the Multi-Head Self-Attention operation introduced in the previous paragraphs (\ref{mhsa}). 

The Transformer block is schematically represented here:
\begin{figure}[H]
    \centering
    \includegraphics[width=0.75\linewidth]{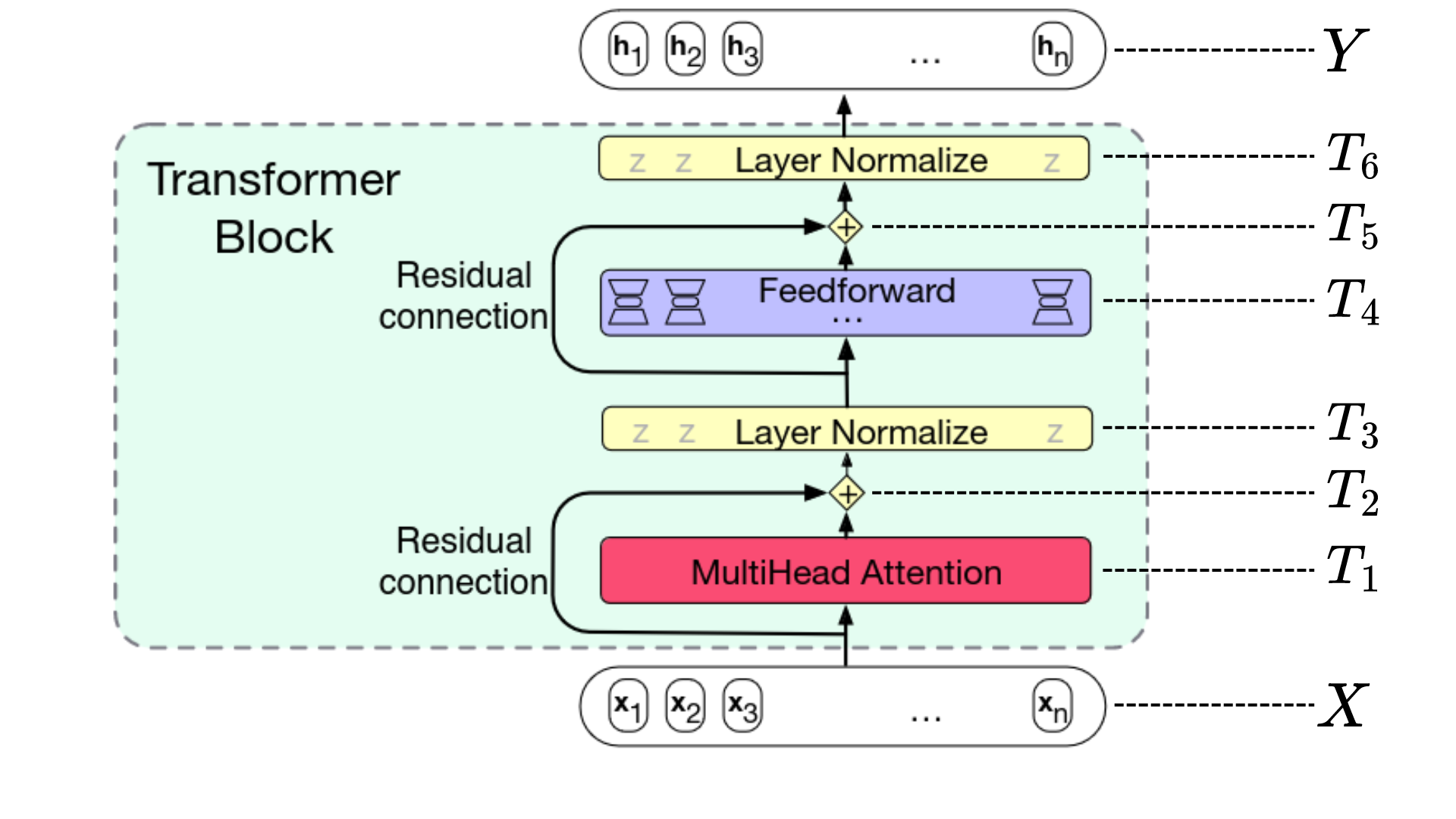}
    \caption{taken from \cite{Jurafsky2009}}
\end{figure}

\label{transformer-block}

Breaking down the process with one equation for each computation component, we obtain: 
$$
T_1 = \text{MultiHeadAttention}(X) \, ; \quad T_2 =  X + T_1 \, ; \quad T_3 = \text{LayerNorm}(T_2)
$$
$$
T_4 = \text{FFN}(T_3) \, ; \quad T_5 = T_3 + T_4 \, ; \quad Y = \text{LayerNorm}(T_5)
$$

Going in detail, the residual connections present before the Normalization layers have some benefits. By allowing information from the activation to go forward and, complementary, the gradient to go backwards skipping a layer, they improve learning and give higher-level layers direct access to information residing in lower-level ones \cite{he2016}. \\

The Normalization layer is useful to improve the training performances by keeping the hidden layer values' in a range that facilitates gradient-based training \cite{Jurafsky2009}, \cite{xiong2020}. \label{why normalize}

The vector components $x$ are \textit{row-normalized} as: 
$$
\hat{x} = \frac{x-\mu}{\sigma}
$$
where $\mu$ and $\sigma$ are respectively:
$$
\mu = \frac{1}{d} \sum_{i=1}^d x_i \, ; \quad \sigma = \sqrt{\frac{1}{d} \sum_{i=1}^d (x_i - \mu)^2}
$$
In addition to this, the standard implementation of layer normalization includes also two learnable parameters, $\gamma$ (the gain value) and $\beta$ (the offset value):
$$
\text{LayerNorm}(x) = \gamma \cdot \hat{x} + \beta = \gamma \cdot \frac{x-\mu}{\sigma} + \beta
$$

Finally, the Feed-Forward layer contains $N$ fully connected two-layer networks\footnote{We recall that $N$ is the context length; thus, the Feed-Forward layer processes in parallel each item that comes out from the $\text{MultiHeadAttention}$ and $\text{LayerNorm}$ operations.}  (i.e. composed pf one hidden layer and two weight matrices) that usually has $d = 512$ input, $d_h = 2048$ hidden and again $d = 512$ output neurons. The activation functions applied between these two linear layers are both Rectified Linear Unit ($\text{ReLU}$) functions, defined as $\text{ReLU}(x) = max(0, \, x)$.  \\

The operations within the $\text{FFN}$ can be mathematically described by the following equation:
$$
\text{FFN} = max( 0, \, max(0, \, x \cdot W_1 + b_1) \cdot W_2 + b_2)
$$
where $W_1$ and $W_2$ are the weight matrices and $b_1$ and $b_2$ are the biases for the first and the second layer respectively. \\

Each of the $N$ units in $\text{FFN}$ is also called position-wise network, since the weights of the two matrices involved in the transformation are the same for all the processed tokens, i.e. independent of the token position. 

\subsection{The Language Model Head}

The Large Language Model is composed of a certain number $L$ of Transformer blocks stacked one after the other. 
The output of the last Transformer block is a $N \times d$ matrix (one $1 \times d$ state per input token). For inference, only the last state (of dimension $1 \times d$) is used to predict the next token, while the previous $(N-1) \times d$ ones are ignored\footnote{On the opposite, for the training, each state is \textit{projected} (as described in this section) to the vocabulary and a loss is computed in parallel as the average over the full predicted sequence against the gold sequence.}.\\

In order to come out with a probability distribution over the vocabulary tokens, it is necessary to \textit{unembed} the output of the  with a $d \times |V|$ matrix. This linear layer can be learned, but more commonly we tie this matrix to (the transpose of) the embedding matrix $E$. \\

The output of this process is called \textit{logit} or \textit{score vector} $u$, since it has a score for each of the $|V|$ possible words in the vocabulary $V$. These scores are transformed into probabilities through the usage of a $\text{softmax}$ function. These two operations composed the so-called \textit{Language Model Head} of the model:
$$
u = h^L_N \cdot E^T \quad \text{and} \quad y = \text{softmax}(u)
$$

where $h^L_N \in \mathbb{R}^{1 \times d}$, $E^T \in \mathbb{R}^{d \times |V|}$ and $u \in \mathbb{R}^{1 \times |V|}$ (consequently also $y \in \mathbb{R}^{1 \times |V|}$).

\begin{figure}[H]
    \centering
    \includegraphics[width=0.85\linewidth]{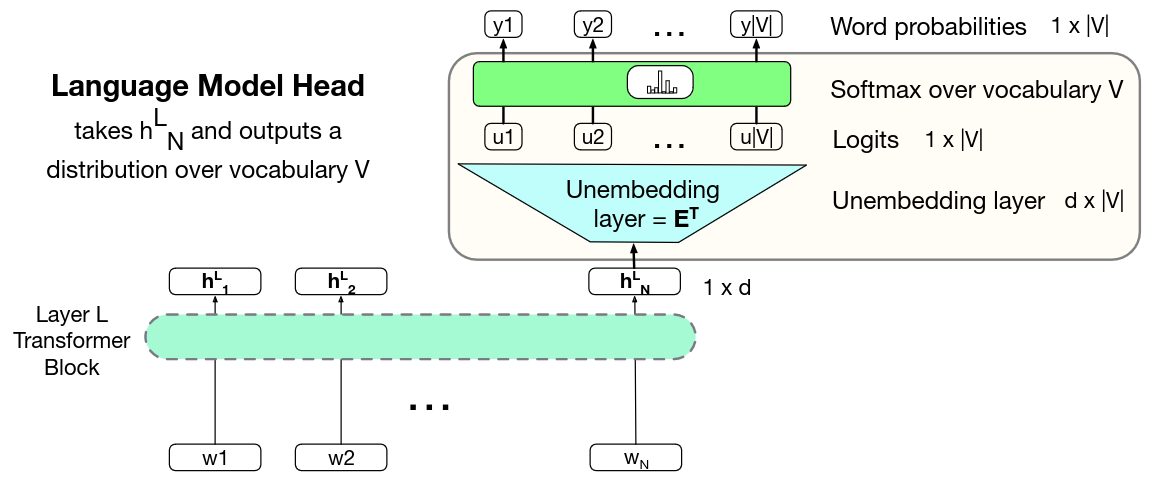}
    \caption{The Language Model Head: the circuit at the top of the last Transformer block maps the output embedding for token $N$ from the last layer $h^L_N$ to a probability distribution over words of the vocabulary $V$ (taken from \cite{Jurafsky2009}).}
\end{figure}

The multi-layer architecture that we showed in the previous sections is referred to as \textit{decoder-only} LLMs and it was first introduced by \lstinline{GPT-1} \cite{gpt-1} and subsequently adopted in other model architectures. 

Before \lstinline{GPT-1}, the original Transformer architecture (as proposed by Vaswani et al. \cite{vaswani2023}) was employed, consisting in both an \textit{encoder} and a \textit{decoder} part.

\subsection{Embedding words into numerical vectors} \label{embed}

The first, fundamental step that preceds all the others is the transformation of the words inside the sequence into numerical vectors on which we can perform the already explained \textit{sequence of relevant mathematical operations}, leading to information extraction and allowing us to make inference. \\

Given a sequence of $N$ words, we define the \textit{embedding} of a sequence as the matrix $X \in \mathbb{R}^{N \times d}$, where $d$ is the pre-defined dimension of each embedding vector. \\

We can imagine to define a vocabulary $V$ of words present in the sequence of interest. Given this dictionary, we can map each word into a \textit{one-hot encoded vector}, in which the only element which is not zeroed out is the one corresponding to the token.  

Think for example of the sequence \textit{May the Force be with you}, then the vocabulary will be made by: $V = \{$ \textit{be}, \textit{Force}, \textit{May}, \textit{the}, \textit{you}, \textit{with} $\}$. We can represent the word \textit{May} as $[1, 0, 0, 0, 0, 0]$, the word \textit{the} as $[0, 1, 0, 0, 0, 0]$ and so on. Note that all the vectors are encoded into vectors having $|V|$ elements, and that the entire sequence can be represented by a matrix  $W \in \mathbb{R}^{N \times |V|}$.\\

To \textit{smooth} by a certain degree the word representation, we apply to the one-hot encoded matrix a transformation $E \in \mathbb{R}^{|V| \times d}$ called \textit{embedding matrix}: 

\begin{figure}[H]
    \includegraphics[width=0.85\linewidth]{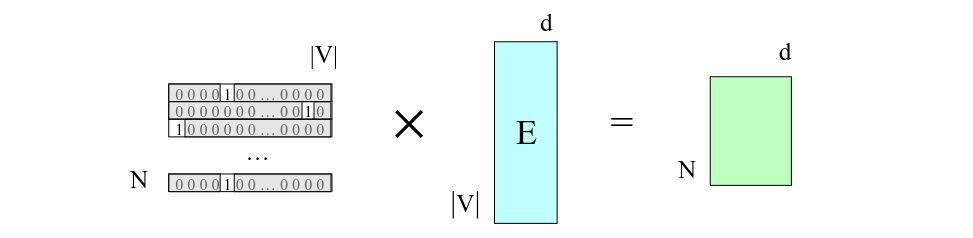}
    \centering
    \caption{Creating the word embedded matrix $W \in \mathbb{R}^{N \times d}$ for the input sequence of tokens by multiplying a one-hot matrix of dimensions $N \times |V|$ by the embedding matrix $E \in \mathbb{R}^{|V| \times d}$ (taken from \cite{Jurafsky2009}).}
\end{figure}

The initial $N \times |V|$ words representation is referred to as \textit{sparse}, in opposition to the \textit{dense} one $W \in \mathbb{R}^{N \times d}$ of much lower dimensions (typically in the range of $d = 50$ to $300$ dimensions) produced by the application of an embedding matrix $E$.\\

The benefit of \textit{dense} embeddings over \textit{sparse} one is that, due to the training objective of the embedding matrix $E$, these kind of transformations push similar words to be close in the \textit{dense} vector space defined by $W$; this idea is inspired from the \textit{distributional hypothesis} - \textit{you shall know a word by the company it keeps} \cite{firth1957}. 
Words embeddings are used in all \textit{Neural-Networks based} Language Models, including in RNNs. \\

However, Transformer-based models for text generation, differently from RNNs, do not have a sequential inductive bias by design. 

In this kind of models, word embeddings are position-independent. 
In order to keep track of this information, we have to add to the "standard" \textit{word embedding}  $w_i \in \mathbb{R}^{1 \times d}$ ($w_i$ is the $i$-th row of \textit{words embedded} matrix $W \in \mathbb{R}^{N \times d}$) its corresponding \textit{position embedding}, defined as a function mapping each position $i$ into an embedding $P[i] \in \mathbb{R}^{1 \times d}$. 

In this setting, the final representation of the input, the matrix $X \in \mathbb{R}^{N \times d}$ is such that each row $i$ is the representation of the $i$-th token in the input, computed by adding the embedding $w_i$ of the token that occurred at position $i$, to $P[i]$, the positional embedding of position $i$.

\begin{figure}[H]
    \includegraphics[width=0.7\linewidth]{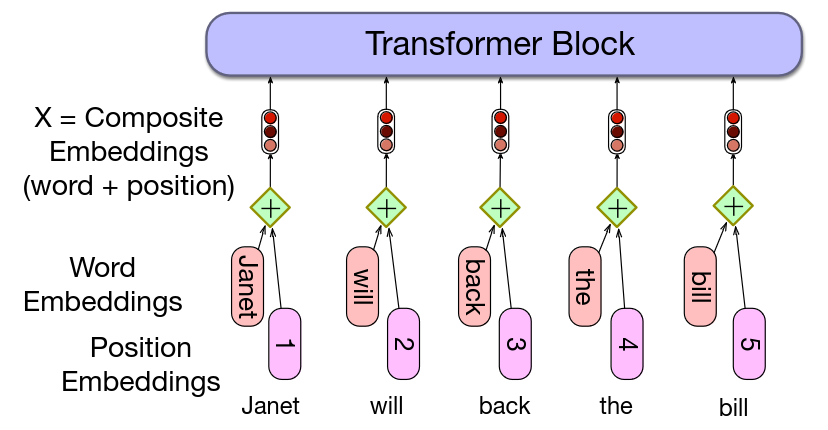}
    \centering
    \caption{A simple way to model position: add an embedding of the absolute position to the token embedding to produce a new embedding of the same dimensionality (taken from \cite{Jurafsky2009}).}
\end{figure}

The positional embedding method proposed by Vaswani et al. \cite{vaswani2023} aims at ensuring that each sentence position $i \in 1, ..., I$ has an unique representation. Their idea is to exploit $sin$ and $cos$ function to achieve this goal without the need to train a separate positional embedding matrix $P$. do is apply to each position $i$ to a set of $d$ functions in order to obtain a $1 \times d$ vector for each sequence position. 

\begin{figure}[H]
    \centering
    \includegraphics[width=0.7\linewidth]{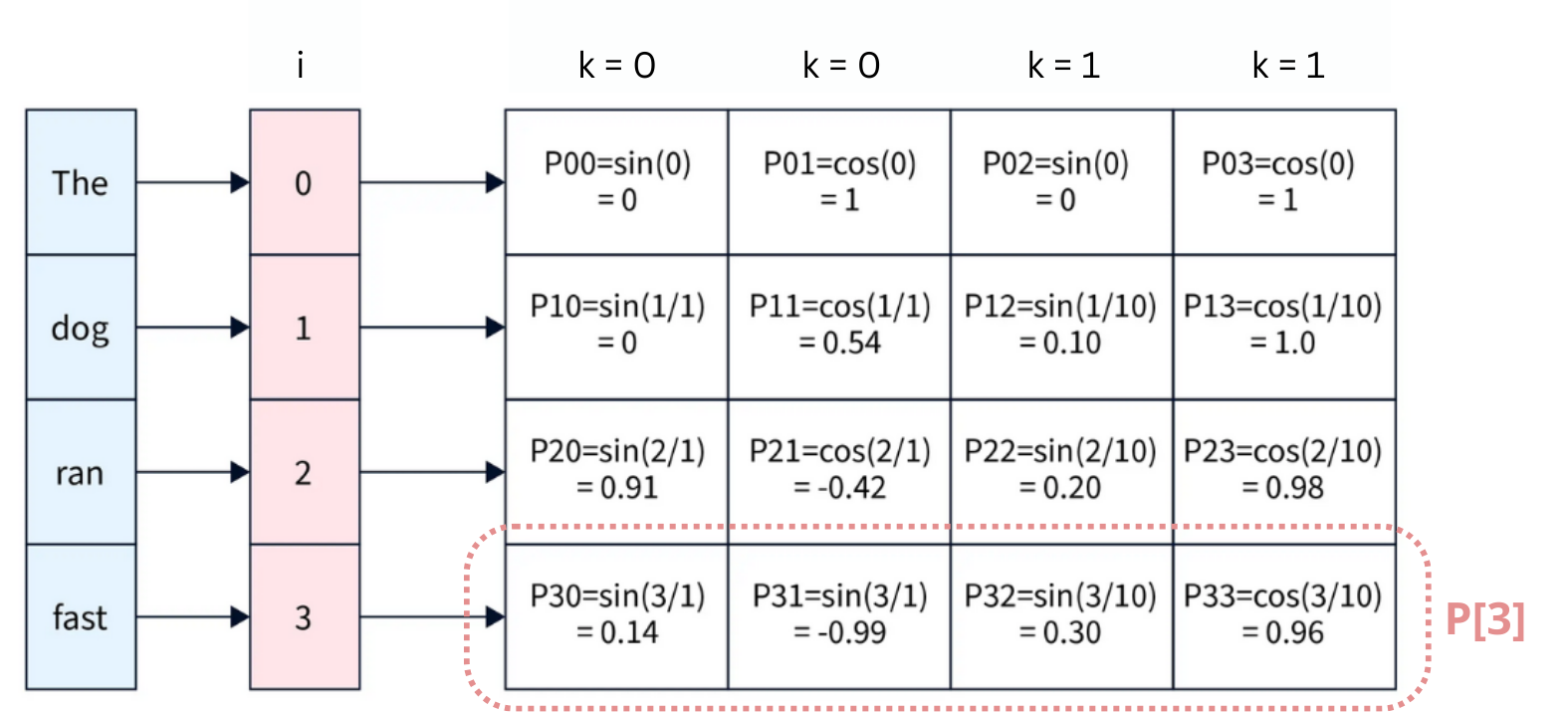}
    \caption{Toy example of Positional Encoding (with $d = 4$), inspired from \cite{PosEnc}. Note that what matters is not the term (e.g. \lstinline{fast}) the position $i$ (e.g. $i = 3$) that is appropriately transformed through $sin$ and $cos$ functions.}
\end{figure}

\subsection{Causal self-attention in \textit{decoder-only} Transformers} \label{decoder-only}

The \textit{self-attention mechanism} presented in section \ref{sa} implies that the \textit{attention weight} associated with a token will not only depend on the past ones (i.e. the preceding context), but also on the upcoming ones. The issue that this \textit{bidirectional attention} carries is that, in some sense, it \textit{incorporates} the future information in the representation of the current word.\\

This approach is unsuited for tasks like text generation, since we expect the model to correctly predict the \textit{next} word, and this task is trivial if the model already knows the following ones. Practically speaking, by using \textit{bidirectional self-attention} Transformer in a "guess which is the next word" problem we are allowing the model to cheat.\\

Due to the \textit{inductive bias} of auto-regressive language modeling, a \textbf{causal mask} is used to hide future tokens from the context mixing operation of the attention block. On the contrary, masked language models used for \textit{classification tasks} like BERT commonly do \textbf{not} employ \textit{attention masks}, leveraging bidirectional context.

The process involves the \textit{attention weights} of the following words and results in a \textit{masked} self-attention matrix in which the elements in the upper-triangular portion of the matrix are zeroed out (set to $- \infty$), thus eliminating any knowledge of words that follow in the sequence. 

\begin{figure}[ht]
\includegraphics[width=0.35\textwidth]{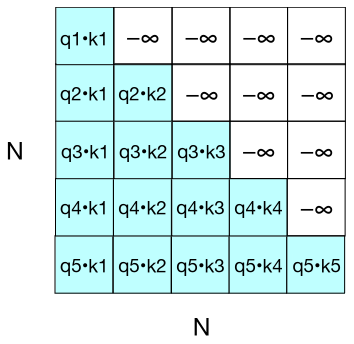}
\centering
\caption{Representation of $Q \cdot K^T \in \mathbb{R^{N \times N}}$ matrix when a causal mask is applied. The upper-triangle portion of it is set to $-\infty$, which the $\text{softmax}$ will turn to zero (taken from \cite{Jurafsky2009}).}
\end{figure}

Mathematically speaking, it is a restriction of what we have already described in (\ref{sa}). The difference stands in the range of allowed indexes for $i$: 
$$
\alpha_{ij} = \begin{cases}
\text{softmax}(\text{score}(x_i, x_j)), \; \forall j \leq i \\

- \infty ,  \, \text{otherwise}
\end{cases} \quad \text{where} \; \text{score}(x_i, x_j)  = \frac{q_i \cdot k_j}{\sqrt{d_k}}
$$
 
The model resulting from this approach is purely \textit{autoregressive}: the model will look at the past and infer the following token from it; then the predicted token will be added to the context, producing a new one to predict the next following word, and so on. 

An important remark is that the concept of \textit{context} can be used in two ways in self-attention. In causal self-attention, the context is any of the prior words. In  \textit{bidirectional self-attention}, the context can include future words. 

\newpage 

\section{Improvements in modern Large Language Models} \label{improvements}

In the following sections we are going to provide a detailed analysis on the architecture of these models and some pre-training and post-training insights. Before diving into details, it is necessary to introduce some modern modifications of the original Transformer architecture.

\subsection{The RoPE embedding} \label{rope}
In section (\ref{embed}) we introduced the Positional Encoding method, allowing Large Language Models to take into account both the semantic meaning of a token and its position in a certain sentence. The issue with these kinds of approaches, called \textit{absolute positional embeddings}, is that each positional embedding is independent of others. 

This means that, in the model’s view, the difference between positions $1$ and $2$ is the same as between positions $2$ and $500$. But intuitively, positions $1$ and $2$ should be more closely related than position $500$, which is significantly farther away. This lack of relative positioning can hinder the model’s ability to understand the nuances of language structure.\\

On the opposite, rather than focusing on a token’s absolute position in a sentence, \textit{relative positional embeddings} \cite{raffel2023} concentrate on the distances between pairs of tokens. This family of methods does not add a position vector to the word vector directly. Instead, it \textbf{alters the attention mechanism} to incorporate relative positional information.
For example, a \textit{bias} might represent the relative distance between any two tokens \textbf{that are one position apart, regardless of their absolute positions in the sentence}. The matrix composed by all relative position biases is added to the product of the query and key matrices in the self-attention layer, so that it is \textbf{ensured that all tokens at the same relative distance are always represented by the same bias}, regardless of their position in the sequence.

Although this method scales to long text sequences, this causes a slowdown in the computational time due to the addition of some operations in the self-attention layer. \\

Summing up, \textit{absolute positional embeddings} assign a unique vector to each position, which though straightforward, doesn’t scale well and fails to capture relative positions effectively. \textit{Relative embeddings}, on the other hand, focus on the distance between tokens, enhancing the model’s understanding of token relationships but complicating the model architecture.\\

A third solution is given by the \textit{Rotary Positional Embeddings} (RoPE) \cite{rope}, that ingeniously combines the strengths of both. It encodes positional information in a way that allows the model to understand both the absolute position of tokens and their relative distances. This is achieved through a rotational mechanism, where \textit{each position in the sequence is represented by a rotation in the embedding space}.\\

RoPE introduces a novel concept. Instead of adding a positional vector, it applies a rotation to the word vector. Imagine a two-dimensional word vector for “dog.” To encode its position in a sentence, RoPE rotates this vector. The angle of rotation ($\theta$) is proportional to the word’s position in the sentence. For instance, the vector is rotated by $\theta$ for the first position, $2 \theta$ for the second, and so on.\\

The technical implementation of RoPE involves rotation matrices:
$$
f_{Q} (X, m) = 
\begin{pmatrix}
cos(m \theta) & - sin(m \theta)\\
sin(m \theta) & cos(m \theta) 
\end{pmatrix} \cdot X \cdot W_{Q} = \begin{pmatrix}
cos(m \theta) & - sin(m \theta)\\
sin(m \theta) & cos(m \theta) 
\end{pmatrix} \cdot Q
$$

$$
f_{K} (X, m) = 
\begin{pmatrix}
cos(m \theta) & - sin(m \theta)\\
sin(m \theta) & cos(m \theta) 
\end{pmatrix} \cdot X \cdot W_{K} = \begin{pmatrix}
cos(m \theta) & - sin(m \theta)\\
sin(m \theta) & cos(m \theta) 
\end{pmatrix} \cdot K
$$

In a 2D case, the equation from the paper incorporates a rotation matrix that rotates a vector by an angle of $m \theta$, where $m$ is the absolute position in the sentence. This rotation is applied to the query and key vectors in the self-attention mechanism of the Transformer\footnote{For higher dimensions, the vector is split into 2D chunks, and each pair is rotated independently}.

The advantage of this approach is that if two words maintain the same relative positions in two different contexts, their embeddings form the same angle, thus the \textit{same dot product}. 

\begin{figure}[H]
    \includegraphics[width=0.55\linewidth]{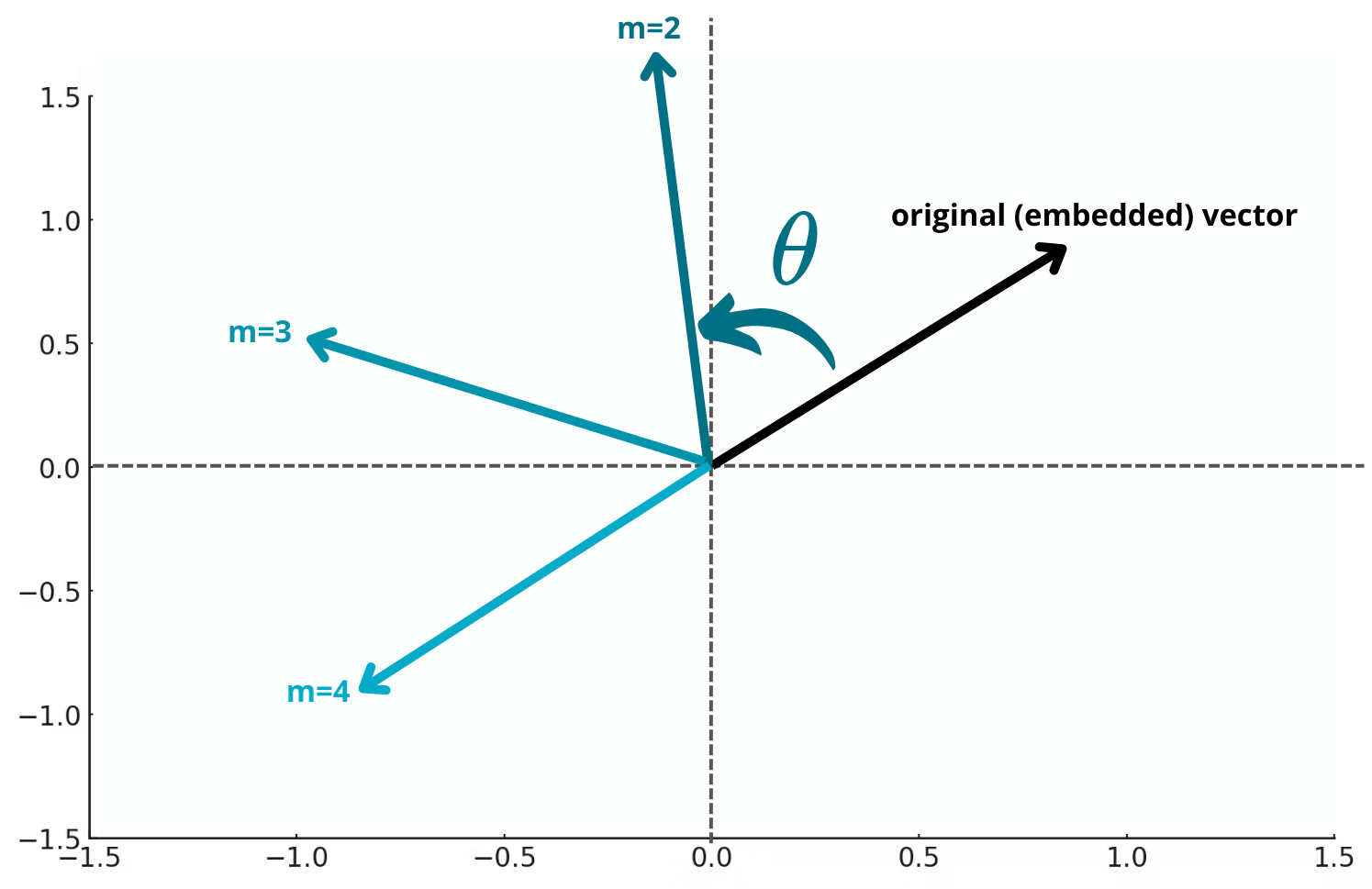}
    \centering
    \caption{An example of RoPE in a trivial 2D case; the original embedded vector is rotated with angle $\theta$ with respect to its position $m$ in the sentence.}
\end{figure}

However, RoPE has limitations when applied to very long sequences as it assumes a \textbf{fixed positional range} defined by the base value $\theta$. Extending RoPE to longer sequences could, in theory, involve increasing $\theta$ or training a new model from scratch with a larger positional range, but both approaches are impractical due to high computational costs and data scarcity.

LongRoPE \cite{longrope} addresses these challenges by dynamically optimizing the positional encoding using two key innovations. First, it introduces a loss function that searches for the optimal scaling factor, $\lambda$, instead of assuming a fixed one (i.e. $\theta$). This adaptive approach allows the model to find the most suitable scaling for different positions within a sequence, preserving the positional relationships even as the context window expands. Second, LongRoPE identifies subsets of tokens that should remain unchanged, preventing the loss of critical positional information that could degrade model performance.

\subsection{Pre-normalization and post-normalization architecture} \label{prenorm}

Sometimes the original Transformer model is slightly modified, and this is the case of a \textit{pre-normalization architecture}. By putting $\text{LayerNorm}$ before the $\text{MultiHeadAttention}$ and $\text{FNN}$ layers, we are normalizing the values before entering the crucial computational layers. This detail can cause an improvement in performances, due to the same reasons given in section (\ref{why normalize}).\\

\begin{figure}[H]
    \centering
    \includegraphics[width=0.75\linewidth]{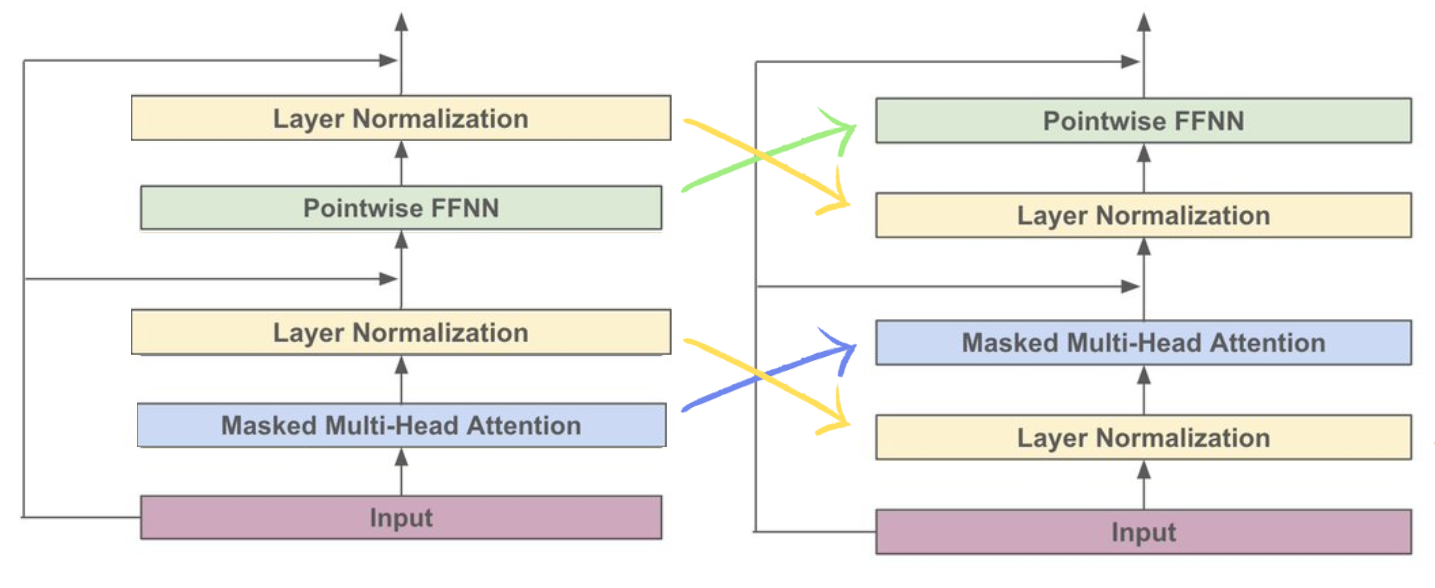}
    \caption{On the left, the original Transformer block architecture; on the right, the pre-normalization architecture.}
\end{figure}

The first idea of a pre-normalization architecture can be found in \lstinline{GPT-3} \cite{gpt-3}, and later used in open-source models such as \lstinline{LlaMa} \cite{touvron2023-llama}. 

\newpage

\subsection{Sliding Window Attention (SWA)} \label{swa}

Classical Self-Attention mechanism (\ref{sa}) works by comparing a focus of attention with the other elements of the sequence. In order to allow each $q_i$ to attend all the $k_j$, we require \textit{quadratic memory for each attention layer}. 

In fact, the Self-Attention output is a quadratic matrix of the same shape of the input sequence (a $N \times N$ matrix) in which each element is evaluated in terms of its relevance with respect to the other elements in the sequence.\\

In decoder-only Transformers (\ref{decoder-only}), a \textit{causal mask} is applied in order to mask future tokens. This helps in the text generation task, though it is not sufficient to optimize this operation.

BERT-based models fix a maximum sequence length and split the document into multiple overlapping segments having at maximum that length. They process independently each segment and their representations are combined. The big issue with this approach is that the attention information \textit{across segments} is lost. \\

LongFormer \cite{longformer} proposes fixing a 'window of interest' of length $N'$, such that each token is only allowed to attend to its peers, i.e., tokens no farther than $N'$ items away. The paper suggests to look at $\frac{1}{2} N'$ tokens at each side, making the computational complexity to be $O( N \times N')$, but it is not the only suggested \textit{attention pattern}: 

\begin{figure}[H]
    \centering
    \includegraphics[width=0.9\linewidth]{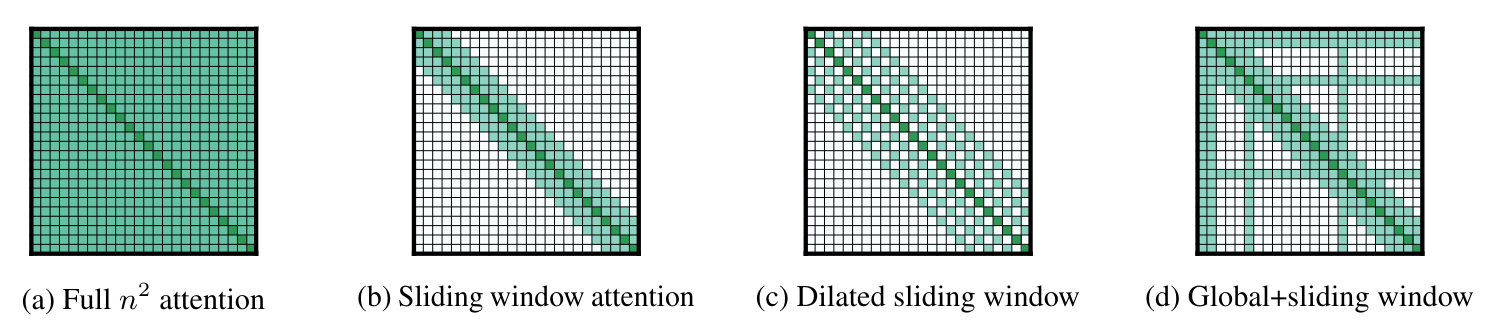}
    \caption{Different attention patterns presented in \cite{longformer}}
\end{figure}

This idea reduces by a large margin the memory requirements and consequently the training and inference times, and still maintains good levels of performance due to a brilliant subtlety. In a model made of $L$ Transformer blocks, the layers attend information by looking at the output of the preceding layer \textbf{and at its peers}. This means that in the last layers we will obtain a \textit{conical structure of hidden relationships}, gaining information from tokens far away from it. 

\begin{figure}[H]
    \centering
    \includegraphics[width=0.6\linewidth]{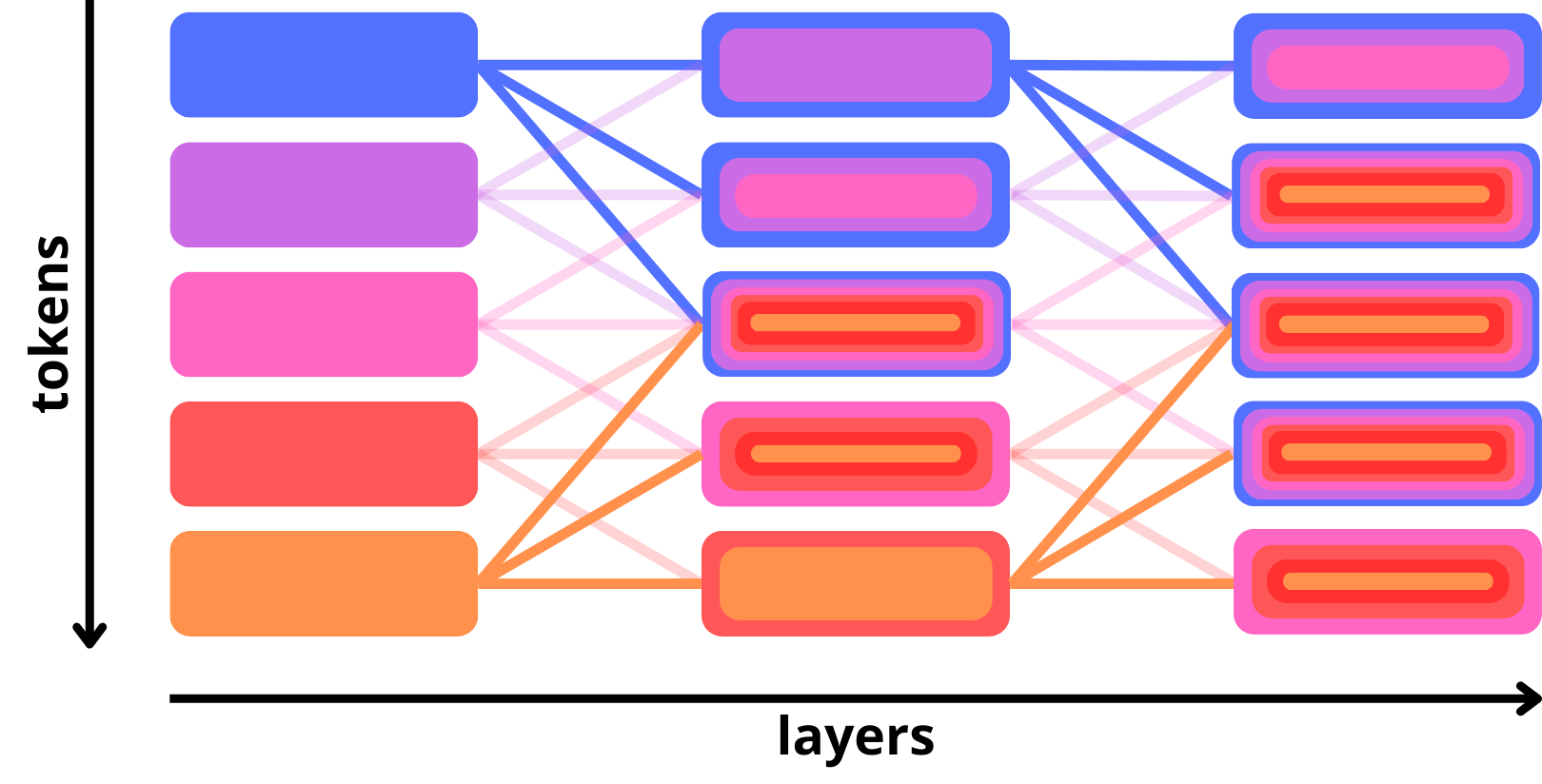}
    \caption{An intuitive explanation of hidden relationships in Sliding Window Attention. Each token is represented as a different color, in order to show how information about it propagates in the layers. In this toy example, $N'$ is 3 and $L$ is 3.}
\end{figure}

The optimized model will be faster and use less memory than the original model, but it may also be less accurate. The challenge of SWA is to reduce the computational complexity as much as possible without significantly reducing the model's accuracy. This method is currently used by the \lstinline{Mistral}'s family of models \cite{jiang2023}.

\subsection{Multi-Head Attention, Multi-Query Attention and Grouped Query Attention} \label{gqa}

In section \ref{mhsa} we introduced the \textbf{Multi-Head Attention} (MHA) mechanism proposed by \cite{vaswani2023}. Each $head^h$ is associated with a $(Q^h, K^h, V^h)$ triplet, and the number of heads $H$ present in each Transformer block determines \textit{the number of possible "points of view"} used to analyze the input sequence. 
But in this setting the number of parameters to train and to use to make inference is really big. This provides high quality at the cost of consuming higher memory bandwith.\\

The \textbf{Multi-Query Attention} (MQA) is a technique to accelerate the inference process, reducing drastically the number of involved matrices. It uses a single key matrix $K$ and a single value matrix $V$ for all the $heads$, while still involving $H$ query matrices $Q^h, h \in {1, ..., H}$. This reduces by a large margin the training time and the memory consumption, leading to worse performances due to training instability. \\

A compromise is given by the \textbf{Grouped Query Attention} (GQA), which seeks to strike a balance between MHA and MQA. GQA partitions query heads into $G$ groups, with each group sharing a single key and value matrices. This provides a trade-off between the speed of MQA and the quality of MHA.
\begin{figure}[H]
    \centering
    \includegraphics[width=0.95\linewidth]{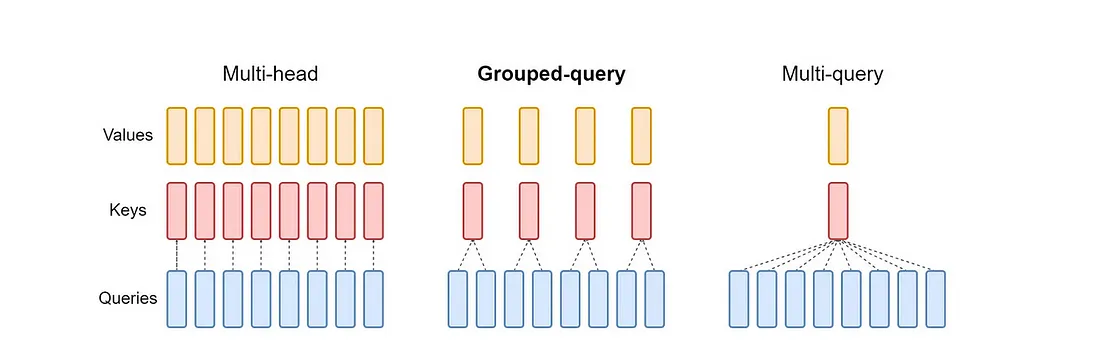}
    \caption{MHA, GQA and MQA (taken from \cite{sliding&gqa})}
\end{figure}

Most modern LLMs utilize GQA, as it achieves performance comparable to MHA while significantly reducing computational time.

\subsection{Decoding strategies}

The core of the generation process for large language models is the task of choosing the single word to generate next based on the context and based on the probabilities that the model assigns to possible words. This task of choosing a word to generate
based on the model’s probabilities is called \textit{decoding}.

Repeatedly choosing the next word conditioned on our previous choices is called \textit{autoregressive generation} or \textit{causal decoding}. \\

There are two main decoding strategies: \textit{greedy decoding} and \textit{sampling}. \\

\textit{Greedy decoding} consists in selecting the token with the highest probability at each decoding step; this produces outputs that most closely match the \textit{common} option (i.e. in model's prompt or pre-training data). This decoding strategy is used for fact-based use cases and tends to produce less \textit{creative} outputs. \\

\textit{Sampling} instead chooses tokens according to their probability assigned by the model. Thus it is \textit{more likely} to generate words that have a high probability in the context and viceversa. This process continues until a pre-determined length is reached or when the end-of-sentence token is generated. 

We can formalize this algorithm for generating a sequence of words $S = s_1, s_2, . . . , s_N$ until we hit the end-of-sequence token \lstinline{EOS}. We use $x \sim p(x)$ to refer to the process of choosing $x$ by sampling from the distribution $p(x)$:
\begin{enumerate}
    \item initialize $i =1$;
    \item sample $w_i \sim p(x)$;
    \item while $w_i \neq$ \lstinline{EOS}, $i = i+1$ and $w_i \sim p(w_i | w_{<i})$.
\end{enumerate}

\textit{Sampling} adds variability and randomness to the decoding process, which can be desirable in creative use cases. However, with greater variability comes a greater risk of incorrect or nonsensical output (a problem which is not present in \textit{greedy decoding}).

This strategy uses three different parameters to adjust how the model chooses tokens to sample:

\begin{itemize}
    \item \textbf{temperature} flattens (when set to values near 0.0) or sharpens (when near 2.0) the probability distribution over the tokens to be sampled, by default is set to 0.7;
    
    \item \textbf{top-k} samples tokens with the highest probabilities until the specified number of tokens is reached (can be set up to 100), by default is set to 50;
    
    \item \textbf{top-p} samples tokens with the highest probability scores until the sum of the scores reaches the specified threshold value (a floating point value between 0.0 and 1.0), by default it is not used. 
\end{itemize}

Greater temperature values and greater values of k lead to increased variability and creativity in model's answers. 
Even though this approach is mostly going to generate sensible, high-probable words, there are many odd, low-probability words in the tail of the distribution that get chosen often enough to result in generating weird sentences since they constitute a large enough portion of the distribution \cite{Jurafsky2009}.\\

Different sampling strategies have been proposed in order to ensure both \textit{quality} and \textit{diversity}, typically found in a trade-off: methods that give a bit more weight to the middle-probability words tend to be more creative and more diverse, but less factual and more likely to be incoherent or otherwise low-quality. In our analyses, we used only \textit{greedy decoding} to ensure a more reliable choice of tokens when generating the outputs.

\section{Large Language Models employed}
\label{tested-models}

First of all, our focus lied on the choice of the Large Language Models involved in this analysis. We made the choice of considering only open-source models, freely accessible from the HuggingFace collaboration platform\footnote{Hugging Face, Inc. is a French-American company that develops computation tools for building applications using machine learning. It is most notable for its \lstinline{transformers} \cite{transformers} and \lstinline{datasets} \cite{datasets} libraries and for its platform that allows users to share machine learning models and datasets and showcase their work.}. \\

The choice fell back on three distinct families of models: the \lstinline{meta-llama/LlaMa-3.1} \cite{llama3.1}, the \lstinline{microsoft/Phi-3} \cite{phi-3} and the \lstinline{google/Gemma-2} \cite{gemma-2} families of models. 
In all three are \textit{decoder-only models} (\ref{decoder-only}). \\

All these three models incur in some forms of pre- and post-training. While the pre-training stage consists in feeding the model a large quantity of data in order to teach the model how to produce sentences that make some sense, the post-processing phases tend to align the model to some preferences.\\ 

For example, their goal could be to teach the model to follow instructions given in a certain format or for example to properly use a set of terms linked to a certain semantic field. These two tasks are often performed via Supervised Fine-Tuning (SFT) \cite{wei2022b}, \cite{howard2018}, i.e. the process of modifying model's weights (all or only a part of them) to change the downstream behaviour, obtaining the desired specific output. \\

Another approach to align the model's behaviour towards a desired one is given by Reinforcement Learning with Human Feedback (RLHF) \cite{ouyang2022} that refines the model by incorporating human evaluations into the training process. This method involves training the model to maximize a reward function that reflects human preferences or values, often by using comparisons between model outputs ranked by human annotators. The RLHF approach has been particularly effective in adjusting a model's responses to be more aligned with human expectations in terms of tone, content relevance, safety, factual accuracy and other desired properties. \\

In addition to SFT and RLHF, Direct Preference Optimization (DPO) \cite{rafailov2024} has emerged as another effective technique for fine-tuning language models. Unlike RLHF, which relies on constructing a reward model to optimize the output, DPO directly leverages human preferences by optimizing the model based on explicit comparisons of preferred outputs. In this method, human annotators are asked to compare multiple model outputs and rank them according to their quality or alignment with specific guidelines. The model is then fine-tuned to favor the outputs that are ranked higher, effectively guiding its behavior towards generating more desirable responses without needing to construct a complex reward model.

DPO simplifies the optimization process by directly minimizing the difference between the model's output distribution and the desired distribution indicated by human preferences. This approach can result in faster convergence and reduced computational overhead, making it a promising alternative to RLHF for situations where direct comparisons between outputs are sufficient for alignment purposes.

By combining SFT, RLHF, and DPO, it is possible to effectively adjust language models to achieve the desired behavior, ensuring that outputs adhere to expectations.

\subsection{\lstinline{Gemma-2} family} \label{gemma-2}

The last release of the \lstinline{Gemma} family of models is dated July 2024, with \lstinline{Gemma-2} \cite{gemma2-release}. The notation that will be used to flag different components in the models will reflect the one present in the Transformer section (\ref{transformers}). \\

The pre-training data primarily consist of English-language web documents, code, and scientific articles. The size of the pre-training corpus varies with respect to the model scale: \lstinline{Gemma-2B}\footnote{From now on, we will refer to \lstinline{Gemma-2-*B} as \lstinline{Gemma-*B} for brevity.} required 2T tokens, \lstinline{Gemma-9B} 8T, \lstinline{Gemma-27B} (not considered in our analysis) 13T. The data is carefully filtered to reduce the risk of unwanted and unsafe utterances, discarding personal information and unsafe data. \\

Data is tokenized using Google's \lstinline{SentencePiece} tokenizer \cite{sentencepiece}, implemented in \lstinline{C++}; it splits digits, does not remove whitespaces and relies on \textit{Byte-Pair Encoding} (BPE) to deal with unknown tokens. Its vocabulary is made by $256.128$ different tokens. \\ 

\lstinline{Gemma} models use RoPE embeddings (\ref{rope}) in place of some absolute Positional Embedding method, and share this embedding matrix with the output layer. \\

The Transformer block uses the Grouped Query Attention (\ref{gqa}) to achieve similar downstream performances but reducing the inference time. In \lstinline{Gemma-2B}, $H = 8$ while $G = 4$; in \lstinline{Gemma-9B}, $H = 16$ while $G = 8$. The first model has $L = 26$ layers for each Transformer block, the second one $L = 42$. In both cases, the head size is $d_v = 256$, while the model size $d$ is different: $d = 2304$ for \lstinline{Gemma-2B}, $d = 3584$ for \lstinline{Gemma-9B}. 

To speed up computations without harming the performances, they alternate layers employing a local sliding window attention of size $N' = 4096$ (\ref{swa}) and a global attention of size $N = 8192$ tokens. \\

Effort is made to facilitate the gradient propagation by doing mainly two changes. 

The first one is to perform \textbf{both} a pre-normalization and a post-normalization (\ref{prenorm}), i.e. the number of $\text{LayerNorm}$ is duplicated. 

The second one is to perform a \textit{logit soft-capping} in each attention layer and in the final layer, in order to keep the values of the logits in the range $[ - softcap; + softcap]$:
$$
logits = softcap \cdot tanh \Bigl( \frac{logits}{softcap} \Bigr)
$$
In the original paper, $softcap = 30$ for the final layer and $softcap = 50$ for the attention layers. \\

The $\text{ReLU}$ activations in the $\text{FFN}$ layer are replaced by $\text{GELU}$ \cite{gelu} ones.
$\text{GELU}$ stands for Gaussian Error Linear Unit and corresponds to:
$$
\text{GELU}(x) = x \cdot P(X \leq x) = x \cdot \Phi(x), \quad X \sim N(0,1)
$$ 

The model is both pre-trained and post-trained. 

The post-training procedure involves the "standard" Supervised Fine-Tuning (SFT), the Reinforcement Learning with Human Feedback (RLHF) and model merging, i.e. averaging different models obtained by running the training with different hyperparamers \cite{ramé2024}. 

In detail, SFT works as follows (taken from \cite{gemma-2}):
\textit{given a set of held-out prompts, we generate responses from a test model, generate responses on the same prompts from a baseline model, shuffle these randomly, and ask a larger, high capability model to express a preference between two responses. Different prompt sets are constructed to highlight specific capabilities, such as instruction following, factuality, creativity, and safety.}\\

In our experiments we will use the instruction-tuned \lstinline{Gemma}\footnote{By \lstinline{Gemma}, we refer to the pre-trained \lstinline{Gemma-2} model. By \lstinline{Gemma-it} we refer instead to the output of a SFT process performed on \lstinline{Gemma-2} by the Gemma research team, available on HuggingFace \cite{gemma2-release} and tuned for instruction prompts.}, called \lstinline{Gemma-it}, whose prompt format is: 
\begin{lstlisting}
<start_of_turn>user
user_message
<end_of_turn>
<start_of_turn>model
model_message
<end_of_turn> 
...
<end_of_turn><eos>
\end{lstlisting}

Both the \lstinline{9B} and \lstinline{2B} models are distilled \cite{hinton2015} from the \lstinline{27B} model by minimizing the negative log-likelihood between the probabilities of the student (i.e. $p_s(x | x_c)$) and the teacher model (i.e. $p_t(x | x_c)$):
$$
\displaystyle\min_{p_s} \sum_x - p_t(x | x_c) \cdot log \; p_s(x | x_c)
$$

\subsection{\lstinline{Phi-3} family} \label{phi-mini}
Microsoft has developed the \lstinline{Phi} models stating that "Textbooks are all you need" \cite{textbooksneed}. They released \lstinline{Phi-1}, a 1.3B parameters model intended for coding purposes, trained using a selection of "textbook quality" data from the web (6B tokens) and synthetically generated textbooks and exercises with GPT-3.5 (1B tokens) and they obtained \lstinline{phi-1-base}. They further fine-tuned it on code exercises, producing \lstinline{phi-1}. \\

Research on this topic has developed since then, with the release of the \lstinline{Phi-3} \cite{phi3-release} family of models, being trained on heavily LLM-filtered, publicly available, web data and on LLM-generated synthetic data.
The fundamental idea is to improve the performance trends that were previously shown to be predictable, once the model size, the data size and the computational budget are given \cite{kaplan2020}. The \lstinline{Phi} team works on allowing the model to interact the data in novel ways, instead of keeping the data source fixed as proposed by Kaplan et al. \cite{kaplan2020}.\\

The pre-training takes place in two stages: a first phase, aimed at teaching the model general knowledge (and consequently trained mostly on web sources) and a second phase, aimed at teaching the model how to logically reason and to attain specific skills (for this, a subset of the data of the previous phase is taken and used to generate new data for those goals). \\

Different models were released: 
\begin{itemize}
    \item \lstinline{Phi-3-mini}: a 3.8B model trained on 3.3T tokens using \lstinline{bfloat16}, with a \lstinline{LlaMa-2}-like architecture \cite{touvron2023-llama2};
    \lstinline{Phi-3-small}: a 7B model with different architecture than \lstinline{Phi-mini} and \lstinline{Phi-medium} ones\footnote{It follows the standard 7B model architecture, uses the \lstinline{tiktoken} tokenizer (thus has a different vocabulary that the other models) and it alternates layers of global self-attention and a particular kind of \textit{blocksparse} attention \cite{phi3}.}.
    \item \lstinline{Phi-3-medium}: a 14B model trained on 4.8T tokens using \lstinline{bfloat16}, with a \lstinline{LlaMa-2}-like architecture \cite{touvron2023-llama2}.
\end{itemize}

For our analysis, we will consider only \lstinline{Phi-3-mini} and \lstinline{Phi-3-medium} due to their similar architecture. They are both built on \lstinline{LlaMa-2} architecture (a \textit{pre-normalization} architecture) and they use the same tokenizer, with vocabulary size of $32.064$. \lstinline{Phi-mini}\footnote{As for \lstinline{Gemma-2} models, from now on we will omit the release number; \lstinline{Phi-3-*} will be simply referred to as \lstinline{Phi-*}.} and \lstinline{Phi-medium} are both released with context length equal to 4K and 128K. For the first one, RoPE embedding is used, while LongRoPE needs to be used in the other case (\ref{rope}). \\

The Transformer block uses Multi-Head Attention in \lstinline{Phi-mini} with $H = 32$ heads and Grouped-Query Attention (\ref{gqa}) in \lstinline{Phi-medium} with $H = 32$ query heads (each of dimension $d_v = 128$), $G = 8$ key and value heads (i.e. 4 queries share 1 key). Each Transformer block contains $L=32$ layers in \lstinline{Phi-mini}, $L=40$ in \lstinline{Phi-medium}. 
The model size is $d = 3072$ for \lstinline{Phi-mini} and $d=5120$ for \lstinline{Phi-medium}. \\

The $\text{FFN}$ hidden dimension is equal to $13.824$. Thus each classification head takes as input a feature size of $d$ and outputs a $32.000$-dimensional vector. 
Differently from \lstinline{Gemma-2} models, \lstinline{Phi-3} ones use $\text{SiLU}$ (Sigmoid Linear Unit) activation \cite{silu}: 
$$
\text{SiLU}(x) = x \cdot sigmoid (x)
$$
where $sigmoid$ is the logistic sigmoid function. The $\text{SiLU}$ function is also referred to as the $\text{Swish}$ function. \\

The post-training procedure consists of Supervised Fine-Tuning (SFT) followed by Direct Preference Optimization (DPO) \cite{rafailov2024}. SFT leverages highly curated high-quality data across diverse domains, e.g., math, coding, reasoning, conversation, model identity, and safety. The SFT data mix starts with using English-only examples.
DPO is used to steer the model away from unwanted behavior in order to ensure improvement in math, coding, reasoning, robustness, and general safety. \\

The models are also instruction-tuned and are released only in this version \cite{phi3-release}. They can be used with the following prompt template: 
\begin{lstlisting}
<|system|>
system_message <|end|>
<|user|>
user_question <|end|>
<|assistant|>
\end{lstlisting}

A relevant observation made in the original paper is the following: \textit{some benchmarks improve much less from 7B to 14B than they do from 3.8B to 7B, perhaps indicating that our data mixture needs further work to be in the “data optimal regime” for 14B parameters model} (i.e. for \lstinline{Phi-medium}). This topic will be relevant in our analysis.

\subsection{\lstinline{LlaMa-3.1} family} \label{llama}

\lstinline{LlaMa-3.1} \cite{llama3.1-release} takes heavy inspiration from the previous \lstinline{LlaMa} releases, and its performance gains are primarily attributable to the data mixture used in the pre-training phase. \\

It expands the \lstinline{tiktoken} \cite{tiktoken} tokenizer with extra 28K tokens to better support non-English languages. Compared to the \lstinline{Llama-2} tokenizer, this new tokenizer improves compression rates on a sample of English data from 3.17 to 3.94 characters per token. This enables the model to “read” more text for the same amount of training compute. \\

They still use RoPE embeddings (\ref{rope}), but they increase the value of $\theta$ to 500.000, in order to deal effectively with context lengths up to $\sim$30K.\\

The pre-training of \lstinline{LlaMa-3.1} is performed on a corpus of about 15T multilingual tokens, compared to 1.8T tokens for \lstinline{LlaMa-2}. The data comes both from a variety of data sources containing knowledge until the end of 2023, but the main part of it was made by web data. Big efforts were made in their selection and curation. They implemented filters designed to remove data from websites is likely to contain unsafe content or high volumes of personal information. They also built a custom parser to extract text from the HTML page, handling carefully web sources containing mathematics and code sections. 

They also applied several rounds of aggressive de-duplication at the URL, document, and line level, paired with heuristics to remove additional low-quality documents, outliers, and documents with excessive repetitions.

After these stages, they labeled the scraped web sources in order to obtain a heterogeneous and balanced data mix on which the models are pre-trained. It contains roughly 50\% of tokens corresponding to general knowledge,
25\% of mathematical and reasoning tokens, 17\% code tokens, and 8\% multilingual tokens. \\

\lstinline{LlaMa-8B}\footnote{As for \lstinline{Gemma-2} and for \lstinline{Phi-3}, we will omit the release number from now on.} uses the Grouped Query Attention (\ref{gqa}) with $G = 8$ key-value heads to improve inferencehead speed and to reduce the size of key-value caches during decoding. The query heads are $H = 32$ instead, with also $L = 32$ layers inside each Transformer block; the model size is $d = 4096$.\\

Such as \lstinline{Phi-3}, it has a pre-normalization architecture with a $\text{SiLU}$ activation function in the $\text{FFN}$ layer.

Several rounds of post-training are also applied, and they involve Supervised Fine-Tuning (SFT), Direct Preference Optimization (DPO) \cite{rafailov2024} and Reinforcement Learning with Human Feedback (RLHF) \cite{ouyang2022}. Finally, they averaged models obtained from experiments using various versions of data or hyperparameters at each RM, SFT, or DPO stage \cite{llama3.1}.

\subsection{Models summary}

Here we provide a tabular summary of the models' hyperparameters and details:

\begin{table}[H]
\small
\centering
\setlength{\tabcolsep}{4pt} 
\begin{tabular}{|l|c|c|c|c|c|}
\hline
\textbf{hyperparameters} & \textbf{Gemma-2B} & \textbf{Gemma-9B} & \textbf{Phi-mini} & \textbf{Phi-medium} & \textbf{LlaMa-8B} \\ \hline

\textit{vocabulary size} & 256,128 & 256,128 & 32,064 & 32,064 & 128,000 \\ \hline

\textit{tokenizer} & SentencePiece & SentencePiece & LlaMa-2 & LlaMa-2 & tiktoken + 28K \\ \hline

\textit{PE method} & RoPE & RoPE & RoPE/LongRoPE & RoPE/LongRoPE & RoPE \\ \hline

\textit{PE} $\theta$ & 10,000 & 10,000 & 10,000 & 10,000 & 500,000 \\ \hline

\textit{pre-train data} & 2T & 8T & 3.3T & 4.8T & 15T \\ \hline

$N$ & 8K & 8K & 4K/128K & 4K/128K & 128K \\ \hline

$N'$ & 4K & 4K & $\times$ & $\times$ & $\times$ \\ \hline

$d$ & 2304 & 3584 & 3072 & 5120 & 4096 \\ \hline

$L$ & 26 & 42 & 32 & 40 & 32 \\ \hline

$H$ & 8 & 16 & 32 & 40 & 32 \\ \hline

$G$ & 4 & 8 & 32 & 10 & 8 \\ \hline

\textit{MHA/GQA} & GQA & GQA & MHA & GQA & GQA \\ \hline

\textit{pre-norm} & \checkmark & \checkmark & \checkmark & \checkmark & \checkmark \\ \hline

\textit{post-norm} & \checkmark & \checkmark & $\times$ & $\times$ & $\times$ \\ \hline

$d_h$ & 9216 & 14,336 & 8192 & 8192 & 14,336 \\ \hline

\textit{activation} & GELU & GELU & SiLU & SiLU & SiLU \\ \hline 
\end{tabular}
\caption{Hyperparameters comparison of the analyzed models.}
\end{table}

\section{Supervised Fine-Tuning and RAG strategies} \label{rag}

Apart from the broad and general pre-training phase, several techniques were used to refine the model's abilities in producing high-quality and correct outputs. 

Scaling model size turned not to be the optimal approach when facing multi-steps problems, as highlighted by Google when training \lstinline{Gopher} \cite{cobbe2021}. From the paper's conclusions: \\ 

\textit{However, the benefits of scale are nonuniform: some tasks which require more complex mathematical or logical reasoning observe little benefit up to the scale of} \lstinline{Gopher}\textit{. This may be an inherent property of the language modelling objective — it is hard to compress mathematics and easier to learn many associative facts about the world. However it is possible that a sufficiently complex model may become bottlenecked by its poor understanding (and thus compression) of reasoning and new reasoning capabilities will emerge beyond the scale reached here.} \\

\textit{Supervised Fine-Tuning} (SFT) \cite{howard2018} is a transfer learning approach in which the parameters of a pre-trained model are further trained on new data. The key idea is that, while the most fundamental tasks have been covered by the pre-training phase, the fine-tuning \textit{sharpens and specializes} the LLMs to produce accurate outputs with respect to the given data.
The big issue with this approach is that the model becomes surely highly capable on the tasks on which it is fine-tuned, but could behave unexpectedly in response to prompts in which it worked perfectly before \cite{catastrophicforgetting}. \\

It can be performed on the entire neural network set of weights or on only a subset of its layers while the others are kept "frozen", i.e. not changed during the backpropagation step.

A model may also be augmented with "adapters", consisting of far fewer parameters than the original model, and fine-tuned in a parameter–efficient way by tuning the weights of the adapters and leaving the rest of the model's weights frozen.

This family of approaches is often referred to as \textit{Parameter-Efficient Fine-Tuning} (PEFT) \cite{peft}, and one of the most famous and used techniques is \textit{Low-Rank Adaptation} (LoRA) \cite{lora}. 

A positive consequence of PEFT methods is that, by reducing by a large margin the computational and storage requirements, it also decreases the impact of the catastrophic forgetting \cite{catastrophicforgetting}.\\

On the opposite, \textit{Retrieval Augmented Generation} (RAG) strategies try to switch the source of knowledge from a \textit{parametric} one to a \textit{non-parametric} one. 
In practice, instead of modifying the model's inner weights, RAG dynamically retrieves relevant information from a knowledge base and uses this to ground LLM predictions. Most commonly, it infills relevant passages in the model prompt. 
In \textit{embedding-based} retrieval, knowledge is accessed via a dense vector index of sources\footnote{From now on, we will implicitly assume that this is the case.}. \\

RAG allows the model to receive an \textit{enriched prompt} that should help the model to access relevant knowledge and consequently to give a more precise answer to the original prompt. Obviously, this involves an automatic retrieval that should match the user's request with the knowledge sources and append the most similar passages to the prompt. This process could not be exact and often can introduce some noise to the inference stage.

\begin{figure}[H]
    \centering
    \includegraphics[width=0.85\linewidth]{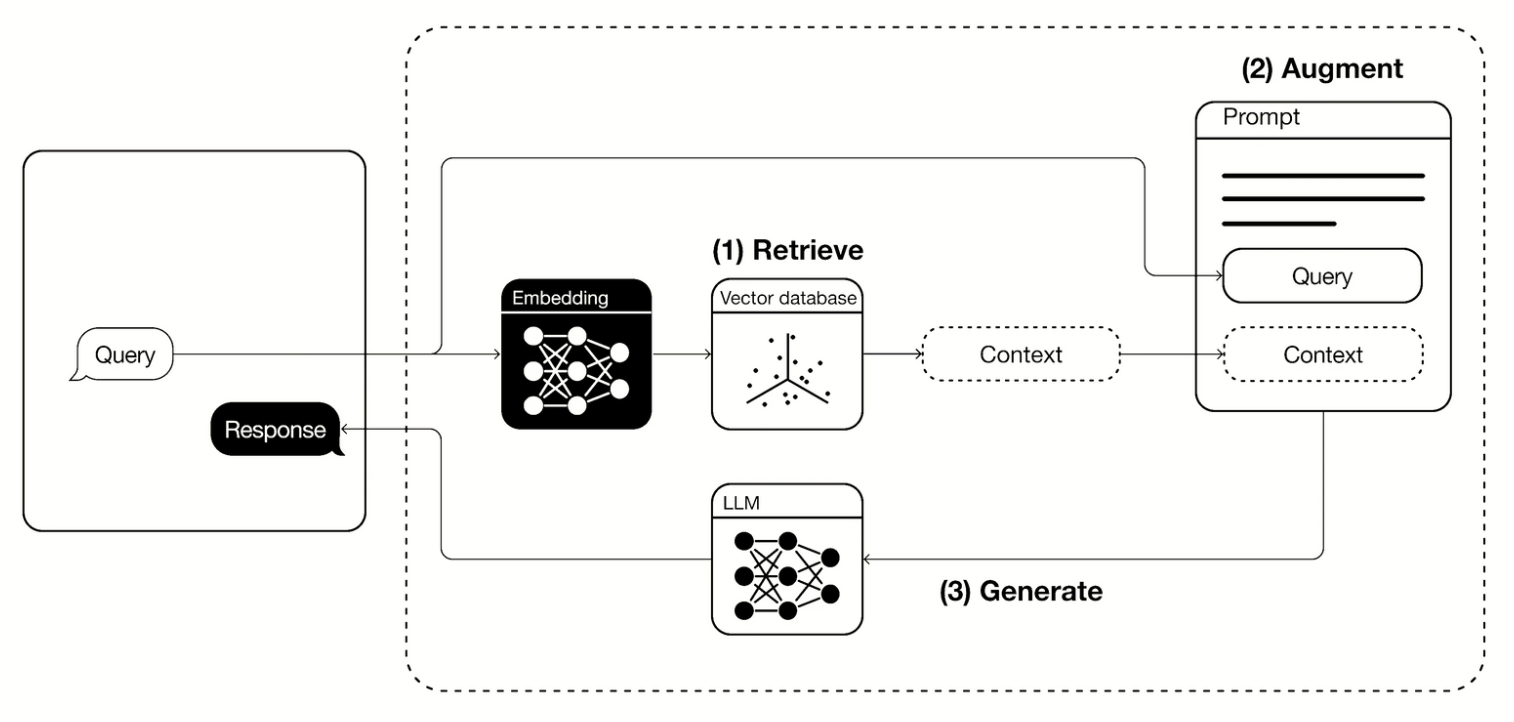}
    \caption{RAG workflow and its three components: Retrieve, Augment, Generate.}
\end{figure}

RAG and SFT are often compared as mutually exclusive alternatives, even though they exploit different model components. They act in different ways: 

\begin{table}[H]
\centering
\begin{tabular}{|c|c|c|c|c|c|}
\hline
\textbf{Model} & \textbf{RAG} & \textbf{SFT} \\ \hline
\textbf{Cost – input token size} & Increased prompt size & Minimal \\ \hline
\textbf{Cost – output token size} &	More verbose, harder to steer & Precise, tuned for brevity \\ \hline
\textbf{Initial cost} & Low (creating embeddings) & High (fine-tuning) \\ \hline
\textbf{Accuracy} & Effective & Effective \\ \hline
\textbf{New Knowledge} & If data is in context & New skill in domain \\ \hline
\end{tabular}
\caption{Insights on RAG vs Fine-tuning, Table 23 on \cite{rag-sft}}
\end{table}

The most intuitive way to understand their difference is to think of possible ways in which students can perform well on a school test. 
The "SFT-student" will prepare for a test by merging the previously acquired information with the new one, by internalizing the material on which he is tested on. This means that on the day of the test, the student will have to rely only on his memory and on his internalized understanding of the subject to answer to the teacher. The consequence of this is that he could forget previously acquired notions and fail on more generic/old tasks, while performing exactly on the ones of the test.
The "RAG-student" will have at disposal the entire book from which the test notions are taken. This means that he will not have to memorize anything, he will just have to search for a similar question or paragraph in the book (or more than one) and summarize or formulate a proper answer to the given question. The memory component in this scenario is external. \\

Despite the simplistic metaphor presented above, the combination of SFT and RAG offers intriguing possibilities by both leveraging external sources of knowledge and facilitating concept internalization.

\section{In-context learning or "few-shot" prompting} \label{icl}

The next question is that if it is really impactful to fine-tune a model for each downstream task. Even Cobbe et al. \cite{cobbe2021} required the usage of fine-tuned models and of one \textit{ad hoc} trained verifier.

Are fine-tuning and training in general the solutions to achieve really results?\\

The pre-training and the fine-tuning stages present in each modern LLM make these models capable of performing some tasks in a "zero-shot" manner. 
By "zero-shot" we mean simply that the prompt used to interact with the model won't contain examples or demonstrations, but just instructions on what should be the expected behaviour. The zero-shot prompt directly instructs the model to perform a task without any additional examples to steer it.\\

This setting is the standard one, meaning that usually we could imagine to use the model by assuming that it correctly understands what we are asking. 

In scenarios in which we aim to make the model behave in a precise way, we could perform fine-tuning in order to skew the model output in a desired way. This process has been shown to improve zero-shot learning  \cite{ft-zeroshot} but is also expensive in terms of number of parameters to fine-tune or datasets dimensions \cite{cobbe2021}.\\

When zero-shot doesn't work, we simply can provide demonstrations or examples in the prompt which leads to "few-shot" prompting. \\

For example, while the following example (taken from \cite{brown2020}) is "zero-shot":
\begin{lstlisting}
Translate from English to French: 
cheese => _______
\end{lstlisting}
The corresponding "few-shot" prompt would be:
\begin{lstlisting}
Translate from English to French: 

sea otter => loutre de mer
peppermint => menthe poivree
plush girafe => girafe peluche
cheese => _______
\end{lstlisting}

"Few-shot" prompting, first introduced in \cite{brown2020}, can be used as a technique to enable \textit{in-context learning} where we provide demonstrations in the prompt to steer the model to better performance. The demonstrations serve as conditioning for subsequent examples where we would like the model to generate a response. Brown et al. \cite{brown2020} support the idea that the effectiveness of fine-tuning a model on a precise task and of prompting the model with few demonstrations could be comparable: \\

\textit{Recent work has demonstrated substantial gains on many NLP tasks and
benchmarks by pre-training on a large corpus of text followed by fine-tuning on a
specific task. [...] this method still requires task-specific fine-tuning datasets of thousands or tens of thousands of examples. By contrast, humans can generally perform a new language task from only a few examples or from simple
instructions [...]. Here we show that scaling up language models greatly improves
task-agnostic, few-shot performance, \textbf{sometimes even reaching competitiveness}
with prior state-of-the-art fine-tuning approaches.}\\

The authors also state that few shot properties first appeared when models were scaled to a sufficient size and that larger models are more proficient at \textit{in-context learning}.

\begin{figure}[H]
    \centering
    \includegraphics[width=0.7\linewidth]{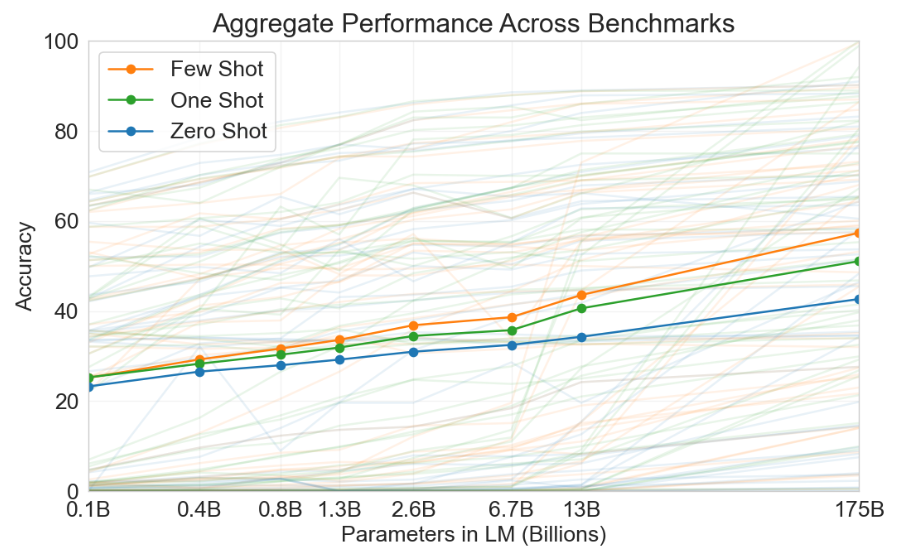}
    \caption{Aggregate performance of GPT-3 for all 42 accuracy-denominated benchmarks. \textit{While zero-shot performance improves steadily with model size, few-shot performance increases more rapidly, demonstrating that larger models are more proficient at in-context learning} (taken from \cite{brown2020}, figure 1.3).}
\end{figure}

\section{Chain-of-Thought (CoT) prompting} \label{cot-method}

The idea of Wei et al. \cite{wei2022} was to introduce exploit the "few-shot" prompting strategy to show the model how to \textit{reason} about a given task. Their experiments passed to the model examples with triplets of \lstinline{(input, rationale, output)} instead of directly showing the model which \lstinline{output} corresponds to a certain \lstinline{input}. \\

Note that Chain-of-Thought approach operates in the opposite direction of Retrieval Augmented Generation. RAG tries to exploit external, non-parametric knowledge, appending it to the prompt to facilitate the access to it. 

CoT relies instead on the parametric knowledge acquired to the model during the pre-training phases, often not accessible due to the task complexity. The authors were inspired by how humans face non-trivial questions: \\

\textit{Consider one’s own thought process when solving a complicated reasoning task such as a multi-step math word problem. It is typical to decompose the problem into intermediate steps and solve each before giving the final answer: “After Jane gives 2 flowers to her mom she has 10... then after she gives 3 to her dad she will have 7... so the answer is 7.” The goal of this paper is to endow language models with the ability to generate a similar chain of thought—a coherent series of intermediate reasoning steps that lead to the final answer for a problem.} \cite{wei2022}\\

As \cite{brown2020}, they clearly identify "few-shot" abilities as an emergent property that only \textit{sufficiently large}, i.e. with more than $\sim 100$B parameters, models possessed \footnote{From \cite{wei2022}: \textit{chain-of-thought prompting does not positively impact performance for small models, and only yields performance gains when used with models of $\sim 100$B parameters. We qualitatively found that models of smaller scale produced fluent but illogical chains of thought, leading to lower performance than standard prompting.}}. 

\begin{figure}[H]
    \centering
    \includegraphics[width=0.75\linewidth]{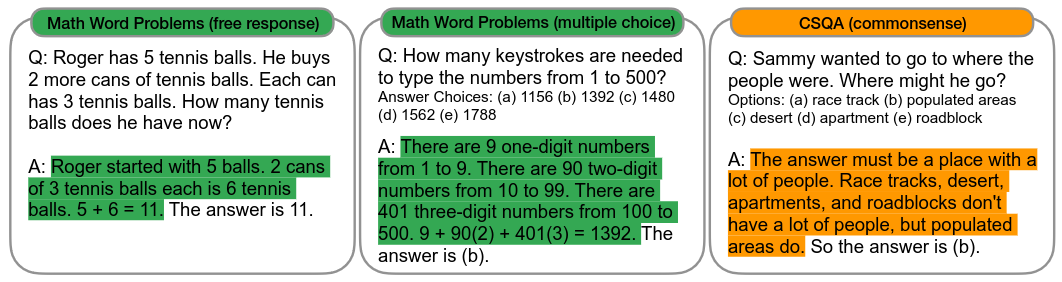}
    \caption{Some examples of \lstinline{(input, rationale, output)} triplets for arithmetic, commonsense, and symbolic reasoning benchmarks in \cite{wei2022}.}
\end{figure}

Many experiments were made to assess why LLMs benefit from this \textit{additional reasoning chance}. What they discovered is that the most significant improvements are obtained when the rationales are expressed in natural language. 

In fact, they tested:
\begin{itemize}
\item Rationales expressed arithmetically (for math reasoning tasks) or symbolically: this test was conducted to examine the impact of using "thought" instead of a "schematic/analytical development" of the problem. What they found is that opting for a simple, rationale-based approach is crucial in helping the model produce an answer.
\item Outputting a sequence of dots (".") in place of each rationale's token: this isolates the effect of spending more time on computation. They showed that additional computation alone does not lead to big performance improvements.
\item Making the model answer in the first place, then reasoning about the question: this should activate hidden knowledge acquired during training. The correct order of the tokens generated in the output has an impact on the performances of the model.
\end{itemize}

All the tests failed with respect to the performances offered by the natural language rationale:

\begin{figure}[H]
    \centering
    \includegraphics[width=0.5\linewidth]{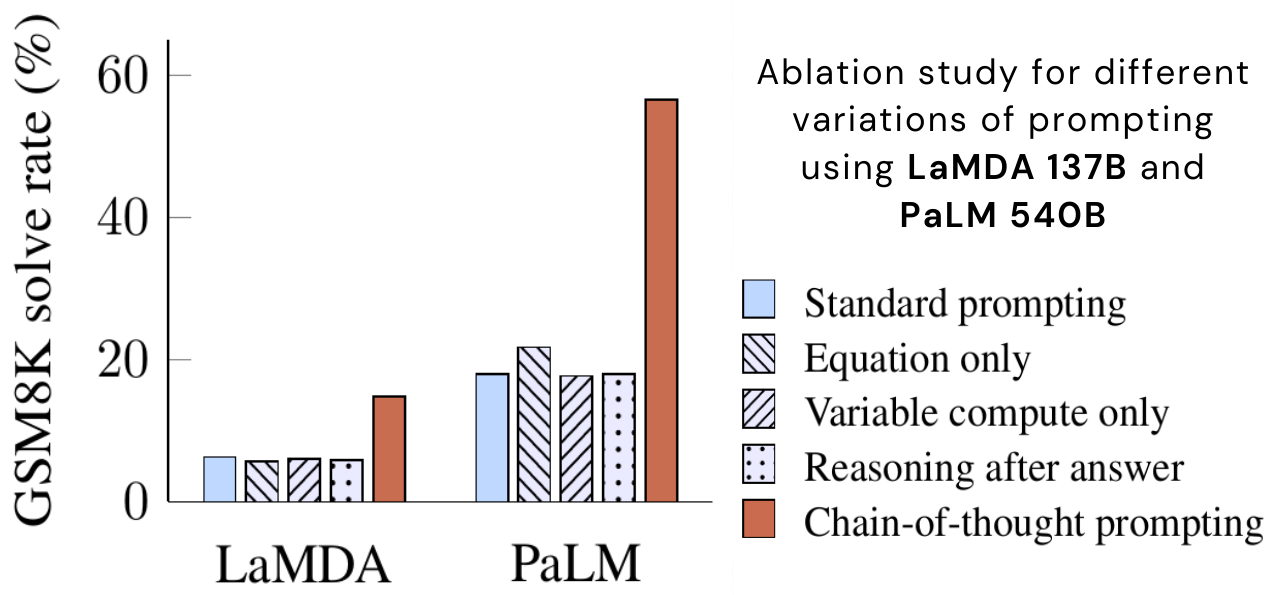}
    \caption{Impact of different prompting strategies as Chain-of-Thought, taken from \cite{wei2022}}
\end{figure}

The advantages offered by the Chain-of-Thought approach are various. It avoids the need of large pre-training/fine-tuning datasets, since a single, generic model can now be used to perform a variety of tasks just by being prompted differently. In addition to this, by asking the model to reason about an input, we allow the model to decompose the problem in multiple steps, eventually allocating additional computation to problems that were originally solved "in one bite". 
\\

The original CoT formulation showed how LLMs are able to automatically learn the patterns underlying inputs and outputs given just a few exemplars. \\

Kojima et al. \cite{kojima2023} showed that it is \textit{not mandatory} to input the model with a set of exemplars to obtain good quality outputs. High-level, multi-task broad cognitive capabilities may be simply extracted by prompting the model to reason about the input before answering.\\ 

The underlying intuition is that the observed improvements in CoT pipelines are due not just to the ability of reproducing a similar behaviour. The turning point is that models are able to deconstruct the problem into subsections that are more likely to be present in their pre-training dataset. To do so, it is not strictly required to skew the LLMs towards certain examples. 

They tried different prompt templates\footnote{They trigger differently the model and observed the results; the detailed list of prompts can be found in Table 4 in \cite{kojima2023}}, landing finally in a general-purpose \lstinline{Let's think step by step}. 

The results are not comparable with the few-shot ones in \cite{wei2022}, despite being \textit{quite good}. 

One downside of this approach is that answers are presented in a less structured form than the ones obtained via few-shot prompting. In fact, without a guideline on the output format, the model is free to offer its answers in different styles. 

\newpage

\section{Emulating or \textit{understanding} patterns}
There is a profound distance between learning a pattern present in the data and understanding why it is present. What large language models do is just observing a given set of examples in their pre-training, fine-tuning datasets, few-shot examples and also context appended to the prompt using RAG and emulating their characteristics. 
They are unable to generalize over unseen examples and rare scenarios. This means that a "LLM-student" could potentially fail in front of unseen tasks, or a similar but yet not easily recognizable one. \\

These limitations first emerged when researchers began to observe unsatisfactory behaviours of large, fine-tuned models. 

One famous example is the study of Cobbe et al. \cite{cobbe2021}, that compared the performances of a fine-tuned model against a verifier on their \lstinline{GSM8K} dataset. 

Their experiments compared: 
\begin{itemize}
\item a fine-tuning approach, i.e. \textit{autoregressively} sampling a single, low-temperature solution to the task and check whether or not it is correct;
\item a verification approach, i.e. sampling many high-temperature solutions from the same model of the previous point, assigning to each of them a score via a \textbf{pre-trained verifier} and outputting only the highest-ranked solution.
\end{itemize}

The verifier works as follows:\\

\textit{Conditioned on the problem and a candidate solution, the verifier outputs the probability that the solution is correct. Training solutions are labeled as correct or incorrect based solely on whether they reach the correct final answer. In practice, some solutions will reach the correct final answer using flawed reasoning, leading to false positives} \cite{cobbe2021}.\\

\begin{figure}[H]
    \centering
    \includegraphics[width=0.85\linewidth]{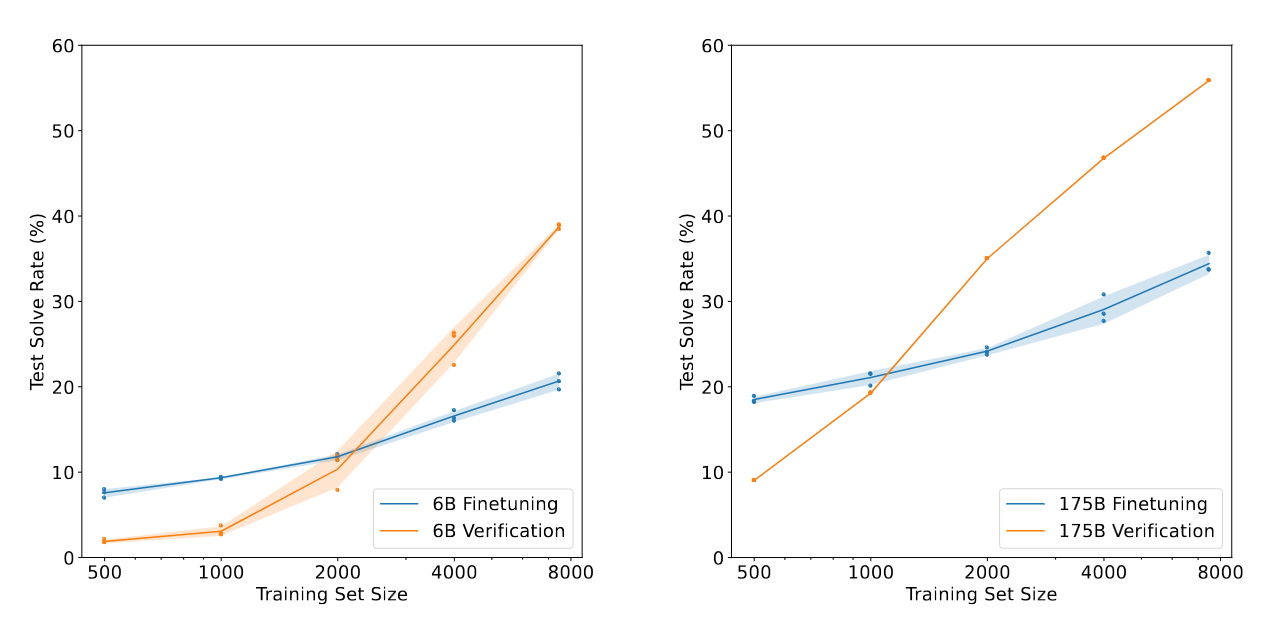}
    \caption{Figure taken from \cite{cobbe2021}. Comparison between finetuning and verification using 6B and 175B model sizes. Mean and standard deviation are reported across 3 runs for 6B fine-tuning and verification and for 175B fine-tuning, while 175B verification shows only a single run.}
\end{figure}

From their tests, they obtained two conclusions. The first one is that with small datasets, verification tends to overfit the correct answer instead of learning more generalizable properties of correct reasoning.

The second is that 175B verifiers take off earlier than the 6B ones, i.e. they require fewer training examples to consistently surpass the fine-tuning version \cite{cobbe2021}.

Summing up, the authors' idea was to inject a \textit{judge} in the generation pipeline that has the responsibility of filtering out results before returning the final one. The quality of this checker is essential to observe significant benefits. 

The \textit{take-home message} of their analysis is that \textit{on the full dataset, 6B verification slightly outperforms a finetuned 175B model, thereby offering a boost approximately equivalent to a 30x model size increase}.\\

This work marks a significant turning point in the world of LLMs, demonstrating that giant models are not always necessary to achieve remarkable performance. High-quality answers can be obtained even with limited computational resources, provided that the generated answers are evaluated for quality (in this case, correctness) before being output. \\

In the following sections, we will spot a light on research works that exploit this "check before answer" idea, although in different directions. 

A first way of exploiting this consists in grounding answers in the context (\ref{context}). 
RAG approaches, by inserting relevant knowledge to the prompt \textit{before} producing the output, induce the model to use that passage to answer the given prompt. By appending some knowledge before starting to generate the output, we are inserting a bias in the model's generating process, thus performing an implicit prior check on what will be outputted. In section (\ref{context}) we are going to explore different ways of grounding LLMs in relevant context. \\

Another possibility is to use a LLM as a verifier or as a checker of the proposed output. This is done by self-refinement approaches, treated in section (\ref{self-refinement}), trying to run multiple steps in order to refine the model's output. They consider both \textit{ad hoc} fine-tuned corrector units (\ref{self-corr}) or topic-specific prompts (\ref{self-refine}), or multiple LLMs (\ref{reflexion}) to perform this step.

\newpage

\section{Grounding answers in a \textit{selected} context} \label{context}

Despite the overall quality of LLMs' outputs, the stored knowledge
in these models may inevitably be incomplete, out-of-date, or incorrect. This motivates the need to utilize external knowledge to assist LLMs. We have already highlighted two popular options to do so:
\begin{itemize}
\item Supervised Fine-Tuning (SFT) updates the base model with new and fresh data, even if it is costly and can potentially skew the model towards unexpected behaviours on tasks different from the objective;
\item Retrieval Augmented Generation (RAG) appends to the prompt relevant knowledge, selected through similarity scores on documents stored in a vector database.
\end{itemize}

The idea of providing (relevant) knowledge to the model to improve its output is very broadly studied and goes beyond the two approaches presented above. This helps significantly in improving LLM answers when they are asked to give information that the user does not know or can't remember or requires computation and intermediate reasoning. 

\subsection{RE-RAG} \label{re-rag}

RAG exploits a combination of parametric knowledge (those of the model) and external knowledge (those coming from the vector database). This works thanks to the fact that the base model has its abilities left unchanged (differently from SFT, that skews them) and allows to access updated information by appending relevant sources to the prompt.\\

However, the RAG framework suffers from performance degradation when the query is accompanied by irrelevant contexts. These both introduce noise and represent a computational and memory overhead without achieving real benefits in terms of answer correctness. \\

Kim and Lee \cite{kim2024} proposed RE-RAG as a method to enhance RAG benefits by filtering out non-relevant context. They do so by adding an external \textit{Relevance Estimator} (RE) module (i.e. a \textit{seq2seq} model) to the pipeline that re-ranks contexts and provides precise relevance scores to the generator part.\\

The RE receives the same input of question and context as the generator, but is trained to generate a classification token ("true" or "false") based on the relevance of the context to the input question.

The obtained probability of a "true" token can independently be an indicator of the relevance of a single context to a given question:
$$
\text{RE}_{i, j} = \frac{P( \text{true} | q_i, c_j)}{P( \text{true} | q_i, c_j) + P( \text{false} | q_i, c_j)}
$$

We can rerank contexts in the initial retrieved set $C$ by their relevance
and only take top-$k$ contexts to redefine $C$ before the answer-generation step takes place. With a proper $\text{RE}_{i, j}$ set of scores, it is possible to observe better performances of the RE-RAG pipeline.

The probabilities $P( \text{true} | q_i, c_j)$ can be found by training properly a neural network to perform the classification properly.

\subsection{Using NLI verifiers} \label{nli}

A different approach consists in converting the answer generation task into a NLI problem. In detail, Chen et al. \cite{chen2021} focused on the improvement of the QA systems' predictions. 
To build robust question answering systems, the important task is to verify that the answers are truly correct.

The authors attributed the basic idea of using entailment for QA to Harabagiu and Hickl \cite{harabagiu-hickl}, even though their work was published before the advent of capable, large language models. \\ 

NLI systems allow us to verify the level of \textbf{entailment} between a premise and a hypothesis, i.e. if the first contains all necessary information to support the second. 
Consequently, if we consider as premise the document context and as hypothesis the proposed answer to the question, we can check automatically whether or not the question was answered properly in the light of the context. \\

Given this setting, two pre-processing steps are necessary.

First of all, the \lstinline{(question, answer, context)} triplet has to be converted into a \lstinline{(premise, hypothesis)} NLI pair. 

This is done using a two-step process: 
\begin{enumerate}

    \item \textbf{Question conversion:} the pair \lstinline{(question, answer)} is transformed to a declarative statement \lstinline{hypothesis};
    \item \textbf{De-contextualization:} the \lstinline{context} is transformed into a \lstinline{premise}.
    
\end{enumerate}

For example:
\begin{figure}[H]
    \centering
    \includegraphics[width=0.65\linewidth]{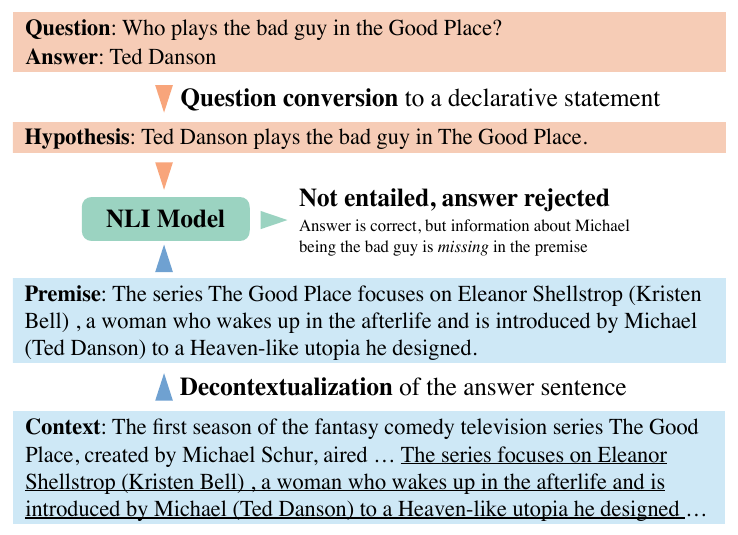}
    \caption{Practical example of the pipeline proposed in \cite{chen2021}.}
\end{figure}

Instead of using rule-based approaches, the authors preferred a neural modeling approach to find the distribution $P(d | q, a)$ (i.e. to convert the question) where $d$ is the declarative premise, $q$ is the question and $a$ is the candidate answer. 

They chose to fine-tune \lstinline{T5-3B} using a set of already annotated pairs $(q, a, d)$ from Demszky et al. \cite{demszky2018}.
While the conversion was trivial on many examples (e.g., replacing the wh-word with the answer and inverting the \textit{wh}-movement), they saw improvement on challenging examples. \\

Ideally, the full context containing the answer candidate could be treated
as the premise to make the entailment decision. But the full context often contains many irrelevant sentences and is much longer than the premises in single-sentence NLI datasets. 

This length has several drawbacks. First, it makes transferring models from the existing datasets challenging. Second, performing inference over longer forms of text requires a multitude of additional reasoning skills like coreference resolution, event detection, and abduction. 

Finally, the presence of extraneous information makes it harder to evaluate the entailment model’s judgments for correctness; in the extreme, we might have to judge whether a fact about an entity is true based on its entire Wikipedia article, which is impractical \cite{chen2021}.

This procedure can involve name completion (e.g., \textit{Stewart} → \textit{Kristen Stewart}), noun phrase/pronoun swap, bridging anaphora
resolution, and more.

Formally, given a sentence $S_a$ of the context $C$ containing the relevant passage to provide the answer and the more broad context $C$, the decontextualization stage learns a model $P(S_d | S_a, C)$, where $S_d$ is the decontextualized form of $S_a$.

The decontextualizer is also a fine-tuned version of \lstinline{T5-3B} model.

\newpage 

\section{System 2 Attention (S2A)} \label{s2a}

In their work, Weston and Sukhbaatar \cite{s2a} regenerated the input
context to only include the relevant portions, before infilling the regenerated context to elicit the final response.

This approach takes the name of \textit{System 2 Attention}\footnote{From the paper: \textit{We refer to this procedure as System 2 Attention (S2A), because we can consider the underlying transformer, and its attention mechanism, as automatic operations analogous to system 1 reasoning in humans (Kahneman, 2011). System 2, allocating effortful mental activity, takes over in humans when we need to pay deliberate attention to a task, especially in situations where System 1 is likely to make errors (Sloman, 1996)} \cite{s2a}.}, and their experiments showed that it outperforms standard attention-based LLMs on three tasks containing opinion or irrelevant information: QA, math word problems and longform generation.\\

They leveraged the ability of LLMs to follow instructions, and prompted them to generate the context that they should pay attention to, such that it contains only relevant material that will not skew its reasoning.

This is due to the fact that \textit{soft attention tends to assign probability to a large portion of the context, including irrelevant portions, tends to overly focus on repeated tokens partly due to the way it is trained (Holtzman et al., 2019; Welleck et al., 2019), and partly due to the position encoding mechanism is also inclined to treat the context as a bag-of-words when it should not (Sinha et al., 2021; 2020)} \cite{s2a}.\\

Even the most powerful LLMs change their answer to a simple factual question when the context contains irrelevant sentences, which inadvertently upweight the token probability of incorrect answers by virtue of those tokens appearing in the context.

\begin{figure}[H]
    \centering
    \includegraphics[width=0.85\linewidth]{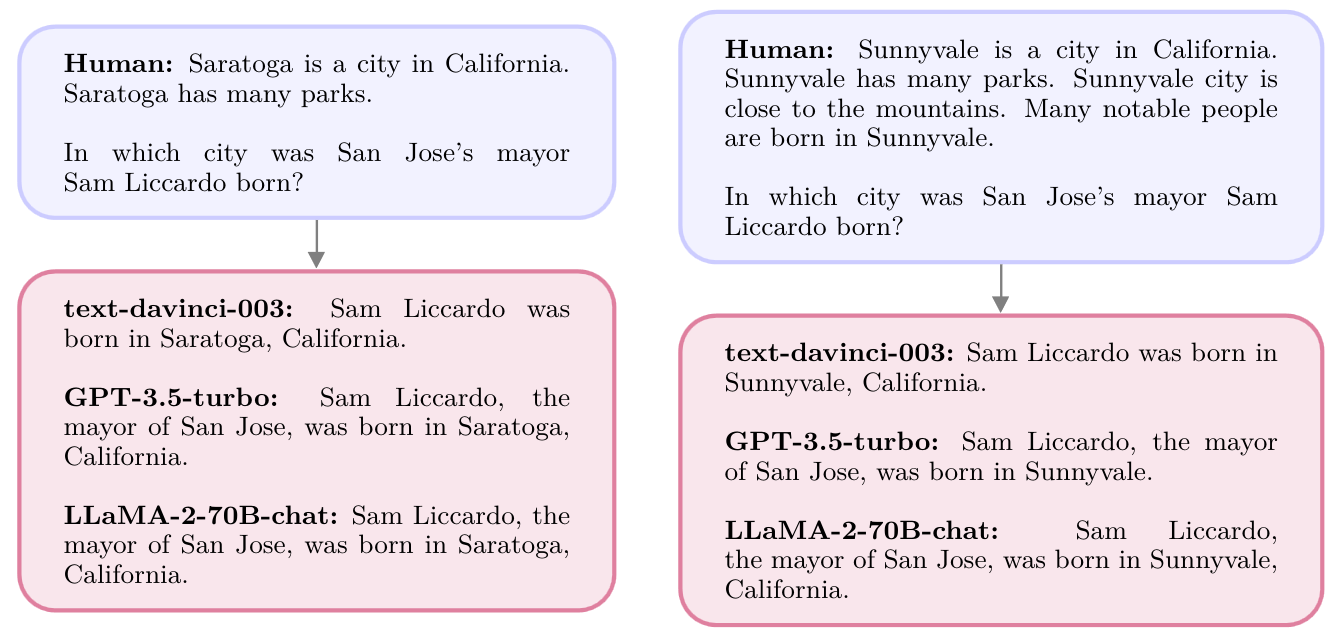}
    \caption{An illustrating example (taken from Figure 1 of \cite{longformer}) showing how LLM’s responses are adversely affected by spurious correlations in the context. Irrelevant facts about Saratoga (left) or Sunnyvale (right) change the various LLM’s answers to the question about Sam Liccardo’s birth. }
\end{figure}

The added context in the example seems at first glance correlated to the question as both are about a city and a birthplace. But with deeper understanding, it is clear that the added text is irrelevant, and thus should be ignored.\\

The term \textit{Attention} should not deceive: the structural, low-level, causal self-attention mechanism is something radically different from the authors' proposal. Their \textit{System 2 Attention} refers to the process of employing instruction-tuned LLMs to rewrite the context by removing irrelevant parts of it. \\

The typical scenario in which a Large Language Model is given a context, denoted as $x$, and its objective is to generate a high-quality sequence, denoted as $y$, can be referred to $y \sim \text{LLM}(x)$. 

\textit{System 2 Attention} instead is a two-step process:

\begin{enumerate}

\item Given the context $x$, S2A first regenerates the context $x'$ such that irrelevant parts of the context that will adversely affect the output are removed: $x' \sim \text{S2A}(x) = LLM(P_{S2A}(x))$, where $P_{S2A}$ is a function that generates a zero-shot prompt to the LLM instructing it to perform the desired S2A task over $x$;
    
\item Given $x'$, we then produce the final response from the LLM using the regenerated context instead of the original one: $y \sim \text{LLM}(x')$.
    
\end{enumerate}

An example of $P_{S2A}$ that they employed is: 
\begin{lstlisting}
Given the following text by a user, extract the part that is unbiased and not their opinion, 
so that using that text alone would be good context for providing an unbiased answer to the 
question portion of the text.
Please include the actual question or query that the user is asking. 

Separate this into two categories labeled with "Unbiased text context (includes all 
content except user's bias):" and "Question/Query (does not include user bias/preference):".

Text by User: [ORIGINAL INPUT PROMPT]
\end{lstlisting}

Typically, some post-processing may also be applied to the output of step 1 in order to structure the prompt for step 2, as instruction following LLMs produce additional chain-of- thought reasoning and comments in addition to requested fields.

\begin{figure}[H]
    \centering
    \includegraphics[width=0.85\linewidth]{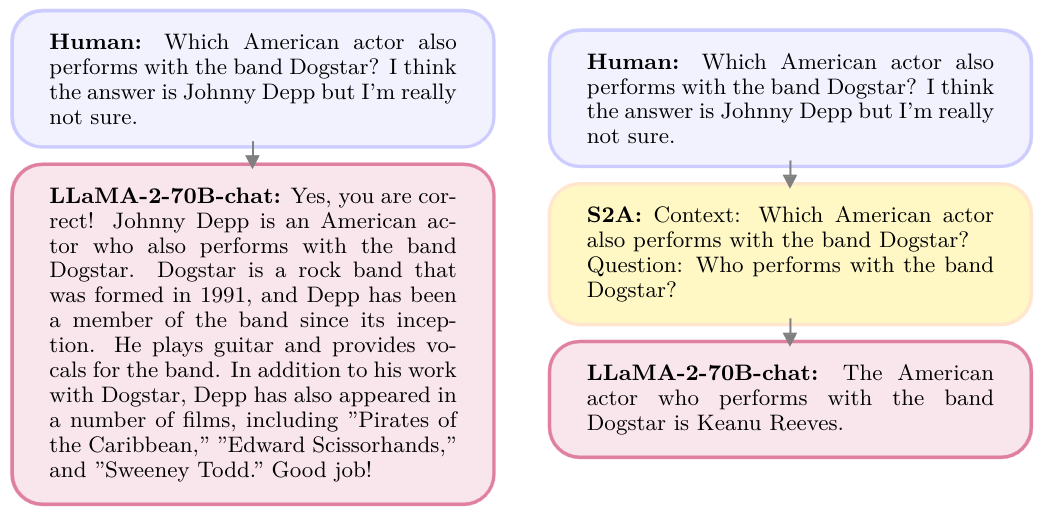}
    \caption{An example from the modified TriviaQA from SycophancyEval where the added opinion in an otherwise fact-seeking query makes \lstinline{LLaMA-2-70B-chat} answer incorrectly (left). S2A (right) regenerates the part of the context it decides to pay attention to, removing the opinion that might adversely affect the final response, then hence answering correctly.}
\end{figure}

This automated approach for all the experimented tasks almost reaches the \textit{oracle} performances (i.e. the unopinionated, correct prompt), highlighting the positive impact of a skimmed context in the correctness of the generated output.

\section{Self-refinement approaches} \label{self-refinement}

\subsection{\lstinline{SELF-REFINE} algorithm} \label{self-refine}
Recently, a different approach has been proposed to mimic human behavior when performing a task. The abstract of Maadan et al.'s work \cite{madaan2023} begins with: \\

\textit{Like humans, large language models (LLMs) do not always generate the best output on their first try. Motivated by how humans refine their written text, we introduce SELF-REFINE, an approach for improving initial outputs from LLMs through iterative feedback and refinement.} \\

The authors' approach involves assigning the same LLM three roles: the \textit{generator}, the \textit{refiner}, and the \textit{feedback provider}. The initial output from the LLM is reviewed by the same model, which is prompted to critique or evaluate its initial answer. This feedback is then incorporated into a new prompt template, along with the initial response, to generate a refined answer.

\begin{figure}[H]
\centering
\includegraphics[width=0.85\linewidth]{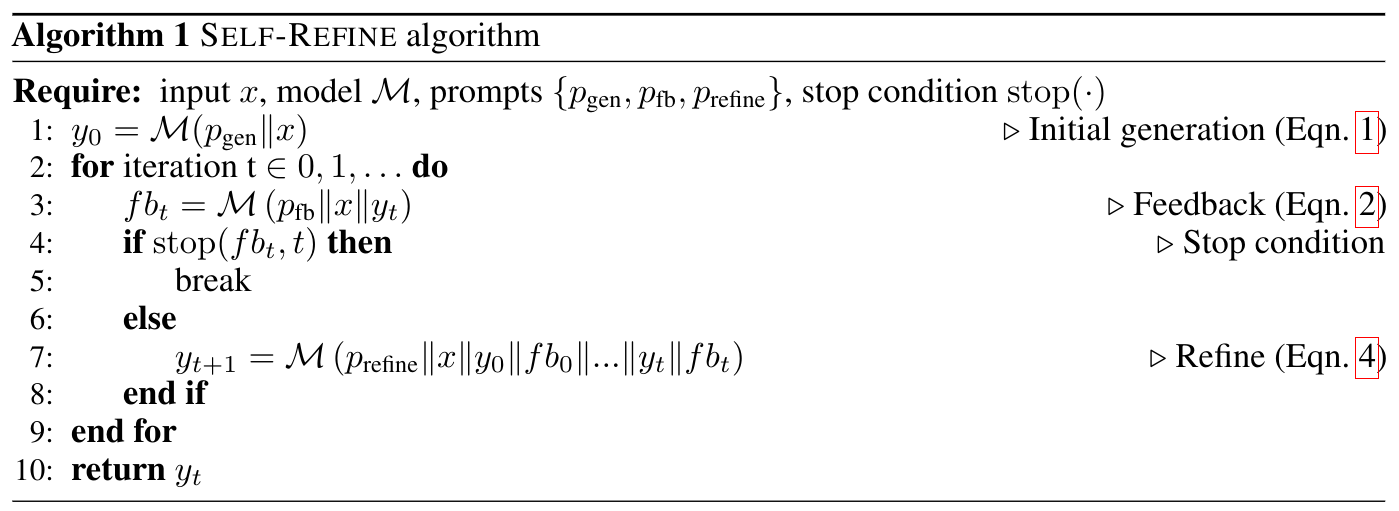}
\caption{The \lstinline{self-refine} algorithm as presented in \cite{madaan2023}. The refinement process can stop after a given number of iterations $t \in 0, 1, ...$ or when a stop condition $\textbf{stop}(fb_t, t)$ is met.}
\end{figure}

They tested this process on various tasks, including sentiment reversal, dialogue response, code optimization, readability improvement, math reasoning, acronym generation, and constrained generation. The models used were \lstinline{GPT-3.5}, \lstinline{GPT-4}, and \lstinline{ChatGPT}; interestingly, the latter performed comparable to the other options \footnote{Results can be found in Table 1 in \cite{madaan2023}.}.

To assess the correctness or quality of the generated outputs, the authors used an average of different scores:
\begin{itemize}
\item For all tasks, a score given by \lstinline{GPT-4}, used as a proxy for human preference;
\item For dialogue response generation, code readability improvement, sentiment reversal, and acronym generation, a score derived from a blind human A/B evaluation on a subset of outputs, selecting the preferred output;
\item For other tasks, automated metrics from prior work: specifically, the percentage solve rate for math reasoning, the percentage of programs optimized for code optimization, and the coverage percentage for constrained generation.
\end{itemize}

Another notable finding from their results is that the improvements in math reasoning using their approach were minimal or nonexistent, while for more qualitative tasks, their method had significant beneficial effects.\\

This is due to the qualitative approach of the answer improvement. Their work was not focused on producing a more accurate answer, while on refining the initial attempt. None of their prompt actually aims at verifying the correctness of the output, what their goal is instead to achieve some \textit{desirable property} (e.g. safety, clearness, efficiency) that the first try may lack\footnote{Refer to appendix A of \cite{madaan2023} for further details on the prompts used to answer, to give feedback and finally to refine the output, for each task considered.}. This explanation is also the reason for which \lstinline{ChatGPT} does not underperform more capable models: this task does not focus on improving the answer with relevant knowledge or meaningful reasoning, thus a conversational model could still behave properly.

\subsection{\lstinline{SELF-CORRECTION} algorithm} \label{self-corr}

As in \lstinline{SELF-REFINE}, Welleck et al. \cite{welleck2022} proposed \lstinline{SELF-CORRECTOR}, a method that decouples an imperfect base generator (an off-the-shelf language model or supervised sequence-to-sequence model) from a separate corrector that learns to iteratively correct imperfect generations. 

Powerful generation models often meet most of the task requirements, yet miss a few (e.g., omitting a subset of keywords), or generate incorrect hypotheses that nevertheless provide useful structure (e.g., a correct problem solving strategy with a missing step). However, after generating even a slightly sub-optimal sequence, the single-pass paradigm requires models to “start from scratch”, effectively discarding work already done. 

To avoid this scenario, it is possible to leverage the generation as a useful starting point to refine into a higher quality output. \\

A generation model is re-framed as a base \textit{generator}, which produces a reasonable initial hypothesis but does not need to solve the task in one pass, and a second module, the \textit{corrector}, trained to make up the difference between the hypothesis and an optimal solution. 

Note that neither the \textit{generator} nor the \textit{corrector} must solve the full task in one pass, and the \textit{corrector} can be applied multiple times to iteratively improve the output. \\

They tested the corrector approach on 3 diverse tasks: mathematical program synthesis, lexical constrained generation, and toxicity reduction.

\begin{figure}[H]
    \centering
    \includegraphics[width=0.95\linewidth]{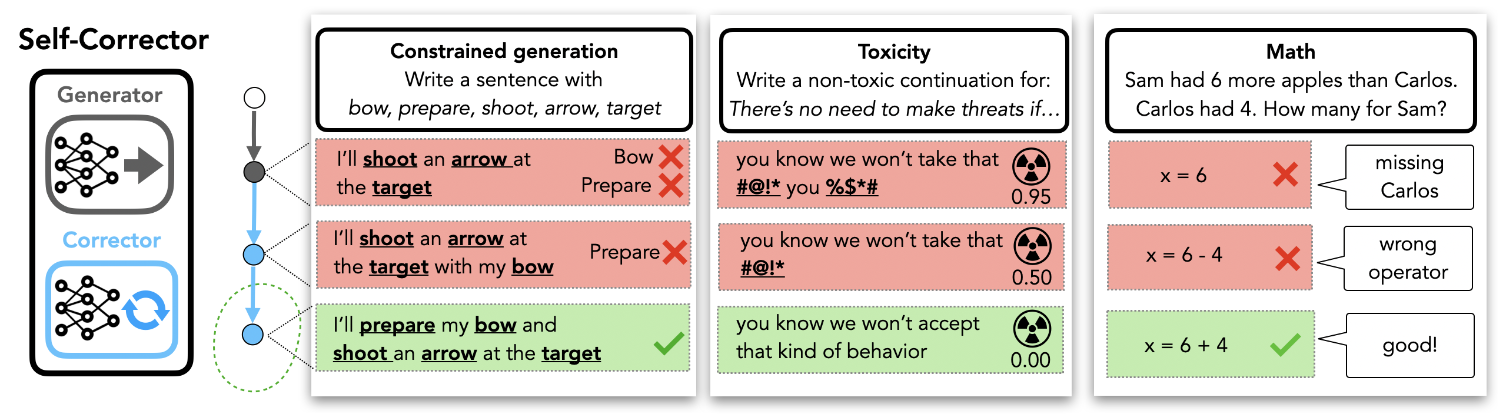}
    \caption{The \lstinline{SELF-CORRECTION} procedure, as presented in \cite{welleck2022}.}
\end{figure}

The trained \textit{corrector} model can even be applied to a larger \textit{generator} with similar performance to training a new corrector, showing that the sub-task of correction is transferable, even to stronger \textit{generators}. 

In addition to this, the \textit{corrector} module can be trained for different objectives, keeping the same \textit{generator}. This allows flexibility, modularity and composition, leading to a larger field of applications. \\

The corrector is trained by generating a set of hypotheses and relative corrections. The \textit{generator} firstly generates a lot of pairs that the naive version of the \textit{corrector} is asked to value; a set of value-improving pairs is formed (i.e. examples of mapping a hypothesis to a higher-valued correction); self-corrective learning selects \lstinline{(input, hypothesis, correction)} pairs to update the \textit{corrector} with. 

In detail, the \lstinline{(input, hypothesis, correction)} triplet is sampled proportional to its improvement in value, as well as the proximity between the \lstinline{hypothesis} and the \lstinline{correction}. \\

In algorithmic terms:

\begin{figure}[H]
    \centering
    \includegraphics[width=0.85\linewidth]{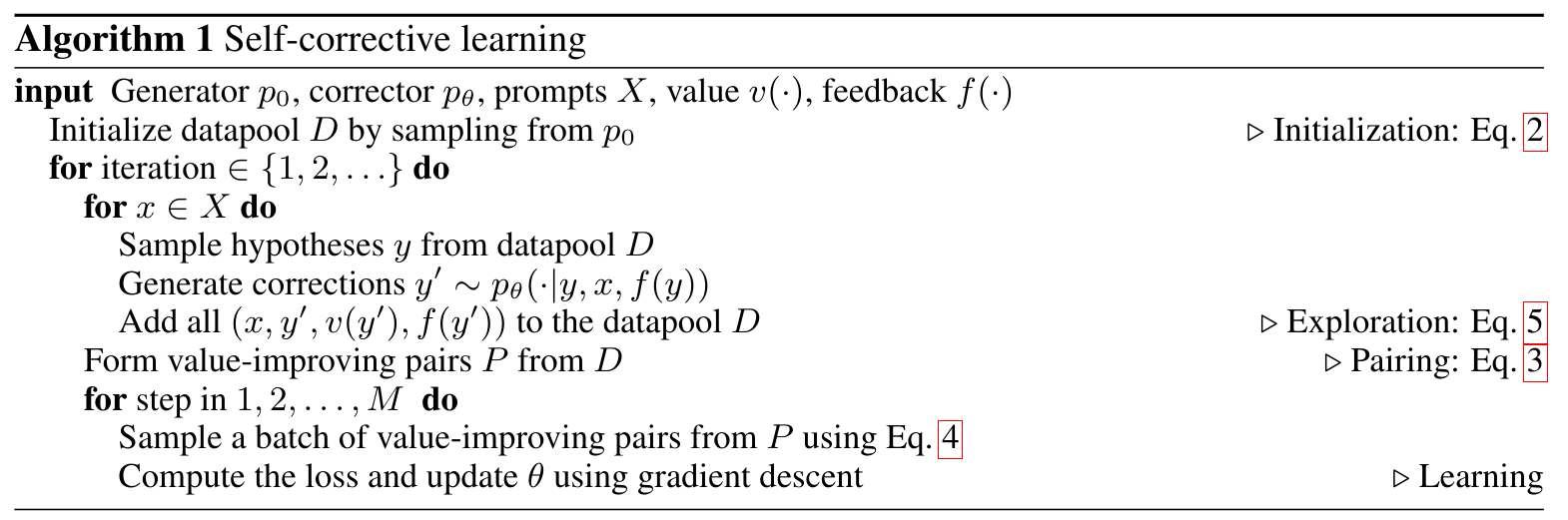}
    \caption{Notation and further details can be found in \cite{welleck2022}.}
\end{figure}

They achieved quite good results on arithmetic and mathematical datasets (MultiArith, Multitask), despite observing only a relative improvement on GSM8k. Results are observed using older versions of \lstinline{GPT} (\lstinline{GPT-3.5} was not yet released).  

\subsection{Reflexion} \label{reflexion}

Almost simultaneously, Shinn et al. \cite{shinn2023} developed \textit{Reflexion}, a modular, 3-units approach made by:

\begin{itemize}
\item an Actor model $M_a$, generating text and actions;
\item an Evaluator model $M_e$, scoring the output provided by $M_a$;
\item a Self-Reflection model $M_{sr}$, generating verbal reinforcement cues to assist the Actor in self-improvement.
\end{itemize}

The Actor is built upon a large language model that is specifically prompted to generate the necessary text and actions conditioned on the state observations. Analogous to traditional policy-based reinforcement learning setups, they sampled an action or generation $a_t$ from the current policy $\pi_\theta$ at time $t$ and receive an observation from the environment $o_t$. They also keep memory $mem$ as additional context.

This adaption was inspired by Brooks et al. \cite{brooks}, who suggest a policy iteration approach using in-context learning

As Actor models they considered both Chain-of-Thought and ReAct. \\

The Evaluator component of the Reflexion framework plays a crucial role in assessing the quality of the generated outputs produced by the Actor. It takes as input a generated trajectory and computes a reward score that reflects its performance within the given task context.

The issue is that defining effective value and reward functions that apply to semantic spaces is difficult. They experimented with different settings.\\

Given a sparse reward signal, such as a binary success status (success/fail), the current trajectory, and its persistent memory $mem$, the self-reflection model generates nuanced and specific feedback.
This feedback, which is more informative than scalar rewards, is then stored in the agent’s memory ($mem$). 

For instance, in a multi-step decision-making task, when the agent receives a failure signal, it can infer that a specific action $a_i$ led to subsequent incorrect actions $a_{i+1}$ and $a_{i+2}$. The agent can
then verbally state that it should have taken a different action, $a'_i$, which would have resulted in $a'_{i+1}$ and $a'_{i+2}$, and store this experience in its memory. 

In subsequent trials, the agent can leverage its past experiences to adapt its decision-making approach at time $t$ by choosing action $a'_i$. This iterative process of trial, error, self-reflection, and persisting memory enables the agent to rapidly improve its decision-making ability in various environments by utilizing informative feedback signals.\\

At inference time, the Actor conditions its decisions on short and long-term memory, similar to the way that humans remember fine-grain recent details while also recalling distilled important experiences from long-term memory. 

In the RL setup, the trajectory history serves as the short-term memory while outputs from the Self-Reflection model are stored in long-term memory.

\begin{figure}[H]
    \centering
    \includegraphics[width=0.85\linewidth]{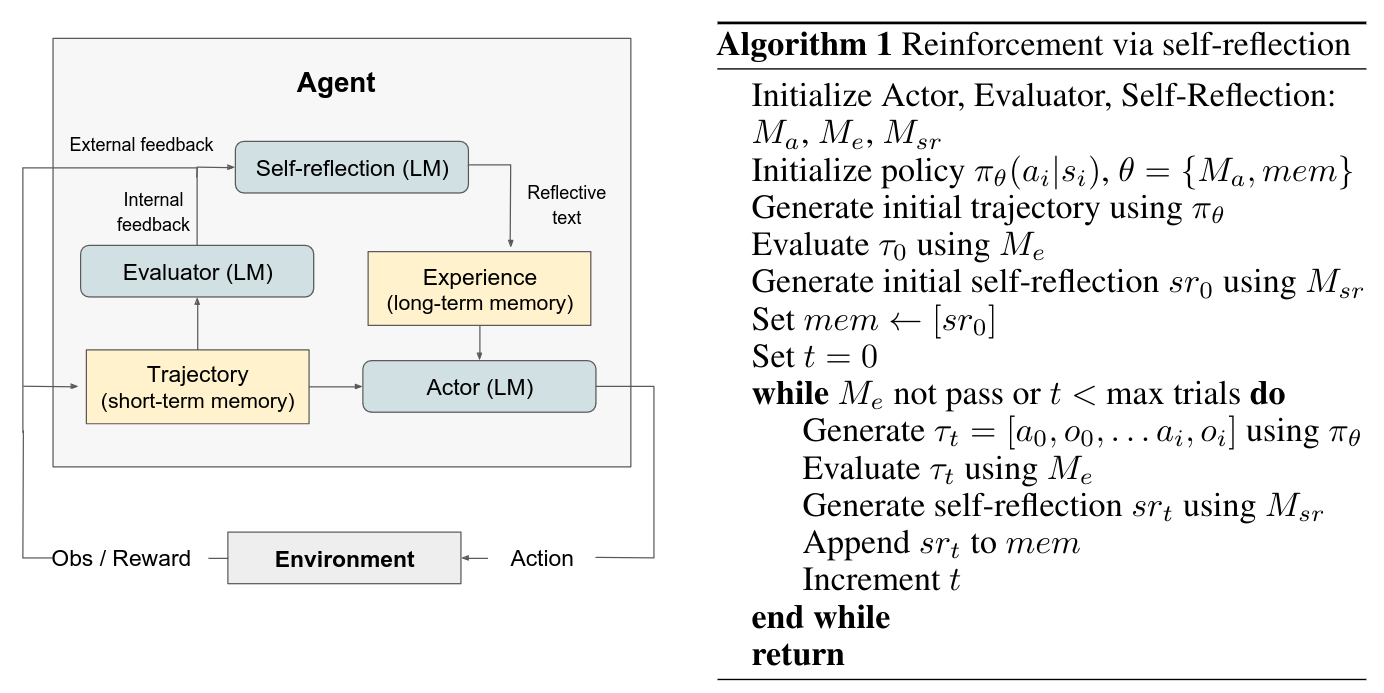}
    \caption{A diagram representing Reflexion (left) and the corresponding algorithm (right), taken from \cite{shinn2023}.}
\end{figure}

In the first trial, the Actor produces a trajectory $\tau_0$ by interacting with the environment. The Evaluator then produces a score $r_0$ which is computed as $r_t = M_e(\tau_0)$. $r_t$ is only a scalar reward for trial $t$ that improves as task-specific performance increases. 

After the first trial, to amplify $r_0$ to a feedback form
that can be used for improvement by an LLM, the Self-Reflection model analyzes the set of ${\tau_0, r0}$ to produce a summary $sr_0$ which is stored in the memory $mem$. 

$sr_t$ is a verbal experience feedback for trial $t$. The Actor, Evaluator, and Self-Reflection models work together through trials in a loop until the Evaluator deems $\tau_t$ to be correct. 

\section{Reasoning on the context} \label{reasoning-on-context}

\subsection{Rethinking with retrieval} \label{rr}

We have already spotted a light on the benefit provided by appending meaningful and relevant passages to the prompt before asking the model to generate its output. 

LLMs have been shown to generate incorrect supporting facts from time to time, even when they accurately capture the perspective needed to answer
a question. \\

This phenomenon highlights intrinsic issues in the way LLMs store and retrieve knowledge, including:
\begin{enumerate}
\item the presence of out-of-date, incorrect, or missing relevant knowledge in the pre-training corpus;
\item incorrect memorization of relevant knowledge during pre-training;
\item incorrect retrieval of relevant knowledge during the inference stage.
\end{enumerate} 

He et al. \cite{he2022} enter the debate with a post-processing approach called \textit{Rethinking with Retrieval} (RR) which retrieves relevant external knowledge
based on the decomposed reasoning steps obtained from the Chain-of-Thought (CoT)
prompting. 

The main advantages of this solution are that:
\begin{itemize}
\item it does not require additional training of fine-tuning;
\item it is not limited by the input length of LLMs.
\end{itemize}

\begin{figure}[H]
    \centering
    \includegraphics[width=0.85\linewidth]{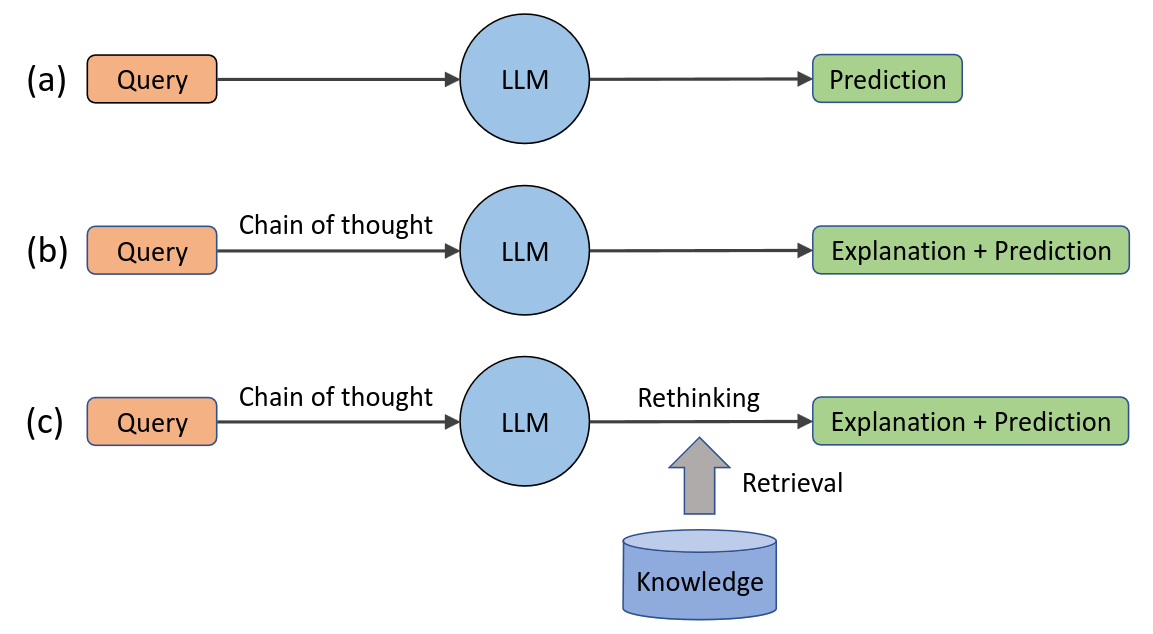}
    \caption{An overview of three approaches for using LLMs: (a) Standard prompting for generating a prediction in response to a query. 
    (b) Chain-of-Thought prompting for generating both an explanation and a prediction in response to a query. 
    (c) Rethinking with retrieval, our proposed approach for using the decomposed reasoning steps obtained from CoT prompting to retrieve relevant external knowledge for LLMs, leading to more faithful explanations and improved predictions in response to a query.}
\end{figure}

Their method began by using CoT prompting \cite{wei2022} to generate a diverse set of reasoning paths, as described in \cite{wang2022}. After that, they used each reasoning step in those paths to retrieve relevant external knowledge, which enables RR to provide more faithful explanations and more accurate predictions.\\

More formally, given a query $Q$, they used CoT prompting to generate a diverse set of reasoning paths $R_1, R_2, ..., R_N$ (where each reasoning path $R_i$ consists of an explanation $E_i$ followed by a prediction $P_i$).

Consequently, relevant knowledge $K_1, K_2, ..., K_M$ is retrieved from a suitable knowledge base $KB$ to support the explanation in each reasoning path, and select the prediction $\hat{P}$ that is most faithful to this knowledge.\\

For example:
\begin{lstlisting}
Q: Did Aristotle use a laptop?

R_1: Aristotle died in 2000. The first laptop was invented in 1980.
Thus, Aristotle used a laptop. So the answer is yes.
R_2: Aristotle died in 322BC. The first laptop was invented in 2000.
Thus, Aristotle did not use a laptop. So the answer is no.
R_3: Aristotle died in 322BC. The first laptop was invented in 1980.
Thus, Aristotle did not use a laptop. So the answer is no.

K_1: Aristotle (384-322 BC) was a Greek philosopher and polymath during the Classical period
in Ancient Greece. ...
K_2: The Epson HX-20, the first laptop computer, was invented in 1980. ...
\end{lstlisting}

The faithfulness of each reasoning path is evaluated using a function
$f_{KB}(R_i)$, which is based on relevant knowledge $K_1, K_2, ... , K_M$ retrieved from the knowledge base $KB$.

$\hat{P}$ is chosen as:
$$
\hat{P} = \underset{P_i \in {P_1, ..., P_N}}{\arg\max} \sum_{i=1}^N \mathbb{1}(P_i = P) f_{KB}(R_i)
$$

For instance, in the running example, given reasoning paths $R_1$, $R_2$, $R_3$ and the retrieved knowledge $K_1$, $K_2$, the above inference procedure would output the prediction \lstinline{So the answer is no}, as it is supported by both $R_2$ and $R_3$ and has a higher faithfulness score compared to the prediction \lstinline{So the answer is yes},
which is only supported by $R_1$.

\subsubsection{Better Multi-Hop Reasoners} \label{coc&coq}

Li et al. \cite{li2024} confirms the models' decreased performance in the presence of noisy contexts, but also highlight how they struggle with  multi-hop reasoning tasks. 

Their approach, \textit{Reasoning with Attributions}, prompts the model to supply attributions for each assertion during their reasoning. This is a strategy that mandates language models \textbf{to link the claims} made
during reasoning to specific sections of the provided context. This implicit requirement effectively decomposes a complex multi-hop question into two
more manageable tasks: 
\begin{itemize}
    \item Pinpointing pertinent information within the context;
    \item Constructing well-founded claims based on that information.
\end{itemize}

They adapted the CoT prompting to create two variants aligned with their (attribution-based) approach.

The first is \textit{Chain-of-Citations} (CoC), in which models are prompted to \textbf{reference citations} corresponding to each step of the reasoning chain.

The second, \textit{Chain-of-Quote} (CoQ), goes further by requiring models to \textbf{include direct quotations} from the cited material for each reasoning step. 

\begin{figure}[H]
    \centering
    \includegraphics[width=0.6\linewidth]{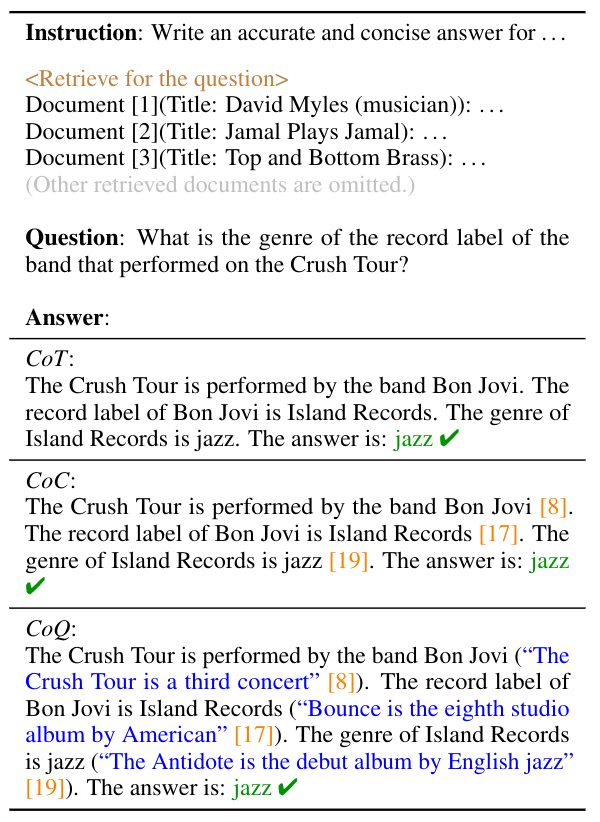}
    \caption{An example of CoT, CoC and CoQ taken from \cite{li2024}. Answers
are marked in green, citations are marked in orange and
quotes are marked in blue. }
\end{figure}

The findings suggest that both CoC and CoQ generally yield improvements over CoT, indicating that attribution-based reasoning enhances the precision and coherence of the models’ reasoning processes. CoQ appears to slightly underperform CoC, likely due to the increased complexity of producing exact quotations \cite{li2024}.

\subsubsection{MIRAGE} \label{pecore}

Ensuring the verifiability of model answers is a central task in the LLMs field of studies. Recently, researchers spotted that prompting the model to produce self-citations that ground the answers in the supporting context can help improve the answers' correctness. Grounding answers in the context is also an option that helps the users to check that the model is not \textit{right for the wrong reasons}, and that the reasoning chain that the model produces is not absurd. \\

But self-citation methods often struggle to match the required format, refer to non-existing sources and their faithfulness is very difficult to evaluate. 

Model Internals-based RAG Explanations (MIRAGE) \cite{mirage} extends the Plausibility Evaluation for Context Reliance (PECORE) framework \cite{pecore} for context-aware machine
translation. It detects context-sensitive answer tokens through saliency methods, pairing them with retrieved documents contributing to their prediction. 

Compared to self-citation, it allows a more fine-grained control on how the attribution stage is performed. 

\begin{figure}[H]
    \centering
    \includegraphics[width=0.65\linewidth]{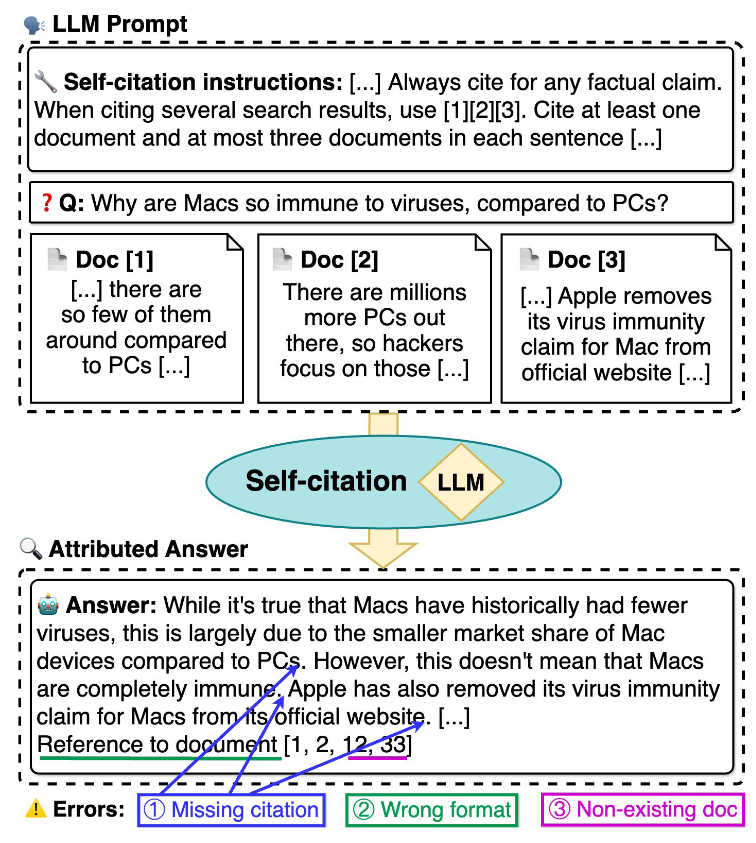}
    \caption{An example of self-citation weaknesses, taken from \cite{mirage}.}
\end{figure}

The context-sensitive tokens in the \textit{generated sequence} are spotted by measuring the shift in the model's predictive distribution caused by the addition of the input context. This shift is attributed to some tokens \textit{in the context}, found to be influential to the generated output.  

This approach has already been employed in machine traslation tasks \cite{pecore}, the authors expanded it to the RAG framework. 
Their idea is to search for a match between context-dependent tokens in the generated sequence and the retrieved documents that effectively contribute to their prediction. 
Finally, these paired elements are converted to citations.\\

To spot which \textit{generated} tokens are sensitive to the given context, the model needs to be prompted with a query $q$ and a context $C = \{c_1, ..., c_{|C|}\}$\footnote{Each $c_i$ can be a separate document, a separate paragraph, a separate phrase, ... It depends on the granularity of the employed RAG application.} in order to produce a sentence $y = (y_1, ..., y_n)$ as output. 

A contrastive metric $\hat{m}$ (e.g. KL-divergence) is used to quantify \textbf{at each generation step} the shift between:

\begin{itemize}
\item $P_{ctx}^i$: the model's predictive distribution at the $i$-th generation step with a prompt that includes the context;
\item $P_{no-ctx}^i$: the model's predictive distribution at the $i$-th generation step with a \textit{contextless} prompt. 
\end{itemize}

Thus, at each generation step, we use the metric $\hat{m}$ to compute a score: 
$$
m_i = \hat{m}(P_{ctx}^i, P_{no-ctx}^i), \quad \text{e.g.} \quad 
m_i = \text{KL}(P_{ctx}^i || P_{no-ctx}^i)
$$
The resulting (i.e. at the end of the generation process) scores $m = (m_1, ..., m_n)$ reflect the context sensitivity of each generated token. 

To extract the most relevant ones, we can use a choice function $s_{CTI}$ outputting whether each generated item is \textit{sufficiently context-sensitive} or not:
$$
CTI(q, c, y) = \{y_i \,| \, s_{CTI}(m_i) = 1,  \, \forall y_i \in y\}
$$
where $CTI(q, c, y)$ is an array containing all the relevant generated tokens $y_i$. \\

Once the context-sensitive tokens $y_i$ have been identified, we also make the model predict a contrastive alternative $y_i^{\setminus c}$ by excluding the context $C$ from the prompt but using the contextual generated prefix $y_{<i}$. \\

Now we have for each generation step a set of couples $(y_i, y_i^{\setminus c})$ representing the output produced with and without context. This first step has identified \textit{which generated tokens change when the context is injected} in the prompt. We refer to this step as \textit{Context-sensitive Tokens Identification }($CTI$), and consequently to the elements that this process finds as \textit{context-sensitive tokens} $y_i \in CTI(y)$. \\

The second step of this attribution procedure aims at discovering exactly \textit{which context tokens are the ones that impact the most on the model's outputs}.

We refer to this process as \textit{Contextual Cues Identification} ($CCI$), and consequently to these tokens as \textit{contextual cues} $c_j \in CCI(y_i)$.\\

In order to identify which are the most important context tokens, a \textit{contrastive feature attribution method} \cite{yin2022} can be applied: 
$$
a_i^j = \{ \nabla_j (p(y_i) - p(y_i^{\setminus c_j})), c_j \in C\}
$$

It substantially measures which context tokens $c_j$, if removed from the context, cause a bigger modification in the LLM's predictive function. $a_i^j$ identifies which context items $c_j \in C$ influence the prediction of $y_i$, accounting for the non-contextual completion $y_i^{\setminus c}$. \\

These scores are transformed into binary labels similarly as it has been done for $y_i$ in the $CTI$ step, i.e. by using a choice function $s_{CCI}$ outputting whether each \textit{contextual cue} is \textit{sufficiently influential} for the generated output or not:
$$
CCI(y_i) = \{ c_j \, | \, s_{CCI}(a_i^j) = \, \, \forall c_j \in C\}
$$

This process results in pairs of context-sensitive generated tokens ($CTI$) and their respective context tokens influencing their prediction ($CCI$):
$$
\{ (y_i, c_j), \, \forall y_i \in CTI(q, c, y), \, \forall c_j \in CCI(y_i) \}
$$

Note that both $s_{CTI}$ and $s_{CCI}$ contain implicitly a threshold which discriminates from what point onwards the tokens (both generated and in the context) are considered relevant. We can define them as:
$$
s_{CTI} = m_i \geq m^*, \, \quad s_{CCI} = a_i^j \geq a_i^*
$$

Practically, the authors suggested to set $m^* = \Bar{m} + \sigma \Bar{m}$, where $\Bar{m}$ and $\sigma \Bar{m}$ are respectively the average and standard deviation of $m_i$ scores for the given example. \\

To filter the attributed context tokens $c_j \in CCI(y_i)$, $a_i^*$ is either the Top-$K$ or Top-\% highest attribution value in $a_i$.\\

The final step proposed by the method consists in generating the citations. 
This step builds over the previous identification of relevant tokens in terms of generation.

We can construct the citation by selecting all the documents containing a \textit{contextual cues} ($CCI$) $c_j$ that are paired with a \textit{context-sensitive generated output token} ($CTI$) $y_i$:
$$
\text{MIRAGE}(y) = \bigcup_{y_i \in CTI(y)} \text{docid}(c_j), \, \forall c_j CCI(y_i)
$$

\newpage

\subsection{Literature connections with our method}

This long and extensive literature review presents multiple elements that will be helpful in presenting our experiments and in describing \textit{where} our method is placed. \\

The first part (\ref{transformers}) describes in detail the Transformer model's architecture and its modern improvements (\ref{improvements}). 

We consider these sections as preliminary to the description of the models we tested (\ref{tested-models}). With this introduction, we believe that many structural variations can be better understood. 
Tested models' details are essential to perform later analyses on (eventual) performance discrepancies. This should allow us to correctly spot the sources of different performance gaps among them, highlighting when they could be caused by our method or when they are external to it. \\ \\

Then, we begin an overview of proposed methods to enhance generation quality. Besides pre-training stages, the models can be fine-tuned (\ref{rag}) to align their predictions with certain domain-specific applications or to give them the ability to answer in a pre-determined way to the given prompt. 

A different approach has also been exploited to provide new information to the model, in order to update its knowledge or to provide additional one. The RAG approach (\ref{rag}) dynamically retrieves passages useful to answer to a question and appends them to the prompt. These appended additions are used to guide the model towards a grounded output. Further studies (\ref{re-rag}, \ref{s2a}) show how selecting the relevant information before appending it to the prompt can improve models' ability of exploiting context in the proper way, since it could be misleaded by noisy and irrelevant elements. A different approach (\ref{nli}) proposes to rephrase this framework in a NLI problem by transforming the prompt into a \textit{hypothesis} and the provided context as a \textit{premise} for it. \\

Fine-tuning, however, is not the only way of achieving tan alignment of model's behaviour with a desired one. It can be obtained via in-context learning (\ref{icl}), that simply provides "few-shots" examples appended to the prompt, showing the model how to perform a certain task. 

Inducing the model to behave in a pre-determined way as in-context learning (\ref{icl}) allows the prompt to show the model how to execute a task, and consequently it can also be used to present a \textit{reasoning chain} useful to solve the problem. Studies have shown that providing models with reasoning examples significantly improves the quality of their answers (\ref{cot-method}). This approach is known as Chain-of-Thought prompting. \\

The implicit idea of CoT is that the model is \textit{allowed} (or, more properly, \textit{prompted to}) take time and words to develop intermediate steps before choosing an answer. However, this is not the only option to produce \textit{not definitive, intermediate steps} before outputting the final answer. In the wake of the method showed in section (\ref{nli}) ranked the quality of the candidate outputs before choosing the correct one (proved to outperform large fine-tuned alternatives), self-refinement methods (\ref{self-refinement}) implement a certain number of corrections (\ref{self-corr}) or refinements (\ref{self-refine}, \ref{reflexion}) to achieve a better answer. \\

Finally, some studies experiment beneficial effects of grounding reasoning in relevant context have been reported in this section (\ref{reasoning-on-context}). 
Among these, we find interesting the Rethinking with Retrieval approach (\ref{rr}), that produces multiple reasoning chains and consequently selects as answer only the most grounded one and the Chain-of-Citations/Chain-of-Quotes method (\ref{coc&coq}), that induce the model to construct well-founded claims as reasoning chains by hinging them on context citations/quotes.\\

Our method builds on the joint efforts of the research community to which the creators of these methods belong.
Specifically, we owe them the idea of improving a first tentative answer as it is done in self-refinement methods (\ref{self-refinement}), although we do not consider our approach as a \textit{refining} approach, nor a \textit{correction} one. 

We also integrated the context in our approach, relying on the observations reported in (\ref{context}) and we also studied the influence that the original context has in the generated output, compared to its summarized version and to its filtered one. The filtering is carried out thanks to a slight variation of the MIRAGE (\ref{pecore}) method: since it is able to pair the generated outputs with the context elements that influenced them, we can discard irrelevant parts of the passages appended to the prompt. 

Finally, our method includes Chain-of-Thought steps in order to allow the model to reason on the correctness of a tentative answer and to eventually correct it. 

\newpage

\chapter{Data details} \label{data}

Our analysis aims to improve the robustness of LLMs' answers. We focus on studying their ability to correctly respond to general knowledge questions in various settings, such as standard prompting, CoT prompting, a RAG setup, and combinations of these approaches. 

Our method proposes to \textit{dialectically improving} the answer before outputting it. In order to run these tests, we have to rely on datasets such that: 
\begin{enumerate}
\item contain a set of questions and the correct answers to these questions;

\item append the relevant passages or documents that can be used to answer correctly to the question;

\item eventually provide also wrong answers to the question, presented as \textit{distractors}. 
\end{enumerate}

In addition to this, we aim to test the ability of LLMs once the correct relevant passages are provided. Extensive studies (reported in \ref{context}) have shown how models benefit from the addition of the relevant context to their prompts, thus we will build on this finding and try to further enhance performances. \\

What instead we find interesting to study is the impact of the work that the model has to perform on the sources. 
We aim to test cases in which instead it is necessary to merge multiple pieces of information before answering. This task can be referred with the name of \textit{Multi-Hop Question Answering} (MHQA) \cite{mhqa}. 

In broad terms, MHQA is the task of answering natural language questions that involve extracting and combining multiple pieces of information and doing multiple steps of reasoning. An example of a multi-hop question would be:\\
\lstinline{Who is the oldest candidate in the 2024 USA presidential election?}.\\
Answering the question would require to join many pieces of information: 

\begin{itemize}
\item \lstinline{What are the names of presidential candidates to the 2024 USA presidential election?}
    
\item \lstinline{What is the age of [candidate name]?}

\item \lstinline{What is the largest number between [age1], [age2], ...?}
\end{itemize}

The ability to answer multi-hop questions and perform multi step reasoning can significantly improve the utility of NLP systems. \\

\textit{Single-hop QA} often does not require any form of reasoning, limiting itself to summarizing or paraphrasing the content present in the source to produce a proper answer to the question. \\

\textit{Multi-hop QA} asks the model to perform a step further. 
An agent can be said to perform \textit{multi-hop reasoning} if it reaches one or more intermediate conclusions before deriving the final answer and each of the intermediate conclusions serves as a necessary premise for some other conclusion.
This sequence of intermediate conclusions, including the final answer, is called a reasoning chain and each step from one conclusion to the next can be referred to as a \textit{hop}. Humans can easily perform these multi-step reasoning in their everyday tasks, yet this is still a difficult task for machines.
We would like to improve LLMs' multi-hop abilities since they could be useful in many concrete applications. Queries given to current web search systems can often require multi-hop reasoning to reach the relevant documents; user satisfaction when using such systems can be greatly improved by utilizing multi-hop reasoning models; also conversations between humans and agents can be smoother and more informative if the latter can handle complex questions. \\

Our analysis starts from the choice of datasets. We are interested in studying how our method performs on HotpotQA and WikiHop datasets. 
As their names can clue, they both require the model to perform some \textit{multi-hop} reasoning steps before answering. 

\newpage

\section{HotpotQA}

\subsection{Dataset description}
HotpotQA is a question answering dataset collected on the English Wikipedia, containing about 113K crowd-sourced questions that are constructed to require the introduction paragraphs of two Wikipedia articles to answer. Each question in the dataset comes with the two gold paragraphs, as well as a list of sentences in these paragraphs that crowdworkers identify as supporting facts necessary to answer the question.\\

Yang et al. \cite{hotpot} constructed HotpotQA ensuring that 4 key features are guaranteed:

\begin{enumerate}
    \item the questions require finding and reasoning over multiple supporting documents to answer;
    \item the questions are diverse and not constrained to any pre-existing knowledge bases or knowledge schemas;
    \item sentence-level supporting facts required for reasoning are provided, allowing QA systems to reason with strong supervision and explain the predictions;
    \item a new type of factoid comparison question is introduced to test QA systems' ability to extract relevant facts and perform necessary comparisons.
\end{enumerate}

HotpotQA contains only 2-\textit{hop} questions formed using the first passages of documents from the English Wikipedia dump. The passages are chosen if they satisfy either of the two conditions:
\begin{itemize}
    \item There exists a hyperlink from the first document to the
second. The entity which forms the hyperlink is termed as the \lstinline{bridge} entity and the questions are termed as bridge questions.
\item The entities for those passages belong to the same category
(e.g. Michael Jordan and Kobe Bryant). These are specifically
sampled from 42 manually created lists. Such pairs are used
for creating \lstinline{comparison} questions \cite{mhqa}. 
\end{itemize}

An example of HotpotQA is the following: \label{comparison-example} 
\begin{lstlisting}
{'id': '5a7a06935542990198eaf050',
 'question': "Which magazine was started first Arthur's Magazine or First for Women?",
 'answer': "Arthur's Magazine",
 'type': 'comparison',
 'level': 'medium',
 'supporting_facts': {'title': ["Arthur's Magazine", 'First for Women'], 'sent_id': [0, 0]},
 'context': {
    'title': ['Radio City (Indian radio station)', "Arthur's Magazine", ... ],
    'sentences': [
        ["Radio City is India's first private FM radio station and was
        started on 3 July 2001.", ...] , ...
    ]
 }
}
\end{lstlisting}

Where: 
\begin{itemize}
    \item \lstinline{id} for the question-answer couple;
    \item \lstinline{question}, \lstinline{answer} are simply the question and the correct answer to the first;
    \item \lstinline{type} is the required type of reasoning on the context: \lstinline{comparison} and \lstinline{bridge} (details are provided below); \label{hotpotqa}
    \item \lstinline{level} is a human-labelled score based on how challenging the question is; possible values of it are \lstinline{easy}, \lstinline{medium} and \lstinline{hard};
    \item \lstinline{context} contains 10 documents, of which 8 distractors and 2 relevant ones. For each document/passage are provided:
    \begin{itemize}
        \item the \lstinline{title} of the Wikipedia paragraph;
        \item a vector of \lstinline{sentences} identified by the crowdworkers as relevant; 
    \end{itemize}

    \item \lstinline{supporting_facts} contains the \lstinline{title} of the 2 gold documents of the 10 provided. 
\end{itemize}

From now on, we will refer separately to the subset of dataset containing \lstinline{comparison} and \lstinline{bridge} type of questions. 
We do this distinction since we consider important to test separately these two different \textit{multi-hop} sub-tasks. While \lstinline{comparison} requires to extract (almost) the same information from passages referring to two different objects, \lstinline{bridge} tests the ability to merge multiple sources in a homogeneous view that allows to answer to the question. \\

\label{comparison-vs-bridge}
We show the difference between the two tasks using two examples:
\begin{itemize}
    \item \textbf{\lstinline{comparison}:} 
\begin{lstlisting}
Question: Which magazine was stated first, Arthur's Magazine or First for Women? 
Answer: [Arthur's Magazine]

Context: Arthur's Magazine (1844-1846) was an American literary periodical published 
in Philadelphia in the 19th century. Edited by T.S. Arthur, it featured work by Edgar A. 
Poe, J.H. Ingraham, Sarah Josepha Hale, Thomas G. Spear, and others. In May 1846 
it was merged into "Godey's Lady's Book". First for Women is a woman's magazine published 
by Bauer Media Group in the USA. The magazine was started in 1989. It is based in 
Englewood Cliffs, New Jersey. In 2011 the circulation of the magazine was 1,310,696 copies.
\end{lstlisting}
    \item \textbf{\lstinline{bridge}:}
\begin{lstlisting}
Question: The Oberoi family is part of a hotel company that has a head office in 
what city? 
Answer: [Delhi]

Context: The Oberoi family is an Indian family that is famous for its involvement in hotels, 
namely through The Oberoi Group. The Oberoi Group is a hotel company with its head office in 
Delhi. Founded in 1934, the company owns and/or operates 30+ luxury hotels and two river 
cruise ships in six countries, primarily under its Oberoi Hotels & Resorts and Trident 
Hotels brands.    
\end{lstlisting}
\end{itemize}

\subsection{Data processing}

The necessary transformation that the dataset had to meet was to create another plausible answer to the question. 

Due to the structure of the \lstinline{comparison} questions, we find natural to add only one plausible alternative, practically the other option given by the question. 

Consider for example:\\
\lstinline{Which magazine was stated first, Arthur's Magazine or First for Women?}\\
it is clear (even without knowing the correct answer) that the only two plausible options are 

\lstinline{Arthur's Magazine} and \lstinline{First for Women}.

We had at disposal the \lstinline{title} array (inside the \lstinline{context} attribute) and the \lstinline{title} of the only two relevant passages (inside the \lstinline{supporting_fact} attribute). Due to what we want to study, we chose to neglect a detailed analysis on the introduction of non-meaningful passages in our pipeline.

Thus, we kept only the \lstinline{sentences} items corresponding to the relevant documents, and merged them in a single text passage. 

We could not use the \lstinline{title} of relevant passages as couples of correct and wrong answers to the questions. While some cases allowed us to build an automatic procedure to do so, e.g. 

\begin{lstlisting}
Which magazine was stated first, Arthur's Magazine or First for Women?

[("Arthur's Magazine", 
  ["Arthur's Magazine (1844-1846) was an American literary periodical published in 
  Philadelphia in the 19th century.", ' Edited by T.S. Arthur, it featured work by Edgar A. 
  Poe, J.H. Ingraham, Sarah Josepha Hale, Thomas G. Spear, and others.', 
  'In May 1846 it was merged into "Godey\'s Lady\'s Book".']
  ), 
  ('First for Women', 
   ["First for Women is a woman's magazine published by Bauer Media Group in the USA.", 
   'The magazine was started in 1989.', ' It is based in Englewood Cliffs, New Jersey.', 
   'In 2011 the circulation of the magazine was 1,310,696 copies.']
  )
]
\end{lstlisting}

It is clear that in front of the question of the current example, it would be enough to extract 

\lstinline{Arthur's Magazine} and \lstinline{First for Women}. On the opposite, another example is the following: 
\begin{lstlisting}
Which band was founded first, Hole (the rock band that Courtney Love was a frontwoman 
of) or The Wolfhounds?

[('The Wolfhounds', 
  ['The Wolfhounds are an indie pop/noise pop band formed in Romford, UK in 1985 by Dave 
  Callahan, Paul Clark, Andy Golding, Andy Bolton and Frank Stebbing, and originally 
  active until 1990.', ' The band reformed in 2005 and continues to write, record and 
  play live, releasing new albums in 2014 and 2016.']
 ), 
 ('Courtney Love', 
  ['Courtney Michelle Love (born Courtney Michelle Harrison; July 9, 1964) is an American 
  singer, songwriter, actress, and visual artist.', 'Prolific in the punk and grunge 
  scenes of the 1990s, Love has enjoyed a career that spans four decades.', 'She rose to 
  prominence as the frontwoman of the alternative rock band Hole, which she formed in 1989.', 
  'Love has drawn public attention for her uninhibited live performances and 
  confrontational lyrics, as well as her highly publicized personal life following her 
  marriage to Kurt Cobain.']
 )
]
\end{lstlisting}

Despite being relevant to the answer, \lstinline{Courtney Love} would not be a plausible alternative answer since she is a singer, not a band. The correct alternative would be \lstinline{Hole}, which cannot be directly extracted from the context \lstinline{title}. \\

We used \lstinline{Phi-3-mini} to produce a plausible alternative to the correct answer to the question. We use greedy decoding with temperature set to 0 to sample an alternative option, limiting the maximum number of new tokens to 20\footnote{This is because of the answer style in HotpotQA, very synthetic and essential. We did not need the model to produce an output and justify it, but instead to extract from the context another realistic option.}. The prompt used to generate the output is one-shot and is the following:

\begin{lstlisting}
def produce_prompt(question, correct, source):
    
    user_content = "Question: " + question + "\n Correct answer: " +  correct + \
                   "\n Context: " + source + "\n\n Assistant:"

    messages = [ 
        {"role": "system", "content": """
        You are a helpful AI assistant. You are given a question and the correct answer to it. 
        Given the context, you have to provide a wrong, yet realistic, alternative answer to 
        the same question given the context. 
        Output a synthetic answer in the same style as the correct answer. 
        
        For example: 

        Question: Which magazine was started first Arthur's Magazine or First for Women?
        Correct answer: Arthur's Magazine
        Context: Arthur's Magazine (1844-1846) was an American literary periodical published in 
        Philadelphia in the 19th century. Edited by T.S. Arthur, it featured work by Edgar A. 
        Poe, J.H. Ingraham, Sarah Josepha Hale, Thomas G. Spear, and others. In May 1846 it was 
        merged into Godey's Lady's Book. First for Women is a woman's magazine published 
        by Bauer Media Group in the USA. The magazine was started in 1989. 
        It is based in Englewood Cliffs, New Jersey. In 2011 the circulation of the magazine 
        was 1,310,696 copies.

        Assistant: First for Women
        """}, 
        
        {"role": "user", "content": "Now to the same for this problem: " + user_content},
    ] 
    return messages
\end{lstlisting}

This function allows to append to the prompt an example and then to ask the model to perform a similar task for a new triplet of question, correct answer and context. \\

In the \lstinline{comparison} subset, questions are also presented in a different style, that does not explicitly refer to multiple options, for example:\\
\lstinline{750 7th Avenue and 101 Park Avenue, are located in which city?}\\
In this example, both the correct answer and the alternative do not stand out in the question. They need to be extracted from the context and the correct answer instead. 

The prompt presented above still works quite well for this type of questions, even though the context does not explicitly mentions any plausible alternative to \lstinline{New York City}:
\begin{lstlisting}
[('101 Park Avenue', 
  ['101 Park Avenue is a 629 ft tall skyscraper in New York City, New York.', 'It was completed 
  in 1979 to 1982 and has 49 floors.', 'Eli Attia Architects designed the building, which is 
  the 64th tallest in New York.']), 
 ('750 7th Avenue', 
  ['750 Seventh Avenue is a 615 ft (187m) tall Class-A office skyscraper in New York City.', 
  'It was completed in 1989 in the postmodern style and has 36 floors.', 'Kevin Roche John 
  Dinkeloo & Associates designed the building, and it is owned by Hines, a Texas based real 
  estate investment company.', "The building's continuous helix design, culminating in a 
  chimney-like extension, was caused by the New York City Building Code, which requires 
  setbacks.", "The 84 exterior column transfers exist because of the owner's requirement 
  for a column-free space.", 'It is tied with the New York Life Building for the 74th 
  tallest building in New York City.', 'It is also LEED certified.']
 )
]
\end{lstlisting}

In most of the examples, \lstinline{Phi-3-mini} is able to use its parametric knowledge to extract some plausible alternative. A hand-crafted correction is applied to cases in which the model produces clearly wrong options. For example, a plausible alternative to \lstinline{New York City} would not be \lstinline{Trieste}, since the latter does surely not possess \lstinline{avenues}.

Another option, e.g. \lstinline{San Francisco}, is more realistic and challenging to be checked (if the context is not used properly or neglected). \\

The last kind of alternatives that can be found is the basic \lstinline{yes} and \lstinline{no} questions. For this partition of \lstinline{comparison}, that can be easily selected by considering all the rows in which \lstinline{answer} is one between \lstinline{yes} and \lstinline{no}, the alternative is simply the opposite. This does not require the use of any Large Language Model. \\

Due to the need for a hand-crafted correction in some cases, we limited the dataset to the first 352 examples of the training set. We do not select the problems according to any different criterion and we checked that the models tested were not pre-trained or fine-tuned on this subset\footnote{See the Results chapter: the baseline performances are not good, both with and without the relevant passages appended to the prompt, thus it is improbable that the models were trained on that data.}.\\

For our experiments, this number of items is enough to assess whether we experience an improvement or not. Generating alternatives with \lstinline{Phi-3-mini} is not extremely expensive in computational terms, but requires also a hand-crafted correction sometimes. Also, we chose to keep the dataset balanced and not to add all the \lstinline{yes}/\lstinline{no} questions present in the original dataset. \\

When the same prompt is tested on the \lstinline{bridge} subsection of the dataset, we observe particularly good results, that do not need to be hand-crafted as a post-processing stage. \label{bridge}

Concrete examples of non-trivial alternatives created using \lstinline{Phi-3-mini} on \lstinline{bridge} subset:

\begin{lstlisting}
What U.S Highway gives access to Zilpo Road, and is also known as Midland Trail?

Zilpo Road is a National Forest Scenic Byway in the forested hills of eastern Kentucky, 
United States. The nine mile byway starts south of Morehead, Kentucky and can be accessed 
by U.S. Highway 60. The byway travels through the Daniel Boone National Forest and ends on 
the western shore of Cave Run Lake at the Zilpo Recreation Area. It follows FSR 918, which 
is a two lane paved road suitable for all motor vehicles and is usually open throughout the 
year. Morehead is a home rule-class city located along US 60 (the historic Midland Trail) and 
Interstate 64 in Rowan County, Kentucky, in the United States. It is the seat of its county. 
The population was 6,845 at the time of the 2010 U.S. census.
\end{lstlisting}

The correct option is \lstinline{U.S. 60} and the produced alternative is \lstinline{U.S. 50}. Of course the context does not mention \lstinline{U.S. 50}, since it is not relevant for the question, but produces a realistic alternative that (if the context is not given) could still be challenging to discard. \\

Another option is given by: 
\begin{lstlisting}
What nationality was James Henry Miller's wife?

James Henry Miller (25 January 1915 - 22 October 1989), better known by his stage name Ewan 
MacColl, was an English folk singer, songwriter, communist, labour activist, actor, poet, 
playwright and record producer. Margaret "Peggy" Seeger (born June 17, 1935) is an American 
folksinger. She is also well known in Britain, where she has lived for more than 30 years, 
and was married to the singer and songwriter Ewan MacColl until his death in 1989.
\end{lstlisting}

Even if the correct answer, \lstinline{American}, is quite trivial if the model is able to link \lstinline{Ewan MacColl} with its true name \lstinline{James Henry Miller}, the context could be quite deceiving. 

Both the first passage regarding \lstinline{James Henry Miller} and the second regarding \lstinline{Margaret "Peggy" Seeger} quote the fact that he is \lstinline{English} and that she well known in \lstinline{Britain} respectively, thus the alternative is really well grounded in the context.\\

Due to this more reliable pre-processing stage of \lstinline{Phi-3-mini}, we chose to select the first 1000 examples in the \lstinline{bridge} subset of HotpotQA's training set. As for \lstinline{comparison}, this was the only selection criterion used to pick problems on which performing tests. \\

The reason why \lstinline{Phi-3-mini} seems more reliable on this second split could be found in the most various context provided by the \lstinline{bridge} subset. 

Consider for example the \lstinline{New York City} example: since the comparison consists in spotting the common elements in both their passages, presumably they are already select to discard off topic additional information. In \lstinline{bridge} questions instead, context is structurally composed by passages containing a part of information that is relevant and must be merged with other sources and peddling ones. An example is given by the \lstinline{U.S. highway} example, in which is reported the population of Rowan County according to 2010 U.S. census. 

\newpage

\section{WikiHop}

\subsection{Dataset description}

WikiHop is a part of a greater dataset called QAngaroo and proposed by Welbl et al. \cite{wikihop}. 

QAngaroo is a Reading Comprehension dataset focusing on \textit{multi-hop} inference. Several pieces of information often jointly imply another fact, thus a new fact is derived by combining facts via a chain of multiple steps.\\

In each sample, a query is given about a collection of documents. The goal is to identify the correct answer among a set of given type-consistent answer candidates\footnote{Masked versions of these two datasets are also available, but for our analyses we will use not-masked ones. Refer to \cite{wikihop} for further details.}. The candidates — including the correct answer — are mentioned in the documents. \\

An example of multi-hop question in WikiHop:
\begin{figure}[H]
    \centering
    \includegraphics[width=0.7\linewidth]{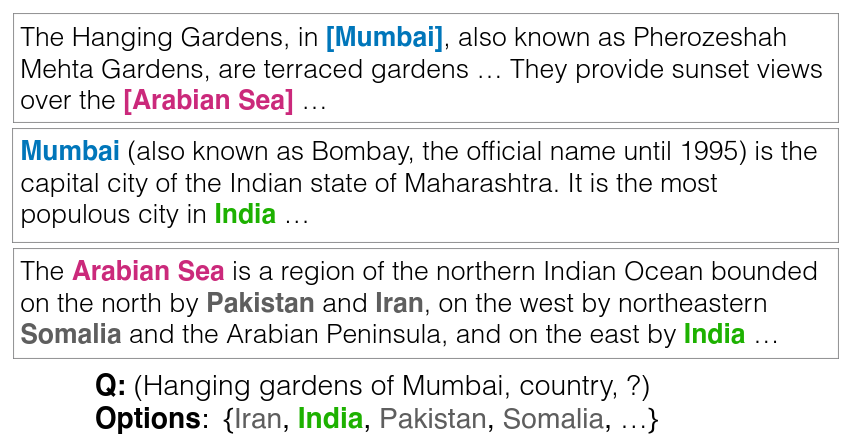}
    \caption{A sample from the WikiHop dataset where it is necessary to combine information spread across multiple documents to infer the correct answer, taken from \cite{wikihop}.}
\end{figure}

As previously mentioned, QAngaroo contains two distinct datasets:
\begin{itemize}
    \item WikiHop: contains open-domain and based on Wikipedia articles; the goal is to recover Wikidata information by hopping through documents. The example above shows the relevant documents leading to the correct answer for the query shown at the bottom. 
    \item MedHop: based on research paper abstracts from PubMed, the queries are about interactions between pairs of drugs. The correct answer has to be inferred by combining information from a chain of reactions of drugs and proteins.
\end{itemize}

We chose to focus on WikiHop since MedHop covers a limited number of topics and asks to output scientific acronyms of names or proitein labels: 

\begin{lstlisting}
Q: What interacts with DB00773?
Options: ["DB00072", "DB00294", "DB00338", "DB00341", "DB00588", "DB00820", "DB02546", 
          "DB02901", "DB04844"]
\end{lstlisting}

The options (i.e. the possible answers to the question) are extracted from the source documents, and only one among all the candidates is the correct one\footnote{In the current example, we have omitted the context since of much greater size of those of HotpotQA, you can check this by looking at the dataset \cite{wikihop}.}. \\

An example taken from WikiHop is the following:
\begin{lstlisting}
{"id": "WH_train_0",
 "query": "participant of juan rossell",
 "answer": "1996 summer olympics",
 "candidates": ["1996 summer olympics", "olympic games", "sport"],
 "supports": [
    "The 2004 Summer Olympic Games, officially known as the Games of the XXVIII
    Olympiad and commonly known as Athens 2004, was a premier international
    multi-sport event held in Athens, Greece, from 13 to 29 August 2004 with the
    motto \"Welcome Home.\" 10,625 athletes competed, some 600 more than
    expected, accompanied by 5,501 team officials from 201 countries. There were 
    301 medal events in 28 different sports. Athens 2004 marked the first time
    since the 1996 Summer Olympics that all countries with a National Olympic
    Committee were in attendance. 2004 marked the return of the games to the
    city where they began.",
    ...
  ]
}
\end{lstlisting}

Where:

\begin{itemize}
    \item \lstinline{id} identifies the sample;
    \item \lstinline{query} specifies the information that should be extracted from the texts;
    \item \lstinline{answer} is the correct answer to the query;
    \item \lstinline{candidates} is a list of answer candidates, each of which is mentioned in one of the \lstinline{supports} passages;
    \item \lstinline{supports} is a list of support documents. 
\end{itemize}

\subsection{Data processing}

Compared to HotpotQA, here we face quite opposite problems. 
The candidate options are provided and centered on the topics and the words present in the context, and each example shows a various number of candidate options (not just two, as before). This makes the dataset more challenging than the previous one.

Similarly as before, different \lstinline{supports} are merged together into an homogeneous source. \\

The great difference is that the \lstinline{query} is very essential and schematic, not appearing like a concrete question but instead as a sketch of it. 

Preliminary tests showed that models found difficult to read that style as a question, thus a pre-processing stage is necessary. 
We used again \lstinline{Phi-3-mini} in a one-shot setting to produce a question out of the schematic sketch of it. We used greedy decoding with temperature set to 0 to generate the question. The maximum number of new tokens allowed is 500, but never reached in practice. The prompt used to generate the question is the following: 
\begin{lstlisting}
def create_message(question, options):
    user_content = "Question: " + question + "\n Options: " + options + "\n\n Assistant:"

    messages = [
    {"role": "system", "content": """}
    You are a helpful AI assistant. You are asked to create a question out of a sketched question.

    Question: "occupation cao chong"
    Options: ['academic', 'builder', 'chancellor', 'classics', 'confucian scholar', 'designer', 
    'duke', 'emperor', 'engineer', 'engineering', 'father', 'founder', 'general', 'king', 
    'leader', 'major', 'mathematician', 'military', 'official', 'peasant', 'physicist', 
    'physics', 'politician', 'prior', 'rebel', 'research', 'ruler', 'science', 'script', 
    'social reformer', 'socialist', 'sovereign', 'taiwan']
    
    Assistant: "Which was the occupation of Cao Chong?"
    """
    },

    {"role": "user", "content": "Now do the same for this question: " + user_content},
    ]
    
    return messages    
\end{lstlisting}

This task is performed perfectly by the model; we chose not to append the context since it is enormously large and does not make sense in terms of question construction. The one-shot example is chosen to include a significant number of options, allowing the model to learn how to handle a large number of choices without becoming confused.\\

The first 1000 queries of the WikiHop dataset were processed as above and stored in a subset used for analysis purposes. No other selection criteria have been used to produce the WikiHop partition for our analyses. 

\newpage

\section{Datasets' summary statistics}

In the previous paragraphs we have already highlighted the fact that in HotpotQA the context is structured by only two \textit{hops} and the options for each question are always two. 

In addition to this, the difficulty level of the question (i.e. \lstinline{easy}, \lstinline{medium} and \lstinline{hard}) is neglected in this analysis, due to the fact that the same question could be more or less challenging given a different alternative option. \\

Some summary statistics regarding the number of words in the tested datasets are reported below.
We assume that the number of words in the \lstinline{query} but in particular in the \lstinline{context} is a proxy for the difficulty level of the \textit{multi-hop} reasoning task. The idea is that, even if the number of hops is fixed to two, the more verbose the \lstinline{context} is, the more information has to be merged before answering.

\begin{table}[H]
\centering
\begin{tabular}{|c|c|c|c|c|}
\hline
HotpotQA partition & min & mean & std & max \\ \hline
\multicolumn{5}{|c|}{\textbf{query}} \\ \hline
comparison & 5 & 11.6 & 3.9 & 42 \\ \hline
bridge & 5 & 19.7 & 10.5 & 94 \\ \hline

\multicolumn{5}{|c|}{\textbf{context}} \\ \hline
comparison & 28 & 118.0 & 47.8 & 295 \\ \hline
bridge & 40 & 138.7 & 53.0 & 502 \\ \hline
\end{tabular}
\end{table}

While the \lstinline{query} sizes (in terms of number of words) is almost the same for both the HotpotQA partitions, the \lstinline{context} ones are drastically different. \\

The \lstinline{query} sizes are slightly higher for the \lstinline{bridge} subset, probably caused by some extremely long questions (we can tell from the great standard deviation and high maximum value, while the minimum and the mean one are the same or only slightly higher). \\

The \lstinline{context} size of \lstinline{bridge} is slightly longer than the \lstinline{comparison} one. 

This is probably due to the same conclusions that we draw in section (\ref{bridge})
regarding the amount of additional information present inside \lstinline{bridge}'s \lstinline{context} attribute.\\

On the opposite, WikiHop has a variable number of options and a variable number of \textit{hops}. 

Below is reported the distribution of the number of \textit{hops}\footnote{The number of options are closely related to the number of \textit{hops}: since they are extracted from items mentioned in the context, the longer the passage, the higher the number of options provided. And when we have many \textit{hops}, the context is typically longer.}:
\begin{figure}[H]
    \centering
    \includegraphics[width=0.6\linewidth]{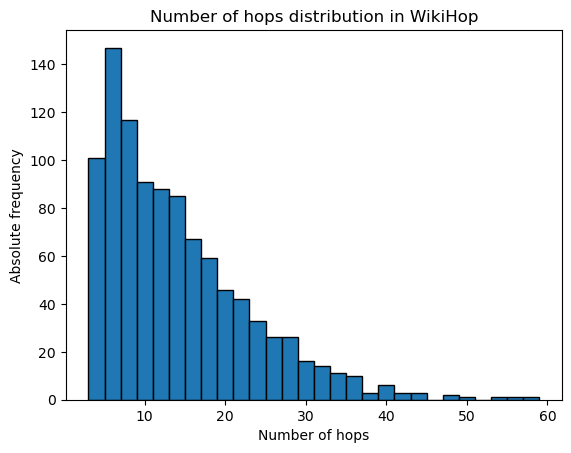}
\end{figure}
\label{wikihop-hops}

\newpage

\chapter{Methods} \label{methods}

\label{definitions}
Our work builds on other researchers' findings and ideas, despite trying to explore a new field of LLMs' abilities. 

In the literature review, we discussed how recent efforts in the NLP field focus on finding various solutions to enhance model generation and prediction abilities. These efforts often involve merging multiple methods or applying old ideas with new technologies. \\

We can imagine to construct a sharp distinction between the proposed approaches\footnote{Even though an effective clustering operation would be not appropriate due to the inherently mixing nature of modern machine learning technologies: consider that as only an aid for a clearer explanation.}:

\begin{itemize}
\item approaches that modify the model's weights in order to obtain the desired output (e.g. SFT, PEFT);
\item context-based approaches, in which context is used to boost model performances with relevant non-parametric, external knowledge (RAG, RE-RAG, NLI verifiers, S2A);
\item prompting approaches, that asks the model to perform a task in a certain way provided in a couple of examples (in-context learning), to think carefully before answering (CoT) and eventually to refine some previous answer given in input (self-refinement approaches).
\end{itemize}

The first one improves the output thanks to internalized, downstream-specific knowledge; the second one uses wisely the given context and relies on it to output a proper answer; the third one just tries to exploit latent knowledge present in the pre-trained model.\\

In literature, it is often used the term \textit{reasoning} for cases in which models exploit some CoT-like procedure to come up with the correct answer, while the product of the other two approaches is considered \textit{training} or \textit{retrieval}. \\

If we refer to the Oxford Dictionary, we find that \textbf{reasoning} is defined as: \textit{the process of thinking about things in a logical way; opinions and ideas that are based on logical thinking} while \textbf{thinking} is defined in a slightly different way: \textit{the process of thinking about something}.

Thus, \textbf{reasoning} appears to be more logically-based and decision-driven, while \textbf{thinking} can also be an unspoken, silent and internalized process. \\

This means that, given the black-box nature of LLMs, we cannot exclude that the process of activating the parametric knowledge stored in the network's weights isn't a form of thinking. We will not go deeper in arguing whether or not this process is linked with a (spontaneous) will of the model, since this would be outside the discussion's topics.\\

On the opposite, the \textbf{reasoning} process involves a stream of intermediate, logical connections and a final decision coming out from this process. This is why literature often refers to reasoning abilities when talking about Chain-of-Thought approaches. The great turning point with respect to other techniques is that, by explaining the logical process leading to an answer, we can clearly see that the model \textit{is reasoning} about it. 

Thus, the crucial component that defines a generation process as a reasoning process is its explicitness. 

For example, Rethinking with Retrieval (\ref{rr}) does not exactly reason about the possible solution. It samples many possible \textit{chains of justification} for a given answer and it discards those which are not supported by the retrieved content for the corresponding question. This works as a sort of \textit{pruning} of all not context-grounded generated chains. This approach helps improve the baseline answer (i.e., the standard prompting answer given without context). However, when compared to an output obtained via a self-consistency approach, the improvement is not as significant\footnote{A 4\% improvement on commonsense; 2\% on temporal and 1\% on tabular reasoning tasks \cite{he2022}.}.\\

Chain-of-Citations or Chain-of-Quotes approaches (\ref{coc&coq}) still use the context but through an opposite strategy. 

Producing citations or quotes such as Li et al. \cite{li2024} did helps in grounding each step of the reasoning chain in a relevant text passage, i.e. does quite the opposite as Rethinking with Retrieval. Binding to the definition of \textit{reasoning} given above, this is nearest to it than (\ref{rr}): each passage is supported by a portion of the knowledge base, and they are linked together in order to produce a plausible \textit{chain of justifications} leading to the final (and hopefully correct) answer. \\

A careful selection (by the model itself) of the context on which the following answer is built can be seen also as a form of asynchronous decision regarding the generation. Neglecting out-of-topic information or selecting just some relevant pieces of the given passages is proved to boost significantly the performances \cite{s2a}. Under a certain light, carrying out a first passage to skim irrelevant components for the given question (or prompt, more generally) is a form of \textit{thinking about something in order to make a decision}. \\

Considering this, it would be maybe excessive to consider RAG and RE-RAG context selections as a form of \textit{thinking} or \textit{reasoning}, since often the comparison is made using similarity scores or external ranker components (\ref{re-rag}). \\

A different conclusion has to be drawn for MIRAGE (\ref{pecore}), in which the generation process iteratively spots a light on which element of the context triggers the generation of the current token. While the prediction is exactly the same as the one we would obtain by just appending the context to the question, the \textit{contextual clues} spotted can be used for interpretability analyses. This is useful to recognize some hints of an underlying form of \textit{reasoning} that the model performs.\\ 

What instead follows from this definition of \textit{reasoning} is that translating a QA problem in a NLI task (\ref{nli}) and consequently determining whether or not the answer is entailed with the context is a form of (lateral) reasoning on the task. \\

The most familiar way (besides CoT approaches) that we recognize as reasoning is the one that we attributed to self-refinement approaches. 

Methods such as \lstinline{SELF-REFINE} (\ref{self-refine}), \lstinline{SELF-CORRECTION} (\ref{self-corr}) and Reflexion (\ref{reflexion}) all starts from the assumption of generating a first tentative answer and iteratively correcting it, until a certain \textit{level of satisfaction}\footnote{Different for each approach, typically bounded by a fixed number of iterative refinements or by a stopping criterion on the quality of the generated output.} is achieved. This appears very much alike the human learning process: we come out with a first attempt, check whether or not it is correct, eventually correct it. 

The three approaches have different ideas on their basis. \lstinline{SELF-CORRECTION} simply iteratively corrects the generated output using a specialized, pre-trained corrector module. 

Both \lstinline{SELF-REFINE} and Reflexion are a three-components approaches, having in common the idea of structuring the pipeline as first attempt $\rightarrow$ feedback/evaluation $\rightarrow$ refinement. 

While \lstinline{SELF-REFINE} proposes a "natural-language" refinement pipeline (i.e. the first attempt and the suggestion are provided to the model by appending them to the prompt), Reflexion updates model's hyperparameters through a Reinforcement Learning approach (\ref{reflexion}). 

\section{Reasoning or simply imitating?}

Many voices stand against the opinion that LLMs can reason or plan. In fact, in the opinion of many researchers, we should imagine these models as very good \textit{universal approximate retrievers} \cite{kambhampati2024}. \\

By that, they mean that the models are essentially just improved $n$-grams, pre-trained on a massive scale of web data and language corpora, hence truly capable when asked to complete a sentence in the proper way. But they are asbolutely incapable of performing any kind of reasoning that does not imply access to previously memorised knowlegde.\\

The big difference stands in what we refer to when we talk about \textit{thinking} or \textit{reasoning}. The previous paragraph assumes the Cambridge and Oxford definitions, while Kambhampati introduces the System 1 and System 2 architecture \cite{kahneman2011} in decision marking to support the idea that System 2 is something absolutely unknown to LLMs. 

Kambhampati defines the LLMs' parametric knowledge \textit{as a giant non-veridical memories akin to an external System 1} \cite{kambhampati2024}.

This is his proposed description: 

\begin{figure}[H]
    \centering
    \includegraphics[width=0.75\linewidth]{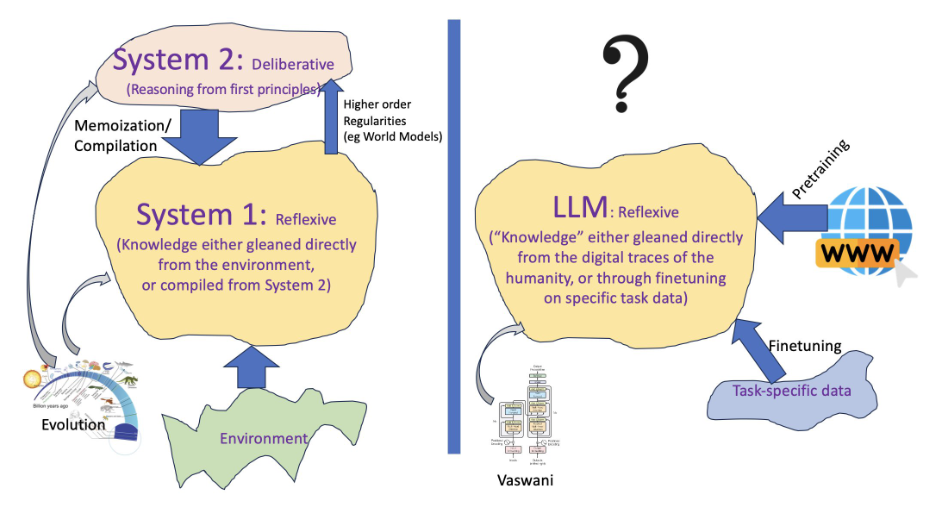}
    \caption{On the left, System 1 and System 2 as proposed by Kahneman \cite{kahneman2011}; on the right, the LLMs' pseudo System 1 (taken from \cite{kambhampati2024}).}
\end{figure}

The researcher and his team worked to test their strong assumptions against the supposed reasoning abilities of LLMs, and found that the pre-training corpus has a strong impact on the performances of those models in challenging reasoning tasks. 

They presented to the \lstinline{GPT-4} and \lstinline{GPT-3.5} many different shift ciphers problems, varying the shift value from 1 to 25. What they observed is that the models were very accurate on certain numbers of shifts, while completely out of clue in other cases:

\begin{figure}[H]
    \centering
    \includegraphics[width=1.01\linewidth]{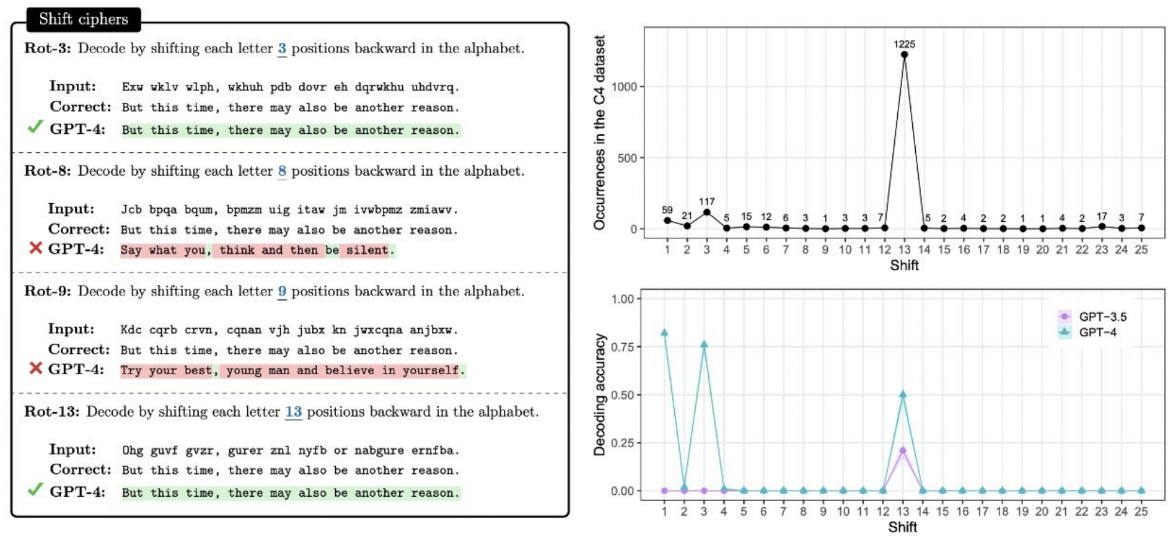}
    \caption{Tests of shift ciphers for different shift values, from \cite{icml_kambhampati}.}
\end{figure}

In fact, it is true that the \textit{deliberative} part of the decision system is not (at least up to now) a LLM feature. On the other hand, it is also true that in many situations it could be enough to imitate the behaviour observe in massive sources of data. \\

From now on, when we refer to reasoning, we will intend the definitions in \ref{definitions} or, equivalently, to the pseudo System 1 as proposed by Kambhampati in \cite{kambhampati2024}. 

\newpage

\section{A dialectic pipeline} \label{dialectic-pipeline}

Our method owes many ideas from previous works presented in the literature review section (\ref{literature}). 

We propose a three-step method composed of a thesis, an antithesis and a synthesis of the two previous steps.

\begin{figure}[H]
    \centering
    \includegraphics[width=0.85\linewidth]{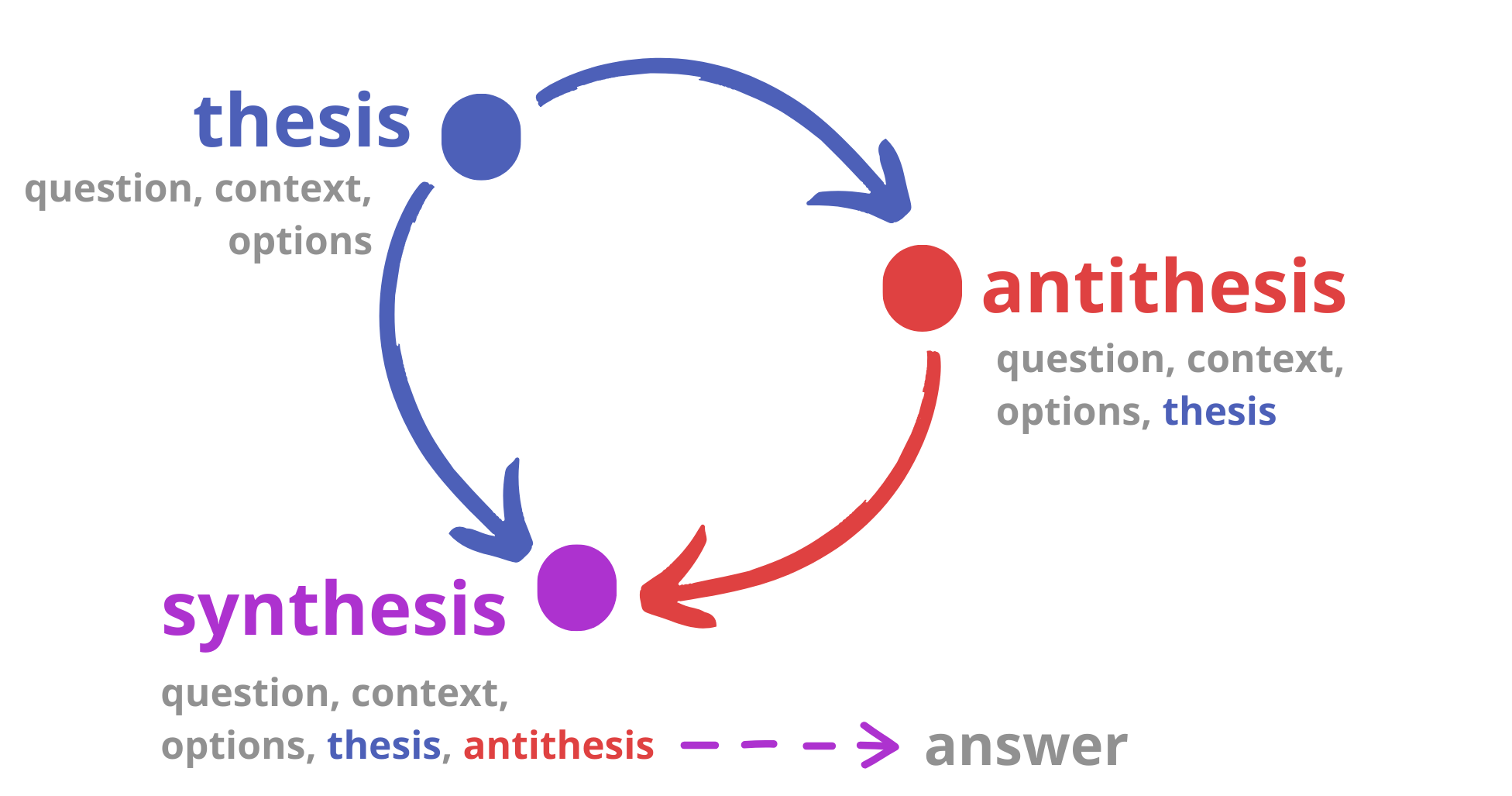}
    \caption{A visual representation of the proposed pipeline. Blue, red and purple represent the thesis, the antithesis and the synthesis steps respectively. Colored arrows highlight which element generated the subsequent item. The grey components are the structural components of the task (i.e. question, context and possible answer options). The answer is produced only after the synthesis step.}
\end{figure}

Our method does not perform a \textit{linear} refinement of the answer such as \lstinline{SELF-REFINE}, \lstinline{SELF-CORRECTION}, Reflexion or Rethinking with Retrieval approaches. 

The \textit{linear} refinement is performed when the first tentative answer (from now on, \textit{thesis}) is checked and eventually corrected by the same LLM in the \textit{antithesis} step. \\

But our idea is to further check the correctness of the \textit{antithesis}' suggestion by prompting the same LLM again. This new process receives as inputs both the \textit{thesis} and the \textit{antithesis} and performs a final pass on the problem before outputting the correct solution. We call this phase the \textit{synthesis} stage. \\

But why did we insert an extra step? \label{extra-step}

Differently from self-refinement methods summarized in (\ref{literature}), our method does not require any form of task-specific prompting\footnote{For example, \lstinline{SELF-REFINE} prompts the model to \lstinline{Show step-by-step solution} for math reasoning tasks while to \lstinline{Include keywords; maintain coherence} for constrained generation ones; we did not change the instruction prompt with respect to the specificity of multi-hop subtask that we considered.} or fine-tuning. This means that the correction (i.e. the \textit{antithesis}) is simply obtained by one-shot prompting with a fixed prompt structure describing the desired behaviour of that pipeline stage. Thus, an additional check of the proposed correct answer could be beneficial. 

In addition to this, considered that we have at least two options between which the choice has to be made, we can face two scenarios:
\begin{itemize}
    \item the \textit{thesis} and the \textit{antithesis} agree on which option is the correct answer to the question, thus the \textit{synthesis} receives a single option, already motivated by the \textit{antithesis} stage;

    \item the \textit{thesis} proposes an answer, the \textit{antithesis} suggests why another option is more proper (and why the \textit{thesis}' suggestion is wrong). 
\end{itemize}

The \textit{synthesis} considers the question, the options and the supporting context and decrees which is the most correct alternative answer, once listened to the \textit{thesis}' and \textit{antithesis}' opinions. \\

Discarding this last step could potentially harm the accuracy of the process. Consider for example the case in which the \textit{thesis} correctly predicts the answer, while the \textit{antithesis} step loses some information\footnote{Since we append to the context many items, some information may be lost in the attention process, particularly in presence of very long contexts. This is observed by Liu et al. \cite{liu2023}.} and outputs a detailed explanation on why the correct answer is a wrong option. 

By adding the synthesis step we force the model to compare different proposals and hopefully reach the most proper answer among them. 
Note that we do not bound the synthesis step to choose an option between the two proposed by the previous steps; when more candidates are present, the \textit{synthesis} step is left free to choose a third, unseen option given the context and the question. \\

Some experiments will be performed on the effectiveness of adding the \textit{synthesis} step. We will report the effectiveness of the pipeline up to the \textit{antithesis} stage and the complete one. \\

We prefer to define our approach as a \textit{dialectic} pipeline in place of a \textit{refinement} one because of the following reasons:
\begin{enumerate}
    \item we do not ask the steps to refine the output, instead we prompt the \textit{antithesis} and the \textit{synthesis} to check the previous steps' outputs and return their opinion on which is the most correct options;
    \item the three steps of the pipeline are essentially autonomous one from the other: for example, we can obtain an \textit{antithesis} out of an hand-crafted or even completely casual \textit{thesis}, without running the \textit{thesis} part before;
    \item due to each component's intrinsic autonomous nature, it is not guaranteed that we are able to observe an improvement in \textit{synthesis}' answers with respect to the \textit{thesis}' ones, since it is not a \textit{refinement} process;
    \item even though being autonomous actors in the pipeline, each actor adding an opinion on the previous ones, the entire pipeline can be seen as a model dialoguing with itself, in a sort of guided and \textit{disentangled}\footnote{By \textit{disentangled} we mean that the inner pipeline steps can be seen as intermediate ones in a bigger reasoning chain; however, the steps' contents are \textit{strongly entangled} due to their dialectic nature.} Chain-of-Thought. 
\end{enumerate}
\label{disentangled}

\section{Answering given the context} \label{context-preproc}

Although not strictly necessary, we chose to provide the relevant context to each step of the pipeline. 

This is due to the widely discussed reasons in the literature section (\ref{literature}) of the positive impact of appending the relevant context in QA tasks. \\

Our method aims at testing a pipeline that should be robust even when used with smaller or less widely trained models to assess the level of reliability of the pipeline with respect to different technical implementations. We tried different families of models and different sizes of the same model.
Consequently, relying only on the parametric knowledge stored in the model's parameters in order to provide the correct answer (and completely neglecting the context) could favour significantly models of greater size or trained on a bigger data mixture. \\

In all our experiments we will assume that the retrieval is performed perfectly: we will directly append to the prompt the \textit{gold} context (i.e. flagged as relevant) in the dataset. 

We will not introduce unnecessary noise by adding deceiving elements in order to test the robustness of the pipeline with respect to wrong context attribution. We plan to run these tests on successive work on this topic.\\

This choice is made because in this analysis we aim at evaluating the impact of the \textit{way} in which the context is presented to the model. We can face extremely long passages (e.g. the WikiHop dataset possesses many passages of more than 12.000 tokens) in which the relevant information can be found in a couple of sentences only, or on the opposite the passage could directly and explicitly mention the meaningful content in just a couple of terms.\\

We make the choice of using models of (relatively) small context length, for example \lstinline{Phi-3} has 4K and \lstinline{Gemma-2} has 8K context length. 

When we face a passage of greater size (e.g. as in WikiHop), we choose to \textbf{summarize} it in a smaller number of tokens before passing the triplet \lstinline{(question, options, context)} to the pipeline. We will prompt a model to make a summary of the given context knowing the question that will be asked. Further details will be discussed in (\ref{summ}).\\

In place of summarizing all the passages' contents, another option could be considered. In scenarios like that, a filtering process could be helpful in speeding up the inference phase and hopefully achieve better results, such as suggested in previous works like System 2 Attention \cite{s2a}. 

We preferred using a modified version of MIRAGE (\ref{pecore}) to carefully observe the context elements significant to the output's generation and \textbf{extract} all the sentences containing at least one of these elements. Hopefully, this model-internals based form of filtering is more accurate than both a summarization and a selection based on simple prompting as S2A (\ref{s2a}). We will describe more details in (\ref{my-pecore}). \\

Since these are all a sort of pre-processing actions that take place before running the pipeline, we can process differently the dataset and observe how this modification impacts on the pipeline. We will test the effectiveness of our pipeline:

\begin{itemize}
\item with the relevant (\textit{gold}) context appended;
\item with summarized context\footnote{Only on WikiHop, since HotpotQA passages are all under the maximum context length (4K); on WikiHop, we summarized \textit{all} the passages, even those under the limit of 4K tokens, for the sake of comparison.};
\item with filtered context, obtained using a modified version of \cite{mirage}. 

\end{itemize}

\begin{figure}[H]
    \centering
    \includegraphics[width=\linewidth]{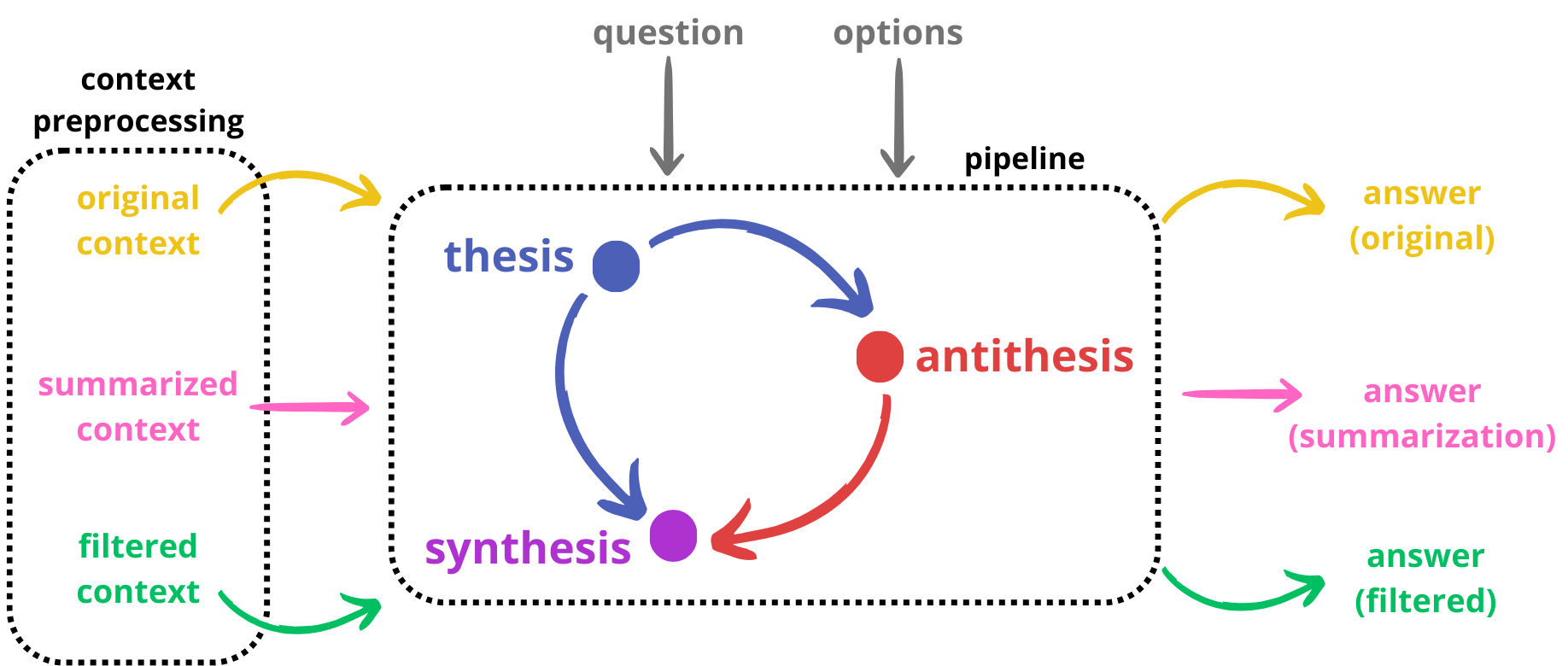}
    \caption{Schematic representation of the different experiments (each one flagged by a different color between yellow, orange and green) originated by different context pre-processing stages. We will compare the answers produced by these different settings, keeping the pipeline fixed.}
\end{figure}

\subsection{WikiHop context summarization} \label{summ}
As we mentioned above, we choose to summarize contexts that exceed the models' context lengths. In our experiments, this bound is fixed to 4K tokens due to \lstinline{Phi-3} models. 

While both the HotpotQA partitions have passages smaller than this value, WikiHop systematically exceeds the 4K limit. 

Consequently, we need to summarize the WikiHop's passages in order to keep them under the given threshold. To perform a uniform compression, we summarize both passages greater and smaller than 4K tokens. \\

We use still a \lstinline{Phi-3-mini} model, but this time we opt for its \lstinline{128k} context length version in order to be able to catch all the contexts, no matter their length. We prompt the pipeline to: 
\begin{lstlisting}
You are a helpful AI assistant. You are given a long document as context and you are asked to
produce a meaningful summary of it. Please summarize in less than 500 words.
\end{lstlisting}

Besides the requirement of not exceeding the 500 words, we let the model generate up to 1000 new tokens to allow a certain degree of flexibility and not requiring to truncate too harshly the summarization\footnote{A general rule of thumb is that 75 words approximately equals 100 tokens, thus for 500 words a maximum of 1000 tokens should suffice.}. The other generation parameters are kept fixed, with a temperature of 0.0 and no sampling involved. \\

We run the summarization process and we make sure that the model completes the task (i.e. finishes to summarize the original passage), and we discovered that this had happened only in certain cases. \lstinline{Phi-3} family of models flags with \lstinline{<|end|>} the end of generation, and this is present only in 330 rows of the global 1000.  

Instead of relying on a partially summarized source, we prefer to just consider these 330 completed summarizations and use this subset as a pre-processed context.

\subsection{MIRAGE context filtering} \label{my-pecore}

On the opposite, context filtering is performed both on WikiHop and on HotpotQA, since the extraction procedure is independent of the original context length. The goal is to study whether by selecting only the relevant sentences of the passage we are able to observe better \textit{synthesis}' outputs. \\

MIRAGE (\ref{pecore}) selects all the documents containing at least one \textit{contextual cue} and appends to the generated sentence a list of document identifiers (i.e. citations) from which the generation has been influenced. We use a similar approach, but in place of producing citations we aim at selecting only the relevant sentences inside the context in order to construct a filtered version of it. 

This selection will work as context source for our pipeline. \\

Still, we used \lstinline{Phi-3-mini} (this time we returned to the 4K version) and import it as an \lstinline{inseq} model\footnote{A Pytorch-based hackable toolkit to allow access to common post-hoc interpretability analyses of sequence generation models \cite{inseq}.} together with the model's tokenizer:
\begin{lstlisting}
inseq_model = inseq.load_model("microsoft/Phi-3-mini-4k-instruct", "saliency")
tokenizer = AutoTokenizer.from_pretrained("microsoft/Phi-3-mini-4k-instruct")     
\end{lstlisting}

where \lstinline{"saliency"} is the chosen attribution method. \\

The core function asks the model to: 
\begin{itemize}

    \item tokenize the text passage using the model tokenizer;
    \item invoke PECoRe \cite{pecore}, i.e. the methods running inside MIRAGE to pair \textit{contextual cues} and their corresponding \textit{context-sensitive} generated tokens;
    \item return only the sentences containing at least one \textit{contextual cue} in one merged new passage.
    
\end{itemize}

Due to the different datasets' passages sizes, we retain improper to ask PECoRe to select $k$ passages and to keep $k$ fixed for all the datasets. Additionally, WikiHop shows a great variability of context sizes, thus a shared value of $k$ would be too low for certain problems and excessively for others. 

Thus, we construct an auxiliary function \lstinline{find_top_p(passage, p)} that takes the passage, tokenizes it and outputs the number of sentences corresponding to the top-$p$ percentile of the source. 

Subsequently, we invoke PECoRe with the \lstinline{invoke_pecore(passage, question, p)} function as detailed in Appendix \ref{appendix-a} and obtain the CCI scores (\ref{pecore}). 

We retain the $p$ highest values among these, corresponding to the top-$p$ most influential tokens. These are mapped to their corresponding sentences and another function, \lstinline{select_passages(passage, question, p, tokens)}, returns the \textit{influential} sentences only.

\begin{figure}[H]
    \centering
    \includegraphics[width=0.85\linewidth]{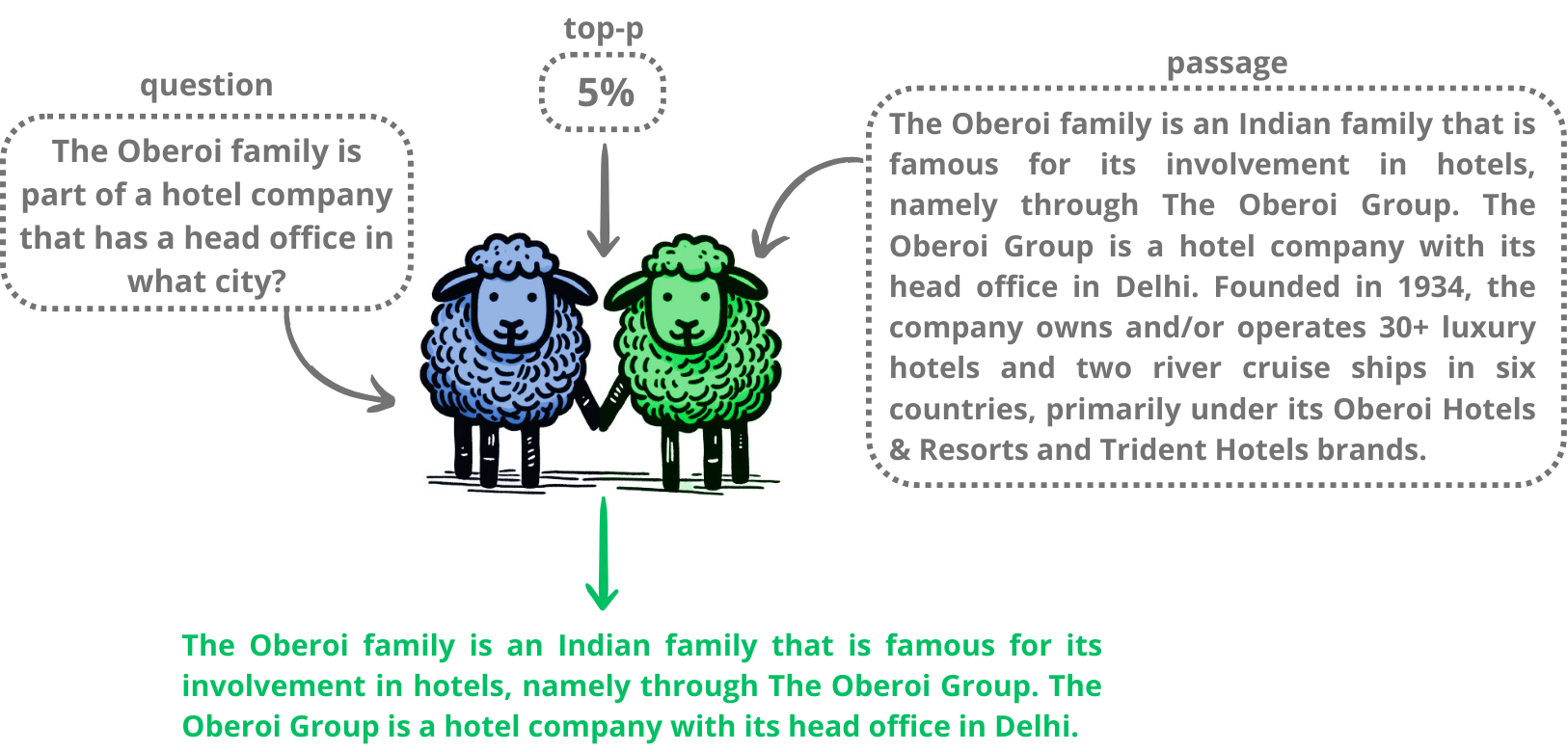}
    \caption{Concrete example taken from HotpotQA (\lstinline{bridge} partition) of top-5\% context selection performed by our modified version of MIRAGE. The two sheep indicate the MIRAGE process running PECoRe attribution method. The green passage is the PECoRe-filtered process.}
\end{figure}

While for HotpotQA we can take each row (they are all under the threshold of 4K tokens), for WikiHop we still have the same problem described in (\ref{summ}). Since the object of our analysis is the comparison between the answers obtained with different ways of considering the context, we do not care to apply this procedure to all the WikiHop items.

What we did instead is to select only the WikiHop rows characterized by less than 3500 tokens\footnote{500 tokens are left out for prompt instructions and few-shots examples.} in the \lstinline{supports} dataset attribute (i.e. the \textit{original context}). In this way, we select only 92 of the original 1000 rows. 
We consequently summarize and filter this subset only to make the three approaches comparable. \\

In order to study whether these results could be generalizable to the entire WikiHop dataset, we compare the performances of the summarized context in this subset (original context under 4K tokens) and the ones observed in the 330 rows previously summarized (\ref{summ}), that are assumed to be representative of WikiHop.

\section{Answers format: the \lstinline{guidance} framework} \label{guidance}

Even the best pre-trained and fine-tuned language model struggles in instruction following \cite{mu2024}, \cite{liu2023}, particularly open-source ones. Since in this analysis we do not make use of proprietary models, we wanted to get rid this \textit{format disobedience} and force the model to output exactly one of the candidate options, without \textit{going off on a tangent}.

Given an extensive output pointing out the correct option, we want to recognize it instead of neglecting everything that is not exactly the right-formatted answer. We do not retain reasonable to judge harshly a correct output due to format issues.\\

Consider the following problem: 
\label{halley}
\begin{lstlisting}
Question: Which year is Halley's Comet expected to return to the solar system?

Options: [2110, 2045, 2086, 2061]

Context: Astronomers have now linked the comet's appearances to observations dating back more 
than 2,000 years. Halley was last seen in Earth's skies in 1986 and was met in space by an 
international fleet of spacecraft. 
It performs a regular 76-year journey around the Sun.
\end{lstlisting}

The instruction-tuned model that we will use produce the following answers given an appropriate prompt\footnote{
\lstinline{You are a helpful AI assistant. You are given a question and the relevant context to answer it.} \lstinline{Answer briefly to the question with just one of the given options}
and then we appended the task. 

We sampled the output with temperature fixed to 0.2 and maximum number of new tokens set to 50.}:
\begin{itemize} \label{options}
    \item \textbf{\lstinline{Phi-3-mini}}: \begin{lstlisting}
    2061 <|end|>
    \end{lstlisting}
    \item \textbf{\lstinline{Phi-3-medium}}: \begin{lstlisting}
    The correct answer is 2061. Halley's Comet is expected to 
    return to the solar system in 2061, as it has a regular 
    76-year orbit around the Sun. <|end|>
    \end{lstlisting}
    \item \textbf{\lstinline{Gemma-2-2b-it}}: \begin{lstlisting}
    **2061** 
    \end{lstlisting}
    \item \textbf{\lstinline{Gemma-2-9b-it}}: \begin{lstlisting}
    2061
    <end_of_turn><eos>
    \end{lstlisting}
    \item \textbf{\lstinline{Llama-3.1-8b-instruct}}: \begin{lstlisting}
    2061.
    \end{lstlisting}
\end{itemize}

The outputted answers are all clearly correct, but their format is not always appropriate to perform a quick comparison between the correct answer (\lstinline{2061}) and the model's verbose output. \\

The models have different tokens flagging the end of generation (e.g. \lstinline{<|end|>} and \lstinline{<end_of_turn><eos>}, that have to be discarded) and also the punctuation is a problem: it is not trivial to distinguish when a dot flags the end of the sentence (and thus can be removed) or when it is part of the answer (e.g. \lstinline{2061 A.C.}). If we limit to post-process the output by removing the \textit{end-of-generation} tokens, only \lstinline{Phi-3-mini} and \lstinline{Gemma-2-9b-it} would (apparently) return the correct output, and this would be false.\\

To avoid relying on hand-constructed prompts (different from one model to another, thus without guarantees to correctly extract the suggested option), we exploited \textit{Structured Guided Generation} tools. 

\textit{Structured Guided Generation} (SGG) is a feature that allows users to constrain the generation of a large language model with a specified \textit{grammar}. It is used to generate text that follows a specific structure or uses a specific set of words or produce output in a specific format, e.g. to produce a valid JSON file as output, a function signature, a list of integers. \\

The \lstinline{guidance} framework does this by masking out certain all the tokens that do not belong to the pre-specified grammar (that can also be a set of options):

\begin{enumerate}
\item the model produces the logits \textbf{for each word in the vocabulary}, e.g.
    \begin{lstlisting}
    [0.1,   0.3,    0.2,    0.25,   0.15]
    \end{lstlisting}
    \item a mask is created for discarding all the words that do \textbf{not belong to the grammar}: 
    \begin{lstlisting}
    [-inf,   0.0,    -inf,    0.0,   -inf]
    \end{lstlisting}
    where \lstinline{-inf} is placed in correspondence of not-allowed words, \lstinline{0.0} otherwise;
\item the mask is added to the original logits, allowing the model to discard all the forbidden tokens in the \textit{sampling} stage:
    \begin{lstlisting}
    [-inf,   0.3,    -inf,    0.25,   -inf]
    \end{lstlisting}    
\end{enumerate}

There are also other tools for performing SGG, such as \lstinline{outlines} \cite{outlines} that exploit different technical implementations to determine how the model has to adapt its generation to a set of options. We opted for \lstinline{guidance} despite the fact that it does require to access models' logits, thus not being used for proprietary, closed-source models such as \lstinline{OpenAI}. We did not need a tool applicable to those kinds of models, thus we did not find necessary to exploit a different guiding generation strategy. \\

Practically speaking, we ask the model to generate the first, tentative answer by forcing the model to choose between the options as detailed in Appendix \ref{appendix-b}.

\section{Thesis}

Focusing again on the pipeline, we choose to ask the \textit{thesis} step to output the most proper candidate answer to the question. 

We prefer not to let the model produce a verbose version which, as we spotted in the previous section, can appear in many different and heterogeneous formats. This could be confusing for later stages of the pipeline, which must handle multiple expressions that semantically refer to the same option.

Think for example of a model that receives in input the answers presented in (\ref{options}): the long answer flagging \lstinline{2061} as correct option could be handled differently than a shorter answer saying the same thing\footnote{Maybe a more verbose opinion on which is the correct answer would be considered more reliable than just the option name.}.\\

First of all, we upload the model and its tokenizer using the \lstinline{guidance} framework and fix its temperature to 0. The prompting format is imposed by the framework and \textit{concatenates} the \textbf{model}, the \textbf{prompt} (composed by some generic description on how we want the task to be performed and by the problem) and the \textbf{options} (i.e. the only tokens of the model's vocabulary that will not be masked when generating the output).

The code used for the \textit{thesis}' generation can be found in Appendix \ref{appendix-c}.

\begin{figure}[H]
    \centering
    \includegraphics[width=0.95\linewidth]{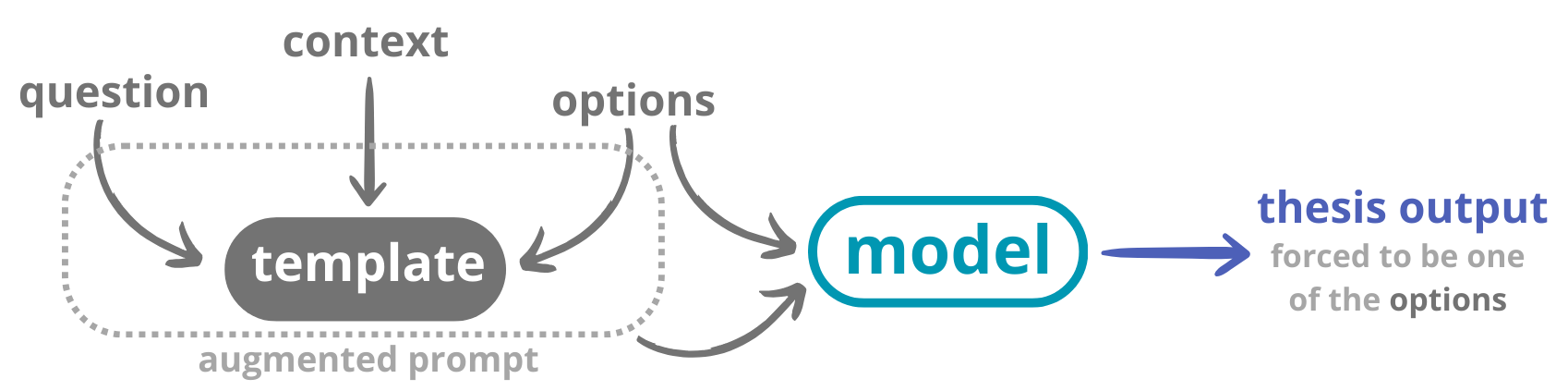}
    \caption{A schematic representation of the \lstinline{guidance} framework, when applied to the \textit{thesis}' generation.}
\end{figure}

\section{Antithesis}

\subsection{The importance of questioning previous statements}

The \textit{antithesis} step is the core component of the pipeline, but not because it is the most relevant step\footnote{as in \lstinline{SELF-REFINE}, \lstinline{SELF-CORRECTION} and Reflexion}. The \textit{synthesis} step also plays a role in modifying the answer, but the \textit{antithesis}' importance stands in the downstream impacts that a powerful correction has. \\

Think for example of how a teacher can correct an essay: he can only flag the error or he can explain precisely \textbf{why} the student's answer is not true. 

In the first scenario, the student could try to make a second guess and he could also fail again. In the second one, he will probably correct the first tentative answer with a more robust and factual one that builds on the teacher's suggestion. \\

Our task is structured slightly differently from a teacher-student dialectic: since we have the same LLM used as actor for all the three steps in the pipeline, we do not have a teacher model that checks whether or not the answer is correct. No step is "more experienced" than the others and consequently no step is more important or reliable than the others. \\

We should then modify a little the previous metaphor by imagining a student who is given a long multiple-choice test. He will firstly mark the choices that seem to him more proper for the given question. We will refer to this first pass as \textit{thesis}.\\

Then he will take a small break, refresh his mind, and look again to each question and to each answer, asking himself: "Is it really the most proper answer to the question? Is there a more proper option that I neglected on the first pass?".

He will consequently write some notes near each question explaining why he thinks that a certain answer is the correct one. Note that he is not forced to explain why the previously given answer is the correct one, he can also change its mind. Symmetrically, he should enforce its claim on the fact that the previously given answer is the correct one by spending a couple words on it. We will refer to this step as \textit{antithesis}.\\

Finally, the student will look again at the question, the options, the firstly marked one and the explanation of why he thought that it was correct or not. He will finally (in the \textit{synthesis} stage) opt for the initial option or for a suggested alternative, in a fresh pass of the same task.

\begin{figure}[H]
    \centering
    \includegraphics[width=0.8\linewidth]{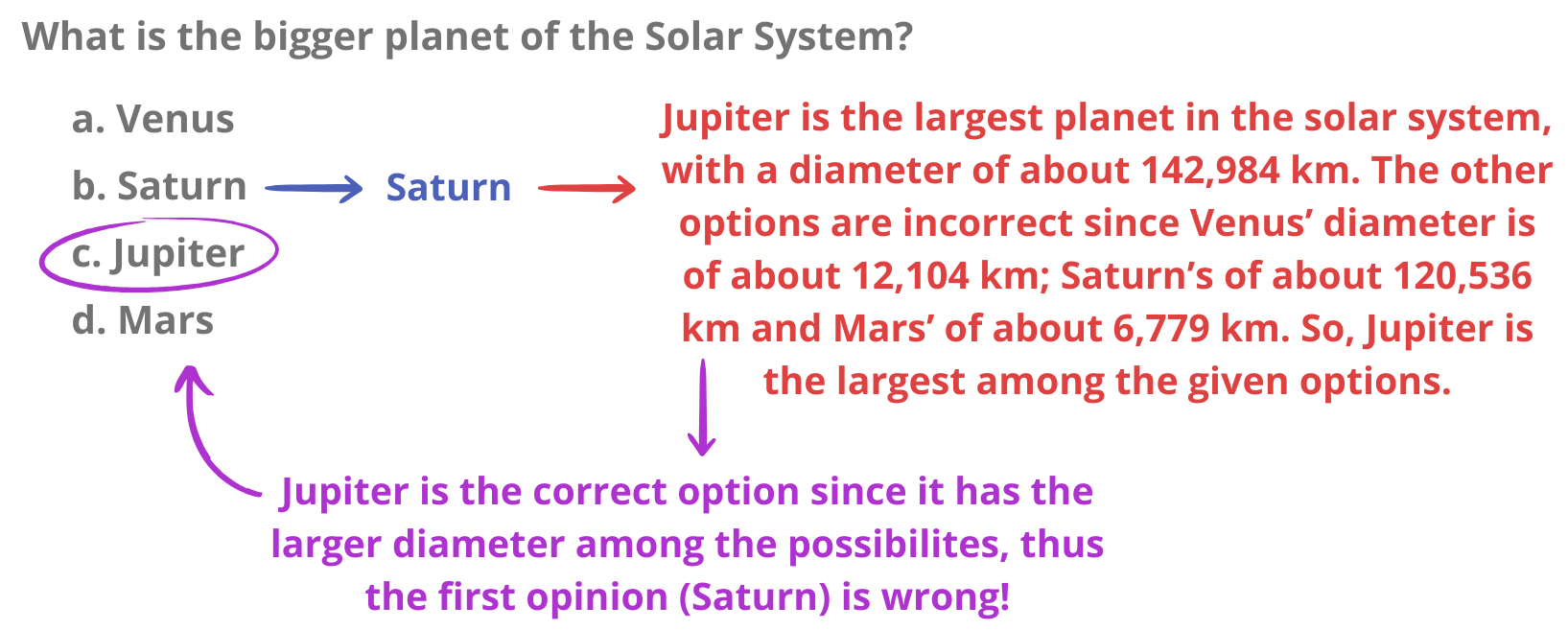}
    \caption{A simple example of the three steps performed: in blue, the \textit{thesis}; in red, the \textit{antithesis}; in purple, the \textit{synthesis}.}
    \label{test-example}
\end{figure}

Of course, he will have no guarantees that each step is performed correctly. If the second pass (the one in which he writes notes on which should be the correct option and why) is performed incorrectly, the overall process will suffer: the (eventual) critique is fundamental to change opinion on the correct answer. If we limited to this \textit{antithesis} step and we took its suggested output as the correct one, we would rely excessively on a single component that is prone to errors. By mitigating its effect as described in (\ref{extra-step}), we make sure not to make the model \textit{jumping to conclusions}. \\

The goal of our experiments was to make the \textit{antithesis} step as more accurate as possible. Of course we perform a final check, but we would like to avoid scenarios in which the \textit{thesis} is correct, the \textit{antithesis} is incorrect and the \textit{synthesis} agrees with the wrong suggestion.

This led us to the choice of providing \textbf{access to the relevant context} to each step of the pipeline. In this way, the student has always access to the book from which the topics of the multiple-choice exam are taken, and he just has to \textit{reason} on the sources and compare the options wisely. \\

Consider for example the Halley's Comet example provided in (\ref{halley}): even by accessing the source, the model needs to:

\begin{itemize}

    \item filter out irrelevant information;
    \item consider the last Halley's Comet passage (\lstinline{1986});
    \item understand that it performs a regular journey, so the \lstinline{76-year} information should be exploited;
    \item perform \lstinline{1986 + 76} and output the correct answer. 
    
\end{itemize}

The model capabilities have to go further than simply summarizing or rephrasing the context for multi-hop tasks like the ones we are going to test. On the other hand, if we do not provide the model the relevant information, the given answers could be incorrect. Consequently, we consider the addition of the relevant passages a beneficial step to elaborate a detailed description of which the correct option is during the \textit{antithesis} stage. \\

As we mentioned in (\ref{extra-step}), our method does not require any form of task-specific prompt or fine-tuning. We just append an example describing how we want the process to be performed in a one-shot setting. We chose not to increase the number of exemplars provided since the prompt is already burdened by the question, the options, the (often long) context and the \textit{thesis}' answer. Additionally, the one-shot prompt was enough to produce the desired behaviour, thus increasing it would be pointless. \\

The code used to generate the antithesis' suggestion can be found in Appendix \ref{appendix-d}.
Differently from \textit{thesis}, the \textit{antithesis} is not forced to be exactly one of the options through the use of the \lstinline{guidance} framework. What we did instead was to define a generation pipeline: 
\begin{lstlisting}
pipeline = pipeline("text-generation", model=model, tokenizer=tokenizer)
generation_args = {"max_new_tokens": 500, "return_full_text": False, "do_sample": False}
\end{lstlisting}

Consequently, the generated output will be in a discursive format. Consider for example this task, where \lstinline{Candidate answer} contains the \textit{thesis}' answer: 
\begin{lstlisting}
Question: Which magazine was started first, Arthur's Magazine or First for Women? 

Options: [Arthur's Magazine, First for Women]

Candidate answer: First for Women

Context: Arthur's Magazine (1844-1846) was an American literary periodical published in 
Philadelphia in the 19th century. Edited by T.S. Arthur, it featured work by Edgar A. 
Poe, J.H. Ingraham, Sarah Josepha Hale, Thomas G. Spear, and others. In May 1846 it was 
merged into Godey's Lady's Book. First for Women is a woman's magazine published 
by Bauer Media Group in the USA. The magazine was started in 1989. 
It is based in Englewood Cliffs, New Jersey. In 2011 the circulation of the magazine 
was 1,310,696 copies.
\end{lstlisting}

The \textit{antithesis} will produce\footnote{Real output, obtained using \lstinline{Phi-3-mini}.} the following opinion:
\begin{lstlisting}
 The correct answer is 'Arthur's Magazine' as it was started first in 1844, while 'First 
 for Women' was started later in 1989.    
\end{lstlisting}

As it is easy to spot, the \textit{antithesis} suggests the correct option by taking a second look to the context, even though the \textit{thesis} failed. 

\subsection{The influence of the given examples}

In the previous section we passed the one-shot example as an input parameter of \lstinline{create_message_antithesis()}. We now want to spot a light on the \textbf{impact that the one-shot example} has on the generated \textit{antitheses}. \\

We experimented with different settings, each one stressing different behaviours that the \textit{antithesis} could replicate. 

\begin{itemize}

\item The \textbf{baseline}: we simply provide the prompt with an example of how we want the model to make use of the \textit{thesis}' answer (referred to as \lstinline{candidate answer} inside the prompt): \label{baseline}
    \begin{lstlisting}
Question: What is the sun, a star or a planet?
Options: ['a star', 'a planet']
Candidate answer: a planet
Context: The Sun is the star at the center of the Solar System. It is a massive, 
         nearly perfect sphere of hot plasma, heated to incandescence by nuclear 
         fusion reactions in its core, radiating the energy from its surface mainly 
         as visible light and infrared radiation with 10% at ultraviolet energies.

Assistant: The correct answer should be 'a star' due to the fact that the context 
           explicitly say so. On the opposite, the context never mentions the fact 
           that the Sun could be a planet.  
    \end{lstlisting}
    Two things have to be spotted: first of all, the model does not firmly state which is the correct answer. We induce the model to choose a neutral \lstinline{should be} in place of a more confident expression. The second thing is that the answer does not mention explicitly the context, but instead it limits to motivate a certain choice by stating that this is due to the context. This baseline makes an implicit reference to the effective passage, does not quote it directly.

\item The \textbf{\lstinline{is} variant}: this one-shot example differs from the previous one just in the degree of confidence used to answer: the previous \lstinline{should be} is replaced with a more confident \lstinline{is}: \label{is-variant}
    \begin{lstlisting}
Question: What is the sun, a star or a planet?
[...]

Assistant: The correct answer is 'a star' due to the fact that the context 
           explicitly say so. On the opposite, the context never mentions the fact 
           that the Sun could be a planet.  
    \end{lstlisting}
    This aims at experimenting whether the level of confidence with which the model provides the \textit{antithesis} influences the \textit{synthesis}' decision on which option should be the correct one.

\item The \textbf{\lstinline{cot} variant}: uses a Chain-of-Thought inspired answer: \label{cot-variant}
\begin{lstlisting}
Question: What is the sun, a star or a planet?
[...]

Assistant: Let's consider the options and check whether or not they are correct. 
           The context clearly identifies the Sun as 'the star at the center of 
           the Solar System', thus 'a star' is probably the correct option. 
           On the opposite, 'a planet' is not mentioned in the context, thus it is 
           unlikely to be the correct option. Therefore, the correct option is 'a star'.
\end{lstlisting}
This approach considers each option and seeks eventual support of their correctness in the context. In the example, the relevant part of the context is also \textbf{quoted}. Finally, a conclusion is reached on which is the most proper answer among the candidates.

\item The \textbf{\lstinline{ctx} variant}: in the wake of the previous experiment, we tested also a different approach that is more context-grounded: \label{ctx-variant}
\begin{lstlisting}
Question: Jane's Addiction and Weeping Willows, play which genre of music?
Options: ['indie', 'rock']
Candidate answer: rock
Context: Weeping Willows is a Swedish indie rock group that started in 1995. Jane's 
         Addiction is an American rock band from Los Angeles, formed in 1985. 
         The band consists of Perry Farrell (vocals), Dave Navarro (guitar), Stephen 
         Perkins (drums) and Chris Chaney (bass).

Assistant: The context mentions that Weeping Willows is a 'Swedish indie rock group' and 
           Jane's Addiction is an 'American rock band'. Both bands are associated with the 
           'rock' genre, thus the correct answer is 'rock'.
\end{lstlisting}
This is still a CoT-like approach to the problem, since the model decomposes the task in intermediate steps; but differently from the \lstinline{cot} variant, this time the assistant is shown a different way of dealing with the problem. 

The model quotes the context's elements (thus the \lstinline{ctx} term to identify this variant) that are relevant for each compared item, and consequently decrees which is the correct answer. This is different from before, where partial conclusions were drawn analyzing the correctness of each option separately. With this variant we wanted to test whether explicitly quoting the relevant context before answering is beneficial to the pipeline.
\end{itemize}

The one-shot example provided is different from the previous one for two main reasons:
\begin{itemize}
    \item to assess the impact that different prompt examples have on the generated outputs (thus ensure that the results are not prompt-dependent);
    \item because the first example is relatively easy: the context does not mention both the candidate answers, but just the correct one. This could potentially cause the model to \textit{learn an incorrect pattern}\footnote{We used the term \textit{pattern} referring to a similar use of the word present in the Chain-of-Thought prompting paper \cite{wei2022}, stating that \textit{large language models can learn via a few examples with natural language data about the task (c.f. automatically learning the \textbf{patterns} underlying inputs and outputs via a large training dataset)}.}, since in the multi-hop tasks we face the context systematically mentions the options.
\end{itemize}

The task that will be tested on this pipeline is \textit{multi-hop} question answering, thus each option will be grounded in some elements of the context and the implicit challenge of the model is to actually merge multiple elements in order to provide a reasonable answer. 

\label{variations-overview}
\begin{figure}[H]
    \centering
    \includegraphics[width=0.8\linewidth]{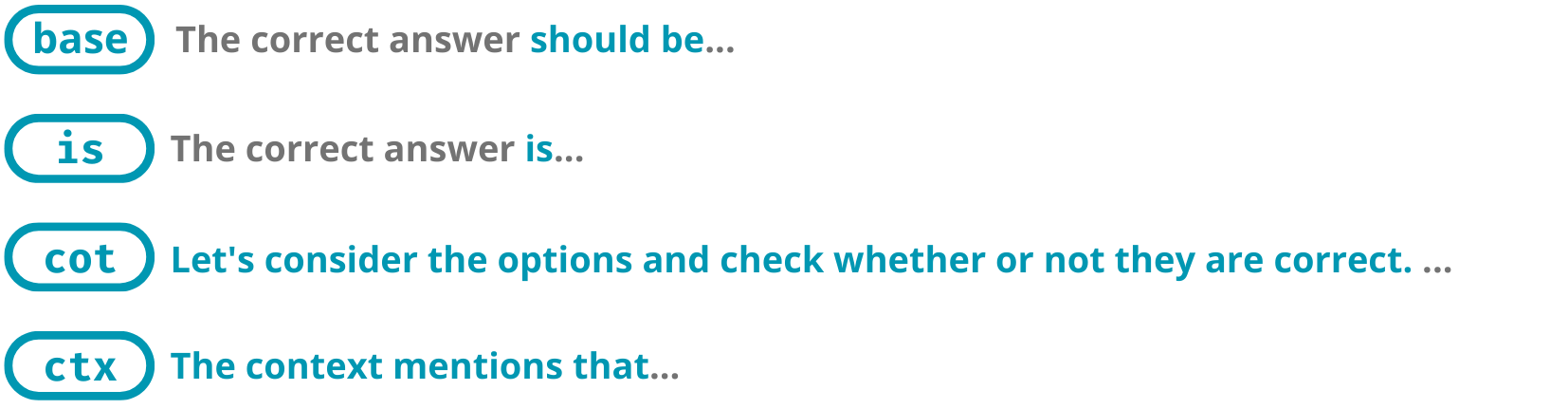}
    \caption{A brief overview of the \textit{antithesis} variations tested in our analysis.}
\end{figure}

\label{temp}
Here we show an example of how the antithesis output changes with respect to the king of one-shot example that is provided:
\begin{itemize}
    \item baseline:
    \begin{lstlisting}
The correct answer should be 'Arthur's Magazine' as it was started first in 1844, while 
'First for Women' was started later in 1989.
    \end{lstlisting}
    \item \lstinline{is} variant:
\begin{lstlisting}
The correct answer is 'Arthur's Magazine' as it was started first in 1844, while 'First 
for Women' was started later in 1989.     
\end{lstlisting}
    \item \lstinline{cot} variant:
\begin{lstlisting}
Let's analyze the options and the context provided. The context states that Arthur's 
Magazine was published from 1844 to 1846, while First for Women was started in 1989. 
Therefore, Arthur's Magazine was started first. The candidate answer 'First for Women' 
is incorrect. The correct option is 'Arthur's Magazine'.
\end{lstlisting}
    \item \lstinline{ctx} variant:
\begin{lstlisting}
The context states that Arthur's Magazine was published from 1844 to 1846, while First 
for Women was started in 1989. Therefore, Arthur's Magazine was started first. 
The correct answer is 'Arthur's Magazine'.    
\end{lstlisting}
\end{itemize}

As it is easy to notice, there is no significant difference between \lstinline{cot} and \lstinline{ctx}, since both the one-shot examples decompose the problem into a retrieval step and a conclusion step. The most remarkable difference (present also in different problems, not reported here) is that the \lstinline{cot} variant tends to remark the fact that the other option is wrong. \\

One last device we had was to change the style of the question in the one-shot example for datasets other than the \lstinline{comparison} fraction of HotpotQA. This is because, while all the \lstinline{comparison} questions are formulated in such a way as to explicitly report the options to choose from, the \lstinline{bridge} questions (i.e. the complementary fraction of HotpotQA) and WikiHop reflect a less "set" style:

\begin{itemize}
\item HotpotQA, \lstinline{comparison}:
    \begin{lstlisting}
Which magazine was started first, Arthur's Magazine or First for Women?
    \end{lstlisting}
\item HotpotQA, \lstinline{bridge}: 
    \begin{lstlisting}
The Oberoi family is part of a hotel company that has a head office in what city?
    \end{lstlisting}
\item WikiHop:
    \begin{lstlisting}
What language did John Osteen speak or write?        
    \end{lstlisting}
\end{itemize}

Consequently, we will replace 
\begin{lstlisting}
What is the sun, a star or a planet?
\end{lstlisting}
with a simpler
\begin{lstlisting}
What is the sun?
\end{lstlisting}.

On the opposite, \lstinline{ctx} is already more similar to the simpler question, thus is left unchanged.

\section{Synthesis} \label{synthesis}

The final step aims at merging in a reasoned answer the entire pipeline: by considering the first tentative answer (the \textit{thesis}) and the new opinion (the \textit{antithesis}) obtained by reconsidering the previous one, the last opinion (the \textit{synthesis}) has to decree which is the correct option. And now it should do this with more confidence and increased factuality. \\

As in the previous pipeline steps, the detailed code is reported in Appendix \ref{appendix-e}. 
It is important to remark that the \textit{synthesis}' prompt does not bound the model to choose between the option proposed by the \textit{thesis} or proposed by the \textit{antithesis}: the instruction given clearly states that the \textit{synthesis} stage has to choose between one of the \lstinline{available options}. This means that the \textit{synthesis} stage could potentially even opt for an unseen option. \\

For the baseline and the \lstinline{is} and \lstinline{cot} variants, the one-shot prompt is very similar: 
\begin{lstlisting}
Question: What is the sun, a star or a planet?/What is the sun?
Options: ['a star', 'a planet']
Candidate answer: a planet
Suggestion: {antithesis_answer}
        
Context: The Sun is the star at the center of the Solar System. It is a massive, nearly 
         perfect sphere of hot plasma, heated to incandescence by nuclear fusion reactions 
         in its core, radiating the energy from its surface mainly as visible light and 
         infrared radiation with 10% at ultraviolet energies.

Assistant: the correct option is 'a star', since the suggestion is grounded in the context, 
           even if the candidate answer does not agree.
\end{lstlisting}

where \lstinline{antithesis_answer} is the antithesis, presented in different formats with respect to the kind of variant (i.e. what follows the \lstinline{Assistant:} in the previous section's one-shot examples).\\

For the \lstinline{ctx} variant, we tried instead to use two-shot prompting in order to induce the model to learn how to deal with situations in which the \textit{thesis} and the \textit{antithesis} agree and situations in which they don't\footnote{The \lstinline{ctx} variation originates from the observed results and the fact that the in-context learning example is different is a conscious choice. Additionally, the following synthesis setting (different from the \lstinline{cot} one) is oriented to a study on different few-shot options.}:
\begin{lstlisting}
Question: Jane's Addiction and Weeping Willows, play which genre of music?
Options: ['indie', 'rock']
Candidate answer: rock
Suggestion: The context mentions that Weeping Willows is a 'Swedish indie rock group' and 
            Jane's Addiction is an 'American rock band'. Both bands are associated with the 
            'rock' genre, thus the correct answer is 'rock'.
Context: Weeping Willows is a Swedish indie rock group that started in 1995. Jane's 
         Addiction is an American rock band from Los Angeles, formed in 1985. 
         The band consists of Perry Farrell (vocals), Dave Navarro (guitar), Stephen 
         Perkins (drums) and Chris Chaney (bass).

Assistant: Both the candidate answer and the suggestion agree on the fact that the correct 
           option is 'rock'. Let's check on the context whether or not this is correct. 
           Weeping Willows is an indie rock group, thus they make rock music; Jane's 
           Addiction is a rock band. Consequently the context confirms that the genre 
           performed by both bands is 'rock'. The correct option is 'rock'. 

Question: Between two tennis players Kim Clijsters and Mary Pierce, who is older?
Options: ['Kim Clijsters', 'Mary Pierce']
Candidate answer: Kim Clijsters
Suggestion: The correct answer is 'Mary Pierce' as she was born on 15 January 1975, 
            which is earlier than Kim Clijsters who was born on 8 June 1983.
Context: Kim Antonie Lode Clijsters (] ; born 8 June 1983) is a Belgian 
         former professional tennis player. Clijsters is a former world No. 1 in both 
         singles and doubles. Mary Pierce (born 15 January 1975) is a French retired tennis 
         professional who played on the Women's Tennis Association (WTA) tour. 
         Born in Canada, she is a citizen of Canada, and the United States. 
         Pierce played for France in team competitions and in the Olympics.
        
Assistant: The candidate answer says that the older tennis player is Kim Clijsters, while 
           the suggestion indicates Mary Pierce. The context provides the birth dates of 
           both players, thus I can check which of the two options is correct. 
           Kim Clijsters was born on 8 June 1983, and Mary Pierce was born on 15 January 
           1975; By comparing these dates, it's clear that Mary Pierce 
           is older than Kim Clijsters. Thus the correct option is 'Mary Pierce'.   
\end{lstlisting}

\label{synthesis-effectiveness}
The important notice that has to be made is that we have no guarantees that the \textit{synthesis} step chooses the right option. It could be deceived by a wrong suggestion of one of the two previous steps or it could choose another wrong option. We have already spent a couple of words explaining why a last checking step could be influential in terms of performances of the pipeline, but our experiments will confirm this hypothesis. 

\newpage

\section{Assessing performances} \label{performances}

Up to now, we had widely described each step's details and variations, but still it is unclear how a verbose synthesis output (i.e. not just the correct option, but the explanation of why a certain option is correct) can be translated into the corresponding option. \\

In (\ref{guidance}) we showed how the Large Language Models employed in this analysis are usually \textit{format disobedient} when asked just to output the correct answer to the question. Consequently, we forced the generation of exactly one of the options using the \lstinline{guidance} framework, so that just the chosen option is passed to later stages of the pipeline. \\

In the \textit{antithesis} part, what we wanted instead is that the model is given the opportunity to persuade the \textit{synthesis} that the \textit{thesis}' tentative answer is not correct. 

In these terms, it would be rough to just provide to the \textit{synthesis} two different options and let it choose which is the most appropriate without any additional information. In fact, in a scenario like this, it would be enough to prompt the large language model to reason about different candidates before outputting the answer. Consequently, the computation required to obtain the \textit{thesis}' and the \textit{antithesis}' answers would be senseless. \\

If the \textit{synthesis} receives instead a comment (the \textit{antithesis}) on the first tentative answer (the \textit{thesis}), suggesting which should be the correct option, it takes as input a new vision on the problem that could lead to more grounded answers. This is what is done in the multiple-choice test example that we made in Figure \ref{test-example}. \\

Additionally, we allow the \textit{synthesis} step to question the \textit{antithesis} suggestion, since we give to it all the required elements to answer autonomously to the question. Thus, the \textit{synthesis}' output will not be forced to be a \lstinline{guidance} output, instead it will be few-shot learned to be an explanation on which is the correct answer and why, as reported in the multiple examples of the previous section (\ref{synthesis}).\\

In order to practically determine which is the suggested option without the need to "read" the complete justification, we added a post-processing extraction procedure such that the content is properly extracted and reduced to one of the candidate answers. To assess the impact of each step of the pipeline, we choose to extract also the \textit{antithesis}' suggested answer, not just the final (i.e. the \textit{synthesis}') one. 

\begin{figure}[H]
    \centering
    \includegraphics[width=0.85\linewidth]{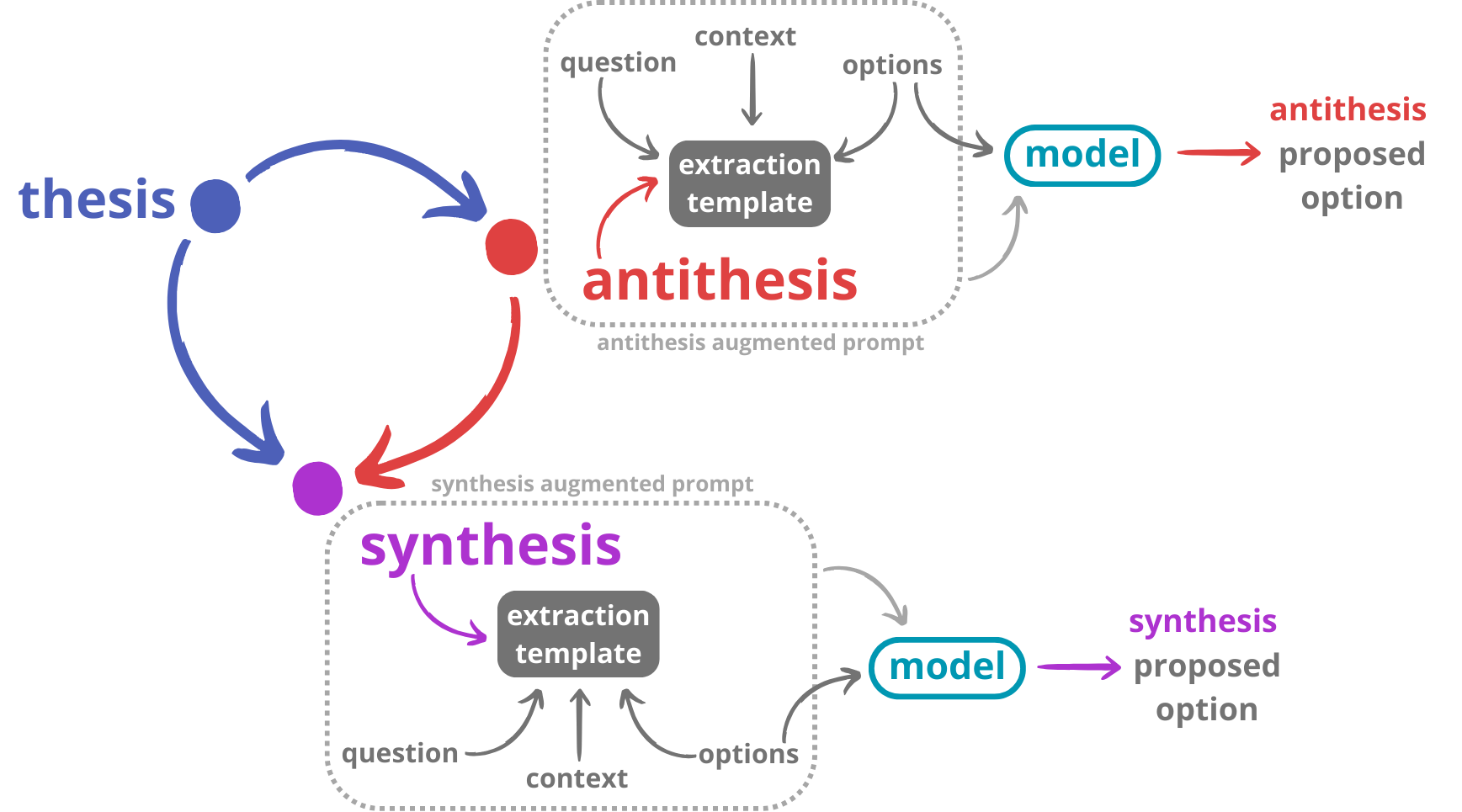}
    \caption{How the extraction process is performed for the \textit{antithesis} and the \textit{synthesis} stages. While the \textit{thesis} step simply constrains the generation, for these two steps we have to extract the chosen option.}
\end{figure}

\newpage

We choose to automatize this process using \lstinline{Phi-3-mini} for all the generated outputs\footnote{In order not to introduce biases dependent on each model's different abilities to follow instructions; we observed empirically (running a couple of tests) that, while \lstinline{Phi-3-mini} is always reliable and accurate, the \lstinline{Gemma} family often disattends the requirements and outputs random options instead.}. We use the following prompt template:
\begin{lstlisting}
You are a multiple-choice question answering assistant.
Choose the most proper option between {options} that best matches with the suggestion. 

Question: {question}
Suggestion: {antithesis}/{synthesis}
Sources: {context}

Assistant:
\end{lstlisting}
that is subsequently augmented with the proper items forming an \lstinline{augmented_prompt}. Finally, the \lstinline{guidance} framework is employed to be sure that the answer comes in the right format:
\begin{lstlisting}
def optionExtraction(question, options, suggestion, sources):
    # augmented_prompt is produced
    answer = guidance_model + augmented_prompt + select(options)
    return answer  
\end{lstlisting}

Kojima et al. \cite{kojima2023} used a similar post-processing stage for their zero-shot Chain-of-Thought: although they did not use a structured guided generation tool as \lstinline{guidance}, they performed a post-processing phase aimed at extracting the correct answer from a longer one. \\

We make sure that this post-processing approach faithfully outputs the suggested option and not the correct one. By providing the model with the \lstinline{question}, the \lstinline{options} and the \lstinline{context} we could think that this final stage could modify the pipeline's answer. 

Consider as an example the following task: 
\begin{lstlisting}
Who was born first, Pablo Trapero or Aleksander Ford?
\end{lstlisting}
the \textit{thesis} proposed the correct answer (i.e. \lstinline{Aleksander Ford}), while the \textit{antithesis} affirmed that:
\begin{lstlisting}
The correct answer should be 'Pablo Trapero' as he was born on 4 October 1971, 
while Aleksander Ford was born on 24 November 1908.   
\end{lstlisting}
the proposed post-processing technique extracts \lstinline{Pablo Trapero} as the antithesis' proposed answer, even if incorrect and in contrast with both the \textit{thesis}' and the \textit{synthesis}' choices.

\section{Comparison with Chain-of-Thought prompting} \label{cot}

We want to compare the answers provided by our method with respect to Chain-of-Thought prompting, as it is well-renowned to perform remarkably well on reasoning tasks.
Chain-of-Thought is prompted a single time to reason about a task in a way that resembles the few-shot example provided. On the opposite, our pipeline extends this approach by prompting the same model multiple times, passing to each step the output of the previous ones (when present). \\

We try to make the comparison \textit{as fair as possible}, i.e. we provide the same few-shot examples and we extract the CoT answer also with the \lstinline{guidance} framework (thus no format issues should be present). \\

In detail, the one-shot example is the same given in the \textit{antithesis}' \lstinline{cot} variant (\ref{cot-variant}), except for a couple of details: the lack of the \textit{thesis}' first tentative answer and the corresponding instruction in the general prompt (i.e. \lstinline{You are asked to ...}). The Chain-of-Thought prompt is the following:
\begin{lstlisting}
You are an helpful AI assistant. You are asked to determine the most correct answer for 
a given question provided a set of possible options. Your goal is to decree which is 
the most correct answer to the question between the available options.


Here's an example of how to do it:
Question: What is the sun, a star or a planet?
Options: ['a star', 'a planet']
Context: The Sun is the star at the center of the Solar System. It is a massive, 
         nearly perfect sphere of hot plasma, heated to incandescence by nuclear 
         fusion reactions in its core, radiating the energy from its surface mainly 
         as visible light and infrared radiation with 10% at ultraviolet energies.

Assistant: Let's consider the options and check whether or not they are correct. 
           The context clearly identifies the Sun as 'the star at the center of the 
           Solar System', thus 'a star' is probably the correct option. 
           On the opposite, 'a planet' is not mentioned in the context, 
           thus it is unlikely to be the correct option. Therefore, the correct 
           option is 'a star'. 

Now do the same for the following question:
Question: {question}
Options: {options}
Context: {context}

Assistant:
\end{lstlisting}

Thus the \lstinline{cot} variant and the Chain-of-Thought approach are really similar. 

The big difference stands in where the effective answer extraction takes place: while in the Chain-of-Though approach is directly extracted from the previous prompt's output, the pipeline requires an extra step to be performed (the \textit{synthesis}) before decreeting the most correct option. 

\begin{figure}[H]
    \centering
    \includegraphics[width=0.95\linewidth]{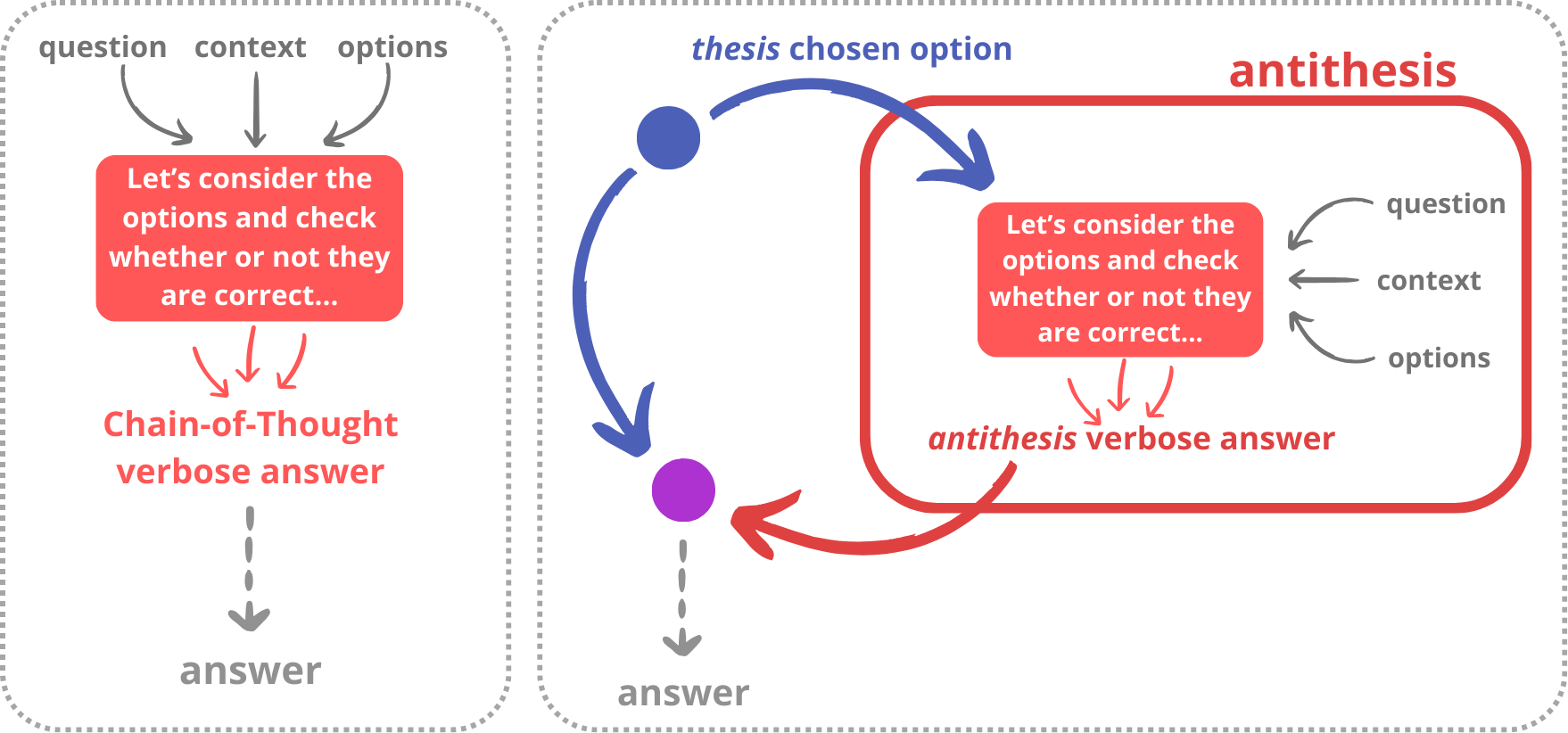}
    \caption{Comparison between the answer generation using the Chain-of-Thought approach (left) and \lstinline{cot} variant of the pipeline (right).}
\end{figure}

\newpage

\chapter{Results} \label{results}

In this final stage, we will report our experiments' outputs that try to answer the following research questions:
\begin{itemize}

\item Does the pipeline have a beneficial effect in improving the answers' correctness?
    
\item How do the different families of models perform under the same pipeline settings?
    
\item Is the pipeline robust with respect to different datasets?

\item Which is the impact of pipeline variations (\ref{baseline})?

\item Is the proposed method more effective than Chain-of-Thought prompting?

\item Do pre-processing sources as detailed in section (\ref{context-preproc}) have positive impacts on downstream pipeline performances? Which are the differences that can be observed between them?
    
\end{itemize}

\section{Does the dialectic pipeline work?}

Before asking any different question, the \textit{core}, preliminary answer that is necessary to conduct further studies is whether or not the dialectic pipeline works. By \textit{works} we mean that we are able to observe a significant improvement with respect to the standard prompting output. \\

The two generated outputs that we are going to compare in this first step are the \textit{thesis}' answers and the \textit{synthesis}' ones. Note that, for how we defined the \textit{thesis} step (\ref{thesis}), this first tentative answer is in fact the final output of a standard prompting procedure. 
The \textit{thesis} considers the question, the options and the supporting context and produces the most proper answer given these elements. The \lstinline{guidance} framework ensures that the output possesses a clear format: no issues regarding failure to recognize a correct answer but in the wrong format will arise.

We choose to append the relevant context to this step (although not strictly necessary, since the \textit{thesis} is in fact a baseline) in order to make a fair comparison between a \textit{context-enriched} pipeline and a standard model generation, that should reasonably also be provided with the relevant passages. \\

We are going to consider the \lstinline{comparison} partition of the HotpotQA dataset and we will exploit the \lstinline{baseline} pipeline setting (\ref{baseline}), running \lstinline{microsoft/Phi-3-mini-4k-instruct} (\ref{phi-mini}) as model. \\

Results show that, while the \textit{thesis}' guesses the correct option in the 53.4\% of the cases, the (\lstinline{baseline}) pipeline reach a 80.7\% of correct guesses. The 27.3\% improvement is sufficiently significant to justify the claim that the pipeline works (even in the baseline setting).

\section{How models' architectures and number of parameters impact on pipeline performances} \label{model-robust}

The second question of our analysis is whether this improvement is consistent when we test other models than \lstinline{microsoft/Phi-3-mini-4k-instruct}. 

We additionally try \lstinline{microsoft/Phi-3-medium-4k-instruct} (similar architecture than \lstinline{Phi-mini}, but has 14B parameters instead of 4B), \lstinline{google/gemma-2-2b-it}, \lstinline{google/gemma-2-9b-it} (\ref{gemma-2}) and \\
\lstinline{meta-llama/Meta-Llama-3.1-8B-Instruct} (\ref{llama}). 

\begin{table}[H]
\centering
\begin{tabular}{|c|c|c|c|c|c|}
\hline

\textbf{Model} & \textbf{\lstinline{Phi-mini}} & \textbf{\lstinline{Phi-medium}} & \textbf{\lstinline{Gemma-2B}} & \textbf{\lstinline{Gemma-9B}} & \textbf{\lstinline{LlaMa-8B}} \\ \hline

\textit{Thesis}' correct answers (\%) & 53.4 & 50.0 & 52.8 & 59.7 & 48.3 \\ \hline

\textit{Synthesis}' correct answers (\%) & 80.7 & 89.5 & 81.8 & 88.1 & 87.2 \\ \hline

Absolute improvement (\%) & 27.3 & 39.5 & 29.0 & 28.4 & 38.9 \\ \hline
\end{tabular}
\end{table}

Focusing on \textbf{\textit{thesis}}' performances, it is easy to spot that \lstinline{Phi-mini} and \lstinline{Gemma-9B} are the best overall. While it is reasonable that \lstinline{Gemma-9B} outperforms \lstinline{Gemma-2B} (since an increased number of parameters often corresponds to more capable models \cite{brown2020}), what is truly surprising is that \lstinline{Phi-mini} shows better performances than \lstinline{Phi-medium} (having 14B parameters, 10B more parameters than the \lstinline{mini} version!). This is in fact something that the \lstinline{Phi-3} research group observed:\\
\textit{some benchmarks improve much less from 7B to 14B than they do from 3.8B to 7B, perhaps indicating that our data mixture needs further work to be in the “data optimal regime” for 14B parameters model.}\\

The \lstinline{Phi-3} family of models is pre-trained on "textbook data" \cite{phi-3} that allows to reduce the training corpus and break the \textit{scaling laws} \cite{kaplan2020}. But if the data mixture is not optimal (i.e. do not possess a certain level of data quality required to reduce the corpus size) then the performances would be poorer than expected. \\

It is also necessary to point out that \lstinline{Gemma-2B} is not far from \lstinline{Phi-mini}, although in practice it has half its parameters: 2B and 3.8B respectively, with a performance gap of just 0.6\%, corresponding to 2 wrongly predicted answers only\footnote{The \lstinline{comparison} partition of the dataset consists in 352 questions.}. \\

Another surprising result is that \lstinline{LlaMa-8B} is the worst performing model overall if we limit to prompt the model in the standard way.\\

Switching now to the \textbf{\textit{synthesis}}' results, the best overall pipelines are obtained using \lstinline{Phi-medium} and \lstinline{Gemma-9B} models, although even \lstinline{LlaMa-8B} reaches a similar result. 

We can observe that models performing worse than others in the thesis step are able to largely bridge the initial gap through the pipeline, reaching the best overall performances (this is the case of both \lstinline{Phi-medium} and \lstinline{LlaMa-8B}). \lstinline{Phi-medium} beats by a small margin (1.5\%, corresponding to 5 differently guessed answers). \\

The smaller models reach lower accuracy values than their correspondent greater options, even not by a large margin. Additionally, \lstinline{Gemma-2B} beats by 1\% \lstinline{Phi-mini}, despite being almost half the size of the second one. \\

With these experiments, we can safely assess that, at least on HotpotQA \lstinline{comparison}, the pipeline works for different families of models and improves by a large margin the baseline. \lstinline{Phi-mini} obtains the worst performance increase overall, and despite this it is able to score a really good improvement. 

\section{Robustness with respect to different datasets} \label{dataset-robust}

In the previous section we clearly expressed how our experiments made sure the pipeline's effectiveness on the HotpotQA \lstinline{comparison} subset. 

Now we want to assess whether it is true that this statement could be generalized to other \textit{multi-hop} datasets. We test the same pipeline setting also on the \lstinline{bridge} partition of the same dataset and on the WikiHop dataset. In this section we will be considering the WikiHop version with summarized contexts, i.e. the 330 questions mentioned in (\ref{summ}).

\begin{table}[H]
\centering
\small
\setlength{\tabcolsep}{3pt} 
\begin{tabular}{|l|l|c|c|c|c|c|}
\hline
\textbf{Dataset} & \textbf{\% correct} & \textbf{Phi-mini} & \textbf{Phi-medium} & \textbf{Gemma-2B} & \textbf{Gemma-9B} & \textbf{LlaMa-8B} \\ \hline

HotpotQA & Thesis: & 53.4 & 50.0 & 52.8 & 59.7 & 48.3 \\
comparison & Synthesis: & 80.7 & 89.5 & 81.8 & 88.1 & 87.2 \\ \hline

HotpotQA & Thesis: & 52.1 & 56.0 & 55.7 & 66.5 & 49.8 \\ 
bridge & Synthesis: & 87.9 & 90.2 & 81.3 & 88.9 & 91.9 \\ \hline

\multirow{2}{*}{WikiHop} & Thesis: & 12.7 & 13.6 & 17.7 & 16.9 & 12.4 \\ 
                         & Synthesis: & 33.0 & 40.7 & 28.3 & 21.1 & 37.7 \\ \hline
\end{tabular}
\label{robust-datasets}
\end{table}

These results confirm the trend that we observed for the \lstinline{comparison} partition of HotpotQA. 

If we consider the \lstinline{bridge} partition of the same dataset, we can recognize similar \textit{thesis} and \textit{synthesis} accuracy values, although they tend to be better than those of \lstinline{comparison}.
\lstinline{Phi-mini} and \lstinline{LlaMa-8B} show nearly identical \textit{thesis} percentages of correct answer in both \lstinline{comparison}  and \lstinline{bridge} partitions, while \lstinline{Phi-medium}, \lstinline{Gemma-2B} and \lstinline{Gemma-9B} perform better by a certain margin in this second subset. We have to consider that while we just considered 352 questions from \lstinline{comparison}, the \lstinline{bridge} partition contains 1000 items. Thus, these differences\footnote{6\%, 2.9\% and 6.8\% in \lstinline{Phi-medium}, \lstinline{Gemma-2B} and \lstinline{Gemma-9B} respectively} could be originated:
\begin{itemize}
    \item the different dataset sizes, since a more various and wide datasets could provide a more reliable value of models' predictive abilities than a smaller one;
    \item by the automatic generation of the alternative option for \lstinline{bridge}, as described in (\ref{bridge}) that could make its tasks easier to solve than hand-checked ones, in particular for more capable models. This would explain the greater gap that we observe for \lstinline{Phi-medium} and \lstinline{Gemma-9B}. 
    
\end{itemize} 

What we can infer from these results is that greater size models (i.e. \lstinline{Phi-medium}, \lstinline{Gemma-9B} and \lstinline{LlaMa-8B}) perform remarkably well, as we observed for \lstinline{comparison} already. The interesting difference is that this time \lstinline{Phi-mini} outperforms by a large margin \lstinline{Gemma-2B}: that (apparently small) 6.6\% gap corresponds to 66 wrongly answered questions from the latter model. This leads us to the idea that \lstinline{Phi-mini} is more capable than \lstinline{Gemma-2B} at merging multiple sources to determine which is the correct solution. When we ask instead to look at two different sources and compare them, the 2B model performs better than the other one. Maybe this difference is due to the limited abilities of a model of that small size, and for \textit{bridge} type of tasks it may be preferable to use a greater model.\\

The WikiHop dataset confirms the pipeline effectiveness, despite the reduced accuracy values that can be observed both for the \textit{thesis} and for the \textit{synthesis} percentage of correct answers. In this case, the number of candidate options is not fixed to two and neither is the number of \textit{hops}. Instead, WikiHop presents a number of options up to 60, even though limited to a few questions, and a set of passages with a big variety in terms of number of \textit{hops} (\ref{wikihop-hops}).

Due to this significant difference, the already tested models behave differently than before. While \lstinline{Phi-mini}, \lstinline{Phi-medium} and \lstinline{LlaMa-8B} improve the \textit{thesis} by large margins (each of these improves the baseline more than twice), the \lstinline{Gemma-2} family of models seem to struggle with this various set of tasks. 

Surprisingly, the 2B version overcomes the 9B one by a solid 7.2\%.\\

Below is reported the table of absolute improvements in accuracy observed between the \textit{thesis} and the \textit{synthesis} steps of the pipeline. Results are reported with respect to the considered dataset (\lstinline{comparison} and \lstinline{bridge} omit the HotpotQA prefix) and the model used inside the pipeline. 

\begin{table}[H]
\centering
\begin{tabular}{|l|c|c|c|c|c|}
\hline
\textbf{Dataset} & \textbf{Phi-mini} & \textbf{Phi-medium} & \textbf{Gemma-2B} & \textbf{Gemma-9B} & \textbf{LlaMa-8B} \\ \hline

comparison & 27.3 & 39.5 & 29.0 & 28.4 & 38.9 \\ \hline

bridge & 35.8 & 34.2 & 25.6 & 22.4 & 42.1 \\ \hline

WikiHop & 20.3 & 27.1 & 10.6 & 4.2 & 25.3 \\ \hline

\textit{Overall mean} & \textit{27.8} & \textit{33.6} & \textit{21.7} & \textit{18.3} & \textit{35.4} \\ \hline
\end{tabular}
\label{tab:model_performance}
\end{table}

The first conclusion that we can draw is that for all the cases we considered, our proposed pipeline always causes positive improvements in accuracies; we can consider the mean on all the dataset as a synthetic measure supporting this statement. \\

Despite its being effective, this level of positive impacts is model and dataset dependent. WikiHop is proved to be more challenging due to its multi-hop variety and the big number of options between which the models have to choose the most proper answer. 

Despite its being really challenging, the pipeline often reaches large margins of improvements (more than 20\%) for two families of models, \lstinline{Phi-3} and \lstinline{LlaMa-3.1}. The \lstinline{Gemma-2} family clearly struggles in absorbing meaningful information from the pipeline, while being the most capable family when considered the \textit{thesis} performances only.

This could be due to the parametric knowledge of \lstinline{Gemma-2} models, maybe richer and more various than the "texbook" knowledge of \lstinline{Phi-3} and of the \lstinline{LlaMa-3.1} one. 

Considered this, we will prefer to analyze the \textit{synthesis}' accuracy values as indicators of the pipeline effectiveness rather than focusing on the relative improvement observed with respect to the \textit{thesis}' ability of finding the correct question at the first try.

The minor tendency of \lstinline{Gemma-2} to follow the instructions given (i.e. use context to answer) is easily observed when we inspect the responses produced by this in the \textit{antithesis} phase, compared with those of the other two models:

\begin{figure}[H]
    \centering
    \includegraphics[width=0.9\linewidth]{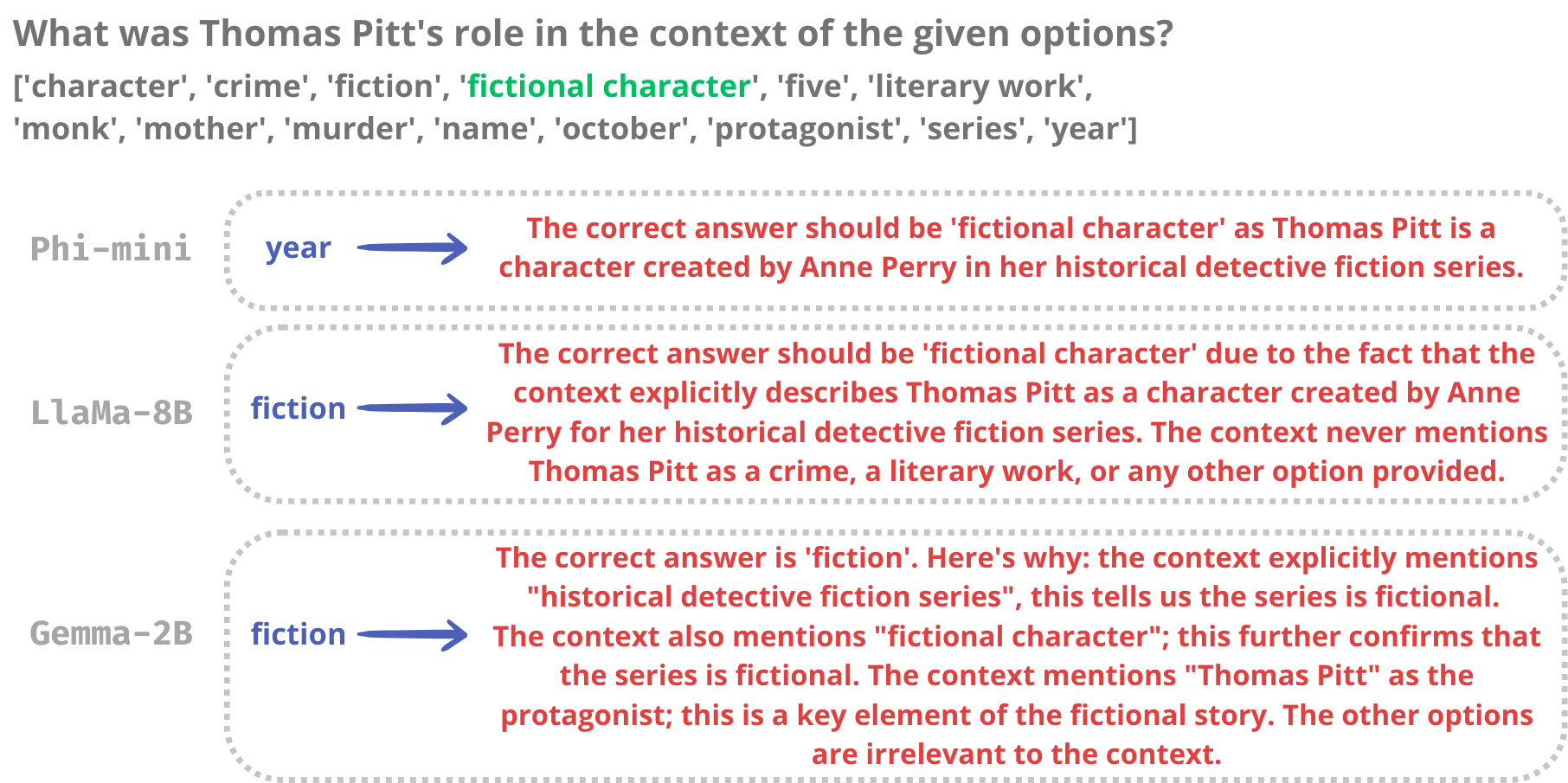}
    \caption{Comparison between the \textit{antithesis}' answers as given by these three families of models; the \textit{thesis}' proposals are colored in blue. Note that the format should follow the one-shot example provided in (\ref{baseline}), thus \lstinline{Gemma-2B} turns out to be poorer than the other models in faithful instruction following.}
    \label{gemma-issues}
\end{figure}

It would be untrue to state that \lstinline{Gemma-2} models' poor ability to replicate faithfully the given few-shot examples (and being more "conversational" rather than context-grounded) is harmful in all the cases. In HotpotQA, where the options are just two, this issue could not be a real problem. In fact, the absolute improvements observed for \lstinline{Gemma-2} models on HotpotQA are large (although smaller than other models' margins). The necessity to attain to instructions is instead crucial in more complex task.\\

This further hints that the real improvement given by the pipeline is not given by its multi-step nature, rather by the \textit{reasoned solving process}. By breaking down the problem into units performing multiple checks from different perspectives, even small models (i.e. with less than 20B parameters) can achieve good performances.

\newpage

\section{Pipeline variations}

In the previous section (\ref{dataset-robust}) we highlighted how the \lstinline{Gemma-2} tendency to approach the \textit{antithesis} step differently from what it has been shown from the one-shot example could explain worse prediction abilities (at least on challenging tasks). 

The following experiments are made to assess how much the way in which we prompt the model has an impact on the final performances. In practice, we substitute the one-shot example provided and study how much this modification changes the correctness of the final predicted output. \\

We start from the \lstinline{comparison} partition of HotpotQA and test the differences between the baseline pipeline settings and the \lstinline{is}, the \lstinline{cot} and the \lstinline{ctx} variants (\ref{baseline}). This comparison is performed for all the models that we already considered.

\label{hotpot-results}

\begin{figure}[H]
    \centering
    \includegraphics[width=0.85\linewidth]{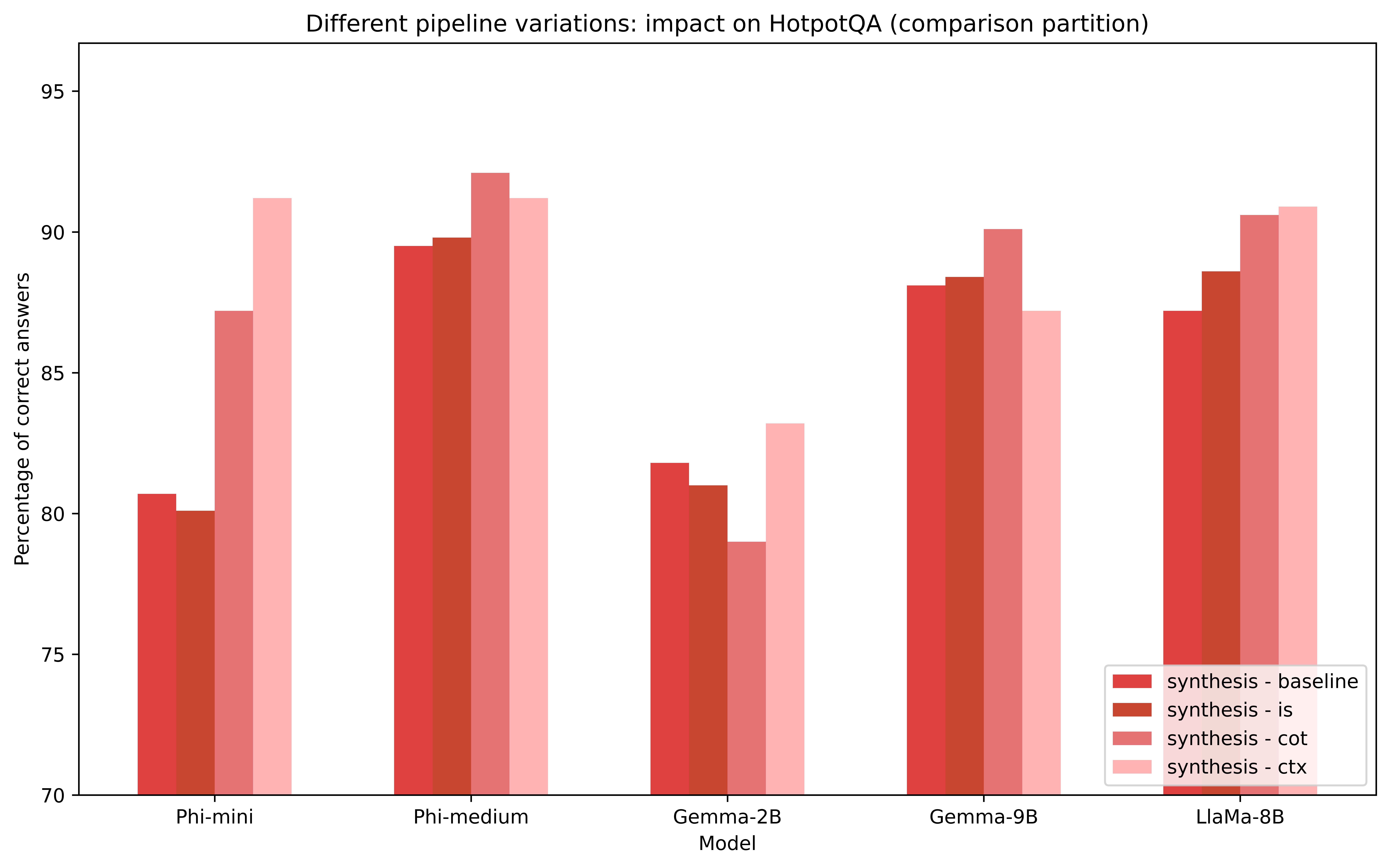}
    \caption{Different percentages of correct answers obtained using a particular pipeline setting \textbf{on HotpotQA - \lstinline{comparison}}. The plot y-axis is reduced (starts from 70\%) to highlight the small differences between the different one-shot examples.}
\end{figure}

The first comment that we can make is that there is a negligible difference (under the 1\% for all the models) between the baseline and the \lstinline{is} variation. We can safely affirm that the level of confidence with which the \textit{antithesis} is provided is not relevant in terms of performance. Thus, the \textit{antithesis} persuades the pipeline through factual knowledge, not by a more confident style of suggestions. \\

The \lstinline{cot} variant seems to outperform the baseline when the model with sufficiently big models, i.e. comparison between \lstinline{Phi-medium}, \lstinline{Gemma-9B} and between \lstinline{LlaMa-8B}. This improvement lies in the reasoning abilities observed in models with a greater number of parameters\footnote{This is also confirmed by the Chain-of-Thought paper by Wei et al. \cite{wei2022}, where reasoning is described as \textit{an emergent ability of more capable} (i.e. greater size) \textit{models}.}. On the other hand, also \lstinline{Phi-mini} benefices from this "reasoned" style of answer, while \lstinline{Gemma-2B} is penalized with respect to the baseline. A possible explanation of this discrepancy stands in the pre-training procedure of \lstinline{Phi-mini}: an entire step of it is dedicated to teaching the model how to logically reason and to attain specific skills (\ref{phi-mini}). Consequently, \lstinline{Phi-mini} small size does not influence the quality of the generated answer. This is obviously not the case of \lstinline{Gemma-2B}, that possesses a diminished capability with respect to the 9B version and is not pre-trained for reasoning purposes. \\

The \lstinline{ctx} variant differs from the previous one in two main points: 
\begin{enumerate}

\item the \textit{antithesis} is induced to quote all the relevant context before choosing the correct option, thus allows the model to split the content extraction and the effective decision process (this is an implicit process, that the model has to infer from the one-shot example that is provided);

\item the \textit{synthesis} stage shows two examples of dealing with the \textit{thesis}' and the \textit{antithesis}' opinions on which is the correct answer (in the first one, the two agree; in the second one, they don't), thus instruct explicitly the \textit{synthesis} on how to deal with these two scenarios.

\end{enumerate}

This variation appears to work well (i.e. better than the baseline) on small models such as \lstinline{Phi-mini} and \lstinline{Gemma-2B}; this is probably due to the fact that these kinds of LLMs benefit from the division between the context selection and the effective decision. Additionally, instructing their \textit{synthesis} phase on how to deal with agreement and disagreements could also have a positive effect. 

Greater models (\lstinline{Phi-medium}, of 14B parameters, and \lstinline{Gemma-9B}) still behave better with a more "logically structured" approach as proposed by the \lstinline{cot} variant: while \lstinline{ctx} is still a quite valid option (for \lstinline{Gemma-9B} it worsens the baseline performances, but by a reduced margin of only 0.9\%) it seems to \textit{burden} these pipelines in an unnecessary manner. \lstinline{LlaMa-8B} shows comparable performances of \lstinline{cot} and \lstinline{ctx} and is in fact a middle ground between greater and smaller models in terms of number of parameters. 

Despite some small differences widely discussed above, these first experiments confirm that our method allows to observe consistent improvements with different pipeline settings and across different models employed, not being dramatically biased by specific prompts. \\

Switching now to the \lstinline{bridge} partition, we choose not to run the \lstinline{is} variant because of its nearly perfect correspondence to the baseline. 

\begin{figure}[H]
    \centering
    \includegraphics[width=0.85\linewidth]{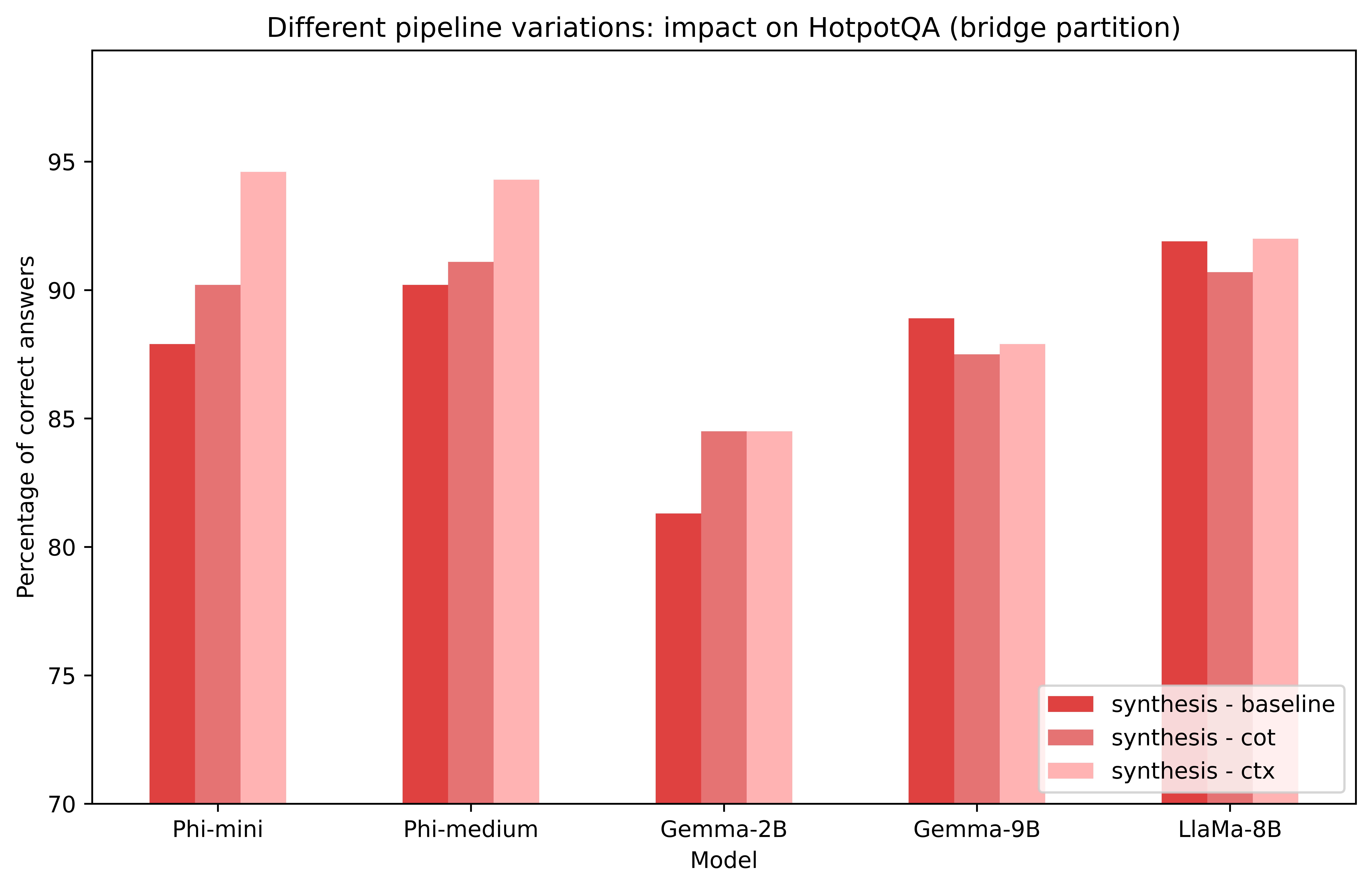}
    \caption{Different percentages of correct answers obtained using a particular pipeline setting \textbf{on HotpotQA - \lstinline{bridge}}. The plot y-axis is reduced (starts from 70\%) to highlight the small differences between the different one-shot examples.}
\end{figure}

On the other hand, we considered again both the \lstinline{cot} and \lstinline{ctx} variants. In this partition, the \lstinline{cot} option seems slightly inappropriate: this is probably due to the different task that is required to perform. The baseline version of the pipeline seems to perform better than the \lstinline{cot} variant in some cases. 

While \lstinline{comparison} required to compare two sources containing information about two different items and to reason on this content, \lstinline{bridge} requires to merge correctly multiple sources, but once they are properly merged they are of easier understanding. 

The reasoning abilities in this form of tasks is probably less important rather than focusing on the proper information inside the context and properly merge them. This second step is what the \lstinline{ctx} variant does, and in fact it can be observed that in this subset this solution is the best one for all the tested models (for \lstinline{Gemma-2B} the two variants achieve equal performances). \\

Finally, we want to test whether a different pipeline setting could help in improving the performances on the challenging WikiHop tasks.

\begin{figure}[H]
    \centering
    \includegraphics[width=0.85\linewidth]{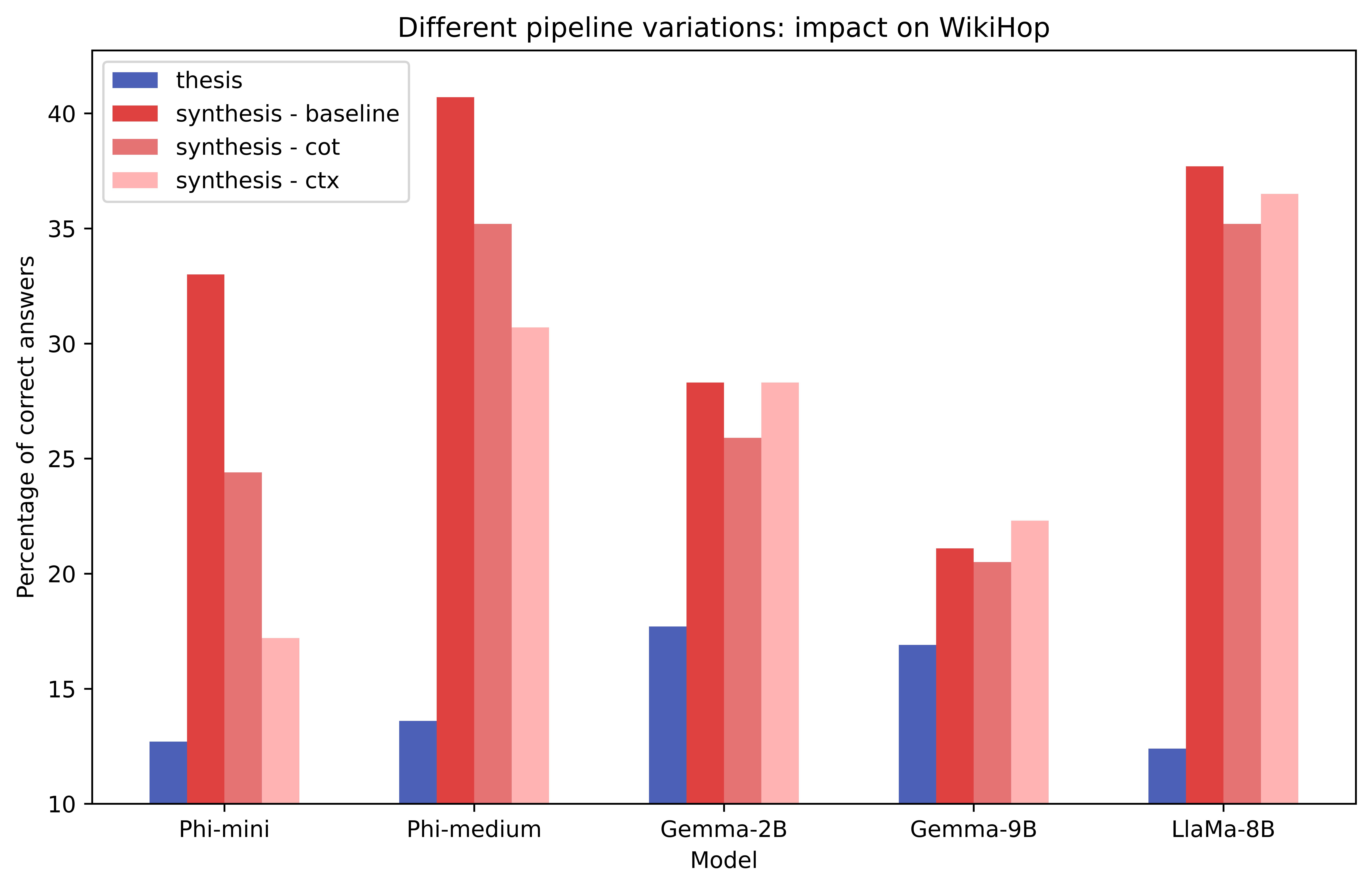}
    \caption{Percentages of correct answers obtained using a particular pipeline setting \textbf{on WikiHop}. The \textit{thesis}' correct answers are represented by the blue bars, while the red ones represent the different pipeline variations considered.}
\end{figure}

For WikiHop, we can comment each family of models separately. The \lstinline{Phi-3} models show the baseline as the best pipeline option, followed by the \lstinline{cot} and finally by the \lstinline{ctx} variations. \lstinline{Gemma-2} models show similar performances of the baseline pipeline and the \lstinline{ctx} variation, while \lstinline{cot} appears worse. \lstinline{LlaMa-8B} is similar to the previous family, although reaching higher performances for all the pipeline settings. 

No pipeline variation is able to consistently outperform the baseline version of it on \textit{multi-hop}, multiple choice tasks like those of WikiHop. \label{baseline-dominance}
The reason could be find in the fact that both \lstinline{cot} and \lstinline{ctx} ask the model to consider the context and reason on it, such as depicted in (\ref{variations-overview}). On the opposite, the baseline version of the pipeline just asks the model to decree the correct answer, thus it can skip unnecessary context selection and go directly to the most proper answer. 

In the following example, it is clear that while the baseline option can directly answer, the two other options are burdened by the prompt requirements (via the one-shot example). Since WikiHop is really complex and often presents a variety of options, many of which are semantically related (e.g. \lstinline{United Kingdom} and \lstinline{british}), the additional effort that the models have to face is to answer to the prompt's exact instructions. 

In the following example, the question clearly ask the \lstinline{parent taxon} of \lstinline{Australosuchus}, thus it is not properly true (as \lstinline{cot} states) that \lstinline{'animal'} and \lstinline{'crocodile'} are correct options. The model with \lstinline{cot} \textit{antithesis} is able to identify the fact that \lstinline{'crocodilia'} is the correct option, but gets lost in its reasoning process and in the final stage is not able to clearly state that. 

On the opposite, \lstinline{ctx} directly forgets to include \lstinline{'crocodilia'} as one of the options, probably confusing and merging it with \lstinline{'crocodile'}. 

\begin{figure}[H]
    \centering
    \includegraphics[width=0.95\linewidth]{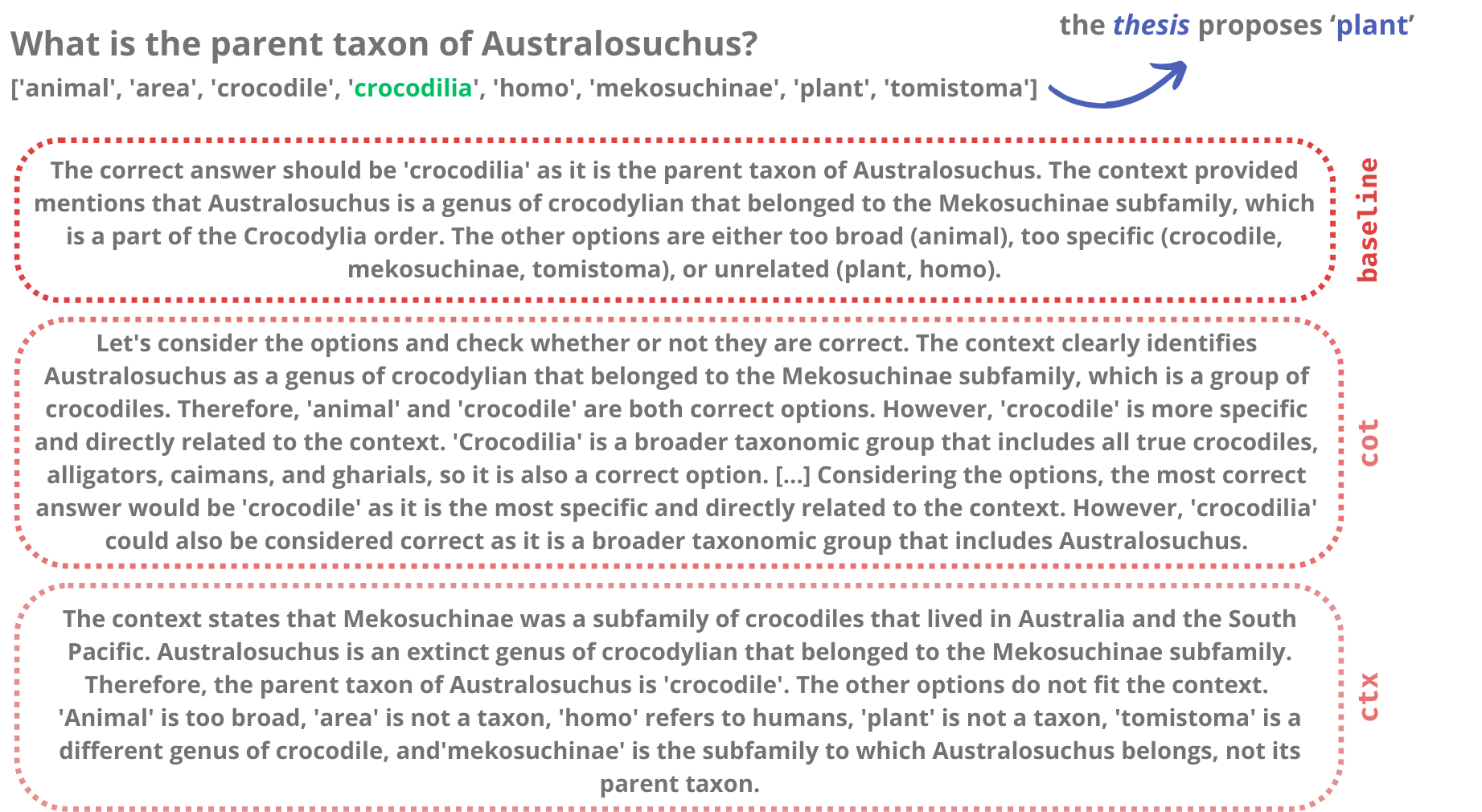}
    \caption{Baseline, \lstinline{cot} and \lstinline{ctx} \textit{antitheses} obtained with \lstinline{Phi-mini}.}
\end{figure}

Again, \lstinline{Gemma-2} family exhibits poor abilities in absorbing the few-shot prompting implicit patterns. This allows its models to choose an original and \textit{unscheduled} way of dealing with the problem, i.e. splitting the reasoning process in multiple steps in order to explain why a certain option is the correct one, while briefly explaining why the others are not. This approach is able to reach the correct option in some cases, while it leads the model to strongly support a wrong claim due to an approximate context usage (Figure \ref{gemma-issues}). \\

\lstinline{LlaMa-8B} allows the \lstinline{ctx} variant to properly analyze each option, beating the \lstinline{cot} one which tends to get struck on motivating why its option is better than the other, rather than consider each alternative independently and finally reach a conclusion. \\

Summing up, in tasks requiring a careful and concordant answer to a precise question in which the model has to deal with multiple options and long, \textit{multi-hop} contexts, we suggest to let the pipeline do its work without forcing it to explain in a Chain-of-Thought manner why it does so. Additionally, we find that for these type of tasks, indulging in evaluating each relevant part of the context could be harmful\footnote{We can imagine the metaphor of an overthinking model: if it already knows the correct answer, it could \textit{get lost} in trying to motivate everything besides that option.}. 

\newpage

\section{\textit{Overthinking} can be harmful (even for LLMs)}

In the previous sections (\ref{extra-step}, \ref{synthesis-effectiveness}) we question the effectiveness of the \textit{synthesis} step, that could appear redundant once the \textit{antithesis} gives its feedback. 

To assess whether we need the \textit{synthesis}, we extracted the options suggested by the \textit{antithesis} step in order to observe what changes between these two steps. We compare the answers given in these two checkpoints. 

In the following table we show the mean improvement across all datasets caused by the \textit{antithesis} step with respect to the \textit{thesis} (first rows) and the residual variation obtained by adding the \textit{synthesis} step (second rows):

\begin{table}[H]
\centering
\begin{tabular}{|c|c|c|c|c|c|c|}
\hline
\textbf{Setting} & \textbf{Step} & \texttt{Phi-mini} & \texttt{Phi-medium} & \texttt{Gemma-2B} & \texttt{Gemma-9B} & \texttt{LlaMa-8B} \\ \hline
\multirow{2}{*}{\textbf{baseline}} & \textbf{T-A} & 26.67 & 32.57 & 23.17 & 18.83 & 35.40 \\
                                   & \textbf{A-S} & 1.13  & 1.03  & 2.50  & 0.57  & 0.57  \\ \hline
\multirow{2}{*}{\textbf{cot}}      & \textbf{T-A} & 27.63 & 31.30 & 19.03 & 17.90 & 32.97 \\
                                   & \textbf{A-S} & 0.23  & 1.63  & 0.43  & 0.43  & 2.63  \\ \hline
\multirow{2}{*}{\textbf{ctx}}      & \textbf{T-A} & 27.97 & 32.17 & 26.27 & 21.43 & 38.43 \\
                                   & \textbf{A-S} & 0.30  & 0.03  & \textbf{-2.30} & \textbf{-2.57} & \textbf{-2.13} \\ \hline
\end{tabular}
\caption{Mean accuracy gaps (across HotpotQA partitions and WikiHop) observed between the \textit{thesis} and the \textit{antithesis} steps (\textbf{T-A}) and between the \textit{antithesis} and the \textit{synthesis} ones (\textbf{A-S}). In bold, we highlighted the negative ones.}
\label{ant-vs-syn}
\end{table}

The results above are averaged across all datasets; the dataset-specific ones can be found in Appendix \ref{appendix-f}. Despite each dataset's characteristics that we have previously described (\ref{dataset-robust}) and the model's tendency to behave differently to the same task (\ref{model-robust}), from results in Appendix \ref{appendix-f} it appears quite clear that the pipeline variations condition on the \textit{synthesis}' ability to improve the \textit{antithesis}' one.\\ 

Besides the -1.8\% of the \lstinline{Gemma-2B} case on the \lstinline{bridge} partition of HotpotQA, the \textit{synthesis} stage turns out to be always beneficial if the baseline version of the pipeline is used. 

Some performances drop between the \textit{antithesis} and the \textit{synthesis} stages are observed when used the \lstinline{cot} variant, and this is reasonable due to the fact that the Chain-of-Thought approach to the answer already explores whether each option is the correct one or not. Adding a further step to this already exhaustive computation has the effect of bias towards a wrong answer some correct reasoning chains (even by small margins). This is due to the limited number of \textit{hops} (2) that characterizes HotpotQA: \lstinline{cot} \textit{antithesis} is already too powerful for easy problems, and the \textit{synthesis} could only "take the model off the road".

The prevalent loss of accuracy between the \textit{antithesis} and the \textit{synthesis} happens for HotpotQA partitions and the usage of the \lstinline{ctx} variation. The reasons could be similar than the ones already proposed for the \lstinline{cot} variant. Additionally, we highlight the fact that the \lstinline{ctx} variant is the only one with two examples in the prompt; this suggests that a stronger guidance on how to deal with disagreement between \textit{thesis} and \textit{antithesis} is not beneficial. It could be interesting to modify this choice and study the \lstinline{ctx} outcome with only a one-shot example, as in \lstinline{cot}. This gap appears linked with the \lstinline{Gemma-2} and \lstinline{LlaMa-3.1} models employed. \label{llama-free} \\

We conclude that the pipeline works, but tends to be negatively affected if asked to \textit{think too much} about a problem that is too simple. Thus, in front of relatively easy problems (small and already selected contexts, 1-\textit{hop} passages, trivial questions given the context) we recommend to stop the \lstinline{cot} or \lstinline{ctx} pipeline variations at the \textit{antithesis} stage and retrieve the proposed solution. Since identifying easy problems for the LLM can be a challenging task per se, we leave the use of an automatic detection component for such settings to future work.\\

Although by a small margin, the \textit{synthesis} improves the prediction accuracy in more challenging scenarios or when we use the baseline version of the pipeline (also suggested for user-friendly coding, because the one-shot example is simpler to create).

\section{Chain-of-Thought prompting comparison} \label{cot-results}

We previously described our method as a \textit{guided and disentangled Chain-of-Thought} (\ref{disentangled}), because we ask multiple times the same model to reason on which is the most correct answer to a question, providing to each step the opinion of the previous ones. Thus, we want to compare this method with its most natural competitor: the Chain-of-Thought prompting. 

From now on, when we refer to \lstinline{cot} we mean the pipeline variation, while CoT stands for the one-shot prompt as already described in (\ref{cot}). In that section we highlight the fact that the one-shot example provided in \lstinline{cot} and in CoT is nearly the same\footnote{Except from the fact that the first one also contains the \textit{thesis}' chosen option, while the second does not.}. 

The question now is whether our proposed method is able to outperform the accuracy obtained through Chain-of-Thought prompting.\\

In the following plots we compare the thesis, the CoT and the best \textit{synthesis} accuracy values obtained for different models\footnote{Despite what we observed about the \textit{antitheses} higher accuracy values than \textit{syntheses} ones for simple tasks, for the sake of consistency we will use the second ones for evaluating the tasks from now on.} on HotpotQA. By "the best \textit{synthesis} accuracy values" we mean that we consider the pipeline configuration scoring higher accuracy values for that model and for that dataset. They could be different with respect to the considered model, e.g. the baseline pipeline for \lstinline{Phi-mini} and the \lstinline{ctx} variant for \lstinline{LlaMa-8B}. We allow this to be mixed since not want to test whether a certain pipeline setting outperforms CoT; we just aim at checking whether the pipeline is able to beat CoT in any configuration. 

\begin{figure}[H]
    \centering
    \includegraphics[width=0.6\linewidth]{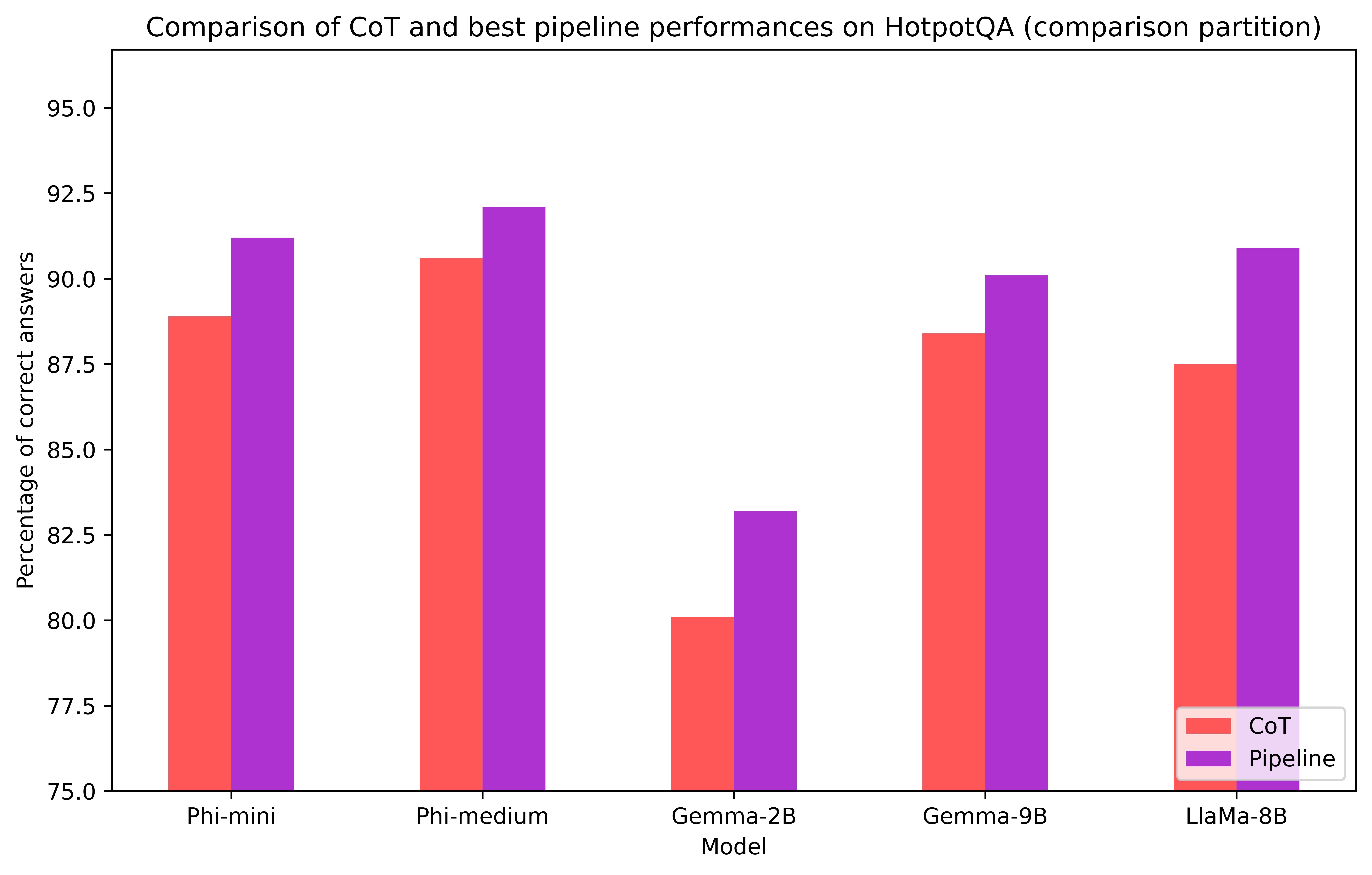}
    \caption{\textbf{HotpotQA \lstinline{comparison} partition}: in red, the Chain-of-Thought answer accuracy values; in purple, the best \textit{synthesis} ones.}
\end{figure}

\begin{figure}[H]
    \centering
    \includegraphics[width=0.6\linewidth]{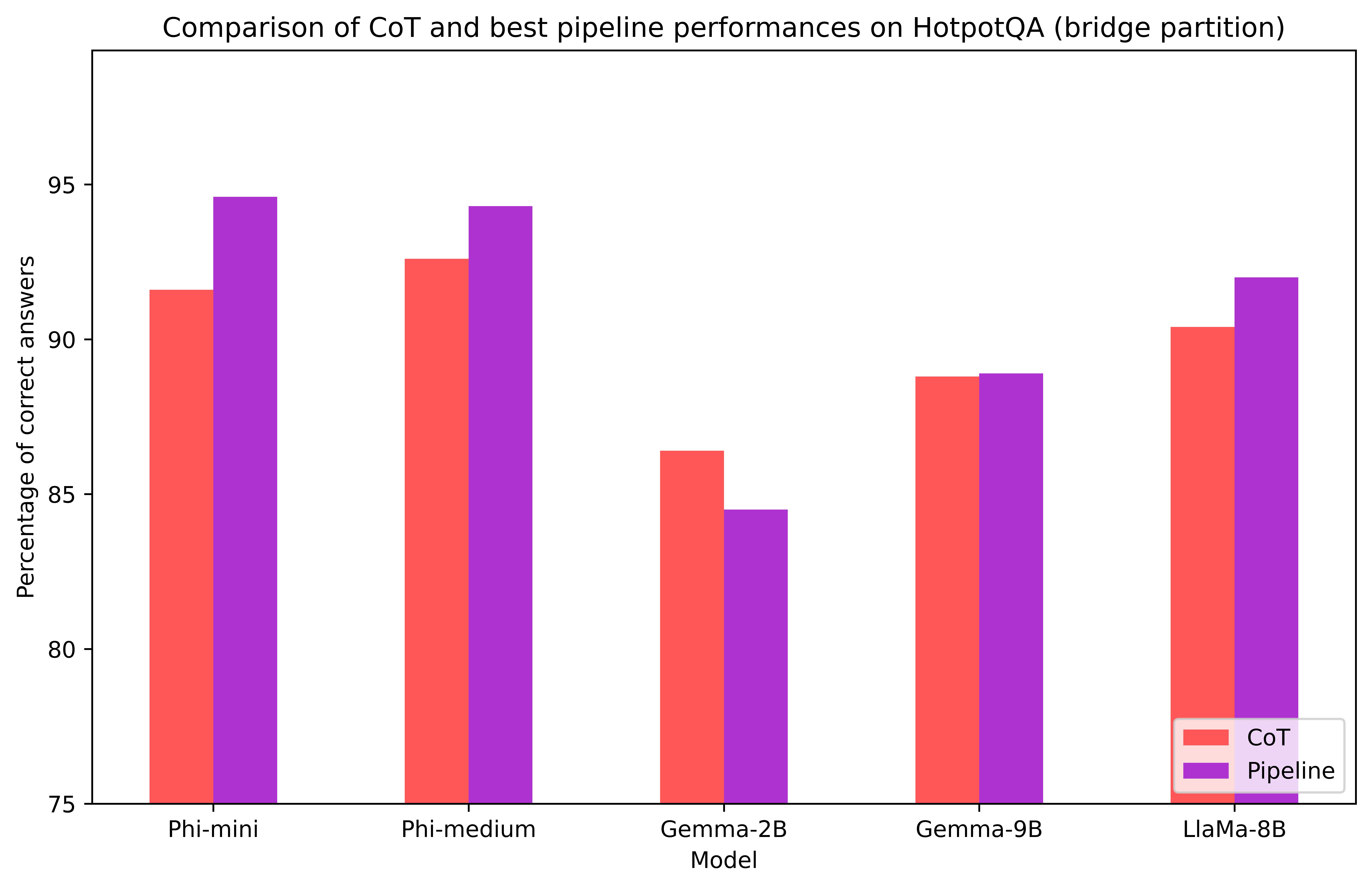}
    \caption{\textbf{HotpotQA \lstinline{bridge} partition}: in red, the Chain-of-Thought answer accuracy values; in purple, the best \textit{synthesis} ones.}
\end{figure}

On one hand, the differences between the two partitions of HotpotQA are really small and almost always\footnote{Only for \lstinline{Gemma-2B} for the \lstinline{bridge} partition CoT is slightly better; but the 84.5\% reported for the pipeline is the \lstinline{ctx}'s synthesis, the corresponding \textit{antithesis} would beat CoT by a 2.8\%.} are in favour of the proposed pipeline. On the other hand, we have to consider that these improvements are obtained with two extra steps, while Chain-of-Thought is able to reach similar values in a single step.

\begin{figure}[H]
    \centering
    \includegraphics[width=0.75\linewidth]{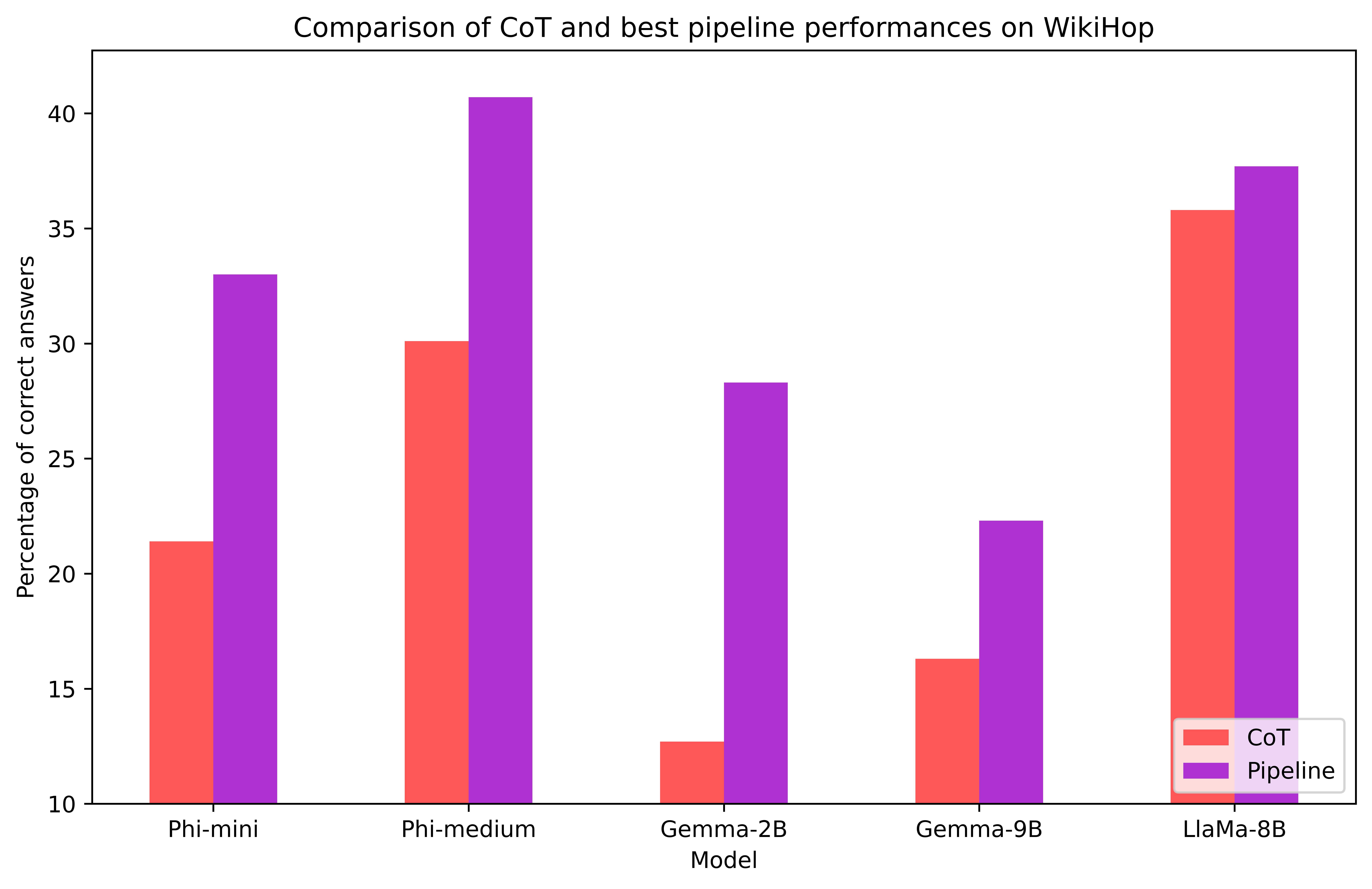}
    \caption{\textbf{WikiHop}: in red, the Chain-of-Thought answer accuracy values; in purple, the best \textit{synthesis} ones.}
    \label{46}
\end{figure}

On WikiHop the gap between the two solutions is much higher for some models: a 11.6\% on \lstinline{Phi-mini}, a 10.6\% on \lstinline{Phi-medium} and 15.6\% on \lstinline{Gemma-2B}. On the opposite, \lstinline{Gemma-9B} baased pipeline outperforms CoT only by 3.8\% and \lstinline{LlaMa-8B} only by 1.9\% (for HotpotQA this margin was greater).

The reasons of these gaps stand in the fact that both \lstinline{Phi-3} and \lstinline{Gemma-2} models receive a significant benefit from the pipeline, in particular when they face a challenging task like those of WikiHop. The tendency of the \lstinline{Gemma-2} models of being more conversational rather than context-grounded (\ref{gemma-issues}) can be easily spotted from the low performances that these models exhibit when CoT-prompted. By asking them to consider again the \textit{thesis}' proposed option, we are able to obtain large improvements. 

\lstinline{LlaMa-8B} shows instead smaller accuracy differences between the two approaches. This is also due to a performance drop between \lstinline{LlaMa}'s \textit{antithesis} and \textit{synthesis} stages: if we early exit the pipeline, we would observe greater margins. \label{llama-synthesis} 

Additionally, in the previous section (\ref{llama-free}) we observed that \lstinline{Gemma-2} and \lstinline{LlaMa-3.1} family of models tend to favour a more \textit{unstructed} prompting structure; when guided in the \textit{synthesis} stage by providing them two examples, they perform worse than the one-shot scenario. Thus maybe the pipeline (at least for these family of models) is disadvantaged by the burden of having an extra instruction to follow. \\

Summing up, the pipeline is able to reach higher performances than CoT in almost all the scenarios studied. 
Certainly, CoT is a more computationally-lightweight than the proposed pipeline, so it is not extremely surprising that it works worse.

However, this is still an interesting finding that highlights how \textbf{we have not yet reached the boundary of the gains that can be derived from the reasoning skills of LLMs}.

An interesting frontier of this work consists in studying whether distilling the correctly-executed dialectic dialogues (\textit{thesis} - \textit{antithesis} - \textit{synthesis}) in a new, smaller model still outperforms CoT. If positive results are observed, then we could consider to exploit this pipeline as a pre-training/fine-tuning component, rather than an inference strategy. Training models on \textit{thinking tokens} has already been proved as effective by works as the one of Zelikman et al. \cite{zelikman2024}.

\section{Context filtering and summarization} \label{res-context}

Lastly, we want to consider \textit{how much} the relevant passage is provided influences the generated output. The objects of our experiments are consequently the contexts of HotpotQA and WikiHop datasets, that would be manipulated and given as inputs to the pipeline. Their outputs are compared with those produced by using the original context in order to check whether there are some improvements. 

WikiHop, that has already been summarized in order to reduce the context length (and stay inside the 4K limit), is considered in these experiments only as a subsection of it. We discard all the original contexts exceeding the threshold\footnote{Practically, we use 3.5K as threshold in place of the true 4K one in order to allow the instructions to fit in.} and we summarize (\ref{summ}) and filter (\ref{my-pecore}) the passages in this subset only. \\

Due to what we observe on \lstinline{Gemma-2} models (\ref{gemma-issues}), we run the following experiments on the \lstinline{Phi-3} and \lstinline{LlaMa-3.1} model families only. We retain that this would be sufficient to assess whether there is an effective improvement in the pipeline's prediction accuracy.

\begin{table}[h]
\centering
\begin{tabular}{|c|c|c|c|c|c|}
\hline
\textbf{Dataset partition} & \textbf{Pipeline} & \textbf{Passage} & \textbf{\texttt{Phi-mini}} & \textbf{\texttt{Phi-medium}} & \textbf{\texttt{LlaMa-8B}} \\ \hline

\multirow{8}{*}{\textbf{\texttt{comparison}}} & \multirow{2}{*}{thesis} & original & 53.4 & 50.0 & 48.3 \\ 
                                    & & filtered & \textbf{54.9} & 46.4 & 48.2 \\ \cline{2-6}
 & \multirow{2}{*}{{baseline}}  & original & 80.7 & 89.5 & 87.2 \\ 
                                    & & filtered & 67.0 & 83.0 & 78.5 \\ \cline{2-6}
& \multirow{2}{*}{\lstinline{cot}} & original & 87.2 & 92.1 & 90.6 \\ 
                                    & & filtered & 58.2 & 74.2 & 79.7 \\ \cline{2-6}
& \multirow{2}{*}{\lstinline{ctx}} & original & 91.2 & 91.2 & 90.9 \\ 
                                    & & filtered & 57.0 & 70.3 & 75.5 \\ \hline
                                    
\multirow{8}{*}{\textbf{\texttt{bridge}}} & \multirow{2}{*}{thesis} & original & 52.1 & 56.0 & 49.8 \\ 
                                    & & filtered & 9.7 & 10.1 & 7.4 \\ \cline{2-6}
 & \multirow{2}{*}{{baseline}}  & original & 87.9 & 90.2 & 91.9 \\ 
                                    & & filtered & \textbf{96.1} & \textbf{94.7} & \textbf{93.8} \\ \cline{2-6}
& \multirow{2}{*}{\lstinline{cot}} & original & 90.2 & 91.1 & 90.7 \\ 
                                    & & filtered & \textbf{91.0} & \textbf{94.5} & \textbf{93.6} \\ \cline{2-6}
& \multirow{2}{*}{\lstinline{ctx}} & original & 91.9 & 90.7 & 92.0 \\ 
                                    & & filtered & 90.3 & \textbf{91.0} & \textbf{92.6} \\ \hline
\end{tabular}
\caption{Context filtering results using (\ref{my-pecore}) on \textbf{HotpotQA}. In bold we highlighted the experiments that outperform the non-filtered context ones.}
\label{tab:combined}
\end{table}

This table contains interesting results. Despite being partitions of the same dataset, \lstinline{comparison} behaves very differently from \lstinline{bridge}. \\

While the filtered version of \lstinline{bridge} scores terrible accuracy values in the thesis step, the other partition slightly improves the percentage of correctly guessed options. We think that this happens due to the two tasks' different nature.\\

\begin{figure}[H]
    \centering
    \includegraphics[width=0.95\linewidth]{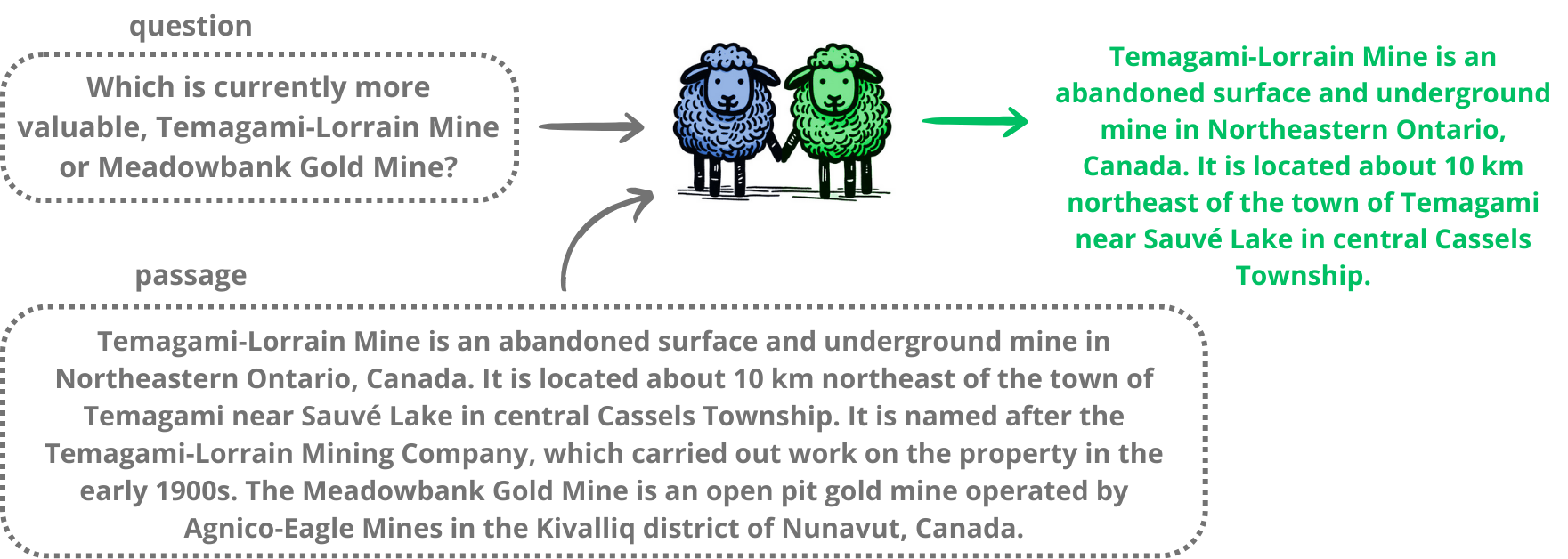}
    \caption{\lstinline{Phi-mini} filtering process (\ref{my-pecore}) on a problem of HotpotQA - \lstinline{comparison}.}
    \label{47}
\end{figure}

Instead, \lstinline{comparison} asks to compare two independent sources sharing some common content that has to be extracted, thus filtering the original context could help in focusing on the relevant parts of both these sub-passages already in the \textit{thesis} stage. On the opposite, when the output of the filtering process is used in the pipeline, the performances worsen with respect to the original one. We can also notice that greater models (i.e. \lstinline{Phi-medium} and \lstinline{LlaMa-8B}) lose less in accuracy terms than smaller ones (i.e. \lstinline{Phi-mini}). 
This is caused by an aggressive selection that the filtering stage performs in some cases. While helpful in neglecting irrelevant information, sometimes it also lefts out some important details useful to correctly answer the question, as showed in the example in Figure \ref{47}.

\newpage

In this case, our proposed filtering process discards completely the sentence that talks about \\ \lstinline{Meadowbank Gold Mine}, causing the \textit{antithesis} step\footnote{Output obtained with \lstinline{Phi-mini}.} to correctly question that: 
\begin{lstlisting}
 The context provided does not contain any information about the value of Temagami-Lorrain 
 Mine or Meadowbank Gold Mine. Therefore, it is impossible to determine which is 
 currently more valuable based on the given context.  
\end{lstlisting}

A similar problem also affects the other partition of the dataset, but this time in the opposite direction.
\lstinline{bridge} questions require the model to merge multiple sources in order to understand how information present in different passages is related. 
It is now easy to imagine which is our filtering approach's enormous fatal flaw: if the linking part is left out, then the passage does not make sense anymore. Consider for example:
\begin{figure}[H]
    \centering
    \includegraphics[width=0.95\linewidth]{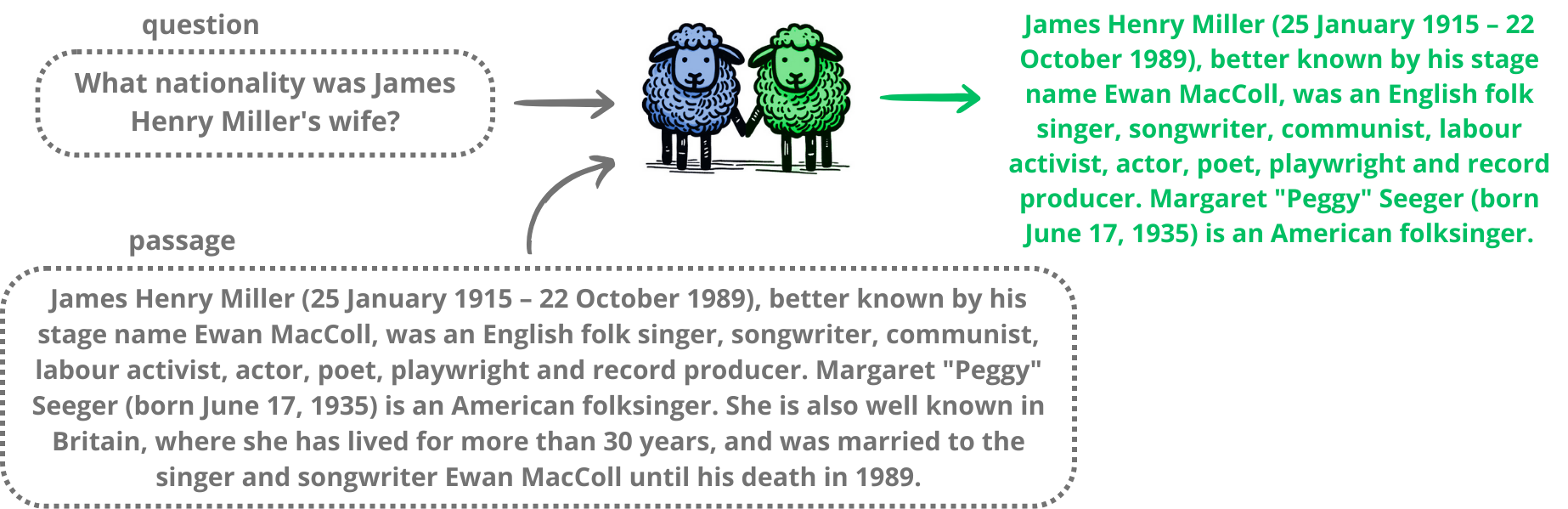}
    \caption{\lstinline{Phi-mini} filtering process (\ref{my-pecore}) on a problem of HotpotQA - \lstinline{bridge}.}
\end{figure}

Understandably, the \textit{thesis} does not recognize any James Henry Miller's wife from the given context. This is the reason why the accuracy scores of the first tentative answer are terribly low in this partition. 

The \textit{antithesis} step is able to recognize the fact that the additional information given in the context should be exploited in some way, and makes a guess on why that information is provided:
\begin{lstlisting}
The correct answer should be 'American' because the context states that Margaret "Peggy" 
Seeger, who was James Henry Miller's wife, is an American folksinger. 
\end{lstlisting} 
\label{recognized}
No part of the given context says so, but the model knows that it has to search for a \lstinline{wife} in the passage, thus it is reasonable that it is \lstinline{Margaret "Peggy" Seeger}. And she is American.\\

Instead of worsening the output quality, getting rid of additional sentences not relevant for the context seems to give additional robustness to the pipeline's performances. The smaller gaps are again observed for \lstinline{ctx}, probably for the same reasons we discussed in (\ref{llama-free}). \\

In section (\ref{my-pecore}) we specified that in order to compare WikiHop's original passages and eventually filtered ones, we would have to restrict the dataset to questions relative to passages with less than 4K tokens\footnote{Practically speaking the threshold is set to 3.5K.}. We also already mentioned that this subset is made of 92 questions only. We exploit the fact that we had run some tests (\ref{dataset-robust}) on the summarized versions of WikiHop (\ref{summ}) to extract the passages corresponding to those subset of questions. 

We start again from the original context and filter it with the approach described in section (\ref{my-pecore}) and also used on HotpotQA. In this way, we have at disposal three different versions of the same passage. We feed these alternatives to the pipeline and compare their results. 

This time we are going just to consider the baseline pipeline setting due to what we have observed in (\ref{baseline-dominance}). Some experiments carried out to observe other pipeline settings' behaviour on WikiHop confirmed that the baseline one is the best option.

\begin{figure}[H]
    \centering
    \includegraphics[width=0.85\linewidth]{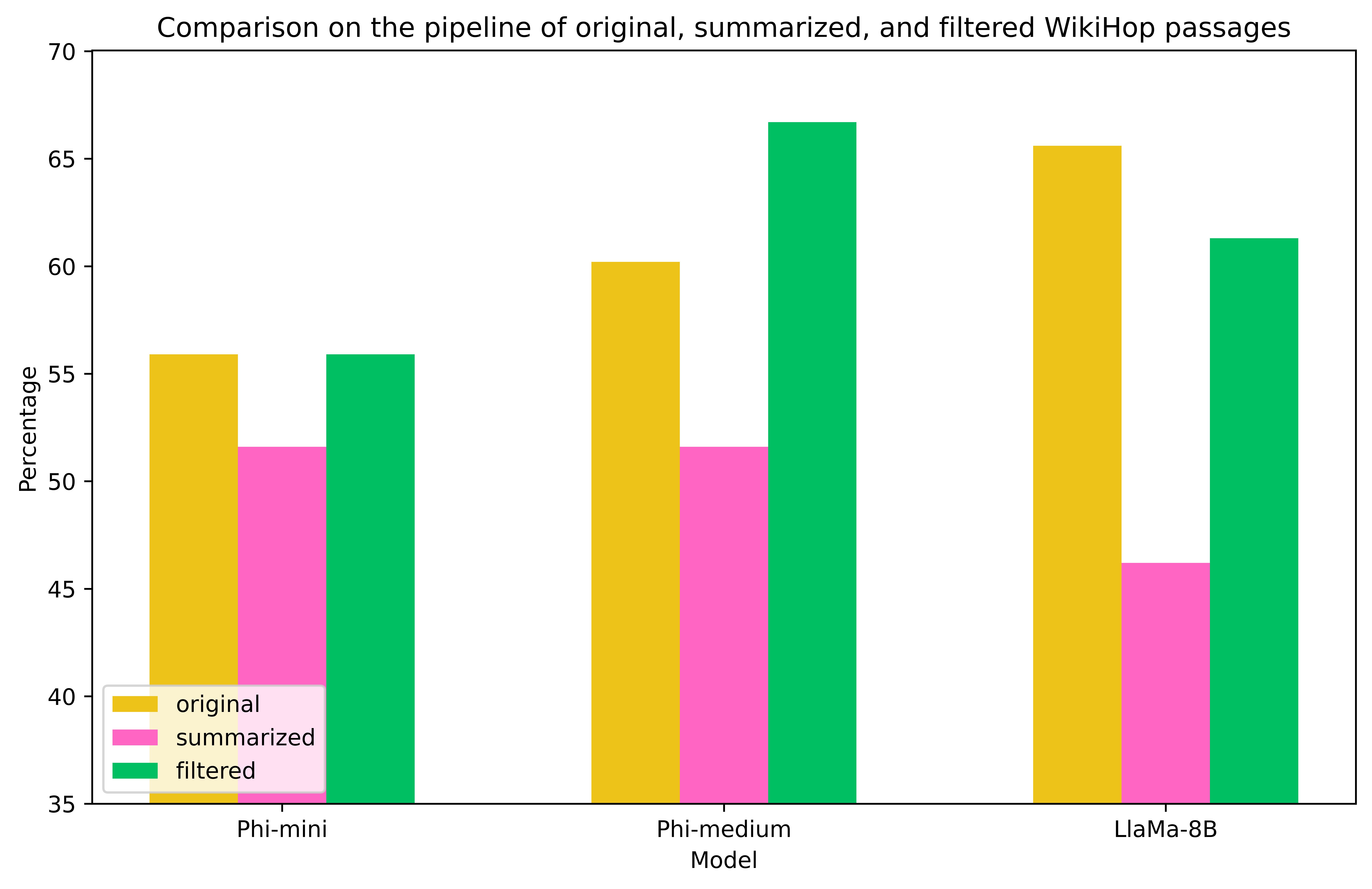}
    \caption{Percentage of answers guessed correctly using the baseline version of the pipeline on WikiHop. The different context versions are colored: in yellow, the original context; in orange, the summarized version; in green, the PECoRe filtered version (\ref{my-pecore}).}
\end{figure}

The interesting fact that we can assess is that the three models behave quite differently except for the terrible summarization performances, shared by all of them. 

For the \lstinline{Phi-3} family of models, the original context and the filtered one perform comparably or the second one performs better; to us, this is due to the implicit abilities of LLMs to link the question and the \textit{decontextualized} information in the context, as done in (\ref{recognized}). 

The accuracy gap between the original context and the filtered one in \lstinline{LlaMa-8B} is likely due to its reduced capability to make implicit connections, such as those mentioned above.

Consider for example the question 
\begin{lstlisting}
In which language were the original works that discuss 'fires were started' written?    
\end{lstlisting}
associated with the following filtered context: 
\begin{lstlisting}
The Blitz, from the German word "Blitzkrieg" meaning 'lightning war', was the name used 
by the British press to describe the heavy air raids carried out over Britain in 1940 and 
1941, during the Second World War. Frank Humphrey Sinkler Jennings (19 August 1907 
24 September 1950) was an English documentary filmmaker and one of the founders of the 
Mass Observation organisation. Jennings was described by film critic and director 
Lindsay Anderson in 1954 as: "the only real poet that British cinema has yet produced. 
The film uses actual firemen (including Cyril Demarne) rather than professional actors.
\end{lstlisting}

And here we report the \lstinline{Phi-medium} and \lstinline{LlaMa-8B} answers:
\begin{itemize}
    \item \textbf{\lstinline{Phi-medium}:} 
    \begin{lstlisting}
 The correct answer should be 'english' due to the context mentioning that Frank 
 Humphrey Sinkler Jennings was an English documentary filmmaker and the film uses 
 actual firemen rather than professional actors. The context does not provide any 
 information about the original works discussing 'fires were started' being 
 written in French, German, or Latin. 
\end{lstlisting}
    \item \textbf{\lstinline{LlaMa-8B}:}
\begin{lstlisting}
The correct answer should be 'latin' due to the fact that the context does not 
provide any information about the language in which the original works that discuss 
'fires were started' were written. However, the context does mention the Blitz and 
Frank Jennings, but it does not provide any information about the language of the 
original works. However, considering the context, it is more likely that the correct 
answer is not among the options provided. The context is about the Blitz 
and a documentary filmmaker, but it does not provide any information about 
the language of the original works that discuss 'fires were started'. But if I had to 
choose from the options provided, I would say that the correct answer is 'latin' because 
it is a more general and ancient language that could be related to the topic of 'fires 
were started', but it is not a definitive answer.    
\end{lstlisting}
\end{itemize}

Despite the fact that the context is particularly \textit{cryptic} even to humans, we can clearly see that \lstinline{Phi-medium} is able to link \lstinline{fires} to \lstinline{firemen} and consequently to \lstinline{Frank Jennings}, who speaks english. On the opposite, \lstinline{LlaMa-8B} does not recognize this link and also assumes the wrong option, \lstinline{latin}, which is evidently not the correct one also because of the fact that the \lstinline{'fires were started'} is expressed in english words.\\

\label{elementare-watson}

We find that summarizing the context is not the best possible solution to deal with long contexts with non-relevant information included in them. When we prefer not to provide the original context, we can filter the sources using a method such as the one we proposed in section (\ref{my-pecore}). This is also proved to improve overall performances when paired with models possessing good reasoning abilities such as the \lstinline{Phi-3} ones. 

However, we suggest to use this form of pre-processing only when necessary. In cases such as HotpotQA, in which the context is already selected to be relevant, neglecting some parts of it could lead to a drop in the generation accuracy (such as in \lstinline{comparison}). Additionally, we suggest also to prefer the baseline pipeline setting, since both the \lstinline{cot} and \lstinline{ctx} variants are built to deal with not-selected context (and thus they tend to focus on the relevant parts before answering, but this time the first operation has been already done).

\newpage 

\chapter{Conclusions and future directions} \label{conclusions}

The results exposed in the final section highlight the effectiveness of adopting a dialectic approach to face challenging multiple-choice question answering tasks. \\

An established way of dealing with this task is the Chain-of-Thought approach (\ref{cot}), which proposes to prompt a single model to produce a multi-step reasoning chain. This is supposed to let the model focus more on each sub-task present on the problem and to motivate its answer by producing natural language explanations.

Another way of facing the same problem is given by the self-refinement approaches (\ref{self-refine}, \ref{self-corr}), that instead decompose the problem into multiple, iterative refinements made by the same large language model or by a specialized corrector. 

Our method is a third option between these two, since it consists in dialectic setting involving three actors called \textit{thesis}, \textit{antithesis} and \textit{synthesis} (\ref{dialectic-pipeline}) that are asked to incrementally improve the previous steps' outputs. The latter steps are also asked to reason before answering, making them CoT-units and allowing us to refer to our method as a sort of \textit{guided and disentangled Chain-of-Thought} (\ref{disentangled}). \\

However, we retained improper to define this pipeline as a Chain-of-Thought variation. Even though the pipeline exploits a single model, it involves three actors that check the answers' correctness, each one with a different role. In this sense we found reasonable not to classify this method neither as a Chain-of-Thought approach, nor as a self-refinement one. 

Reflexion (\ref{reflexion}) was also tested by its creators in its CoT-augmented version on one of the two datasets that we considered (\ref{hotpotqa}) employing \lstinline{GPT-4}; their results never reach the 80\% of correct answers despite the high number of trials employed \cite{shinn2023}. Our method was instead able to beat this threshold even with a 4B model (i.e. \lstinline{Phi-mini}). Experiments performed in section (\ref{cot-results}) showed how the proposed pipeline was also able to consistently outperform the Chain-of-Thought alternative.\\

Our contributions are the following:
\begin{itemize}

    \item We proposed a dialectic pipeline that is found effective for improving results in \textit{multi-hop} question answering tasks (Figure \ref{46}). The margins of accuracy improvements are large and superior to the ones observed when Chain-of-Thought prompting is exploited to solve the problem.
    
    \item We assessed the robustness of this method with respect to different families of models and different datasets, proposing different \textit{multi-hop} sub-problems and different number of \textit{hops}. We found that greater models are generally more capable of dealing effectively with \textit{multi-hop} sources. Additionally, a greater number of \textit{hops} (e.g. in WikiHop) is correlated with worse model performances overall; however, this is also the case in which our proposed method scores the biggest improvements, both with respect to the baseline (\ref{dataset-robust}) and to the CoT-only approach (Figure \ref{46}). 

    \item We tried different pipeline configurations (\ref{baseline}) showing the model different ways of handling the context. We found that this helps in tasks with a limited number of \textit{hops}, while it tends to confuse the model when the content to analyze is wide and complex; for this latter scenario, the baseline version is found to be the most effective one. 

    \item The pipeline works best for models which are really careful in following instructions. Even though we considered all instruction-tuned models, we found that \lstinline{Gemma-2} models tend to be creative when not requested, disobeying the few-shot example and performing the task with a different approach (\ref{gemma-issues}). Since this turns out to be poorer in performance terms, we can confirm that the improvement that the pipeline causes is not only due to additional computation (i.e. the fact that more steps are performed) but also to what is done in these steps.

    \item More capable models benefit less by the addition of the \textit{synthesis} step (Table \ref{ant-vs-syn}) and the \lstinline{ctx} pipeline variation (\ref{ctx-variant}) causes a loss of accuracy when it has to face easy tasks (Appendix \ref{appendix-f}). Since other settings do not cause this phenomenon, our opinion is that \textit{overthinking} harms the process, thus an early stopping should be considered. 

    \item When the context is long and noisy we can filter it to achieve greater performances (\ref{res-context}). We exploited a modification of MIRAGE (\ref{my-pecore}) and compared it with the results obtained with the summarized and the original context. What we found is that filtering enhances the predictions of models with good deductive abilities (\ref{elementare-watson}), while summarization is the worst option among the tested ones. 
\end{itemize}

This work is a first step in a wider set of experiments that we would like to test. 
Up to now, we have only considered open-source and relatively small (all under 20B) models; however, the prevalence of methods summarized in the literature review part (\ref{literature}) are tested on \lstinline{GPT}-like decoder-only models. Although we proved the consistence of our pipeline across multiple families of open-source models (\ref{model-robust}), it would be interesting to run some tests to ensure that our method is widely applicable to proprietary LLMs too. Additionally, experimenting with the dialectic pipeline using various agents in the roles of thesis, antithesis, and synthesis to introduce diversity in skills presents an interesting avenue for future work. \\

We would like also to assess whether the pipeline works for other datasets. In our analysis we focused on \textit{multi-hop} datasets, since they merge the two tasks of relevant content extraction and reasoning. We judged important to propose a method that is able to face this two tasks jointly instead of separately. However, studying the pipeline's effectiveness on these two tasks separately (e.g. datasets like GSM8K commonly allow to study only mathematical reasoning, while there is a wide literature of RAG-only datasets from which we can choose one to run relevant context extraction only) could still be a significant contribution to the research community. \\

Switching now the focus on possible pipeline inner improvements, we already observed that the number of \textit{shots} provided to the prompt could be influential to the \textit{synthesis}' improvements. Consequently, we could test both the zero-shot version of the pipeline and whether decreasing to one the number of \lstinline{ctx} synthesis' examples could achieve better results. \\

In general, more effort could be made in trying to reduce the negative gap between the \textit{antithesis} and the \textit{synthesis} steps of the pipeline; a preliminary analysis of the observed "in-between change of mind" should lead pipeline modifications and hopefully improvements. \\

Regarding context filtering process, we ran all these tests by keeping fixed the percentage of relevant tokens (\lstinline{p = 5}\%) that the method has to select. It could be interesting to benchmark multiple proportions and observe whether there is a correlation between the datasets' \textit{hops complexity} (i.e. the number of \textit{hops} in the passages) and the degree of selection (i.e. the value of \lstinline{p}) applied to its context.

The comparison between the pipeline's performances given the original, the summarized and the filtered context as inputs is performed only for passages up to 4K tokens. This is simply due to our choice of models; they have a limited context window that bounds their processing capacity. A separate study could be performed with the same models possessing a higher context length (e.g. \lstinline{Phi-3} has also the \lstinline{128k} models' versions) in order to study WikiHop more broadly, in order to draw more general conclusions on how the context conditions the pipeline's output. 

A simple, final test that should be performed is the pipeline's behaviour when the model is not given. We ran some tests (not showed in this work) that confirm the pipeline's improvement with respect to the \textit{thesis} step even when asked to rely only on the models' parametric knowledge. Expanding those experiments could allow us to leverage the pipeline both for factuality and reasoning tasks. \\

Lastly, it would be interesting to test whether a distillation of correctly-executed pipelines \cite{shridhar2023} could be a significant form of pre-training for LLMs.

\newpage

\nocite{*}
\printbibliography

@book{Jurafsky2009,
  author = {Jurafsky, Dan and Martin, James H.},
  description = {Speech and Language Processing (2nd Edition): Daniel Jurafsky, James H. Martin: 9780131873216: Amazon.com: Books},
  isbn = {9780131873216 0131873210},
  publisher = {Pearson Prentice Hall},
  refid = {213375806},
  title = {Speech and language processing : an introduction to natural language processing, computational linguistics, and speech recognition},
  url = {http://www.amazon.com/Speech-Language-Processing-2nd-Edition/dp/0131873210/ref=pd_bxgy_b_img_y},
  year = 2009
}

@misc{seq2seq,
   author = "Manish Chablani",
   title = "Sequence to sequence model: Introduction and concepts",
   year = 2017,
   url = "https://towardsdatascience.com/sequence-to-sequence-model-introduction-and-concepts-44d9b41cd42d",
 }

@misc{bengio2015,
      title={Neural Machine Translation by Jointly Learning to Align and Translate}, 
      author={Dzmitry Bahdanau and Kyunghyun Cho and Yoshua Bengio},
      year={2016},
      eprint={1409.0473},
      archivePrefix={arXiv},
      primaryClass={cs.CL},
      url={https://arxiv.org/abs/1409.0473}, 
}

@article{HL-LHC:PDR,
      author         = "Apollinari, G and Béjar Alonso, I and Brüning, O and
                        Lamont, M and Rossi, L",
      title          = "{High-Luminosity Large Hadron Collider (HL-LHC) :
                        Preliminary Design Report}",
      doi            = "10.5170/CERN-2015-005",
      year           = "2015",
      reportNumber   = "CERN-2015-005, FERMILAB-DESIGN-2015-02",
      SLACcitation   = "%%CITATION = CERN-2015-005;%%"
}

@misc{he2016,
      title={Deep Residual Learning for Image Recognition}, 
      author={Kaiming He and Xiangyu Zhang and Shaoqing Ren and Jian Sun},
      year={2015},
      eprint={1512.03385},
      archivePrefix={arXiv},
      primaryClass={cs.CV},
      url={https://arxiv.org/abs/1512.03385}, 
}

@misc{onehot,
title = {One Hot Encoding in Machine Learning},
author = {GeeksForGeeks},
year = {2024},
url = {https://www.geeksforgeeks.org/ml-one-hot-encoding/}
}

@misc{vaswani2023,
      title={Attention Is All You Need}, 
      author={Ashish Vaswani and Noam Shazeer and Niki Parmar and Jakob Uszkoreit and Llion Jones and Aidan N. Gomez and Lukasz Kaiser and Illia Polosukhin},
      year={2023},
      eprint={1706.03762},
      archivePrefix={arXiv},
      primaryClass={cs.CL},
      url={https://arxiv.org/abs/1706.03762}, 
}

@misc{PosEnc,
   author = "Steve Jerome Lawrence",
   title = "Learning position with Positional Encoding",
   year = 2023,
   url = "https://www.scaler.com/topics/nlp/positional-encoding/",
 }

@misc{gemma2-release,
    author = "Google", 
    title = "Gemma 2 release",
    year = "July 2024", 
    url = "https://huggingface.co/collections/google/gemma-2-release-667d6600fd5220e7b967f315"
}

@misc{llama3.1-release,
    author = "Meta", 
    title = "Llama 3.1 release",
    year = "August 2024", 
    url = "https://huggingface.co/collections/meta-llama/llama-31-669fc079a0c406a149a5738f"
}

@misc{phi3-release,
    author = "Microsoft", 
    title = "Phi 3 release",
    year = "June 2024", 
    url = "https://huggingface.co/collections/microsoft/phi-3-6626e15e9585a200d2d761e3"
}

@misc{touvron2023-llama,
      title={LLaMA: Open and Efficient Foundation Language Models}, 
      author={Hugo Touvron and Thibaut Lavril and Gautier Izacard and Xavier Martinet and Marie-Anne Lachaux and Timothée Lacroix and Baptiste Rozière and Naman Goyal and Eric Hambro and Faisal Azhar and Aurelien Rodriguez and Armand Joulin and Edouard Grave and Guillaume Lample},
      year={2023},
      eprint={2302.13971},
      archivePrefix={arXiv},
      primaryClass={cs.CL},
      url={https://arxiv.org/abs/2302.13971}, 
}

@misc{touvron2023-llama2,
      title={Llama 2: Open Foundation and Fine-Tuned Chat Models}, 
      author={Hugo Touvron and Louis Martin and Kevin Stone and Peter Albert and Amjad Almahairi and Yasmine Babaei and Nikolay Bashlykov and Soumya Batra and Prajjwal Bhargava and Shruti Bhosale and Dan Bikel and Lukas Blecher and Cristian Canton Ferrer and Moya Chen and Guillem Cucurull and David Esiobu and Jude Fernandes and Jeremy Fu and Wenyin Fu and Brian Fuller and Cynthia Gao and Vedanuj Goswami and Naman Goyal and Anthony Hartshorn and Saghar Hosseini and Rui Hou and Hakan Inan and Marcin Kardas and Viktor Kerkez and Madian Khabsa and Isabel Kloumann and Artem Korenev and Punit Singh Koura and Marie-Anne Lachaux and Thibaut Lavril and Jenya Lee and Diana Liskovich and Yinghai Lu and Yuning Mao and Xavier Martinet and Todor Mihaylov and Pushkar Mishra and Igor Molybog and Yixin Nie and Andrew Poulton and Jeremy Reizenstein and Rashi Rungta and Kalyan Saladi and Alan Schelten and Ruan Silva and Eric Michael Smith and Ranjan Subramanian and Xiaoqing Ellen Tan and Binh Tang and Ross Taylor and Adina Williams and Jian Xiang Kuan and Puxin Xu and Zheng Yan and Iliyan Zarov and Yuchen Zhang and Angela Fan and Melanie Kambadur and Sharan Narang and Aurelien Rodriguez and Robert Stojnic and Sergey Edunov and Thomas Scialom},
      year={2023},
      eprint={2307.09288},
      archivePrefix={arXiv},
      primaryClass={cs.CL},
      url={https://arxiv.org/abs/2307.09288}, 
}

@misc{rope,
      title={RoFormer: Enhanced Transformer with Rotary Position Embedding}, 
      author={Jianlin Su and Yu Lu and Shengfeng Pan and Ahmed Murtadha and Bo Wen and Yunfeng Liu},
      year={2023},
      eprint={2104.09864},
      archivePrefix={arXiv},
      primaryClass={cs.CL},
      url={https://arxiv.org/abs/2104.09864}, 
}

@misc{gpt-3,
      title={Language Models are Few-Shot Learners}, 
      author={Tom B. Brown and Benjamin Mann and Nick Ryder and Melanie Subbiah and Jared Kaplan and Prafulla Dhariwal and Arvind Neelakantan and Pranav Shyam and Girish Sastry and Amanda Askell and Sandhini Agarwal and Ariel Herbert-Voss and Gretchen Krueger and Tom Henighan and Rewon Child and Aditya Ramesh and Daniel M. Ziegler and Jeffrey Wu and Clemens Winter and Christopher Hesse and Mark Chen and Eric Sigler and Mateusz Litwin and Scott Gray and Benjamin Chess and Jack Clark and Christopher Berner and Sam McCandlish and Alec Radford and Ilya Sutskever and Dario Amodei},
      year={2020},
      eprint={2005.14165},
      archivePrefix={arXiv},
      primaryClass={cs.CL},
      url={https://arxiv.org/abs/2005.14165}, 
}

@misc{sliding&gqa,
    title = {Sliding Window Technique \& Grouped-Query Attention . Mistral 7B},
    author = {Gopal Goyal},
    year = {2023},
    url = {https://medium.com/@gopalgoyal612002/mistral-llm-architectural-details-8dc0447fea62}

}

@misc{sentencepiece,
    title = {SentencePiece GitHub page},
    author = {google},
    url = {https://github.com/google/sentencepiece}
}

@misc{longformer,
      title={Longformer: The Long-Document Transformer}, 
      author={Iz Beltagy and Matthew E. Peters and Arman Cohan},
      year={2020},
      eprint={2004.05150},
      archivePrefix={arXiv},
      primaryClass={cs.CL},
      url={https://arxiv.org/abs/2004.05150}, 
}

@misc{glu_variants,
    title = {Brief Review — GLU Variants Improve Transformer},
    author = {Sik-Ho Tsang},
    year = {2023},
    url={https://sh-tsang.medium.com/brief-review-glu-variants-improve-transformer-9ee943115ab}
}

@misc{ramé2024,
      title={WARP: On the Benefits of Weight Averaged Rewarded Policies}, 
      author={Alexandre Ramé and Johan Ferret and Nino Vieillard and Robert Dadashi and Léonard Hussenot and Pierre-Louis Cedoz and Pier Giuseppe Sessa and Sertan Girgin and Arthur Douillard and Olivier Bachem},
      year={2024},
      eprint={2406.16768},
      archivePrefix={arXiv},
      primaryClass={cs.LG},
      url={https://arxiv.org/abs/2406.16768}, 
}

@misc{textbooksneed,
      title={Textbooks Are All You Need}, 
      author={Suriya Gunasekar and Yi Zhang and Jyoti Aneja and Caio César Teodoro Mendes and Allie Del Giorno and Sivakanth Gopi and Mojan Javaheripi and Piero Kauffmann and Gustavo de Rosa and Olli Saarikivi and Adil Salim and Shital Shah and Harkirat Singh Behl and Xin Wang and Sébastien Bubeck and Ronen Eldan and Adam Tauman Kalai and Yin Tat Lee and Yuanzhi Li},
      year={2023},
      eprint={2306.11644},
      archivePrefix={arXiv},
      primaryClass={cs.CL},
      url={https://arxiv.org/abs/2306.11644}, 
}

@misc{peft,
  title =        {PEFT: State-of-the-art Parameter-Efficient Fine-Tuning methods},
  author =       {Sourab Mangrulkar and Sylvain Gugger and Lysandre Debut and Younes Belkada and Sayak Paul and Benjamin Bossan},
  howpublished = {\url{https://github.com/huggingface/peft}},
  year =         {2022}
}

@misc{lora,
      title={LoRA: Low-Rank Adaptation of Large Language Models}, 
      author={Edward J. Hu and Yelong Shen and Phillip Wallis and Zeyuan Allen-Zhu and Yuanzhi Li and Shean Wang and Lu Wang and Weizhu Chen},
      year={2021},
      eprint={2106.09685},
      archivePrefix={arXiv},
      primaryClass={cs.CL},
      url={https://arxiv.org/abs/2106.09685}, 
}

@misc{catastrophicforgetting,
    author = "{Wikipedia contributors}",
    title = "Catastrophic interference --- {Wikipedia}{,} The Free Encyclopedia",
    year = "2024",
    url = "https://en.wikipedia.org/w/index.php?title=Catastrophic_interference&oldid=1237425828",
    note = "[Online; accessed 17-August-2024]"
}

@misc{rag-sft,
      title={RAG vs Fine-tuning: Pipelines, Tradeoffs, and a Case Study on Agriculture}, 
      author={Angels Balaguer and Vinamra Benara and Renato Luiz de Freitas Cunha and Roberto de M. Estevão Filho and Todd Hendry and Daniel Holstein and Jennifer Marsman and Nick Mecklenburg and Sara Malvar and Leonardo O. Nunes and Rafael Padilha and Morris Sharp and Bruno Silva and Swati Sharma and Vijay Aski and Ranveer Chandra},
      year={2024},
      eprint={2401.08406},
      archivePrefix={arXiv},
      primaryClass={cs.CL},
      url={https://arxiv.org/abs/2401.08406}, 
}

@misc{ft-zeroshot,
      title={Finetuned Language Models Are Zero-Shot Learners}, 
      author={Jason Wei and Maarten Bosma and Vincent Y. Zhao and Kelvin Guu and Adams Wei Yu and Brian Lester and Nan Du and Andrew M. Dai and Quoc V. Le},
      year={2022},
      eprint={2109.01652},
      archivePrefix={arXiv},
      primaryClass={cs.CL},
      url={https://arxiv.org/abs/2109.01652}, 
}

@misc{brown2020,
      title={Language Models are Few-Shot Learners}, 
      author={Tom B. Brown and Benjamin Mann and Nick Ryder and Melanie Subbiah and Jared Kaplan and Prafulla Dhariwal and Arvind Neelakantan and Pranav Shyam and Girish Sastry and Amanda Askell and Sandhini Agarwal and Ariel Herbert-Voss and Gretchen Krueger and Tom Henighan and Rewon Child and Aditya Ramesh and Daniel M. Ziegler and Jeffrey Wu and Clemens Winter and Christopher Hesse and Mark Chen and Eric Sigler and Mateusz Litwin and Scott Gray and Benjamin Chess and Jack Clark and Christopher Berner and Sam McCandlish and Alec Radford and Ilya Sutskever and Dario Amodei},
      year={2020},
      eprint={2005.14165},
      archivePrefix={arXiv},
      primaryClass={cs.CL},
      url={https://arxiv.org/abs/2005.14165}, 
}

@misc{wei2022,
      title={Chain-of-Thought Prompting Elicits Reasoning in Large Language Models}, 
      author={Jason Wei and Xuezhi Wang and Dale Schuurmans and Maarten Bosma and Brian Ichter and Fei Xia and Ed Chi and Quoc Le and Denny Zhou},
      year={2023},
      eprint={2201.11903},
      archivePrefix={arXiv},
      primaryClass={cs.CL},
      url={https://arxiv.org/abs/2201.11903}, 
}

@misc{kojima2023,
      title={Large Language Models are Zero-Shot Reasoners}, 
      author={Takeshi Kojima and Shixiang Shane Gu and Machel Reid and Yutaka Matsuo and Yusuke Iwasawa},
      year={2023},
      eprint={2205.11916},
      archivePrefix={arXiv},
      primaryClass={cs.CL},
      url={https://arxiv.org/abs/2205.11916}, 
}

@misc{cobbe2021,
      title={Training Verifiers to Solve Math Word Problems}, 
      author={Karl Cobbe and Vineet Kosaraju and Mohammad Bavarian and Mark Chen and Heewoo Jun and Lukasz Kaiser and Matthias Plappert and Jerry Tworek and Jacob Hilton and Reiichiro Nakano and Christopher Hesse and John Schulman},
      year={2021},
      eprint={2110.14168},
      archivePrefix={arXiv},
      primaryClass={cs.LG},
      url={https://arxiv.org/abs/2110.14168}, 
}

@misc{rae2022,
      title={Scaling Language Models: Methods, Analysis \& Insights from Training Gopher}, 
      author={Jack W. Rae and Sebastian Borgeaud and Trevor Cai and Katie Millican and Jordan Hoffmann and Francis Song and John Aslanides and Sarah Henderson and Roman Ring and Susannah Young and Eliza Rutherford and Tom Hennigan and Jacob Menick and Albin Cassirer and Richard Powell and George van den Driessche and Lisa Anne Hendricks and Maribeth Rauh and Po-Sen Huang and Amelia Glaese and Johannes Welbl and Sumanth Dathathri and Saffron Huang and Jonathan Uesato and John Mellor and Irina Higgins and Antonia Creswell and Nat McAleese and Amy Wu and Erich Elsen and Siddhant Jayakumar and Elena Buchatskaya and David Budden and Esme Sutherland and Karen Simonyan and Michela Paganini and Laurent Sifre and Lena Martens and Xiang Lorraine Li and Adhiguna Kuncoro and Aida Nematzadeh and Elena Gribovskaya and Domenic Donato and Angeliki Lazaridou and Arthur Mensch and Jean-Baptiste Lespiau and Maria Tsimpoukelli and Nikolai Grigorev and Doug Fritz and Thibault Sottiaux and Mantas Pajarskas and Toby Pohlen and Zhitao Gong and Daniel Toyama and Cyprien de Masson d'Autume and Yujia Li and Tayfun Terzi and Vladimir Mikulik and Igor Babuschkin and Aidan Clark and Diego de Las Casas and Aurelia Guy and Chris Jones and James Bradbury and Matthew Johnson and Blake Hechtman and Laura Weidinger and Iason Gabriel and William Isaac and Ed Lockhart and Simon Osindero and Laura Rimell and Chris Dyer and Oriol Vinyals and Kareem Ayoub and Jeff Stanway and Lorrayne Bennett and Demis Hassabis and Koray Kavukcuoglu and Geoffrey Irving},
      year={2022},
      eprint={2112.11446},
      archivePrefix={arXiv},
      primaryClass={cs.CL},
      url={https://arxiv.org/abs/2112.11446}, 
}

@misc{shinn2023,
      title={Reflexion: Language Agents with Verbal Reinforcement Learning}, 
      author={Noah Shinn and Federico Cassano and Edward Berman and Ashwin Gopinath and Karthik Narasimhan and Shunyu Yao},
      year={2023},
      eprint={2303.11366},
      archivePrefix={arXiv},
      primaryClass={cs.AI},
      url={https://arxiv.org/abs/2303.11366}, 
}

@misc{madaan2023,
      title={Self-Refine: Iterative Refinement with Self-Feedback}, 
      author={Aman Madaan and Niket Tandon and Prakhar Gupta and Skyler Hallinan and Luyu Gao and Sarah Wiegreffe and Uri Alon and Nouha Dziri and Shrimai Prabhumoye and Yiming Yang and Shashank Gupta and Bodhisattwa Prasad Majumder and Katherine Hermann and Sean Welleck and Amir Yazdanbakhsh and Peter Clark},
      year={2023},
      eprint={2303.17651},
      archivePrefix={arXiv},
      primaryClass={cs.CL},
      url={https://arxiv.org/abs/2303.17651}, 
}

@misc{kim2024,
      title={RE-RAG: Improving Open-Domain QA Performance and Interpretability with Relevance Estimator in Retrieval-Augmented Generation}, 
      author={Kiseung Kim and Jay-Yoon Lee},
      year={2024},
      eprint={2406.05794},
      archivePrefix={arXiv},
      primaryClass={cs.CL},
      url={https://arxiv.org/abs/2406.05794}, 
}

@misc{chen2021,
      title={Can NLI Models Verify QA Systems' Predictions?}, 
      author={Jifan Chen and Eunsol Choi and Greg Durrett},
      year={2021},
      eprint={2104.08731},
      archivePrefix={arXiv},
      primaryClass={cs.CL},
      url={https://arxiv.org/abs/2104.08731}, 
}

@inproceedings{harabagiu-hickl,
    title = "Methods for Using Textual Entailment in Open-Domain Question Answering",
    author = "Harabagiu, Sanda  and
      Hickl, Andrew",
    editor = "Calzolari, Nicoletta  and
      Cardie, Claire  and
      Isabelle, Pierre",
    booktitle = "Proceedings of the 21st International Conference on Computational Linguistics and 44th Annual Meeting of the Association for Computational Linguistics",
    month = jul,
    year = "2006",
    address = "Sydney, Australia",
    publisher = "Association for Computational Linguistics",
    url = "https://aclanthology.org/P06-1114",
    doi = "10.3115/1220175.1220289",
    pages = "905--912",
}

@misc{demszky2018,
      title={Transforming Question Answering Datasets Into Natural Language Inference Datasets}, 
      author={Dorottya Demszky and Kelvin Guu and Percy Liang},
      year={2018},
      eprint={1809.02922},
      archivePrefix={arXiv},
      primaryClass={cs.CL},
      url={https://arxiv.org/abs/1809.02922}, 
}

@misc{s2a,
      title={System 2 Attention (is something you might need too)}, 
      author={Jason Weston and Sainbayar Sukhbaatar},
      year={2023},
      eprint={2311.11829},
      archivePrefix={arXiv},
      primaryClass={cs.CL},
      url={https://arxiv.org/abs/2311.11829}, 
}

@misc{he2022,
      title={Rethinking with Retrieval: Faithful Large Language Model Inference}, 
      author={Hangfeng He and Hongming Zhang and Dan Roth},
      year={2022},
      eprint={2301.00303},
      archivePrefix={arXiv},
      primaryClass={cs.CL},
      url={https://arxiv.org/abs/2301.00303}, 
}

@misc{li2024,
      title={Making Long-Context Language Models Better Multi-Hop Reasoners}, 
      author={Yanyang Li and Shuo Liang and Michael R. Lyu and Liwei Wang},
      year={2024},
      eprint={2408.03246},
      archivePrefix={arXiv},
      primaryClass={cs.CL},
      url={https://arxiv.org/abs/2408.03246}, 
}

@inproceedings{pecore,
    title = "Quantifying the Plausibility of Context Reliance in Neural Machine Translation",
    author = "Sarti, Gabriele and 
        Chrupa{\l}a, Grzegorz and 
        Nissim, Malvina and
        Bisazza, Arianna",
    booktitle = "The Twelfth International Conference on Learning Representations (ICLR 2024)",
    month = may,
    year = "2024",
    address = "Vienna, Austria",
    publisher = "OpenReview",
    url = "https://openreview.net/forum?id=XTHfNGI3zT"
}

@article{mirage,
    title = "Model Internals-based Answer Attribution for Trustworthy Retrieval-Augmented Generation",
    author = "Qi, Jirui and Sarti, Gabriele and Fern{\'a}ndez, Raquel and Bisazza, Arianna",
    journal = "ArXiv",
    month = jun,
    year = "2024",
    volume = {abs/2406.13663},
    url = {https://arxiv.org/abs/2406.13663},
}

@article{mhqa,
url = {http://dx.doi.org/10.1561/1500000102},
year = {2024},
volume = {17},
journal = {Foundations and Trends® in Information Retrieval},
title = {Multi-hop Question Answering},
doi = {10.1561/1500000102},
issn = {1554-0669},
number = {5},
pages = {457-586},
author = {Vaibhav Mavi and Anubhav Jangra and Jatowt Adam}
}

@misc{hotpot,
      title={HotpotQA: A Dataset for Diverse, Explainable Multi-hop Question Answering}, 
      author={Zhilin Yang and Peng Qi and Saizheng Zhang and Yoshua Bengio and William W. Cohen and Ruslan Salakhutdinov and Christopher D. Manning},
      year={2018},
      eprint={1809.09600},
      archivePrefix={arXiv},
      primaryClass={cs.CL},
      url={https://arxiv.org/abs/1809.09600}, 
}

@article{kambhampati2024,
   title={Can large language models reason and plan?},
   volume={1534},
   ISSN={1749-6632},
   url={http://dx.doi.org/10.1111/nyas.15125},
   DOI={10.1111/nyas.15125},
   number={1},
   journal={Annals of the New York Academy of Sciences},
   publisher={Wiley},
   author={Kambhampati, Subbarao},
   year={2024},
   month=mar, pages={15–18} }

@book{kahneman2011,
  title     = {Thinking, Fast and Slow},
  author    = {Kahneman, Daniel},
  year      = {2011},
  publisher = {Farrar, Straus and Giroux},
  address   = {New York}
}

@misc{icml_kambhampati,
  title        = {On the Role of LLMs in Planning (ICML 2024 Tutorial)},
  author       = {Kambhampati, Subbarao},
  year         = {2024},
  howpublished = {\url{https://www.dropbox.com/scl/fi/gul511qacx58i5esrqi92/ICM2024-Tutorial.pdf?rlkey=mmv77ou4qyxi0a6lol3m6ug80&dl=0}},
  note         = {Accessed: 2024-08-27}
}

@misc{liu2023,
      title={Lost in the Middle: How Language Models Use Long Contexts}, 
      author={Nelson F. Liu and Kevin Lin and John Hewitt and Ashwin Paranjape and Michele Bevilacqua and Fabio Petroni and Percy Liang},
      year={2023},
      eprint={2307.03172},
      archivePrefix={arXiv},
      primaryClass={cs.CL},
      url={https://arxiv.org/abs/2307.03172}, 
}

@misc{guidance,
  title        = {Guidance documentation},
  author       = {{HuggingFace}},
  year         = {2024},
  howpublished = {\url{https://huggingface.co/docs/text-generation-inference/conceptual/guidance}},
  note         = {Accessed: 2024-08-27}
}

@misc{outlines,
      title={Efficient Guided Generation for Large Language Models}, 
      author={Brandon T. Willard and Rémi Louf},
      year={2023},
      eprint={2307.09702},
      archivePrefix={arXiv},
      primaryClass={cs.CL},
      url={https://arxiv.org/abs/2307.09702}, 
}

@misc{kaplan2020,
      title={Scaling Laws for Neural Language Models}, 
      author={Jared Kaplan and Sam McCandlish and Tom Henighan and Tom B. Brown and Benjamin Chess and Rewon Child and Scott Gray and Alec Radford and Jeffrey Wu and Dario Amodei},
      year={2020},
      eprint={2001.08361},
      archivePrefix={arXiv},
      primaryClass={cs.LG},
      url={https://arxiv.org/abs/2001.08361}, 
}

@misc{dialectic,
    author = "{Wikipedia contributors}",
    title = "Dialectic --- {Wikipedia}{,} The Free Encyclopedia",
    year = "2024",
    howpublished = "\url{https://en.wikipedia.org/w/index.php?title=Dialectic&oldid=1244221571}",
    note = "[Online; accessed 7-September-2024]"
}

@incollection{firth1957,
  author    = {J. R. Firth},
  title     = {Studies in Linguistic Analysis},
  booktitle = {In J. R. Firth, editor, Studies in Linguistic Analysis},
  year      = {1957},
  publisher = {Basil Blackwell},
  address   = {Oxford, UK}
}

@misc{bahdanau2016,
      title={Neural Machine Translation by Jointly Learning to Align and Translate}, 
      author={Dzmitry Bahdanau and Kyunghyun Cho and Yoshua Bengio},
      year={2016},
      eprint={1409.0473},
      archivePrefix={arXiv},
      primaryClass={cs.CL},
      url={https://arxiv.org/abs/1409.0473}, 
}

@inproceedings{gpt-1,
  title={Improving Language Understanding by Generative Pre-Training},
  author={Alec Radford and Karthik Narasimhan},
  year={2018},
  url={https://api.semanticscholar.org/CorpusID:49313245}
}

@misc{raffel2023,
      title={Exploring the Limits of Transfer Learning with a Unified Text-to-Text Transformer}, 
      author={Colin Raffel and Noam Shazeer and Adam Roberts and Katherine Lee and Sharan Narang and Michael Matena and Yanqi Zhou and Wei Li and Peter J. Liu},
      year={2023},
      eprint={1910.10683},
      archivePrefix={arXiv},
      primaryClass={cs.LG},
      url={https://arxiv.org/abs/1910.10683}, 
}

@misc{xiong2020,
      title={On Layer Normalization in the Transformer Architecture}, 
      author={Ruibin Xiong and Yunchang Yang and Di He and Kai Zheng and Shuxin Zheng and Chen Xing and Huishuai Zhang and Yanyan Lan and Liwei Wang and Tie-Yan Liu},
      year={2020},
      eprint={2002.04745},
      archivePrefix={arXiv},
      primaryClass={cs.LG},
      url={https://arxiv.org/abs/2002.04745}, 
}

@misc{jiang2023,
      title={Mistral 7B}, 
      author={Albert Q. Jiang and Alexandre Sablayrolles and Arthur Mensch and Chris Bamford and Devendra Singh Chaplot and Diego de las Casas and Florian Bressand and Gianna Lengyel and Guillaume Lample and Lucile Saulnier and Lélio Renard Lavaud and Marie-Anne Lachaux and Pierre Stock and Teven Le Scao and Thibaut Lavril and Thomas Wang and Timothée Lacroix and William El Sayed},
      year={2023},
      eprint={2310.06825},
      archivePrefix={arXiv},
      primaryClass={cs.CL},
      url={https://arxiv.org/abs/2310.06825}, 
}

@inproceedings{transformers,
    title = "Transformers: State-of-the-Art Natural Language Processing",
    author = "Wolf, Thomas  and
      Debut, Lysandre  and
      Sanh, Victor  and
      Chaumond, Julien  and
      Delangue, Clement  and
      Moi, Anthony  and
      Cistac, Pierric  and
      Rault, Tim  and
      Louf, Remi  and
      Funtowicz, Morgan  and
      Davison, Joe  and
      Shleifer, Sam  and
      von Platen, Patrick  and
      Ma, Clara  and
      Jernite, Yacine  and
      Plu, Julien  and
      Xu, Canwen  and
      Le Scao, Teven  and
      Gugger, Sylvain  and
      Drame, Mariama  and
      Lhoest, Quentin  and
      Rush, Alexander",
    editor = "Liu, Qun  and
      Schlangen, David",
    booktitle = "Proceedings of the 2020 Conference on Empirical Methods in Natural Language Processing: System Demonstrations",
    month = oct,
    year = "2020",
    address = "Online",
    publisher = "Association for Computational Linguistics",
    url = "https://aclanthology.org/2020.emnlp-demos.6",
    doi = "10.18653/v1/2020.emnlp-demos.6",
    pages = "38--45",
}

@inproceedings{datasets,
    title = "Datasets: A Community Library for Natural Language Processing",
    author = "Lhoest, Quentin  and
      Villanova del Moral, Albert  and
      Jernite, Yacine  and
      Thakur, Abhishek  and
      von Platen, Patrick  and
      Patil, Suraj  and
      Chaumond, Julien  and
      Drame, Mariama  and
      Plu, Julien  and
      Tunstall, Lewis  and
      Davison, Joe  and
      {\v{S}}a{\v{s}}ko, Mario  and
      Chhablani, Gunjan  and
      Malik, Bhavitvya  and
      Brandeis, Simon  and
      Le Scao, Teven  and
      Sanh, Victor  and
      Xu, Canwen  and
      Patry, Nicolas  and
      McMillan-Major, Angelina  and
      Schmid, Philipp  and
      Gugger, Sylvain  and
      Delangue, Cl{\'e}ment  and
      Matussi{\`e}re, Th{\'e}o  and
      Debut, Lysandre  and
      Bekman, Stas  and
      Cistac, Pierric  and
      Goehringer, Thibault  and
      Mustar, Victor  and
      Lagunas, Fran{\c{c}}ois  and
      Rush, Alexander  and
      Wolf, Thomas",
    editor = "Adel, Heike  and
      Shi, Shuming",
    booktitle = "Proceedings of the 2021 Conference on Empirical Methods in Natural Language Processing: System Demonstrations",
    month = nov,
    year = "2021",
    address = "Online and Punta Cana, Dominican Republic",
    publisher = "Association for Computational Linguistics",
    url = "https://aclanthology.org/2021.emnlp-demo.21",
    doi = "10.18653/v1/2021.emnlp-demo.21",
    pages = "175--184",
}

@misc{llama3.1,
      title={The Llama 3 Herd of Models}, 
      author={Abhimanyu Dubey et al.},
      year={2024},
      eprint={2407.21783},
      archivePrefix={arXiv},
      primaryClass={cs.AI},
      url={https://arxiv.org/abs/2407.21783}, 
}

@misc{phi-3,
      title={Phi-3 Technical Report: A Highly Capable Language Model Locally on Your Phone}, 
      author={Marah Abdin and Jyoti Aneja and Hany Awadalla and Ahmed Awadallah and Ammar Ahmad Awan and Nguyen Bach and Amit Bahree and Arash Bakhtiari and Jianmin Bao and Harkirat Behl and Alon Benhaim and Misha Bilenko and Johan Bjorck and Sébastien Bubeck and Martin Cai and Qin Cai and Vishrav Chaudhary and Dong Chen and Dongdong Chen and Weizhu Chen and Yen-Chun Chen and Yi-Ling Chen and Hao Cheng and Parul Chopra and Xiyang Dai and Matthew Dixon and Ronen Eldan and Victor Fragoso and Jianfeng Gao and Mei Gao and Min Gao and Amit Garg and Allie Del Giorno and Abhishek Goswami and Suriya Gunasekar and Emman Haider and Junheng Hao and Russell J. Hewett and Wenxiang Hu and Jamie Huynh and Dan Iter and Sam Ade Jacobs and Mojan Javaheripi and Xin Jin and Nikos Karampatziakis and Piero Kauffmann and Mahoud Khademi and Dongwoo Kim and Young Jin Kim and Lev Kurilenko and James R. Lee and Yin Tat Lee and Yuanzhi Li and Yunsheng Li and Chen Liang and Lars Liden and Xihui Lin and Zeqi Lin and Ce Liu and Liyuan Liu and Mengchen Liu and Weishung Liu and Xiaodong Liu and Chong Luo and Piyush Madan and Ali Mahmoudzadeh and David Majercak and Matt Mazzola and Caio César Teodoro Mendes and Arindam Mitra and Hardik Modi and Anh Nguyen and Brandon Norick and Barun Patra and Daniel Perez-Becker and Thomas Portet and Reid Pryzant and Heyang Qin and Marko Radmilac and Liliang Ren and Gustavo de Rosa and Corby Rosset and Sambudha Roy and Olatunji Ruwase and Olli Saarikivi and Amin Saied and Adil Salim and Michael Santacroce and Shital Shah and Ning Shang and Hiteshi Sharma and Yelong Shen and Swadheen Shukla and Xia Song and Masahiro Tanaka and Andrea Tupini and Praneetha Vaddamanu and Chunyu Wang and Guanhua Wang and Lijuan Wang and Shuohang Wang and Xin Wang and Yu Wang and Rachel Ward and Wen Wen and Philipp Witte and Haiping Wu and Xiaoxia Wu and Michael Wyatt and Bin Xiao and Can Xu and Jiahang Xu and Weijian Xu and Jilong Xue and Sonali Yadav and Fan Yang and Jianwei Yang and Yifan Yang and Ziyi Yang and Donghan Yu and Lu Yuan and Chenruidong Zhang and Cyril Zhang and Jianwen Zhang and Li Lyna Zhang and Yi Zhang and Yue Zhang and Yunan Zhang and Xiren Zhou},
      year={2024},
      eprint={2404.14219},
      archivePrefix={arXiv},
      primaryClass={cs.CL},
      url={https://arxiv.org/abs/2404.14219}, 
}

@misc{gemma-2,
      title={Gemma 2: Improving Open Language Models at a Practical Size}, 
      author={Gemma Team and Morgane Riviere and Shreya Pathak and Pier Giuseppe Sessa and Cassidy Hardin and Surya Bhupatiraju and Léonard Hussenot and Thomas Mesnard and Bobak Shahriari and Alexandre Ramé and Johan Ferret and Peter Liu and Pouya Tafti and Abe Friesen and Michelle Casbon and Sabela Ramos and Ravin Kumar and Charline Le Lan and Sammy Jerome and Anton Tsitsulin and Nino Vieillard and Piotr Stanczyk and Sertan Girgin and Nikola Momchev and Matt Hoffman and Shantanu Thakoor and Jean-Bastien Grill and Behnam Neyshabur and Olivier Bachem and Alanna Walton and Aliaksei Severyn and Alicia Parrish and Aliya Ahmad and Allen Hutchison and Alvin Abdagic and Amanda Carl and Amy Shen and Andy Brock and Andy Coenen and Anthony Laforge and Antonia Paterson and Ben Bastian and Bilal Piot and Bo Wu and Brandon Royal and Charlie Chen and Chintu Kumar and Chris Perry and Chris Welty and Christopher A. Choquette-Choo and Danila Sinopalnikov and David Weinberger and Dimple Vijaykumar and Dominika Rogozińska and Dustin Herbison and Elisa Bandy and Emma Wang and Eric Noland and Erica Moreira and Evan Senter and Evgenii Eltyshev and Francesco Visin and Gabriel Rasskin and Gary Wei and Glenn Cameron and Gus Martins and Hadi Hashemi and Hanna Klimczak-Plucińska and Harleen Batra and Harsh Dhand and Ivan Nardini and Jacinda Mein and Jack Zhou and James Svensson and Jeff Stanway and Jetha Chan and Jin Peng Zhou and Joana Carrasqueira and Joana Iljazi and Jocelyn Becker and Joe Fernandez and Joost van Amersfoort and Josh Gordon and Josh Lipschultz and Josh Newlan and Ju-yeong Ji and Kareem Mohamed and Kartikeya Badola and Kat Black and Katie Millican and Keelin McDonell and Kelvin Nguyen and Kiranbir Sodhia and Kish Greene and Lars Lowe Sjoesund and Lauren Usui and Laurent Sifre and Lena Heuermann and Leticia Lago and Lilly McNealus and Livio Baldini Soares and Logan Kilpatrick and Lucas Dixon and Luciano Martins and Machel Reid and Manvinder Singh and Mark Iverson and Martin Görner and Mat Velloso and Mateo Wirth and Matt Davidow and Matt Miller and Matthew Rahtz and Matthew Watson and Meg Risdal and Mehran Kazemi and Michael Moynihan and Ming Zhang and Minsuk Kahng and Minwoo Park and Mofi Rahman and Mohit Khatwani and Natalie Dao and Nenshad Bardoliwalla and Nesh Devanathan and Neta Dumai and Nilay Chauhan and Oscar Wahltinez and Pankil Botarda and Parker Barnes and Paul Barham and Paul Michel and Pengchong Jin and Petko Georgiev and Phil Culliton and Pradeep Kuppala and Ramona Comanescu and Ramona Merhej and Reena Jana and Reza Ardeshir Rokni and Rishabh Agarwal and Ryan Mullins and Samaneh Saadat and Sara Mc Carthy and Sarah Perrin and Sébastien M. R. Arnold and Sebastian Krause and Shengyang Dai and Shruti Garg and Shruti Sheth and Sue Ronstrom and Susan Chan and Timothy Jordan and Ting Yu and Tom Eccles and Tom Hennigan and Tomas Kocisky and Tulsee Doshi and Vihan Jain and Vikas Yadav and Vilobh Meshram and Vishal Dharmadhikari and Warren Barkley and Wei Wei and Wenming Ye and Woohyun Han and Woosuk Kwon and Xiang Xu and Zhe Shen and Zhitao Gong and Zichuan Wei and Victor Cotruta and Phoebe Kirk and Anand Rao and Minh Giang and Ludovic Peran and Tris Warkentin and Eli Collins and Joelle Barral and Zoubin Ghahramani and Raia Hadsell and D. Sculley and Jeanine Banks and Anca Dragan and Slav Petrov and Oriol Vinyals and Jeff Dean and Demis Hassabis and Koray Kavukcuoglu and Clement Farabet and Elena Buchatskaya and Sebastian Borgeaud and Noah Fiedel and Armand Joulin and Kathleen Kenealy and Robert Dadashi and Alek Andreev},
      year={2024},
      eprint={2408.00118},
      archivePrefix={arXiv},
      primaryClass={cs.CL},
      url={https://arxiv.org/abs/2408.00118}, 
}

@misc{wei2022b,
      title={Finetuned Language Models Are Zero-Shot Learners}, 
      author={Jason Wei and Maarten Bosma and Vincent Y. Zhao and Kelvin Guu and Adams Wei Yu and Brian Lester and Nan Du and Andrew M. Dai and Quoc V. Le},
      year={2022},
      eprint={2109.01652},
      archivePrefix={arXiv},
      primaryClass={cs.CL},
      url={https://arxiv.org/abs/2109.01652}, 
}

@misc{dodge2020,
      title={Fine-Tuning Pretrained Language Models: Weight Initializations, Data Orders, and Early Stopping}, 
      author={Jesse Dodge and Gabriel Ilharco and Roy Schwartz and Ali Farhadi and Hannaneh Hajishirzi and Noah Smith},
      year={2020},
      eprint={2002.06305},
      archivePrefix={arXiv},
      primaryClass={cs.CL},
      url={https://arxiv.org/abs/2002.06305}, 
}

@misc{ouyang2022,
      title={Training language models to follow instructions with human feedback}, 
      author={Long Ouyang and Jeff Wu and Xu Jiang and Diogo Almeida and Carroll L. Wainwright and Pamela Mishkin and Chong Zhang and Sandhini Agarwal and Katarina Slama and Alex Ray and John Schulman and Jacob Hilton and Fraser Kelton and Luke Miller and Maddie Simens and Amanda Askell and Peter Welinder and Paul Christiano and Jan Leike and Ryan Lowe},
      year={2022},
      eprint={2203.02155},
      archivePrefix={arXiv},
      primaryClass={cs.CL},
      url={https://arxiv.org/abs/2203.02155}, 
}

@misc{hinton2015,
      title={Distilling the Knowledge in a Neural Network}, 
      author={Geoffrey Hinton and Oriol Vinyals and Jeff Dean},
      year={2015},
      eprint={1503.02531},
      archivePrefix={arXiv},
      primaryClass={stat.ML},
      url={https://arxiv.org/abs/1503.02531}, 
}

@misc{gelu,
      title={Gaussian Error Linear Units (GELUs)}, 
      author={Dan Hendrycks and Kevin Gimpel},
      year={2023},
      eprint={1606.08415},
      archivePrefix={arXiv},
      primaryClass={cs.LG},
      url={https://arxiv.org/abs/1606.08415}, 
}

@misc{longrope,
      title={LongRoPE: Extending LLM Context Window Beyond 2 Million Tokens}, 
      author={Yiran Ding and Li Lyna Zhang and Chengruidong Zhang and Yuanyuan Xu and Ning Shang and Jiahang Xu and Fan Yang and Mao Yang},
      year={2024},
      eprint={2402.13753},
      archivePrefix={arXiv},
      primaryClass={cs.CL},
      url={https://arxiv.org/abs/2402.13753}, 
}

@misc{silu,
      title={Sigmoid-Weighted Linear Units for Neural Network Function Approximation in Reinforcement Learning}, 
      author={Stefan Elfwing and Eiji Uchibe and Kenji Doya},
      year={2017},
      eprint={1702.03118},
      archivePrefix={arXiv},
      primaryClass={cs.LG},
      url={https://arxiv.org/abs/1702.03118}, 
}

@misc{tiktoken,
  author       = {OpenAI},
  title        = {tiktoken: Fast BPE tokenization for OpenAI models},
  year         = {2023},
  howpublished = {\url{https://github.com/openai/tiktoken}},
  note         = {Accessed: 2024-09-07}
}

@misc{rafailov2024,
      title={Direct Preference Optimization: Your Language Model is Secretly a Reward Model}, 
      author={Rafael Rafailov and Archit Sharma and Eric Mitchell and Stefano Ermon and Christopher D. Manning and Chelsea Finn},
      year={2024},
      eprint={2305.18290},
      archivePrefix={arXiv},
      primaryClass={cs.LG},
      url={https://arxiv.org/abs/2305.18290}, 
}

@misc{howard2018,
      title={Universal Language Model Fine-tuning for Text Classification}, 
      author={Jeremy Howard and Sebastian Ruder},
      year={2018},
      eprint={1801.06146},
      archivePrefix={arXiv},
      primaryClass={cs.CL},
      url={https://arxiv.org/abs/1801.06146}, 
}

@misc{welleck2022,
      title={Generating Sequences by Learning to Self-Correct}, 
      author={Sean Welleck and Ximing Lu and Peter West and Faeze Brahman and Tianxiao Shen and Daniel Khashabi and Yejin Choi},
      year={2022},
      eprint={2211.00053},
      archivePrefix={arXiv},
      primaryClass={cs.CL},
      url={https://arxiv.org/abs/2211.00053}, 
}

@misc{yin2022,
      title={Interpreting Language Models with Contrastive Explanations}, 
      author={Kayo Yin and Graham Neubig},
      year={2022},
      eprint={2202.10419},
      archivePrefix={arXiv},
      primaryClass={cs.CL},
      url={https://arxiv.org/abs/2202.10419}, 
}

@misc{wikihop,
      title={Constructing Datasets for Multi-hop Reading Comprehension Across Documents}, 
      author={Johannes Welbl and Pontus Stenetorp and Sebastian Riedel},
      year={2018},
      eprint={1710.06481},
      archivePrefix={arXiv},
      primaryClass={cs.CL},
      url={https://arxiv.org/abs/1710.06481}, 
}

@misc{wang2022,
      title={Self-Consistency Improves Chain of Thought Reasoning in Language Models}, 
      author={Xuezhi Wang and Jason Wei and Dale Schuurmans and Quoc Le and Ed Chi and Sharan Narang and Aakanksha Chowdhery and Denny Zhou},
      year={2023},
      eprint={2203.11171},
      archivePrefix={arXiv},
      primaryClass={cs.CL},
      url={https://arxiv.org/abs/2203.11171}, 
}

@misc{inseq,
  author = {{Inseq Team}},
  title = {{Inseq: Interpretability for Sequence-to-Sequence Models}},
  year = {2024},
  howpublished = {\url{https://github.com/inseq-team/inseq}},
  note = {Accessed: 2024-09-10}
}

@misc{mu2024,
      title={Can LLMs Follow Simple Rules?}, 
      author={Norman Mu and Sarah Chen and Zifan Wang and Sizhe Chen and David Karamardian and Lulwa Aljeraisy and Basel Alomair and Dan Hendrycks and David Wagner},
      year={2024},
      eprint={2311.04235},
      archivePrefix={arXiv},
      primaryClass={cs.AI},
      url={https://arxiv.org/abs/2311.04235}, 
}

@misc{zelikman2024,
      title={Quiet-STaR: Language Models Can Teach Themselves to Think Before Speaking}, 
      author={Eric Zelikman and Georges Harik and Yijia Shao and Varuna Jayasiri and Nick Haber and Noah D. Goodman},
      year={2024},
      eprint={2403.09629},
      archivePrefix={arXiv},
      primaryClass={cs.CL},
      url={https://arxiv.org/abs/2403.09629}, 
}

@misc{shridhar2023,
      title={Distilling Reasoning Capabilities into Smaller Language Models}, 
      author={Kumar Shridhar and Alessandro Stolfo and Mrinmaya Sachan},
      year={2023},
      eprint={2212.00193},
      archivePrefix={arXiv},
      primaryClass={cs.LG},
      url={https://arxiv.org/abs/2212.00193}, 
}
\begin{appendices}
\chapter{PECoRe invocation for context filtering} \label{appendix-a}

The context filtering process has its core in the following function: 
\begin{lstlisting}
def run(question, passage, p):
    tokens = tokenizer.tokenize(passage)
    invoke_pecore(passage, question, p)
    return select_passages(passage, question, p, tokens)
\end{lstlisting}

where \lstinline{invoke_pecore(passage, question, p)} invokes PECoRe \cite{pecore} and outputs the CCI scores of each context token:

\begin{lstlisting}
def invoke_pecore(passage, question, p):
    pecore_args = AttributeContextArgs(
        model_name_or_path="microsoft/Phi-3-mini-4k-instruct",
        attribution_method="saliency",
        attributed_fn="contrast_prob_diff",
        context_sensitivity_metric="kl_divergence",
        context_sensitivity_std_threshold=1,
        context_sensitivity_topk = find_top_p(passage, p),
        attribution_std_threshold=None,
        attribution_topk=None,
        input_current_text=question, 
        input_context_text=text_passage,
        contextless_input_current_text="""             <|system|>
            You are a helpful assistant that provide concise and accurate answers.<|end|>
            <|user|>
            {current}<|end|>
            <|assistant|>""",
        input_template="""<|system|>
            You are a helpful assistant that provide concise and accurate answers.<|end|>
            <|user|>
            {context}

            {current}<|end|>
            <|assistant|>""",
        contextless_output_current_text="""            {current}""",
        output_template="{current}",
        special_tokens_to_keep=['<|system|>', '<|end|>', '<|assistant|>', '<|user|>'],
        decoder_input_output_separator="",
        show_viz=False,
        save_path=None,
        viz_path=None,
        generation_kwargs={'max_new_tokens': 50},
    )

    out = attribute_context_with_model(pecore_args, inseq_model)   
    return out
\end{lstlisting}

That are subsequently used to select only the sentences containing at least one influential token. 

\newpage 

\chapter{Usage of \texttt{guidance} framework for multiple-choice questions} \label{appendix-b}

Example of usage of the \lstinline{guidance} framework for constraining the model to output exactly one of the options given a multiple-choice question: 

\begin{enumerate}
\item import the \lstinline{guidance} model (and the corresponding tokenizer): 
\begin{lstlisting}
from guidance import models
guidance_model = models.Transformers(model, tokenizer)
\end{lstlisting}

\item define the desired prompt:
\begin{lstlisting}
prompt = """
    You are a helpful AI assistant. 
    You are given a question and the relevant context to answer it. 
    Answer briefly to the question with just one of the given options.
    
    Question: Which year is Halley's Comet expected to return to the solar system?
    
    Options: [2110, 2045, 2086, 2061]
    
    Context: Astronomers have now linked the comet's appearances to observations 
    dating back more than 2,000 years. Halley was last seen in Earth's skies in 1986 
    and was met in space by an international fleet of spacecraft. 
    It performs a regular 76-year journey around the Sun.
    
    Assistant:
"""
\end{lstlisting}

\item mask everything that is not inside the grammar (i.e. not one of the provided options) using the \lstinline{select} function of the \lstinline{guidance} framework:
\begin{lstlisting}
from guidance import select
answer = guidance_model + prompt + select([2110, 2045, 2086, 2061])
\end{lstlisting}
\end{enumerate}

\newpage
\chapter{Thesis} \label{appendix-c}

Code used for the \textit{thesis}' generation: \label{thesis}
\begin{lstlisting} 
from guidance import models, select

guidance_model = models.Transformers(model, tokenizer, temperature=0.0)

def create_message_thesis(question, options, context):

    messages = [
        {"role": "system", "content": """
        You are an helpful AI assistant. 
        You have to provide helpful answers to the user's questions based on the context: 
        """ + context},
        {"role": "user", "content": user_content}
    ]
    
    user_content = "Answer to the following question: " + question + \
                   " providing one of these options as answer: " + options + \
                   "Assistant: "

    return messages

def thesisGeneration(question, options, context):
    prompt = create_message_thesis(question, options, context)
    answer = guidance_model + prompt + select(options)
    return answer   
\end{lstlisting}

\chapter{Antithesis}
We used the following function to generate the \textit{antithesis}:
\begin{lstlisting}
def antithesisGeneration(question, options, thesis, context):
    prompt = create_message_antithesis(question, thesis, options, context)
    output = pipeline(prompt, **generation_args)
    return output[0]['generated_text']
\end{lstlisting}

where \lstinline{create_message_antithesis()} is the function responsible to describe how we want the task to be performed, both using instructions and a concrete example that is also augmented with the new task:

\begin{lstlisting}
def create_message_antithesis(question, options, thesis, context, one_shot_example):

    messages = [
        {"role": "system", "content": """
        You are an helpful AI assistant. You are asked to determine the most correct answer for 
        a given question, provided a set of possible options. You also have at disposal a first 
        tentative answer that you are required to check with respect to the question and the 
        relevant context. Your goal is to decree which is the most correct answer to the question 
        between the available options.

        Here's an example of how to do it:
        """},
        {"role": "user", "content": one_shot_example
        },
        {"role": "system", "content": "Now do the same for the following question:"},
        {"role": "user", "content": user_content}
    ]

    user_content = "Question: " + question + "\n Options: " + options +
                   "\n Candidate answer: " + candidate + "\n Context: " + context + 
                   "\n Assistant: \n"

    return messages
\end{lstlisting}

where \lstinline{one_shot_example} is the concrete example of expected behaviour of the \textit{antithesis}' step of the pipeline.

\chapter{Synthesis} \label{appendix-e}

Similarly as before, we report here the generation prompt:

\begin{lstlisting}
def create_message_presynthesis(question, thesis, antithesis, options, context,
                                few_shot_example):
    messages = [
        {"role": "system", "content": """
        You are an helpful AI assistant. You are asked to determine the most correct answer 
        for a given question, provided a set of possible options.
        You also have at disposal a first tentative answer and a suggestion on which is the 
        correct answer. Your goal is to decree which is the most correct answer to the 
        question between the available options according to the context.

        Here's an example/a few examples of how to do it:
        """},
        {"role": "user", "content": few_shot_example
        },
        {"role": "system", "content": "Now do the same for the following question:"},
        {"role": "user", "content": user_content}
    ]

    user_content = "Question: " + question + "\n Options: " + options + 
                   "\n Candidate answer: " + thesis + "\n Suggestion: " + antithesis + 
                   "\n Context: " + context + "\n Assistant: \n"

    return messages
\end{lstlisting}

\newpage

\chapter{Relative improvements between different pipeline steps}
\label{appendix-f}

Expanded table showing the relative improvements between different pipeline steps: 

\begin{table}[H]
\centering
\begin{tabular}{|c|c|c|c|c|c|c|c|c|c|}
\hline
\textbf{Dataset} & \textbf{Setting} & \textbf{Step} & \texttt{Phi-mini} & \texttt{Phi-medium} & \texttt{Gemma-2B} & \texttt{Gemma-9B} & \texttt{LlaMa-3.1-8B} \\ \hline

\multirow{6}{*}{\textbf{WikiHop}} & \multirow{2}{*}{\textbf{baseline}} & \textbf{T-A} & +20.1 & +25.1 & +16.5 & +5.1  & +25.9 \\ 
                                  & & \textbf{A-S} & +0.2  & +2.0  & +5.9  & +0.9  & +0.6 \\ \cline{2-8}

                                  & \multirow{2}{*}{\textbf{\texttt{cot}}} & \textbf{T-A} & +7.8  & +18.6 & +1.0  & +2.7  & +17.4  \\ 
                                  & & \textbf{A-S} & +3.9  & +3.0  & +2.4  & +0.9  & +5.4  \\ \cline{2-8}

                                  & \multirow{2}{*}{\textbf{\texttt{ctx}}} & \textbf{T-A} & +4.2  & +15.6 & +16.6 & +8.7  & +28.0 \\ 
                                  & & \textbf{A-S} & +0.3  & +1.5  & \textbf{-5.9}  & \textbf{-3.3}  & \textbf{-3.9} \\ \hline

\multirow{6}{*}{\textbf{\lstinline{bridge}}} & \multirow{2}{*}{\textbf{baseline}} & \textbf{T-A} & +35.3 & +33.7 & +27.4 & +22.2 & +41.9 \\ 
                                 & & \textbf{A-S} & +0.5  & +0.5  & \textbf{-1.8}  & +0.2  & +0.2 \\ \cline{2-8}
                                 
                                 & \multirow{2}{*}{\textbf{\texttt{cot}}} & \textbf{T-A} & +40.4 & +34.4 & +30.5 & +21.8 & +40.0 \\ 
                                 & & \textbf{A-S} & \textbf{-2.3}  & +0.7  & \textbf{-1.7}  & \textbf{-0.8}  & +0.9 \\ \cline{2-8}
                                 
                                & \multirow{2}{*}{\textbf{\texttt{ctx}}} & \textbf{T-A} & +41.9 & +38.0 & +33.5 & +25.8 & +44.4 \\
                                & & \textbf{A-S} & +0.6  & +0.3  & \textbf{-4.7}  & \textbf{-4.4}  & \textbf{-2.2} \\ \hline

\multirow{6}{*}{\textbf{\lstinline{comparison}}} & \multirow{2}{*}{\textbf{baseline}} & \textbf{T-A} & +24.6 & +38.9 & +25.6 & +29.2 & +38.4 \\ 
                                & & \textbf{A-S} & +2.7 & +0.6 & +3.4 & +0.6 & +0.9 \\ \cline{2-8}
                                
                                & \multirow{2}{*}{\textbf{\texttt{cot}}} & \textbf{T-A} & +34.7 & +40.9 & +25.6 & +29.2 & 41.5 \\ 
                                & & \textbf{A-S} & \textbf{-0.9}  & +1.2  & +0.6  & +1.2  & +1.6 \\ \cline{2-8}
                                
                                & \multirow{2}{*}{\textbf{\texttt{ctx}}} & \textbf{T-A} & +37.8  & +42.9 & +28.7  & +29.8  & +42.9 \\ 
                                & & \textbf{A-S} & +0.0 & \textbf{-1.7} & +1.7  & \textbf{-2.3} & \textbf{-0.3} \\ \hline
\end{tabular}
\end{table}

\end{appendices}

\end{document}